\documentclass[twoside,11pt]{article}
\usepackage[utf8]{inputenc}
\usepackage{fancyhdr}
\usepackage{array}
\usepackage{amsmath,amsfonts,amssymb,graphicx}    
\usepackage{fontenc}
\usepackage{subfigure}   
\usepackage{float} 
\usepackage{indentfirst} 
\usepackage{bm}          
\usepackage{multicol}    
\usepackage{indentfirst} 
\usepackage{abstract}    
\usepackage{amsthm}      
\usepackage{booktabs}    
\usepackage{titlesec}
\usepackage{times}
\usepackage{wasysym}
\usepackage{multirow}
\usepackage{algorithm}

\usepackage{algpseudocode}
\usepackage{color}
\usepackage{threeparttable}
\usepackage[font=small,labelfont=bf,labelsep=none]{caption}
\usepackage{epstopdf}
\usepackage{extarrows}
\usepackage{appendix}
\usepackage[english]{babel}
\usepackage{natbib}
\usepackage{diagbox}

\newcommand\hc{\hat{c}}

\newcommand\hp{\hat{p}}

\newcommand\hs{\hat{s}}

\newcommand\hsigma{\hat{\sigma}}

\newcommand\hC{\hat{C}}

\newcommand\bbC{\mathbb{C}}

\newcommand\bbR{\mathbb{R}}

\newcommand\calD{\mathcal{D}}

\newcommand\calN{\mathcal{N}}

\newcommand\cvp{\stackrel{p}{\longrightarrow}}

\newcommand\cvd{\stackrel{d}{\longrightarrow}}
\newcommand\cvas{\stackrel{a.s.}{\longrightarrow}}
\newcommand\Def{\stackrel{\Delta}{=}}

\DeclareMathOperator{\ave}{Ave}

\usepackage[figuresright]{rotating}
\usepackage{geometry}
\usepackage{jmlr2e}

\jmlrheading{}{}{}{12/02}{}{Xuran00a}{Xuran Meng and Jianfeng Yao}

\ShortHeadings{Impact of classification difficulty on the weight matrices spectra in Deep Learning}{Meng and Yao}
\firstpageno{1}

\begin{document}

\title{Impact of classification difficulty on the weight matrices spectra in Deep Learning  and application to early-stopping}

\author{\name Xuran Meng \email u3007800@connect.hku.hk \\
       \addr Department of Statistics and Acturial Science\\
       University of Hong Kong\\
       Eastern and Western, HongKong
       \AND
       \name Jianfeng Yao \email jeffyao@hku.hk \\
       \addr Department of Statistics and Acturial Science\\
       University of Hong Kong\\
       Eastern and Western, HongKong}

\editor{~}

\maketitle

\begin{abstract}
Much research effort has been devoted to explaining the success of deep learning.  Random Matrix Theory (RMT) provides an emerging way to this end: spectral analysis of large random matrices involved in a trained deep neural network (DNN) such as weight matrices or Hessian matrices with respect to the stochastic gradient descent algorithm.  To have more comprehensive understanding of weight matrices spectra, we conduct extensive experiments on weight matrices in different modules, e.g., layers, networks and data sets. Following the previous work of \cite{martin2018implicit}, we classify the spectra in the terminal stage  into three main types: Light Tail (LT), Bulk Transition period  (BT) and Heavy Tail(HT). These different types, especially HT, implicitly indicate some regularization in the DNNs. A main contribution from the paper is that we identify the difficulty of the classification problem as a driving factor for the appearance of heavy tail in weight matrices spectra.
Higher the classification difficulty, higher 
the chance for HT to appear. 
Moreover, the classification difficulty can be affected by the signal-to-noise ratio of the dataset, or by the complexity of the classification problem (complex features, large number of classes) as well. 
Leveraging on this finding, we further propose 
a spectral criterion to detect the appearance of heavy tails and use it to early stop the training process without testing data.   Such early stopped DNNs have the merit of avoiding overfitting and   unnecessary extra training while preserving a much comparable generalization ability.
These findings from the paper are validated in several NNs, using Gaussian synthetic data and real data sets (MNIST and CIFAR10).

\end{abstract}

\begin{keywords}
  Deep Learning, Weight matrices, Heavy tailed spectrum, Early stopping
\end{keywords}

\section{Introduction}

In the past decade, deep learning \citep{lecun2015deep} has achieved
impressive success in numerous areas. Much research effort has since been concentrated on providing a rational explanation of the success. The task is difficult, particularly because the training of most successful deep neural networks  (DNNs)  relies on a collection of expert choices that determine the final structure of the DNNs. These expert choices include nonlinear activation, hidden layer architecture, loss function, back propagation algorithm and canonical datasets. Unfortunately, these empirical choices  usually bring nonlinearity into  the model,  and nonconvexity of optimization into the training process.  As a matter of consequence, practitioners of deep learning are facing certain lack of general guidelines about the ``right choices'' to design and train an effective DNN for their own machine learning problem.

 To make progress on the understanding of existing trained and
 successful DNNs, it is important to explore their properties in some 
 principled way.  To this end, 
 a popular way has recently emerged in the literature,  namely
 spectral analysis of various large characteristic random matrices of
 the DNNs, such as the  Hessian matrices of the  back-propagation
 algorithm,   weight matrices between different layers, and covariance
 matrices of output features.  
 Actually, such  spectral analysis  helps to gain insights into
 the behavior of DNNs, and many researchers believe that these
 spectral properties, once better understood, will provide clues to
 improvements in deep learning training
 \citep{dauphin2014identifying,papyan2019spectrum,PapyanHierarchical,sagun2017eigenvalues,Yaozhewei,granziol2020random,Pennington2019nonlinear,ge2021largedimensional}.
 Recently, \cite{martin2018implicit} studied the empirical spectra
 distributions (ESD) of weight matrices in different neural networks,
 and observed a  ``Phase Transition 5+1" phenomenon  in these ESDs. Interestingly, the phenomenon highlights
 signatures of traditionally regularized statistical models
 even though there is no set-up of any traditional regularization in
 the DNNs. Here, traditional  regularization refers to the minimization
 of an explicitly defined and penalized loss function of the form  $L(\theta)+\alpha\cdot
 p(\theta)$ with some tuning parameter $\alpha$ ($\theta$ denotes all the parameters in the DNN).  However,  those well-known expert choices  such as early stopping  also
 produces a regularization effect  in  DNNs, and this is the reason
 why such expert choices are recommended for practitioners. Actually,
 \cite{kukacka2017regularization} presented about  50 different
 regularization techniques which may improve DNN's generalization. 
 Among them, 
 batch normalization, early stopping, dropout, and weight decay are a
 few commonly used ones. 

A main finding from  \cite{martin2018implicit} is that the effects of these regularization practices can be identified through the spectra of different weight matrices of a DNN. Moreover, the forms of these spectra in the ``5+1 phase transition" help assess the existence of some regularization in the DNN. For instance, if  these  spectra are far away from the Mar\v{c}enko-Pastur (MP) law, or the largest eigenvalue departs from the Tracy-Widom (TW) Law (see in Appendix\ref{sec: RMT}), there is strong evidence for the onset of more regular structures in the weight matrices. A connection between implicit regularization in a DNN and the forms of the spectra of its  weight matrices is thus established. Particularly, they considered  the evolution of weight matrices spectra during the training process of a DNN from its  start to its final stage (usually 200-400 epochs), and pointed out that in the late stage of the training, the deviation between the spectra and MP Law (namely the emergence of Heavy Tail) indicates some regularization technique which will improve the generalization error in some way. Indeed, the emergence of Heavy Tail is due to the complex correlations which are generated from such regularization technique. Some theoretical analysis of Heavy Tail is proposed to illustrate such correlations. Recently, 
\cite{gurbuzbalaban2021heavytail} pointed out that SGD can bring out heavy tails in linear regression settings, \cite{hodgkinson2021multiplicative} prove the phenomenon under a more generalized case.


A generally accepted definition of the concept of 
{\em Regularization} in a DNN refers to any supplementary mechanism that aims at making the model generalize better (see in \cite{kukacka2017regularization}). In the next work from \cite{martin2021predicting}, they claimed that the ``Heavy Tail based metrics can do much better—quantitatively better at discriminating among series of well-trained models with a given architecture; and qualitatively better at discriminating well-trained versus poorly trained models." An important issue we notice from spectral analysis is HT plays a more significant role for guiding the practice of Deep Learning . 
In our synthetic data experiments, an interesting phenomenon is that HT emerges when the training data quality is low. Unlike the 
factors proposed by \cite{hodgkinson2021multiplicative} such as increasing the step size/decreasing the batch size or increasing $L_2$ regularization to achieve Heavy Tail, we emphasize a special factor that affect the emergence of Heavy Tail--classification difficulty in data sets. More difficult to classify, higher possibility HT appears. We say the factor is specical in the reason that it includes two different cases when HT appears. One case is the poor data quality, the emergence of HT indicates an attempt for DNNs to extract more features and increase testing accuracy. Another case is modern data sets with complex features and hard for the information extraction, the emergence of HT indicates an attempt for DNNs to identify data information and improve generalization error. Both cases are in high classification difficulty, but the training results could be entirely different. In the second case, the emergence of HT indicates the feature extraction and some regularization for improving the test accuracy. However, in the first case, HT indicates some excessive information extraction and thus overfitting happens.

The paper is aim to give a more comprehensive understanding of HT phenomenon.  Intuitively, the classification difficulty is a metric for how difficult the data sets will get classification or get prediction under certain model architectures. Nevertheless the classification difficulty is vague and hard for a clear definition, there are a lot of indicators that give a significant impact. For example, the decrease of signal-to-noise ratio (SNR), the increment of class numbers $K$ or the complex features in the data sets would all increase the difficult level for the data sets to get classification. The factor we find is indeed consistent with the statement in \cite{martin2018implicit}. In line with \cite{martin2021predicting}, the weight matrix spectrum could be regarded as information encoder during training procedure. Leveraging on this findings, a spectral criterion is proposed in section \ref{sec:specCriterion} to guide the early stopping in practice. To our best knowledge, it is the first quantitative criterion based on the spectra to guide early stopping in practice.  Without prior information, HT indicates some regularization at play or some problematic issues such as overfitting in the training procedure. The spectral criterion depends on the detection of appearance of HT. We summarize our contributions below:
\begin{enumerate}
    \item We identify that the classification difficulty is a driving factor for the appearance of heavy tail in weight matrices spectra. Experiments are conducted on both synthetic and real data sets. Decrease of SNR or increment of class numbers $K$ in Gaussian data experiments, which both bring in higher classification difficulty, will generate heavy tail at the end of training. In real data experiments, heavy tails appears more in experiments with CIFAR10 than with MNIST due to the complex features in CIFAR10.
    
    \item Following the previous work in \cite{martin2018implicit}, the terminal spectrum bulk type in our paper are divided into three parts: Light Tail (LT), Bulk Transition period (BT) and Heavy Tail (HT). With decrease of classification difficulty, the terminal weight matrix spectrum has the phase transition from HT $\to$ BT $\to$ LT.
    
    \item Leveraging on this finding, we propose a spectral criterion to guide the early stopping without access of testing data. The HT(BT)-based spectral criterion could not only cut off a large training time with just a little drop of test accuracy, but also avoid over-fitting even when the training accuracy is increasing. 
\end{enumerate}

The paper summarizes experimental results in the sections \ref{sec:Gaussian} and \ref{sec:realdata}, and the details of spectral criterion for early stopping could be refered in section \ref{sec:specCriterion}.

\section{Experiments with Gaussian Data}
\label{sec:Gaussian}

Most deep neural networks are applied to real world data. These networks are
however too complex in general for developing a  rigorous
theoretical analysis on the spectral behavior. For a theoretical investigation, a widely used
and simplified model is the Gaussian input model
\citep{lee2018deep}. By examining a well-defined Gaussian model for
classification, we establish the evidence for a classification difficulty driving regularization via the confirmation of
a transition phenomenon in the spectra of network's weight matrices
in the order of HT $\to$ BT $\to$ LT.
Moreover, the transition is quantitatively controlled by a single parameter of data quality in the Gaussian model, namely its signal-to-noise ratio (SNR).

\bigskip
\noindent \textbf{Empirically Results:} Signal-to-noise ratio (SNR) is a common indicator to measure data quality and greatly impacts the classification difficulty from a Gaussian model. We empirically examine the spectra by
tuning SNR in different architectures:
\begin{enumerate}
\item
  Different NN structures: wider but shallower, or narrower but deeper. These structures are similar to the various well known NNs' fully connected denser layers, such as LeNet and MiniAlexNet, etc.;
\item
  Different layers in neural networks: all weight matrices in different layers have spectrum transition driven by SNR;
\item
  Different class numbers in input data: the spectrum transition is always observed in different class numbers, and HT is more likely to emerge with increment of $K$. 
\end{enumerate}

Table~\ref{Summary} gives a short  summary of the findings.

\begin{table}[htbp]
\centering
\caption{~~Summary of spectrum transition in a controlled Gaussian  model with $K$ classes and various SNRs.}
\begin{tabular}{|c|c|c|}
\hline
\diagbox{SNR}{~}    & Type of spectra & Number of spikes   \\ \hline
Weak   & Heavy Tail                                                                                    & $K-1$ or $K$ \\ \hline
Middle & \begin{tabular}[c]{@{}c@{}}Heavy Tail\\ $\downarrow$\\ MP+Bleed out\end{tabular} & $K-1$ or $K$ \\ \hline
  Strong & Light Tail (MP Law)                                                                           & $K-1$ or $K$ \\ \hline
\end{tabular}
\label{Summary}
\end{table}

We empirically observe the spectrum transition in all settings.
The transition is fully driven by the classification difficulty.
Therefore, in this Gaussian model,  the indicated implicit
regularization in the trained  DNN is data-effective,  directly
determined by the difficulty.  Precisely,
under low level SNR or high class numbers, the weight matrices of a DNN  deviates far away from
the common  MP model. Instead, they are connected to very different
complex Random Matrix models. The decrease of  classification difficulty drives the
weight matrices from Heavy Tailed model into  MP models at the final training epoch.



\subsection{Gaussian Data Sets}
\label{sec:Gausssiandesign}
In the multi-classification task, Gaussian model is a commonly used
model for assessing theoretical properties of a learning system \citep{lee2018deep}.
In this model with $K$ classes, data from a class $k\in\{1,\ldots,K\}$
are $p$-dimensional vector of  the form
\begin{equation}\label{eq:hik}
  h_{i,k}=\mu_k+\varepsilon_{i,k},\quad 1\le i\le n_k,
\end{equation}
where $\mu_k\in\bbR^p$ is the class mean,
$\varepsilon_{i,k}\stackrel{iid}{\sim} \calN(0,\sigma^2I_p)$ are
Gaussian noise, $n_k$ is the total number of observation from class
$k$. (This Gaussian data model is referred to as the $K$-way ANOVA model  in the statistics literature.) The signal-to-noise ratio (SNR) for this $K$-class  Gaussian model is defined as 
\begin{equation}\label{eq:SNR}
  \text{SNR}=\mathop{\ave}_{  \{k,k'\} }\frac{||\mu_k-\mu_{k'}||}{\sigma}.
\end{equation}
Here $||\cdot||$ denotes the  Euclidean norm in $\bbR^p$, and the
average is taken over the ${K \choose 2}$ pairs of classes. 

We aim to examine the impact of classification difficulty on the weight matrices spectra in a trained NN for such Gaussian data.  We thus consider two settings for the class means $\{\mu_k\}$ which lead to different families of SNRs. In all the remaining discussions, we will take $\sigma=1$. 

\subsubsection*{Dataset $\calD_1(\delta)$: class means with randomly shuffled locations}

\noindent Consider a base mean vector
$u=(m,\ldots,m,m+\delta,\ldots,m+\delta)^T\in\bbR^p$ where half of the
components are  $m$, and the other half, $m+\delta$.
For the class means $\mu_k$, we reshuffle the locations of these
components randomly (and independently).  Formally, 
for each class $k$, we pick a random subset $I_k\subset \{1,...,p\}$,
of size $p/2$, and define the mean for this class as
\begin{equation}\label{eq:muk}
  \mu_k=m\mathbf{1}_{I_k}+(m+\delta)\mathbf{1}_{I_k^c}.
\end{equation}
Here for a subset $A\subset \{1,...,p\}$, $\mathbf{1}_A$ is the indicator vector of $A$ with coordinates $\mathbf{1}_A(i) = \mathbf{1}_{\{i\in A\}}$  ($1\le i\le p$).

This setting with randomized locations is motivated by an essential empirical finding from exploring a few classical trained  DNNs such as MiniAlexNet and LeNet. Indeed, we found that in these DNNs, the global histograms of the features from all  the neurons are pretty similar, with very comparable means and variances, for various NNs; the differences across the  NNs are that high and  low values of the features appear in different neurons (locations). The randomly shuffled means used in our experiments are designed to imitate these working mechanisms observed in real-world NNs.

It follows that for the difference $\mu_k-\mu_{k'}=(z_j),~ 1\leq j\leq p$   from two classes
$k\ne k'$, its coordinates $z_j$ take on the values $-\delta$, 0 and
$\delta$ with probability $\frac14$, $\frac12$ and $\frac14$,
respectively. Clearly, the model SNR will depend on the tuning
parameter $\delta$.
By Hoeffding inequality, we first conclude that 
$$
P\left(\left|\frac{||\mu_k-\mu_{k'}||^2}{p}-\frac{\delta^2}{2}\right|\leq \epsilon\delta^2\right)\geq 1-\exp{(-2\epsilon^2p)},
$$
or equivalently, 
$$
P\left(\frac{\delta}{\sqrt 2}\sqrt{1-2\epsilon}\leq \frac{||\mu_k-\mu_{k'}||}{\sqrt p}\leq \frac{\delta}{\sqrt 2}\sqrt{1+2\epsilon}\right)\geq 1-\exp{(-2\epsilon^2p)}
$$
Note that $\sqrt{1+x}\leq1+x$, $\sqrt{1-x}\geq1-x$ when $0<x<1$.  By
taking  $\epsilon=\sqrt{\log p/p}$, we conclude that with probability at least $1-1/p^{2}$, 
$$
\left|||\mu_k-\mu_{k'}||-\delta\sqrt{\frac{p}{2}}\right|\leq\delta\sqrt{2\log p}.
$$
Therefore at a first-order approximation, the SNR~\eqref{eq:SNR} in
this Gaussian model is (with $\sigma=1$),
\begin{equation}\label{eq:SNR2}
  \text{SNR}=\mathop{\ave}_{  \{k,k'\} }\frac{||\mu_k-\mu_{k'}||}{\sigma}\sim \delta\sqrt{\frac{p}{2}}.
  \qquad \square
\end{equation}

\subsubsection*{Dataset $\calD_2(t)$: class means of ETF type}

\noindent Consider the family of vectors $\{v_k\}_{1\le k\le K}$ where $v_k$ is
defined by 
$$
v_k= {\mathbf{1}}_{\{i=k\}} -\frac1K {\mathbf{1}}_{\{1\le i\le K\}},\quad 1\leq i\leq p.
$$
So $v_k$ has support on $\{1,\ldots,K\}$ and $||v_k||=\sqrt{(K-1)/K}$.
The normalized family $\{v_k/||v_k||\}$ is called a  $K$-standard
ETF structure \citep{PapyanNeuralCollapse}.

We define the $k$-th class
mean as $\mu_k=tv_k$, and use  the scale parameter  $t>0$ to tune the SNR
of the model. It is easy to see that $||\mu_k-\mu_{k'}||=\sqrt{2}t$
so that  the model SNR is
\begin{equation}
  \label{eq:SNR3}
  \text{SNR}=\mathop{\ave}_{  \{k,k'\}}\frac{||\mu_k-\mu_{k'}||}{\sigma}=||\mu_k-\mu_{k'}||= \sqrt{2}t.
\end{equation}
\citep{PapyanNeuralCollapse}  has  shown that the  ETF structure is
an optimal position for the final training outputs.
Many experiments on real data sets lead to ETF structure for final engineered features. From a layer-peered perspective as mentioned in \citep{ji2021gradient}, each layer in NN can be regarded as an essential part of feature engineering, and the feature is extracted layer by layer. The ETF structure model considers that the first Dense layer behind the convolution layer is already close to the end of feature extraction.

\begin{table}[htbp]
  \centering
  \caption{~~Ranges of SNRs observed in various   datasets/ networks combinations.\label{tbl:SNR-ranges}}
  \label{tb: TPrange}
  \begin{tabular}{clcclc}\toprule
    & \multicolumn{2}{c}{$\calD_1(\delta)$}   &\quad&   \multicolumn{2}{c}{$\calD_2(t)$}          \\ 
    &   $\delta$  & SNR interval    & & $t$  &  SNR interval    \\ \cline{2-3} \cline{5-6} \\[-4mm]
    NN1 &     0.01      & [0.01, 1.19]  &&   0.08  &  [0.08, 4.80]  \\ 
    &     0.05      & [1.20, 2.00]  &     &&      \\
    NN2 &     0.005     & [0.005, 0.4]  && 0.08   & [0.08, 4.80] \\ \bottomrule
  \end{tabular}
\end{table}

In our experiments, we take $m=-0.2$ (and $\sigma=1$).  
The size of each class $k$ is $n_k=7500$ in the training dataset, and
$n_k=800$ in test dataset.
The number of classes $K$ takes on the values $\{2,5,8\}$  on all datasets.
Table~\ref{tbl:SNR-ranges} gives the ranges of the model SNR
observed  in different dataset/NN combinations with the chosen values
of tuning parameters $\delta$ and $t$.


\subsubsection{Structure of neural networks}

\begin{figure}[htbp]
\centering
\subfigure[NN1]{
\begin{minipage}[t]{0.4\linewidth}
\centering
\includegraphics[width=2in]{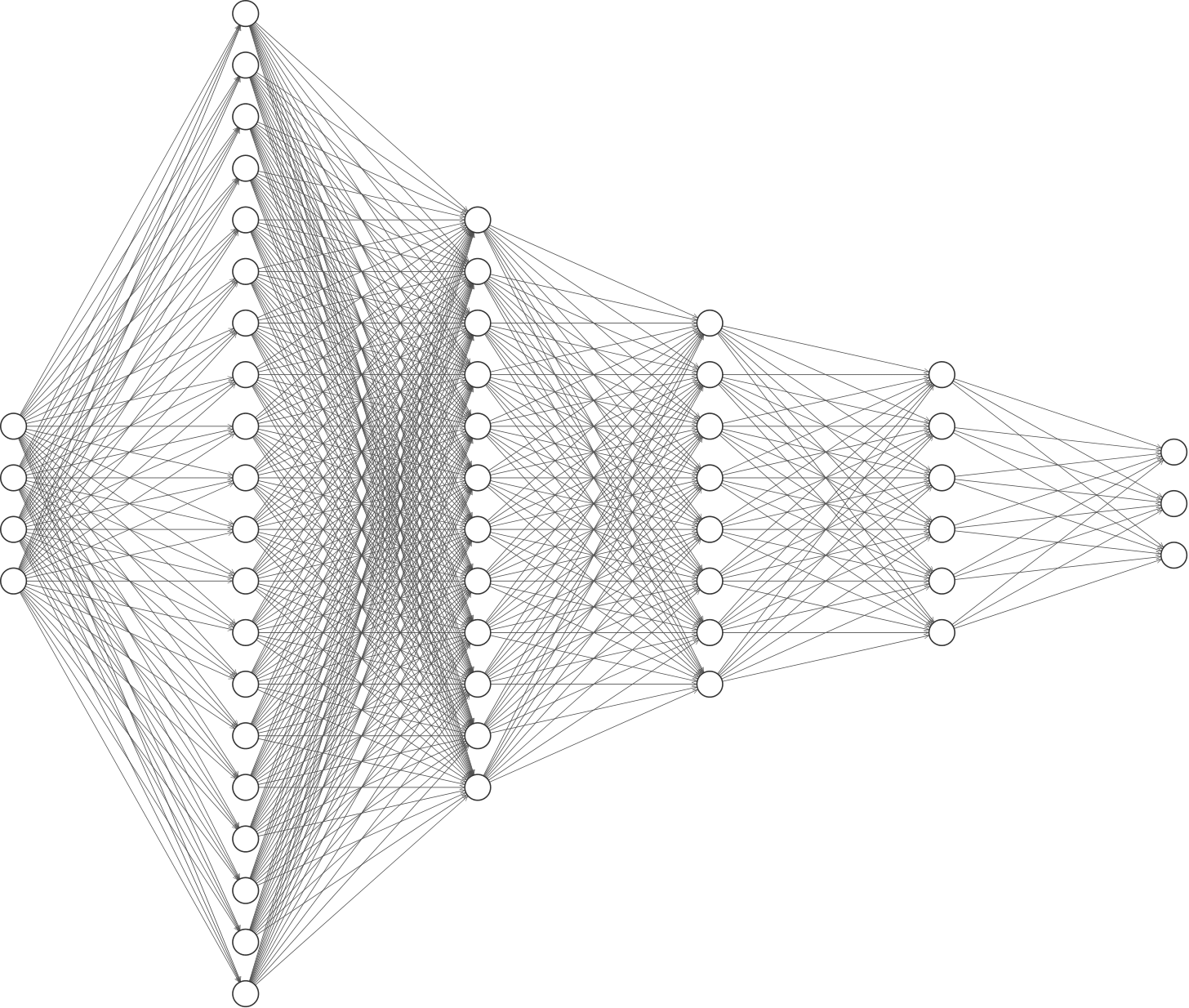}
\end{minipage}%
}%
\subfigure[NN2]{
\begin{minipage}[t]{0.4\linewidth}
\centering
\includegraphics[width=1.5in]{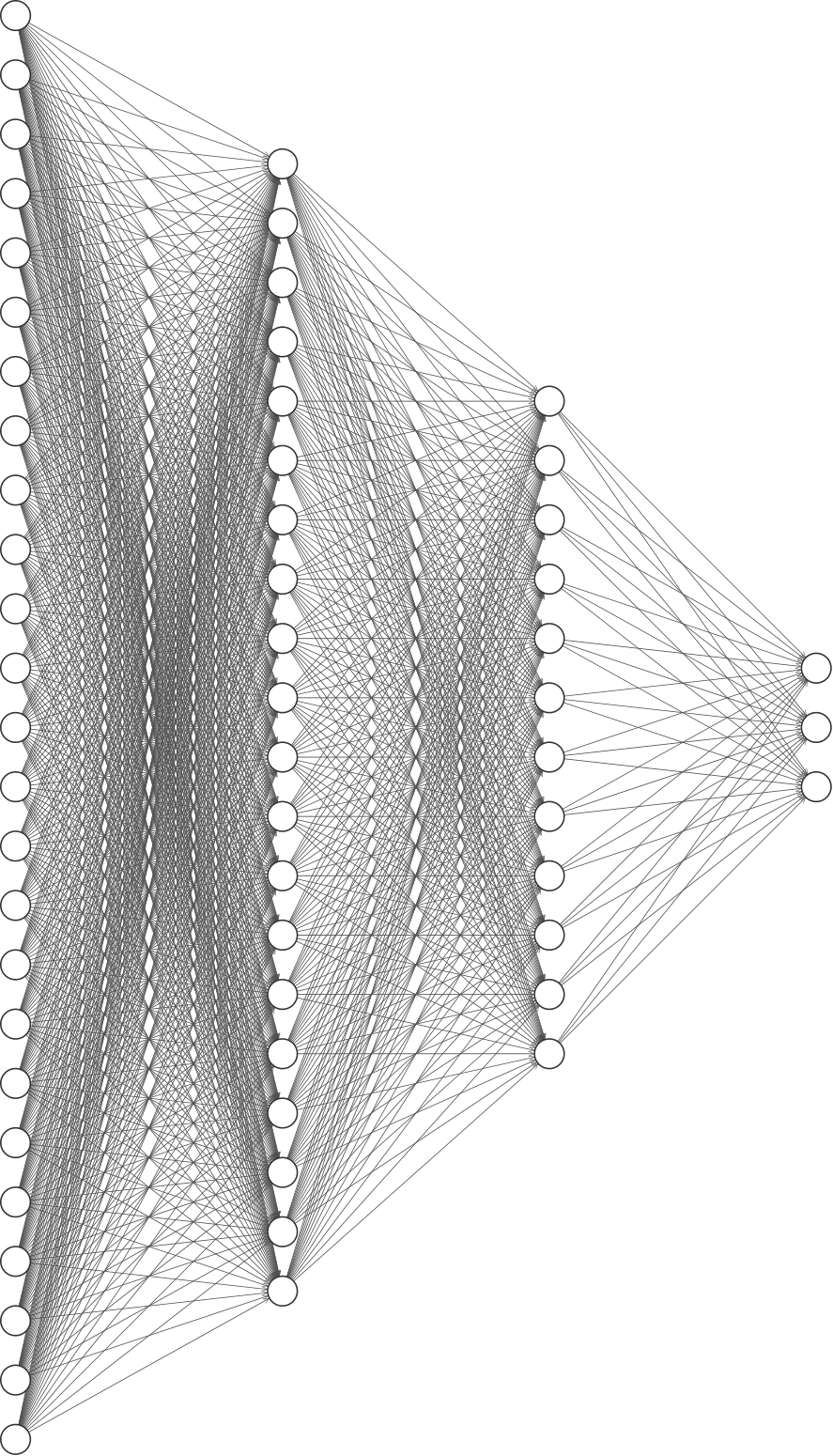}
\end{minipage}%
}%
\caption{~~The two  NNs considered which imitate the dense
  layer in well-known NNs such as MiniAlexNet, VGG and LeNet.\label{fig:NN12}}
\end{figure}

We consider two different neural networks, a narrower but deeper NN1, and a wider but shallower NN2.  The number of layers and their dimensions are shown in Figure~\ref{fig:NN12}: \\
\mbox{{} \hskip3cm}  NN1: $100\to1024\to512\to384\to192\to K,$\\
\mbox{{} \hskip3cm}  NN2: $2048\to1024\to512\to K.$\\[1mm]
The activation function is $\text{ReLU}(x)=\max(x,0)$. We do not apply any activation function on the last layer. 

\subsubsection{Optimization Methodology}\label{sec:optimization}
Following common practice, we minimize the cross-entropy loss using stochastic gradient descend with momentum $0.9$. All the datasets are trained with batch size =64 on a single GPU, for 248 epochs. Trained NNs are saved for the first 10, and then every four epochs. The total number of saved NNs is $(136+60+80+60)\times3\times70=70560$. The initialization is Pytorch's default initialization, which follows a uniform distribution. The learning rate is $0.01$. 


\subsection{Results on synthetic data experiments}
To investigate the influence of the data SNR on the whole training 
process, we first report synthetic data experiments results.

\subsubsection{Three types of spectrum bulk}
\label{sec:threetypes}
We use SNR to measure the data quality and focus on the non-zero eigenvalues of the matrix $WW^T$. The SNR could directly reflect the classification difficulty. The  weight matrices $W$ we consider in this section are those  at the
final epoch (248th). 
In the Gaussian data sets, with different values of
SNR, we have observed the following three typical types for the bulk spectrum of
the weight matrices:
$$\begin{aligned}
\text{\textbf{HT}}&:\text{ Heavy Tail}\\
\text{\textbf{BT}}&: \text{ Bulk Transition}\\
\text{\textbf{LT}}&:\text{ Light Tail (MP Law)}
\end{aligned}
$$

\begin{figure}[htbp]
\centering
\subfigure[HT(0,0)]{
\begin{minipage}[t]{0.27\linewidth}
\centering
\includegraphics[width=1.5in]{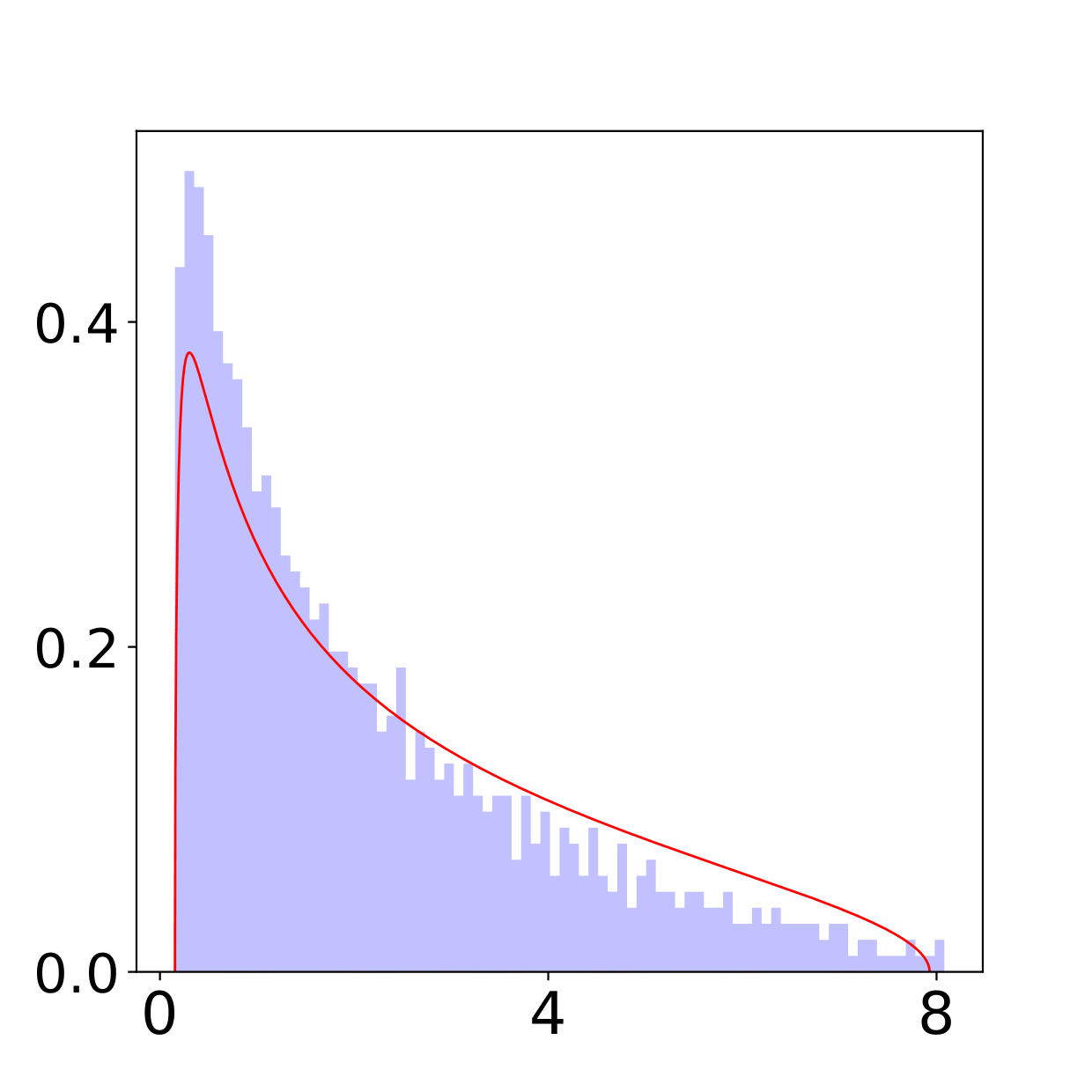}
\label{HT00}
\end{minipage}%
}%
\subfigure[HT(0,1)]{
\begin{minipage}[t]{0.27\linewidth}
\centering
\includegraphics[width=1.5in]{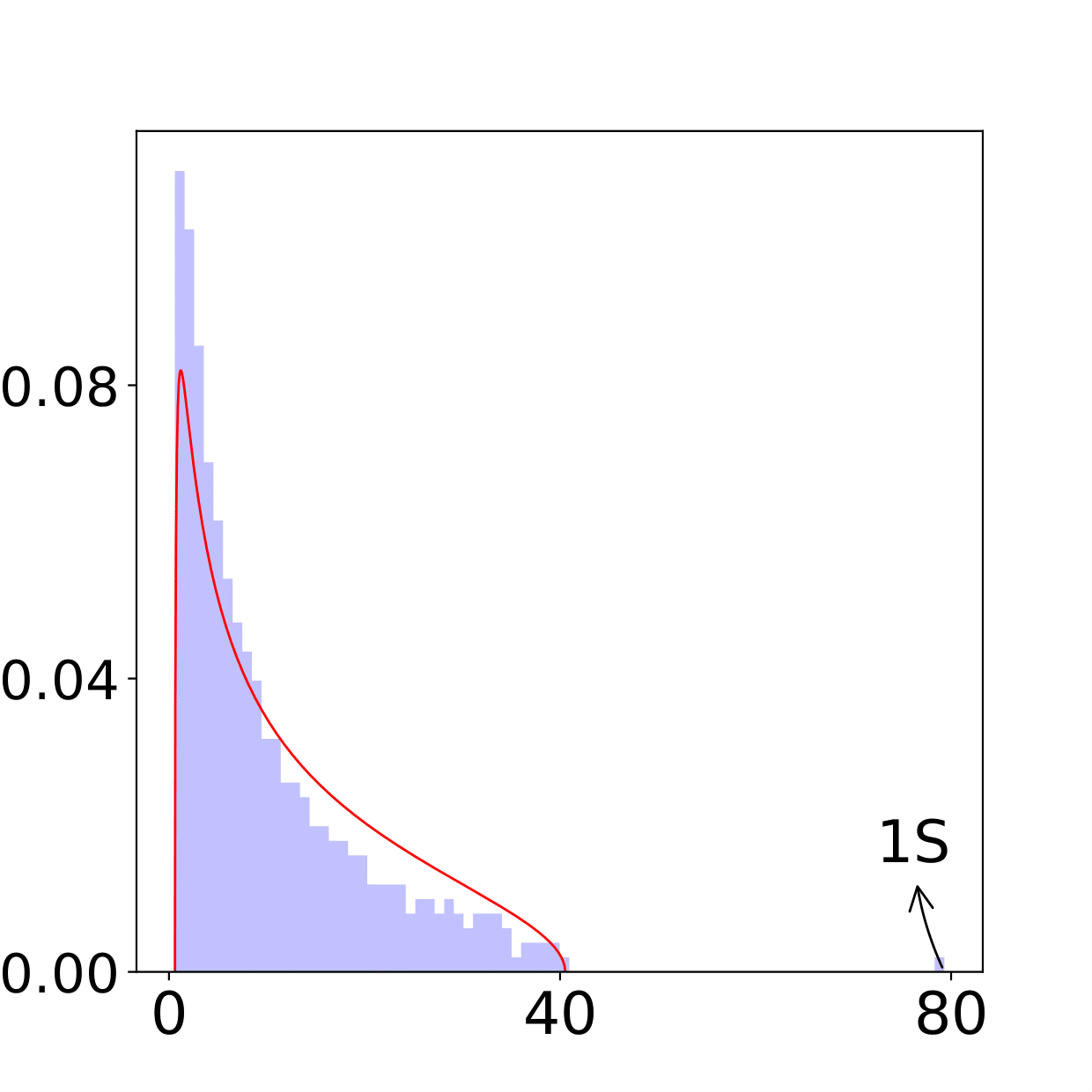}
\label{HT01}
\end{minipage}%
}%
\subfigure[HT(7,1)]{
\begin{minipage}[t]{0.27\linewidth}
\centering
\includegraphics[width=1.5in]{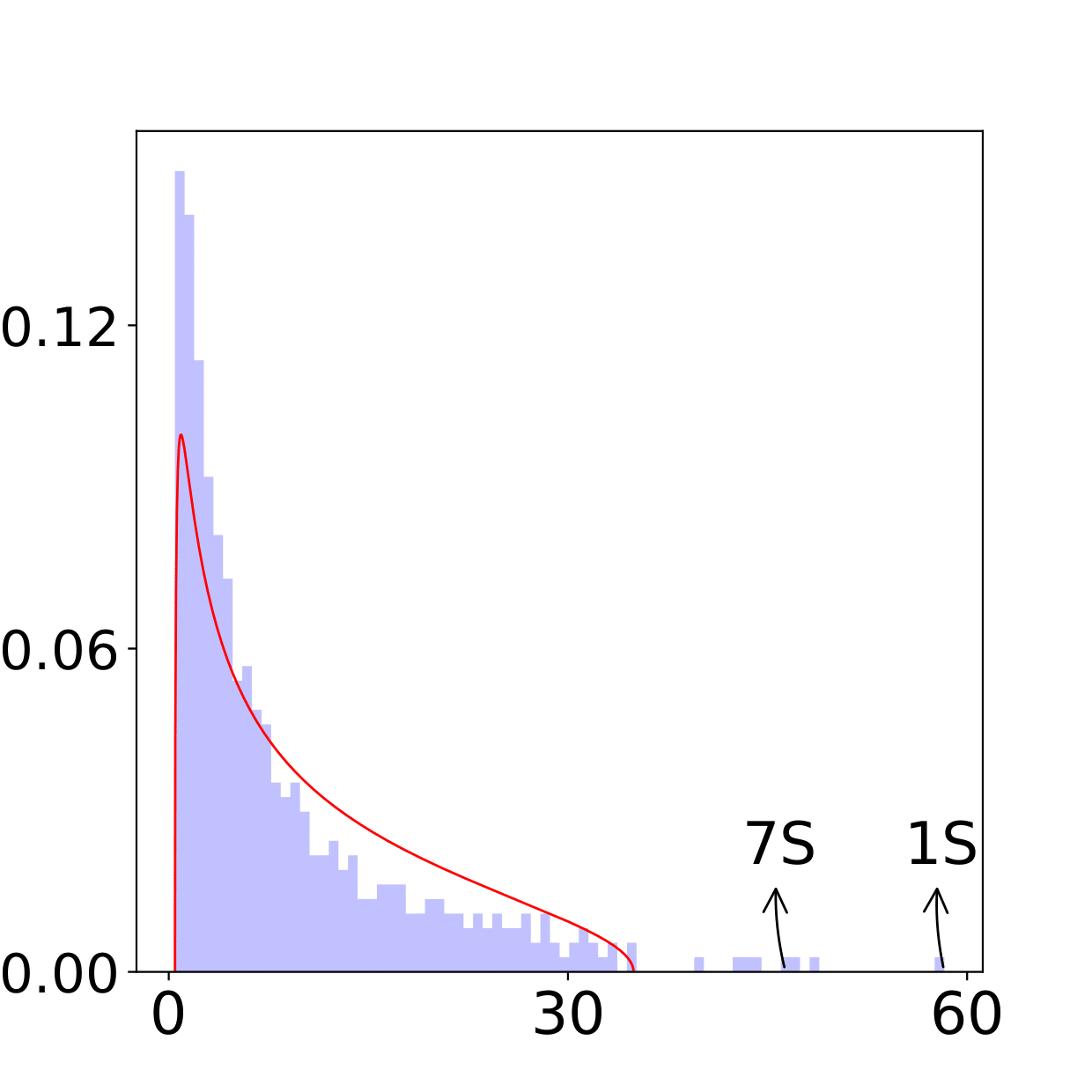}
\label{HT71}
\end{minipage}
}%
\subfigure[HT(0,8)]{
\begin{minipage}[t]{0.27\linewidth}
\centering
\includegraphics[width=1.5in]{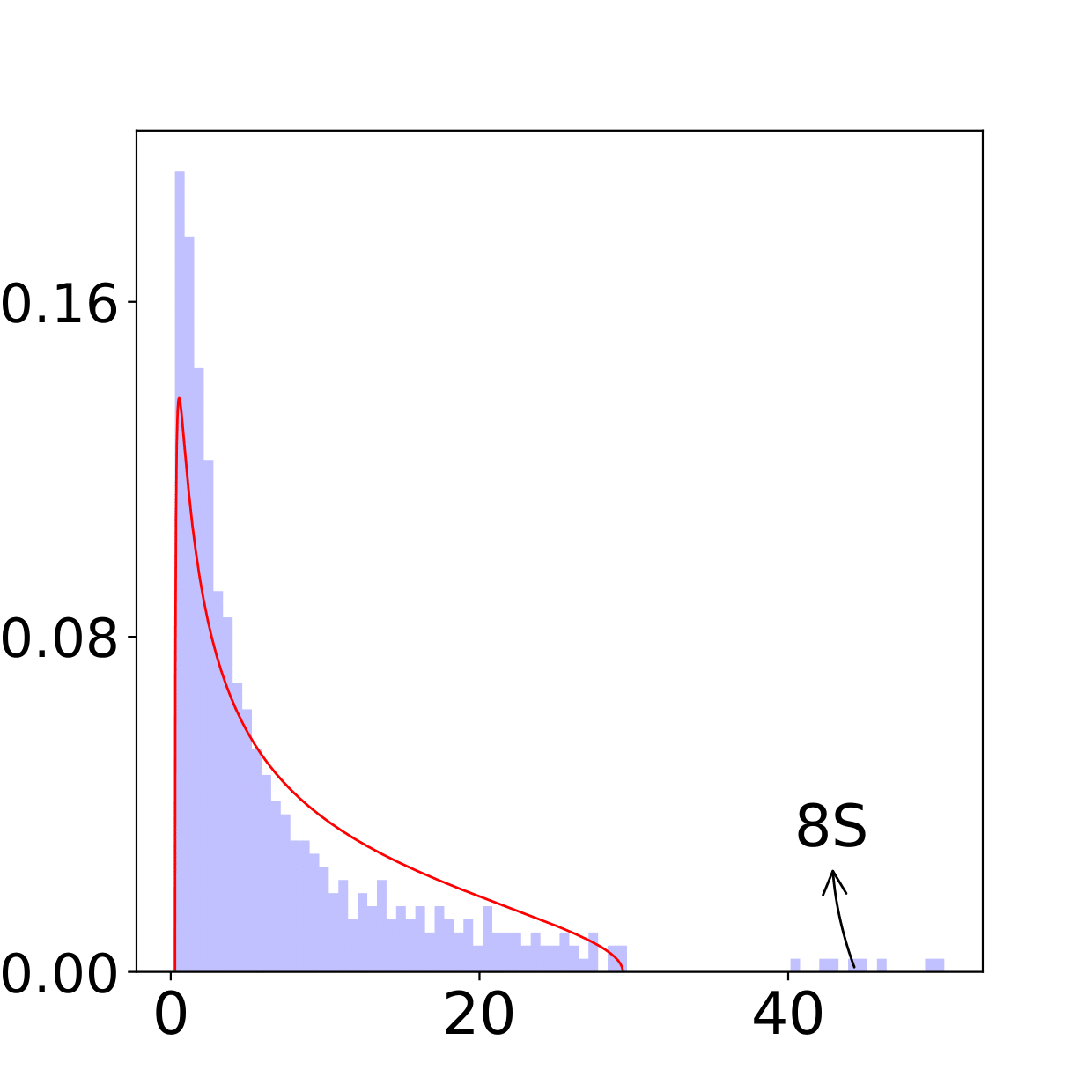}
\label{HT08}
\end{minipage}
}%
\centering
\caption{~~ Examples of observed HT type spectrum bulks.}
\label{Heavy Tail}
\end{figure}

\begin{figure}[htbp]
\centering
\subfigure[BT(1,7)]{
\begin{minipage}[t]{0.27\linewidth}
\centering
\includegraphics[width=1.5in]{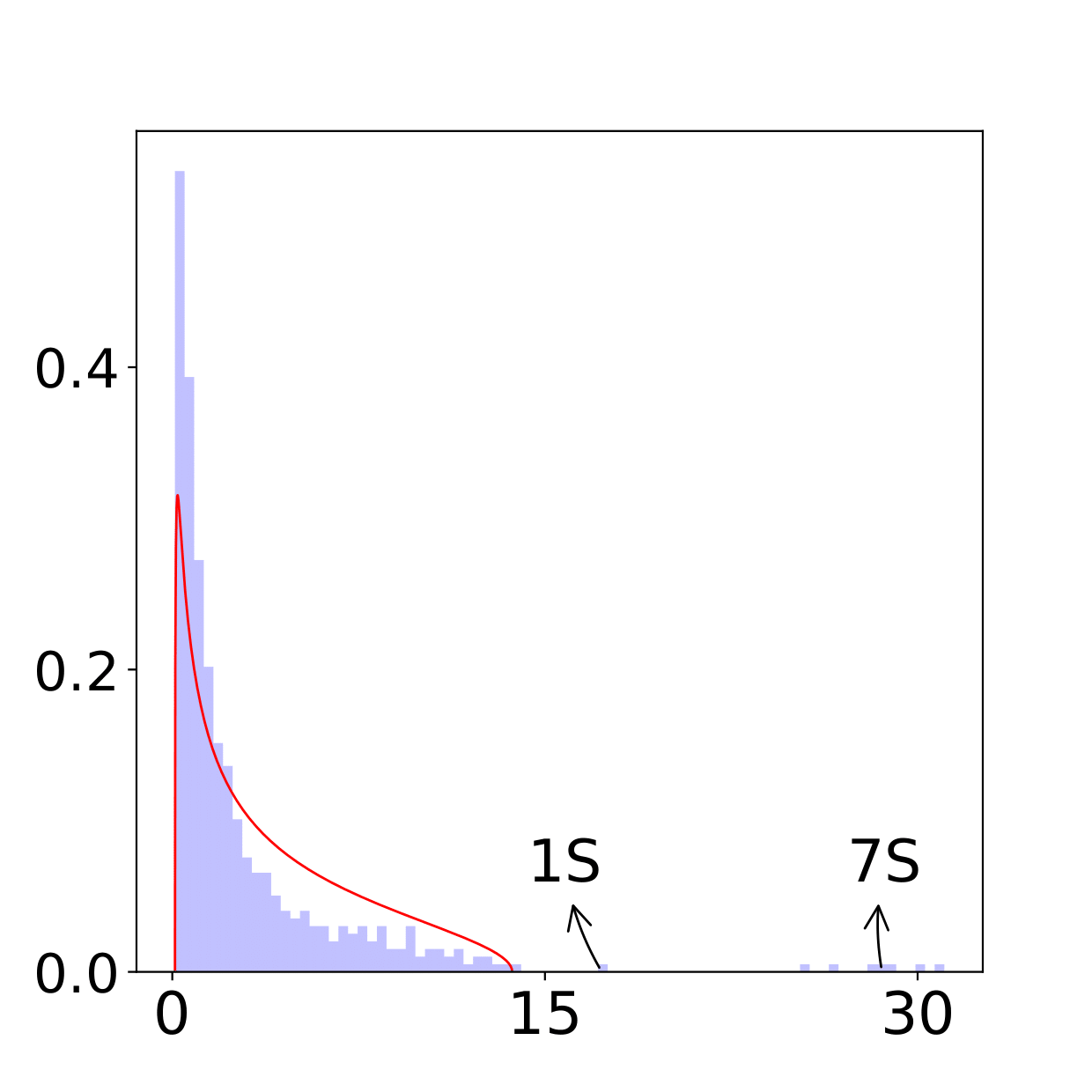}
\label{HT17}
\end{minipage}%
}%
\subfigure[BT(0,7)]{
\begin{minipage}[t]{0.27\linewidth}
\centering
\includegraphics[width=1.5in]{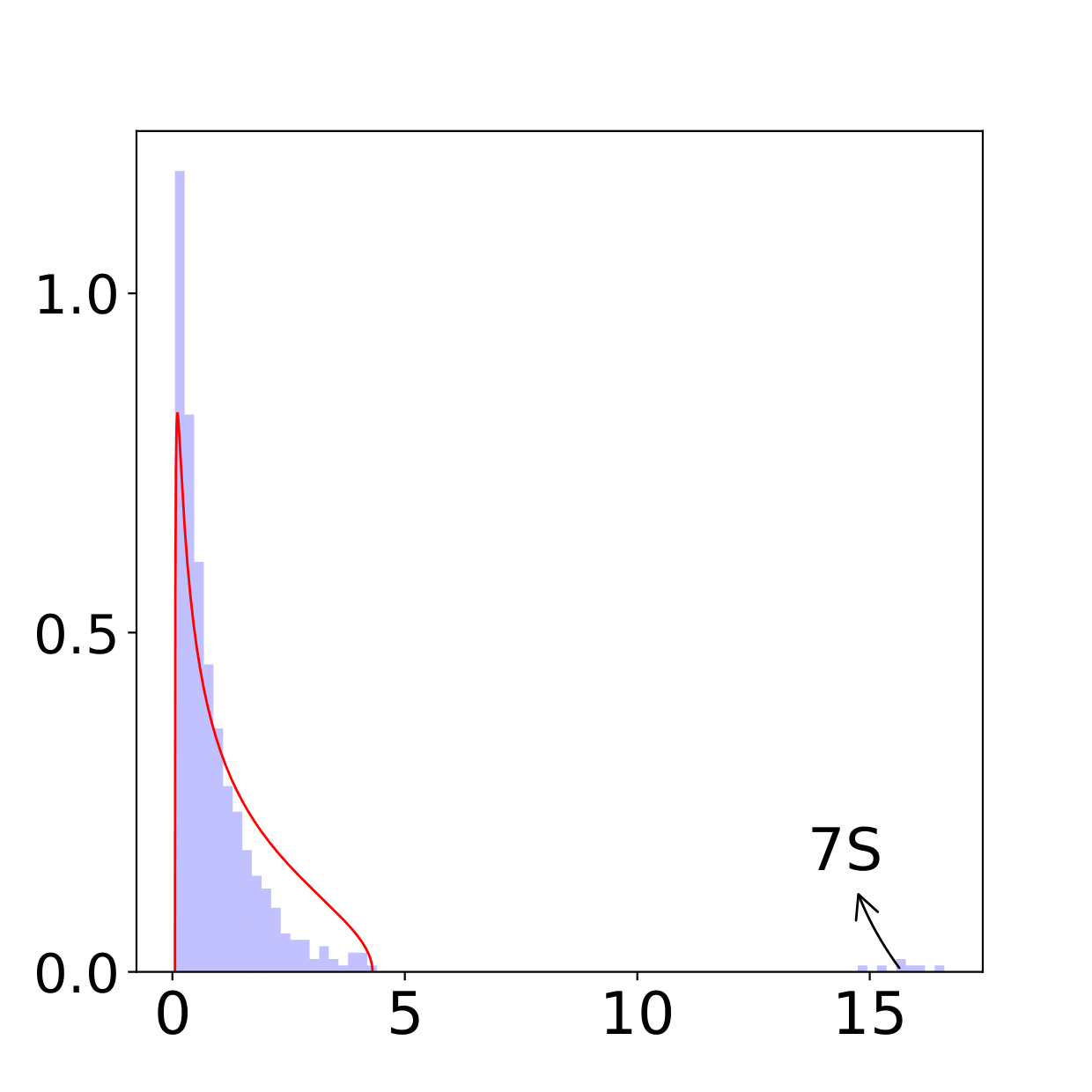}
\label{HT07}
\end{minipage}%
}%
\subfigure[BT(0,7)]{
\begin{minipage}[t]{0.27\linewidth}
\centering
\includegraphics[width=1.5in]{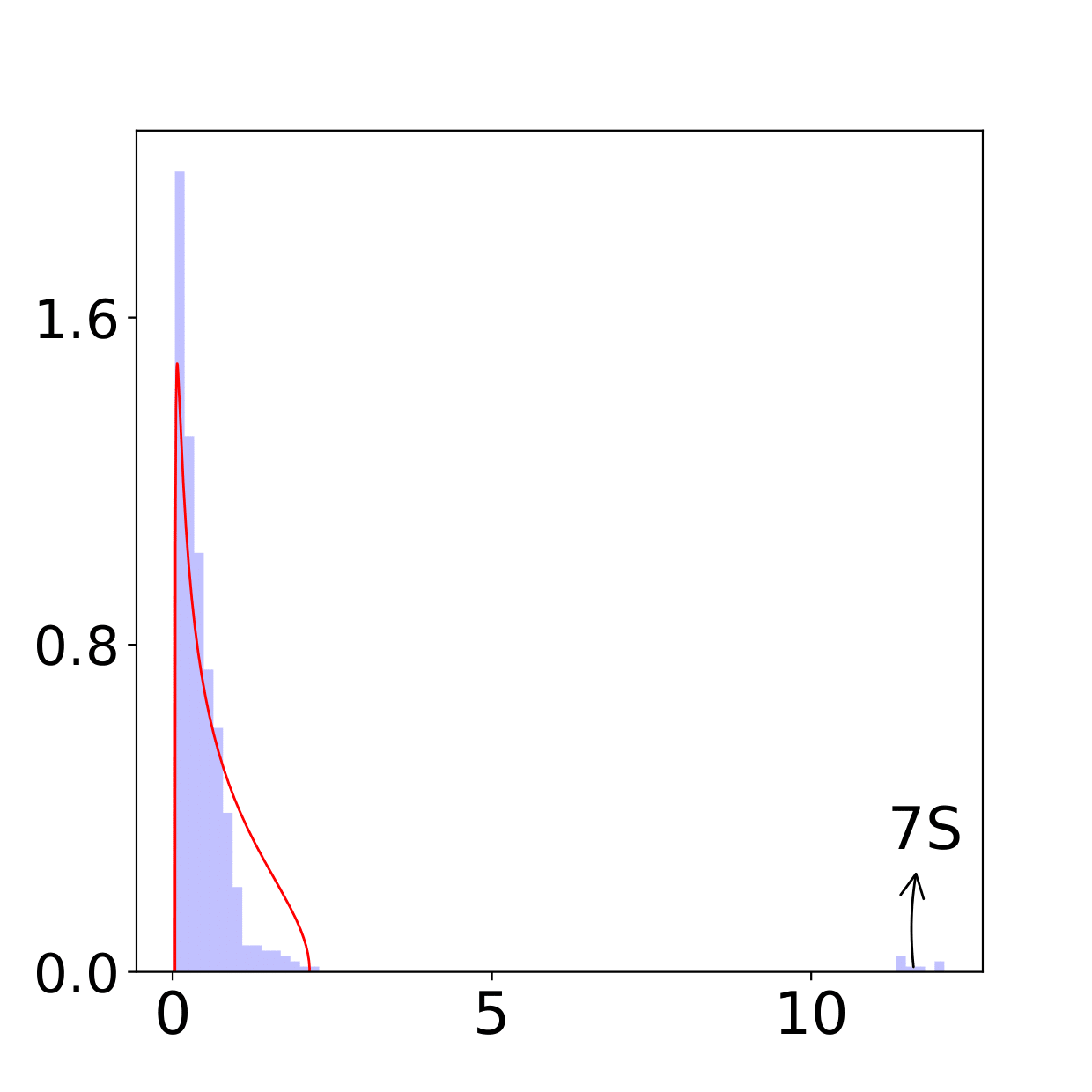}
\label{MPB07}
\end{minipage}
}%
\subfigure[BT(1,7)]{
\begin{minipage}[t]{0.27\linewidth}
\centering
\includegraphics[width=1.5in]{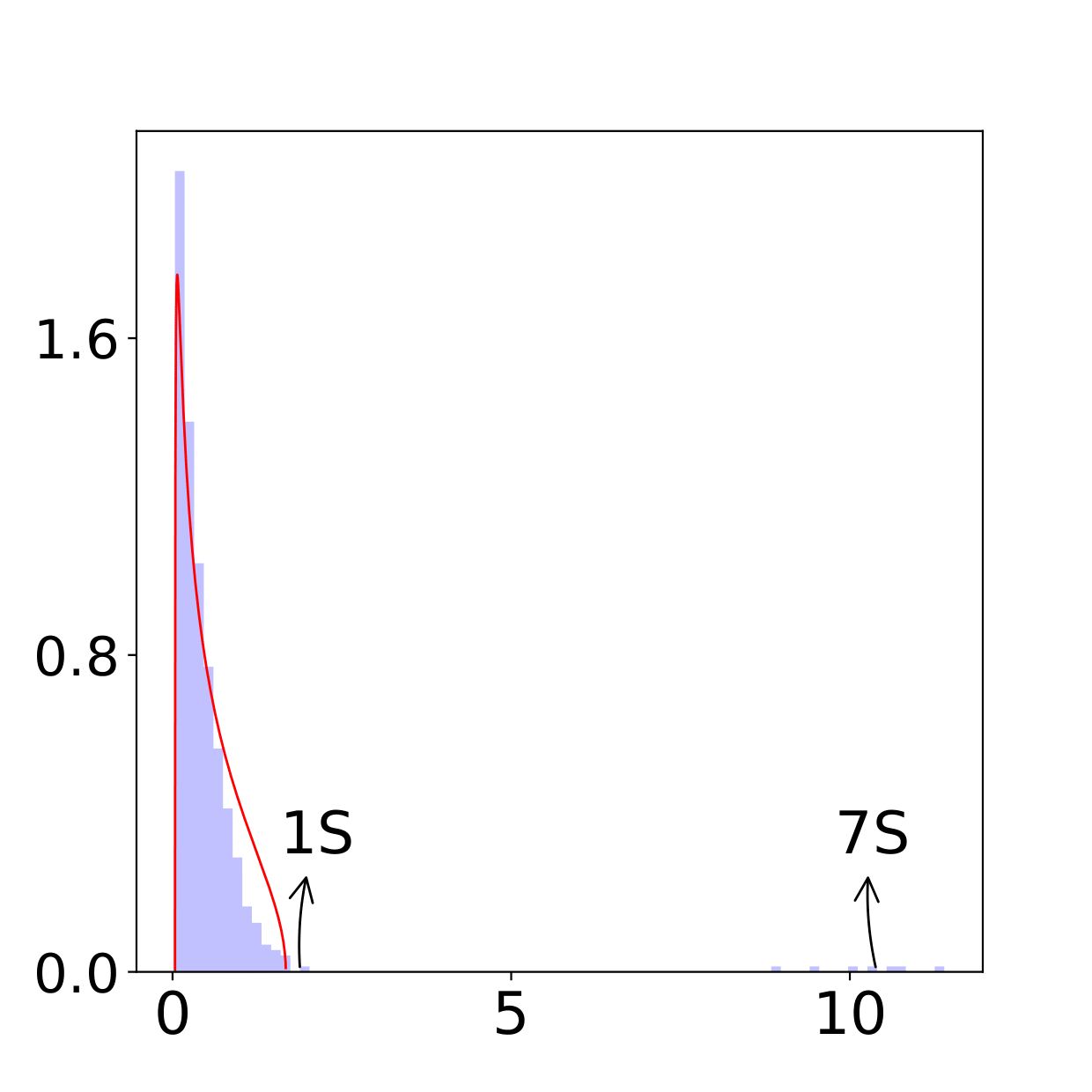}
\label{MPB17}
\end{minipage}
}%
\centering
\caption{~~ Examples of observed BT type spectrum bulks.}
\label{MPB}
\end{figure}

\begin{figure}[htbp]
\centering
\subfigure[LT(1,7)]{
\begin{minipage}[t]{0.27\linewidth}
\centering
\includegraphics[width=1.5in]{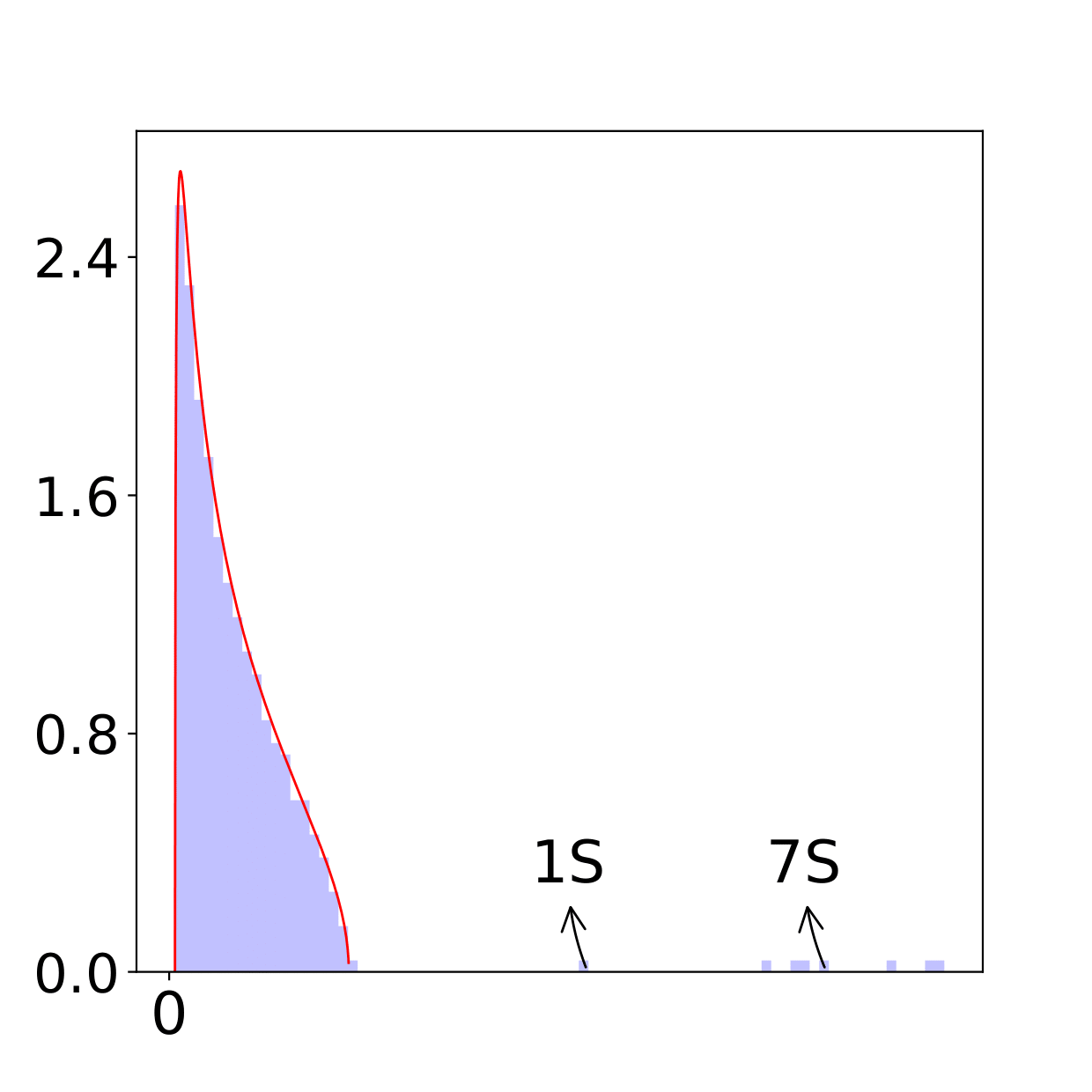}
\label{MP17}
\end{minipage}%
}%
\subfigure[LT(0,8)]{
\begin{minipage}[t]{0.27\linewidth}
\centering
\includegraphics[width=1.5in]{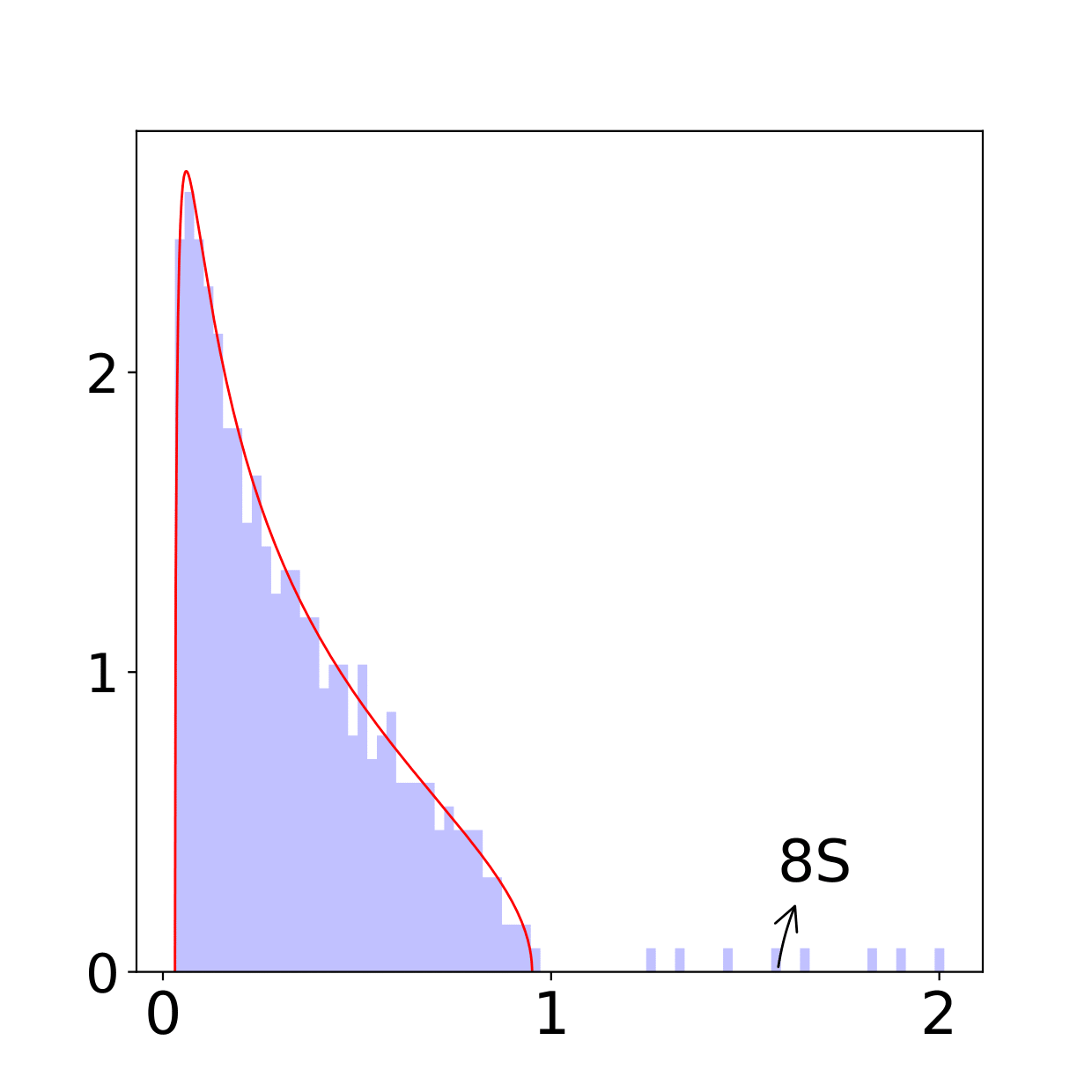}
\label{MP08}
\end{minipage}%
}%
\subfigure[LT(0,7)]{
\begin{minipage}[t]{0.27\linewidth}
\centering
\includegraphics[width=1.5in]{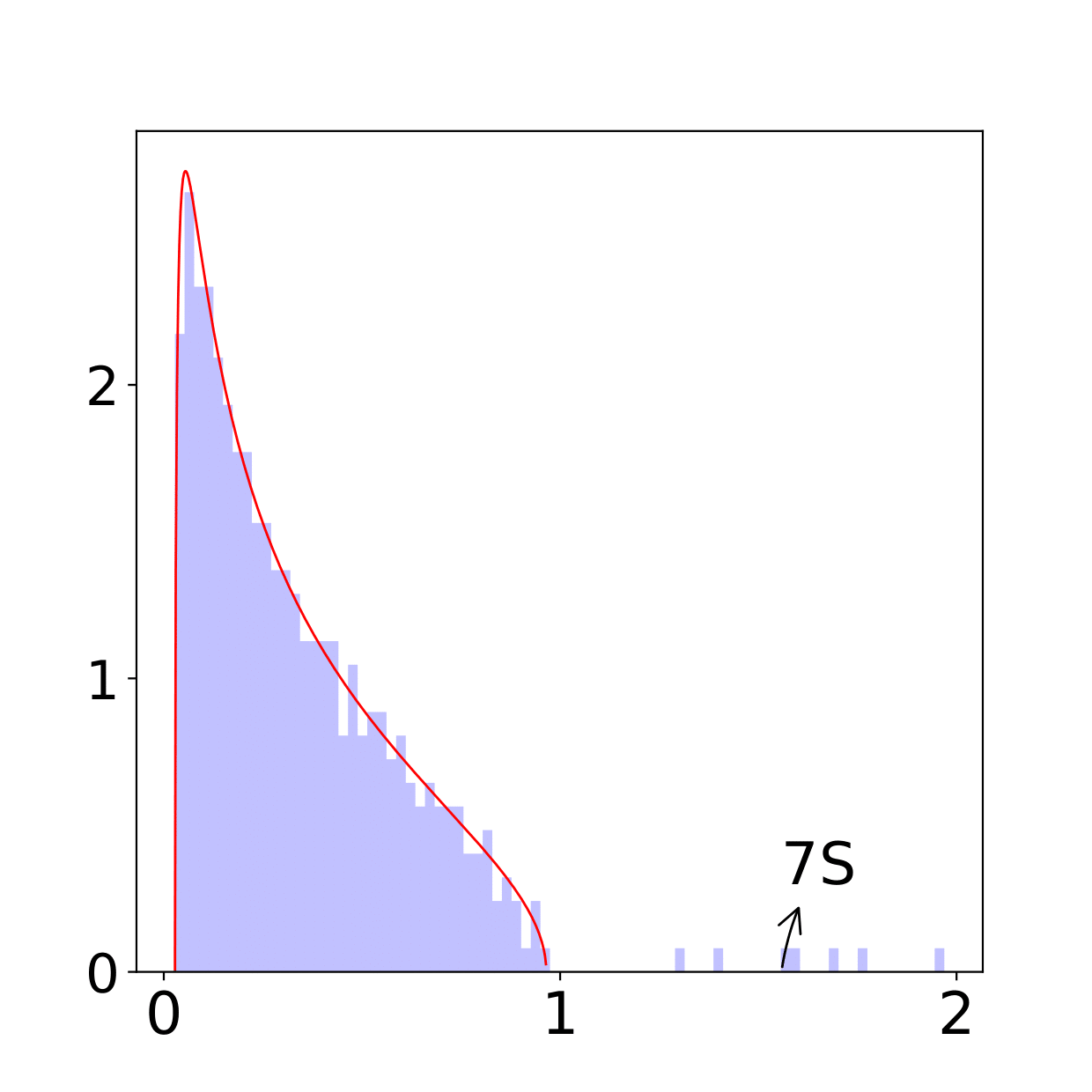}
\label{MP07}
\end{minipage}
}%
\subfigure[LT(0,0)]{
\begin{minipage}[t]{0.27\linewidth}
\centering
\includegraphics[width=1.5in]{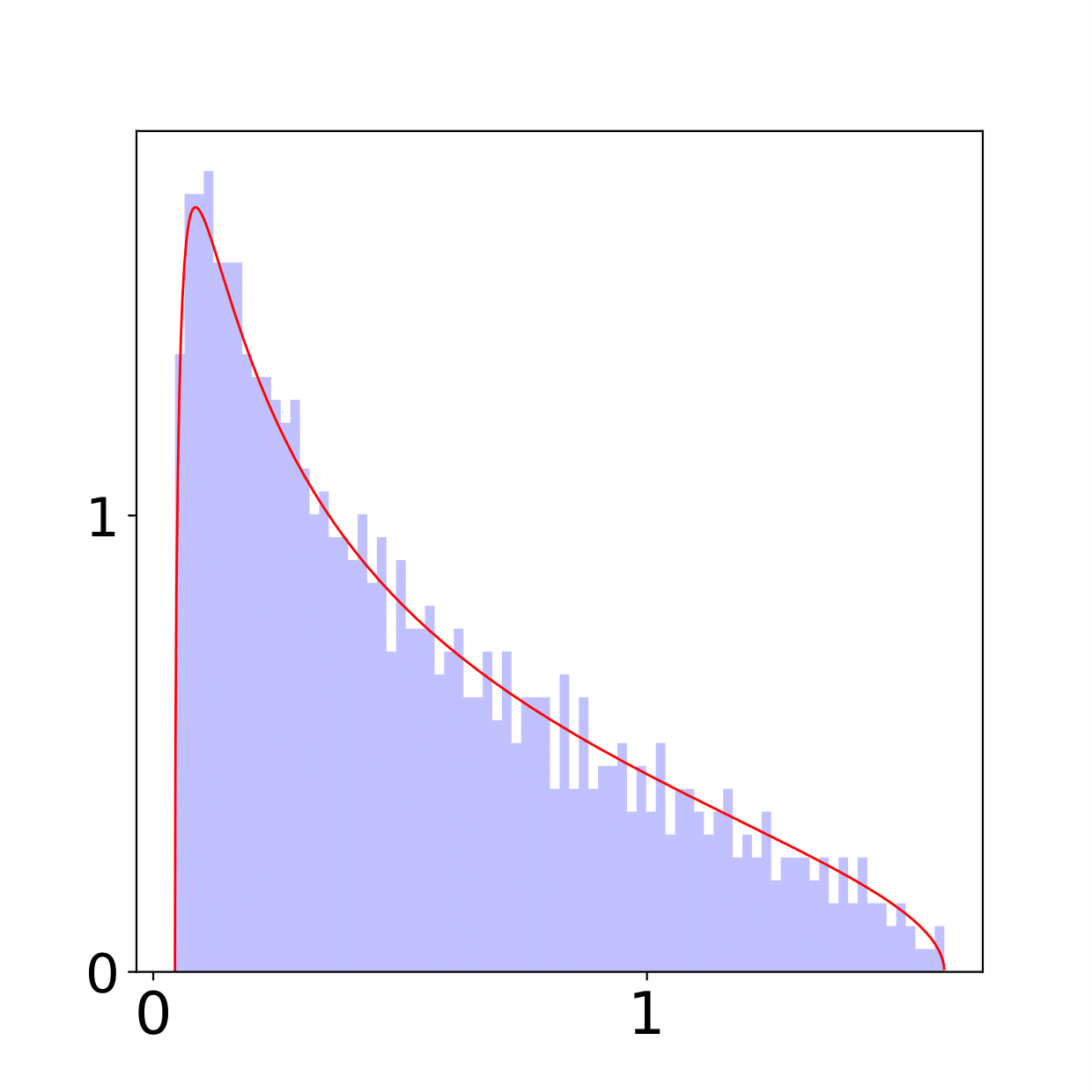}
\label{MP00}
\end{minipage}
}%
\centering
\caption{ ~~ Examples of observed LT type spectrum bulks.}
\label{Light Tail}
\end{figure}

We increase gradually the SNR of the Gaussian model and report in the figures \ref{Heavy Tail}-\ref{Light Tail} the  spectra of weight matrices at the end of training (in each figure, the SNR is increasing from (a) to (d)). 
As we can see in Figure \ref{Heavy Tail} with relatively low SNR,  spectra from weight matrices (in blue) shows significant departure of the spectrum from a typical MP spectrum (in red).  This defines the class HT.
In  contrary, spectra in Figure \ref{Light Tail} with relatively high SNR,  well match an MP spectrum, and this corresponds to the LT class.  More  complex structures appear in figure \ref{MPB}, corresponding to medium values of the SNR.  A transition is in place from Figure \ref{HT17}, which is closer to the HT spectra, to Figure \ref{MPB17}, which is now closer to a MP spectrum.  We call this a bulk transition class BT.   

We now describe the evolution of spikes during the full transition from type HT to BT and LT.  
It is reported in \cite{papyan2020traces} that the total $K= 8$ spikes are grouped in two clusters with $K-1=7$ spikes (determined by the between-class covariance matrix)  and the singleton one (determined by the general mean), respectively.  
At the begining (Figure \ref{HT00}), all the spikes are hidden in the bulk. When the SNR increases, the group of 7 spikes  emerge from the bulk and stay outside the spectrum for ever. The movement of the single spike is more complex, hiding in and leaving the bulk repeatedly. 
There is particular moment where the two groups meet and stay close each other: we then see a group of 8 spikes. 

We use ``XX(m,n)" to describe the whole ESD type of the spectrum of
weight matrices. Here  ``XX" means one of the three bulk types in
\{HT, BT, LT\}. The number 
``m" or ``n" gives us position information of the two groups of the spikes, numbered in increasing order. 
For instance, BT(1,7) displayed in Figure~\ref{MPB17},
means the bulk type is BT, the single spike lays between the bulk and the group of $K-1$ spikes; HT(0,8) means two groups of spikes are mixed; HT(0,7) means we see only the group of $K-1$ spikes.

\begin{remark}
    The spectrum transition from HT to BT and LT can also be assessed by more quantitative criteria. 
    (i) The transition from HT to BT is related to the distance between  the group of $K-1$ spikes and the bulk edge.  When this distance is large enough, the HT type ends and the BT phase starts. Note that here the bulk type is heavy-tailed in both regimes HT and BT.
    (ii) 
    The transition from BT to LT can be directly detected by comparing the bulk spectrum to the reference MP spectrum. Precisely, this can be achieved using our spectral distance statistic  $\hs_n$ defined in section \ref{sec:specCriterion}.  
\end{remark}

\newgeometry{top=2.5cm}

\vspace{0pt}

\begin{figure}[htbp]
    \centering
    \includegraphics[width=1.5in]{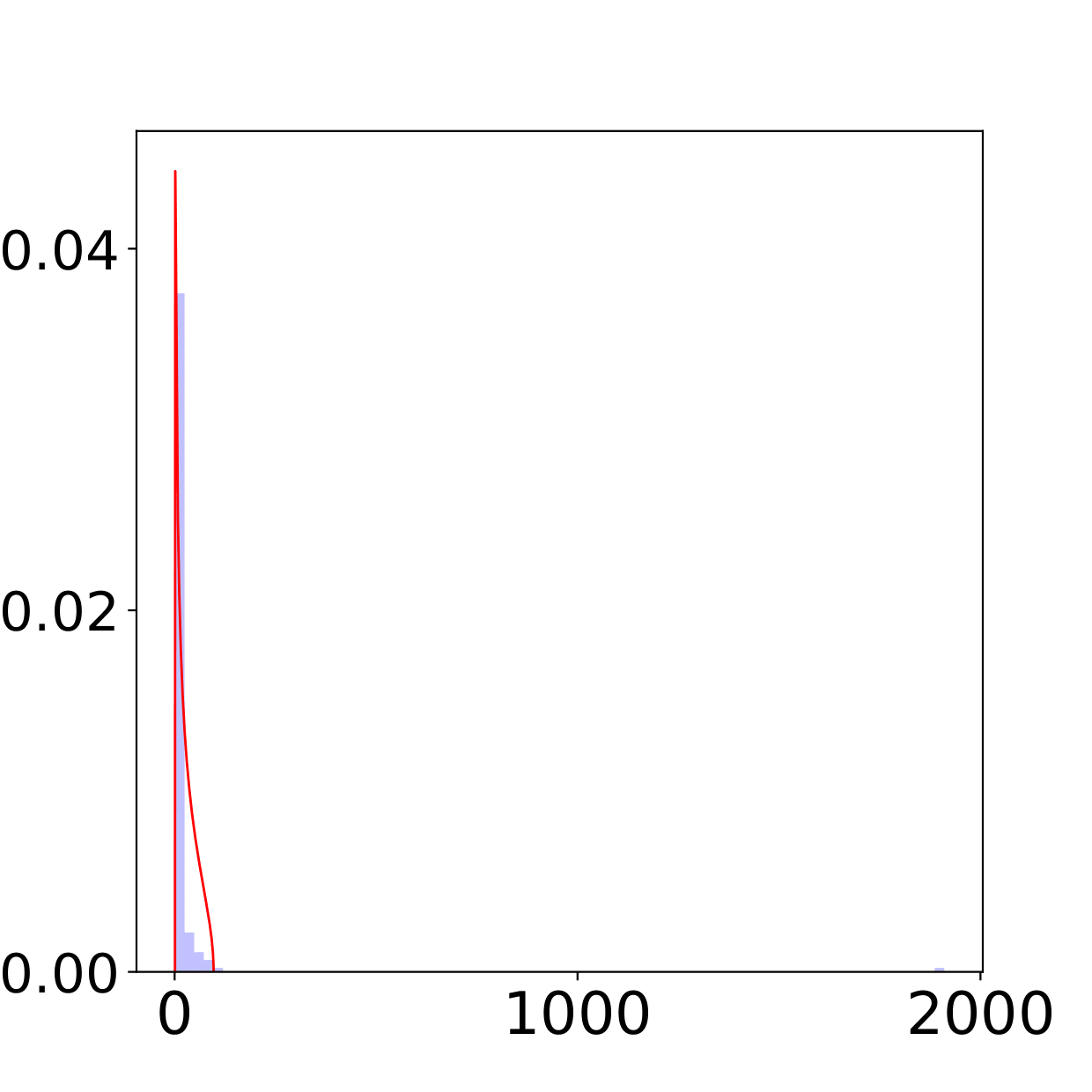}
    \caption{~~Example of spectrum with Rank Collapse.}
    \label{RC}
\end{figure}

\noindent\textbf{Rank Collapse:} One special case, Rank Collapse, occasionally emerges in our experiments especially when SNR is small. Rank Collapse is the phenomenon that the spike is huge, making the bulk in the picture 'needle like' as shown in Figure \ref{RC}. By tuning up SNR, Rank collapse gradually disappears and turns into Light Tail or Heavy Tail.

\subsubsection{Phase Transition}

\begin{figure}[htbp]
\centering

\begin{tabular}{cc}
\multicolumn{2}{c}{\includegraphics[width=3in]{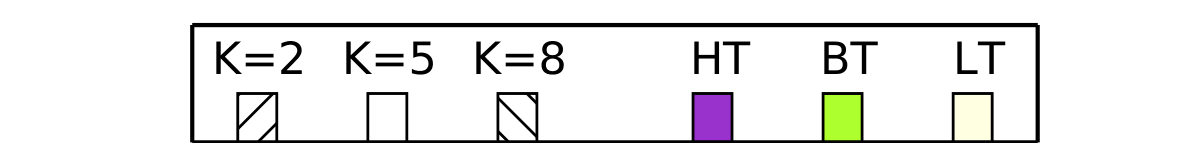}}\\
\includegraphics[width=2.4in]{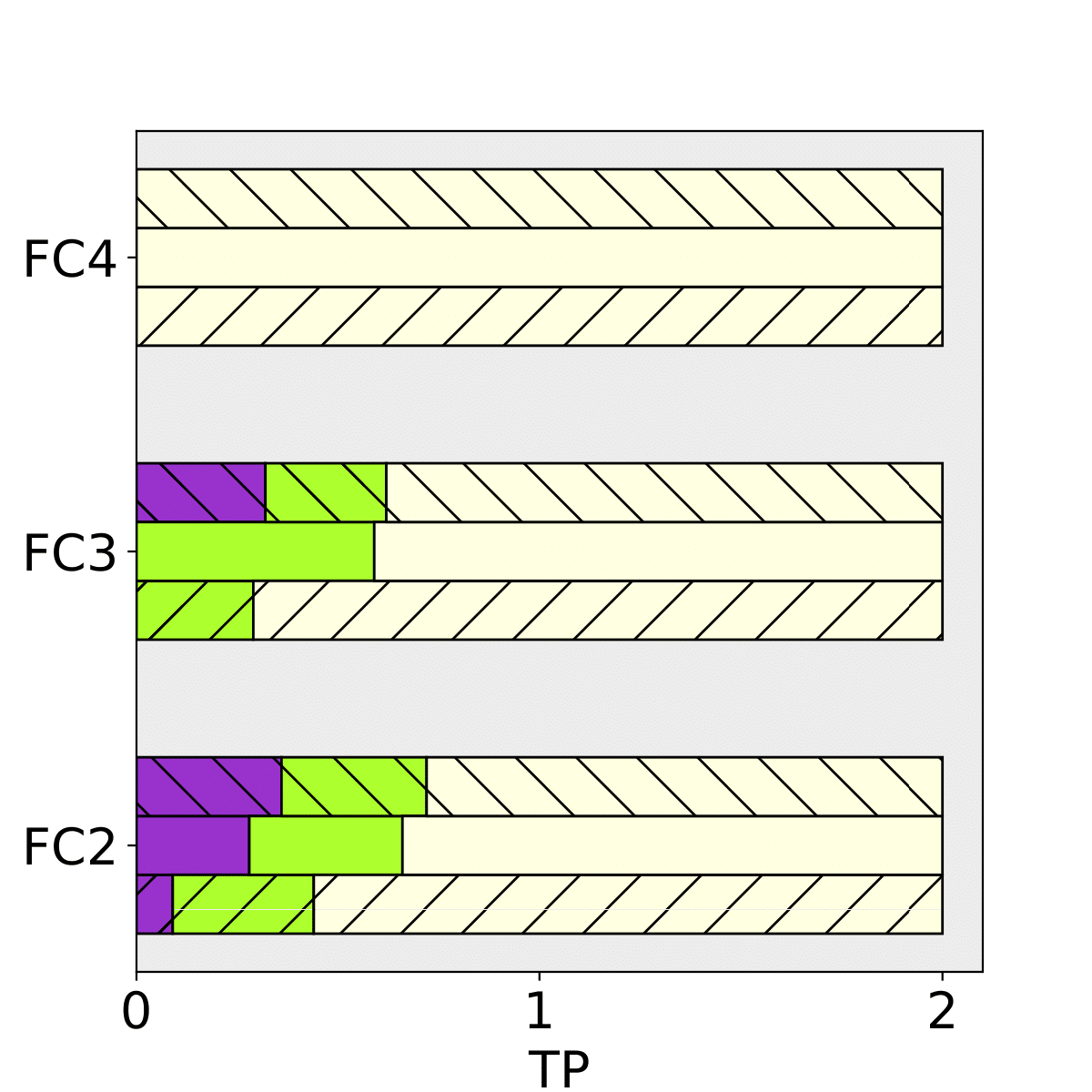}
&
\includegraphics[width=2.4in]{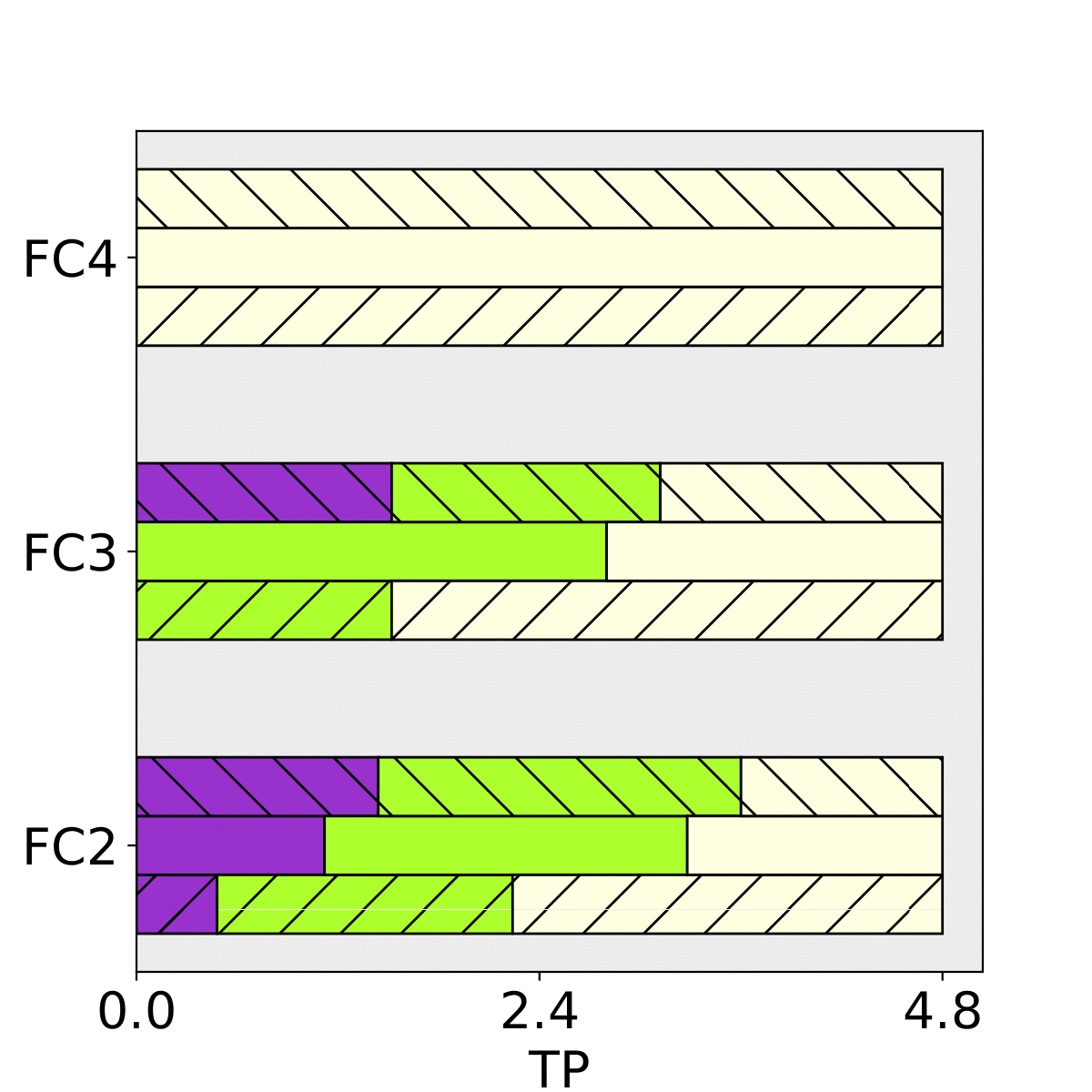}\\
(a) NN1+$\calD_1$  &    (b) NN1+$\calD_2$ \\
\includegraphics[width=2.4in]{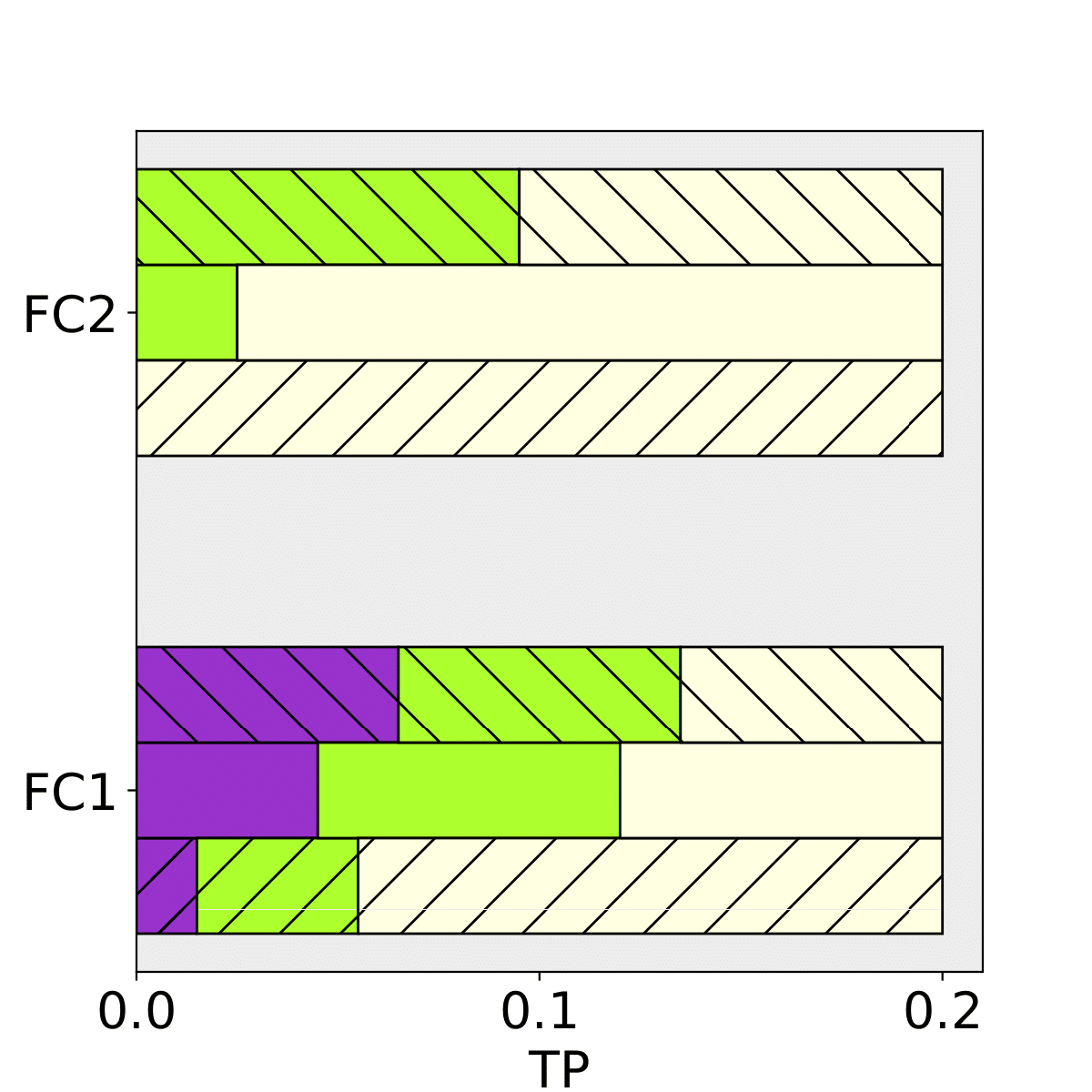}
& 
\includegraphics[width=2.4in]{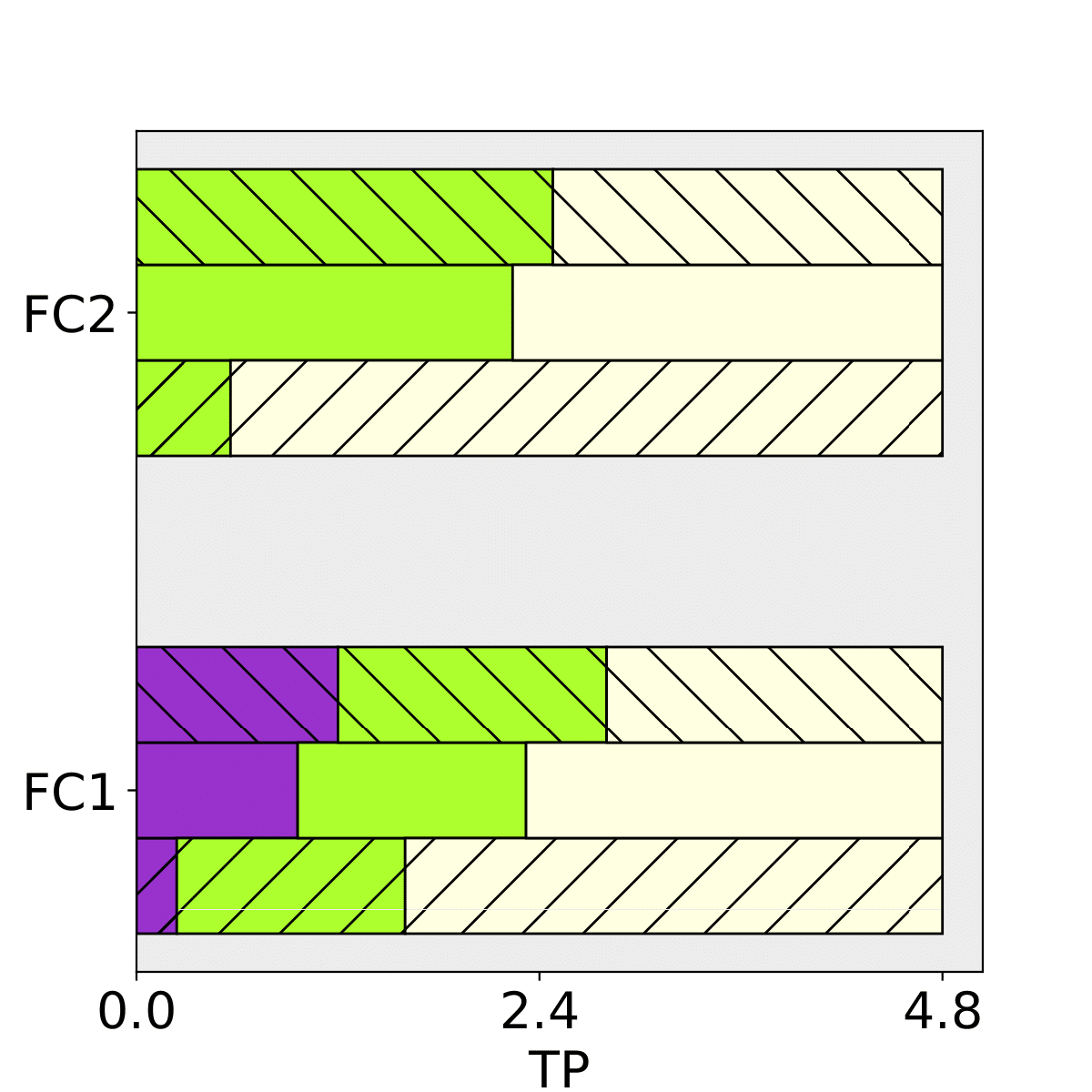}\\
(c) NN2+$\calD_1$  &  (d)  NN2+$\calD_2$ \\
\end{tabular}

\caption{~~ Transition Period with the four NNs/datasets combinations. The x-axis is the tuning parameter (TP) to tune the SNR level with the range given in Table \ref{tb: TPrange}. Each block of three lines in a given layer corresponds to the cases $K=8$ (Topline), $K=5$ (Middle line) and $K=2$ (Bottom line). Different colors represent different spectra types.}
\label{Transition}
\end{figure}

We now describe evidence that the spectrum bulks of weight matrices undergo a phase transition controlled by the data SNR. The phase transition operates in the direction of 
$$
\text{HT}\to\text{BT}\to\text{LT}
$$
when the SNR increases. The complete experimental results, with recorded phase transition periods (in terms of intervals of SNR values) in all NN layers, are given in Table \ref{tb:NN1}-\ref{tb:NN2}, for the four NNs/datasets combinations respectively. These tables are summarized in Figure \ref{Transition} as a graphical summary. 

The main findings from these results are as follows:
\vskip 2mm
\noindent (1) When data SNR increases, all spectrum bulks in weight matrices fall into the LT type at final epoch. The four subfigures for the NNs/datasets combinations all contain the \textbf{Same Transition Period Direction}:
$$
\text{Heavy Tail}\to\text{Bulk Transition}\to\text{Light Tail}.
$$
In our experiments, the bulk transition (BT) period starts when the block $K-1$ eigenvalues become distant, and the Heavy Tailed spectrum bulk gradually changes into BT and LT with an increment of SNR. It is also noted that some weight matrices start from BT type and LT type. The transition from HT to LT all starts from low SNR to high SNR, in line with less and less difficulty to classify. All the transitions above provide us evidence on data-effective regularization in the training, and strong impact of classification difficulty on the weight matrices spectra.

\vskip 2mm

\noindent (2) For a given layer in the neural network at the same SNR level, HT has higher possibility to emerge as the number of classes $K$ increases. When the SNR level is the same, the more classes number the data set has, the higher difficulty to classify them all clearly. The emergence of HT indicates some implicit regularization that attempt to improve generalization error. The phenomenon that HT emerges with an increased number of classes $K$ also gives direct evidence on data-effective regularization and strong impact of classification difficulty on weight matrices spectra. 

\vskip 2mm

\newgeometry{top=2.5cm}

\noindent (3) One interesting phenomenon is that when one travels from the initial layer to deeper layers (FC2$\to$FC4 in NN1, FC1$\to$FC2 in NN2), the layers become narrower and the tails of spectrum bulks become lighter. This is true for both NNs and all SNR levels. In line with the previous work in \cite{hodgkinson2021multiplicative}, the statement that the wider layers will more likely to bring in HT is validated in our synthetic data experiments.  The total factors cause HT are complex, we emphasize data classification difficulty and model architectures here to convince that the spectra in weight matrices could be regarded as a training information encoder. Without prior information during training procedure, spectra analysis inspires us to design wide fully connected layers to encode the training information instead of narrow layers. The benefits are not for the testing error improvement but for the training procedure monitoring. Indeed, HT usually emerges if the data classification is difficult. It could be regarded as an alarm on the hidden and problematic issues in the bad cases such as the training data quality is poor, or an indication of the emergence of implicit regularization which improves generalization error, especially in the case that the features in data are complex  for extraction. The spectra in weight matrices encode training information to some extent, and bring new insights into Deep Learning. Based on these, we propose spectral criterion in section \ref{sec:specCriterion} to guide early stopping without access of testing data.
\subsubsection{Additional experiments on different batch sizes}
Batch size also has great impact on the training stability together with data features. \citep{batchsizeNitish,batchsizePriya} observe that different batch sizes may give different influence on the training dynamics. To give more comprehensive evidence of the impact of classification difficulty on weight matrices spectra, we change data SNR on different batch sizes, and conduct experiments on NN1+$\calD_1$ and NN2+$\calD_2$ with $K=8$, and two additional batch sizes $256,32$ (previous experiments all used a batch size of 64).





\begin{figure}[htbp]
\centering

\begin{tabular}{cc}
\multicolumn{2}{c}{\includegraphics[width=3in]{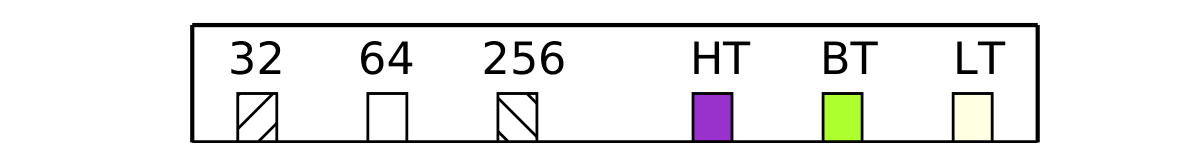}}\\
\includegraphics[width=2.4in]{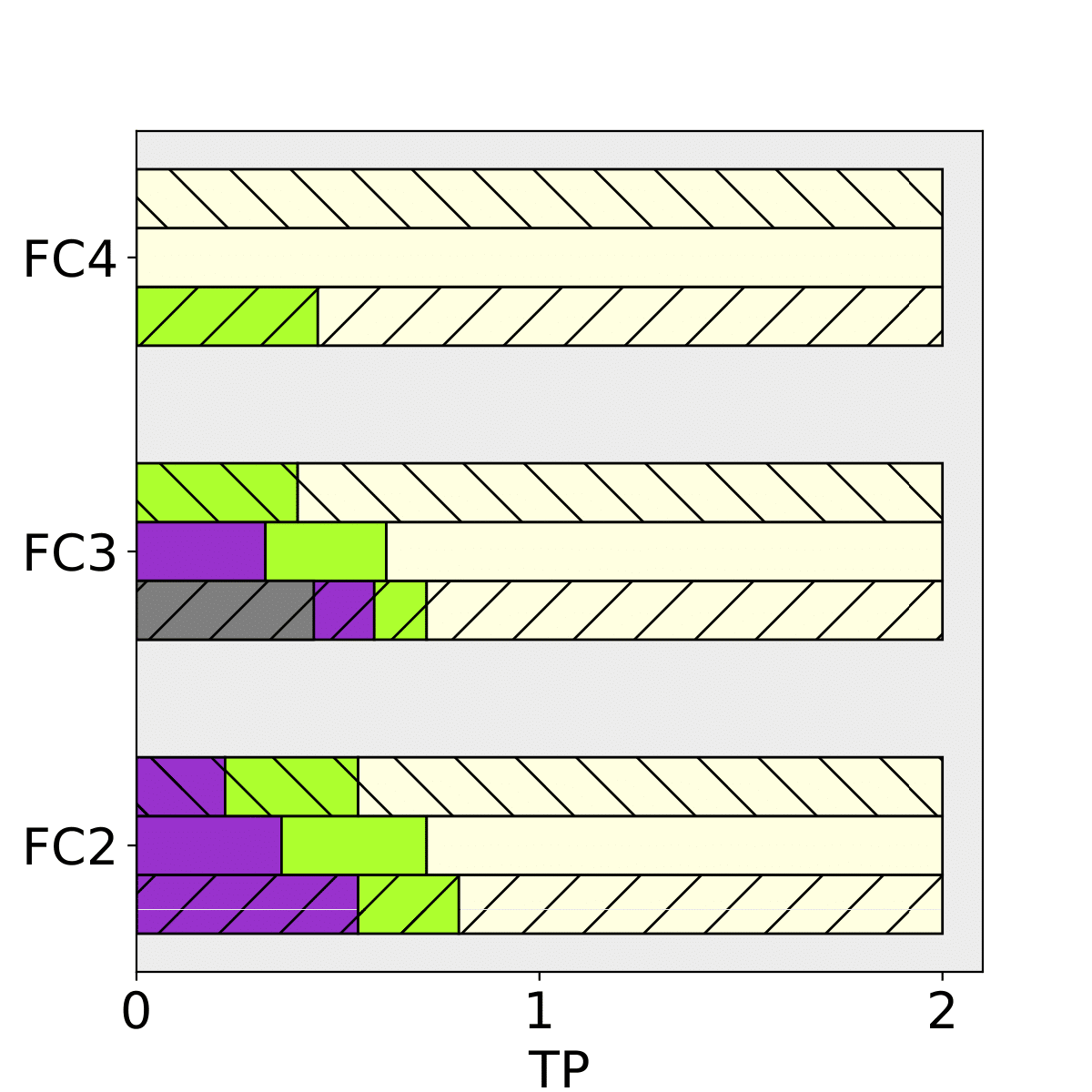}
&
\includegraphics[width=2.4in]{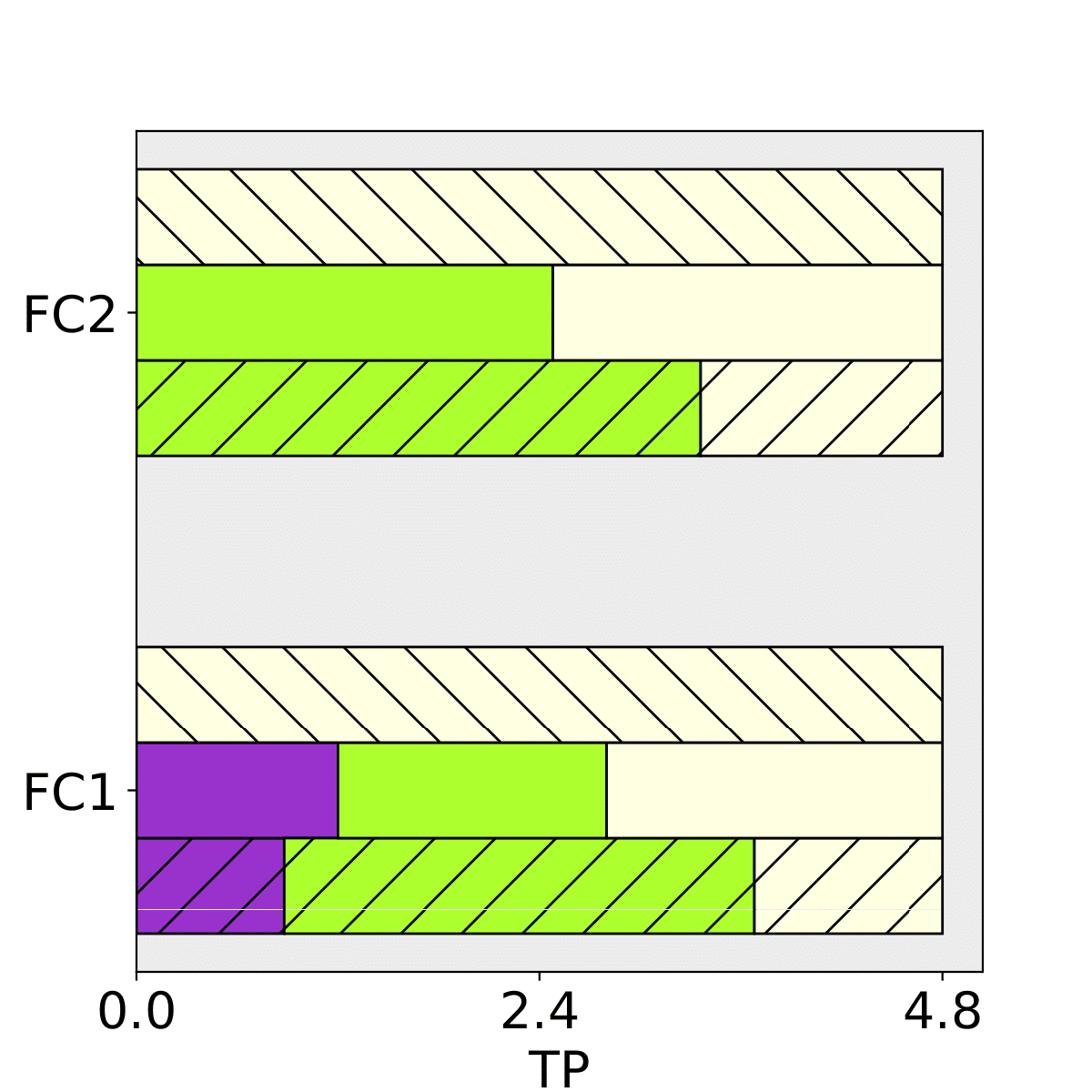}\\
(a) NN1+$\calD_1$  &    (b) NN1+$\calD_2$ \\

\end{tabular}

\caption{~~ Transition Period in different batch sizes. The gray part in NN1+$\calD_1$ at FC3 represents Rank Collapse.}
\label{TransitionBatch}
\end{figure}

We check the transition. The phase transition of the spectrum bulk is still observed in the same direction as previously (HT$\to$BT$\to$LT). Full results are given in Table \ref{tb:batch}. Again, a summarized visualization is given in Figure \ref{TransitionBatch}. The settings are all similar to those used previously.

\begin{sidewaystable}[htbp]
\caption{~~ The results of ESD types in different layers with different SNR in \textbf{NN1}. }
\label{tb:NN1}

\begin{tabular}{ccccccclcccccc}\toprule
\multicolumn{6}{c}{$K=2$ (NN1+$\calD_1$)}                                                                                                                                                    &  &  & \multicolumn{6}{c}{$K=2$ (NN1+$\calD_2$)}                                                                                                        \\
\multicolumn{6}{c}{}                                                                                                                                                                         &  &  & \multicolumn{6}{c}{}                                                                                                                             \\
\multicolumn{2}{c}{FC2}                        & \multicolumn{2}{c}{FC3}                                                          & \multicolumn{2}{c}{FC4}                                  &  &  & \multicolumn{2}{c}{FC2}                            & \multicolumn{2}{c}{FC3}                            & \multicolumn{2}{c}{FC4}                \\
{[}0.01,0.03{]} & \multicolumn{1}{c|}{HT(0,1)} & \multicolumn{2}{c|}{\multirow{3}{*}{-}}                                          & \multicolumn{2}{c}{\multirow{3}{*}{-}}                   &  &  & {[}0.08,0.24{]} & \multicolumn{1}{c|}{HT(0,1)}     & \multicolumn{2}{c|}{\multirow{3}{*}{-}}            & \multicolumn{2}{c}{\multirow{3}{*}{-}} \\
{[}0.04,0.09{]} & \multicolumn{1}{c|}{HT(1,1)} & \multicolumn{2}{c|}{}                                                            & \multicolumn{2}{c}{}                                     &  &  & {[}0.32,0.4{]}  & \multicolumn{1}{c|}{HT(1,1)}     & \multicolumn{2}{c|}{}                              & \multicolumn{2}{c}{}                   \\
                & \multicolumn{1}{c|}{}        & \multicolumn{2}{c|}{}                                                            & \multicolumn{2}{c}{}                                     &  &  &                 & \multicolumn{1}{c|}{}            & \multicolumn{2}{c|}{}                              & \multicolumn{2}{c}{}                   \\
{[}0.1,0.44{]}  & \multicolumn{1}{c|}{BT(1,1)} & {[}0.01,0.29{]}                  & \multicolumn{1}{c|}{BT(1,1)}                  & \multicolumn{2}{c}{-}                                    &  &  & {[}0.48,2.16{]} & \multicolumn{1}{c|}{BT{[}1,1{]}} & {[}0.08,1.44{]} & \multicolumn{1}{c|}{BT{[}1,1{]}} & \multicolumn{2}{c}{-}                  \\
{[}0.45,2{]}    & \multicolumn{1}{c|}{LT(1,1)} & {[}0.3,0.38{]}                   & \multicolumn{1}{c|}{LT(1,1)}                  & {[}0.01,2{]}                  & LT(0,1)                  &  &  & {[}2.24,4.8{]}  & \multicolumn{1}{c|}{LT(1,1)}     & {[}1.52,2.64{]} & \multicolumn{1}{c|}{LT(1,1)}     & {[}0.48,4.8{]}        & LT(0,1)        \\
                & \multicolumn{1}{c|}{}        & {[}0.39,2{]}                     & \multicolumn{1}{c|}{LT(0,1)}                  &                               &                          &  &  &                 & \multicolumn{1}{c|}{}            & {[}2.72,4.8{]}  & \multicolumn{1}{c|}{LT(0,1)}     &                       &                \\
\multicolumn{6}{c}{}                                                                                                                                                                         &  &  & \multicolumn{6}{c}{}                                                                                                                             \\
\multicolumn{6}{c}{$K=5$ (NN1+$\calD_1$)}                                                                                                                                                    &  &  & \multicolumn{6}{c}{$K=5$ (NN1+$\calD_2$)}                                                                                                        \\
\multicolumn{6}{c}{}                                                                                                                                                                         &  &  & \multicolumn{6}{c}{}                                                                                                                             \\
\multicolumn{2}{c}{FC2}                        & \multicolumn{2}{c}{FC3}                                                          & \multicolumn{2}{c}{FC4}                                  &  &  & \multicolumn{2}{c}{FC2}                            & \multicolumn{2}{c}{FC3}                            & \multicolumn{2}{c}{FC4}                \\
{[}0.01,0.15{]} & \multicolumn{1}{c|}{HT(0,1)} & \multicolumn{2}{c|}{\multirow{3}{*}{-}}                                          & \multicolumn{2}{c}{\multirow{3}{*}{-}}                   &  &  & {[}0.08,0.48{]} & \multicolumn{1}{c|}{HT(0,1)}     & \multicolumn{2}{c|}{\multirow{3}{*}{-}}            & \multicolumn{2}{c}{\multirow{3}{*}{-}} \\
{[}0.16,0.2{]}  & \multicolumn{1}{c|}{HT(4,1)} & \multicolumn{2}{c|}{}                                                            & \multicolumn{2}{c}{}                                     &  &  & {[}0.56,0.64{]} & \multicolumn{1}{c|}{HT(4,1)}     & \multicolumn{2}{c|}{}                              & \multicolumn{2}{c}{}                   \\
{[}0.21,0.28{]} & \multicolumn{1}{c|}{HT(0,5)} & \multicolumn{2}{c|}{}                                                            & \multicolumn{2}{c}{}                                     &  &  & {[}0.72,1.04{]} & \multicolumn{1}{c|}{HT(0,5)}     & \multicolumn{2}{c|}{}                              & \multicolumn{2}{c}{}                   \\
{[}0.29,0.66{]} & \multicolumn{1}{c|}{BT(1,4)} & {[}0.01,0.59{]}                  & \multicolumn{1}{c|}{BT(1,4)}                  & \multicolumn{2}{c}{-}                                    &  &  & {[}1.12,3.2{]}  & \multicolumn{1}{c|}{BT(1,4)}     & {[}0.08,2.72{]} & \multicolumn{1}{c|}{BT(1,4)}     & \multicolumn{2}{c}{-}                  \\
{[}0.67,1.04{]} & \multicolumn{1}{c|}{LT(1,4)} & {[}0.6,1.9{]}                    & \multicolumn{1}{c|}{LT(1,4)}                  & {[}0.01,2{]}                  & LT(0,4)                  &  &  & {[}3.28,4.8{]}  & \multicolumn{1}{c|}{LT(1,4)}     & {[}2.8,4.8{]}   & \multicolumn{1}{c|}{LT(1,4)}     & {[}0.08,4.8{]}        & LT(0,4)        \\
{[}1.05,2{]}    & \multicolumn{1}{c|}{LT(0,5)} & {[}1.95,2{]}                     & \multicolumn{1}{c|}{LT(0,5)}                  &                               &                          &  &  &                 & \multicolumn{1}{c|}{}            &                 & \multicolumn{1}{c|}{}            &                       &                \\
\multicolumn{6}{c}{}                                                                                                                                                                         &  &  & \multicolumn{6}{c}{}                                                                                                                             \\
\multicolumn{6}{c}{$K=8$ (NN1+$\calD_1$)}                                                                                                                                                    &  &  & \multicolumn{6}{c}{$K=8$ (NN1+$\calD_2$)}                                                                                                        \\
\multicolumn{6}{c}{}                                                                                                                                                                         &  &  & \multicolumn{6}{c}{}                                                                                                                             \\
\multicolumn{2}{c}{FC2}                        & \multicolumn{2}{c}{FC3}                                                          & \multicolumn{2}{c}{FC4}                                  &  &  & \multicolumn{2}{c}{FC2}                            & \multicolumn{2}{c}{FC3}                            & \multicolumn{2}{c}{FC4}                \\
{[}0.01,0.19{]} & \multicolumn{1}{c|}{HT(0,1)} & \multirow{3}{*}{{[}0.01,0.32{]}} & \multicolumn{1}{c|}{\multirow{3}{*}{HT(0,8)}} & \multicolumn{2}{c}{\multirow{3}{*}{-}}                   &  &  & {[}0.08,0.56{]} & \multicolumn{1}{c|}{HT(0,1)}     & {[}0.08,1.12{]} & \multicolumn{1}{c|}{HT(7,1)}     & \multicolumn{2}{c}{\multirow{3}{*}{-}} \\
{[}0.2,0.25{]}  & \multicolumn{1}{c|}{HT(7,1)} &                                  & \multicolumn{1}{c|}{}                         & \multicolumn{2}{c}{}                                     &  &  & {[}0.64,0.96{]} & \multicolumn{1}{c|}{HT(7,1)}     & {[}1.2,1.44{]}  & \multicolumn{1}{c|}{HT(0,8)}     & \multicolumn{2}{c}{}                   \\
{[}0.26,0.36{]} & \multicolumn{1}{c|}{HT(0,8)} &                                  & \multicolumn{1}{c|}{}                         & \multicolumn{2}{c}{}                                     &  &  & {[}1.04,1.36{]} & \multicolumn{1}{c|}{HT(0,8)}     &                 & \multicolumn{1}{c|}{}            & \multicolumn{2}{c}{}                   \\
{[}0.37,0.72{]} & \multicolumn{1}{c|}{BT(1,7)} & {[}0.32,0.63{]}                  & \multicolumn{1}{c|}{BT(1,7)}                  & \multicolumn{2}{c}{-}                                    &  &  & {[}1.44,3.52{]} & \multicolumn{1}{c|}{BT(1,7)}     & {[}1.52,3.12{]} & \multicolumn{1}{c|}{BT(1,7)}     & \multicolumn{2}{c}{-}                  \\
{[}0.73,1.01{]} & \multicolumn{1}{c|}{LT(1,7)} & {[}0.64,1.15{]}                  & \multicolumn{1}{c|}{LT(1,7)}                  & \multirow{3}{*}{{[}0.01,2{]}} & \multirow{3}{*}{LT(0,7)} &  &  & {[}3.6,4.64{]}  & \multicolumn{1}{c|}{LT(1,7)}     & {[}3.2,4.8{]}   & \multicolumn{1}{c|}{LT(1,7)}     & {[}0.08,4.8{]}        & LT(0,7)        \\
{[}1.02,1.9{]}  & \multicolumn{1}{c|}{LT(0,8)} & {[}1.16,1.95{]}                  & \multicolumn{1}{c|}{LT(0,8)}                  &                               &                          &  &  & {[}4.72,4.8{]}  & \multicolumn{1}{c|}{LT(0,8)}     &                 & \multicolumn{1}{c|}{}            &                       &                \\
{[}1.95,2{]}    & \multicolumn{1}{c|}{LT(0,7)} & {[}2,2{]}                        & \multicolumn{1}{c|}{LT(0,7)}                  &                               &                          &  &  &                 & \multicolumn{1}{c|}{}            &                 & \multicolumn{1}{c|}{}            &                       &               
\\\bottomrule\\
\end{tabular}
\footnotesize{* The interval is the range of tuning parameters which tune data SNR.}
\end{sidewaystable}

\begin{sidewaystable}[htbp]

\caption{~~ The results of ESD types in different layers with different SNR in \textbf{NN2}. }
\label{tb:NN2}
\begin{tabular}{ccccclcccc}\toprule
\multicolumn{4}{c}{$K=2$ (NN2+$\calD_1$)}                                                 &  &  & \multicolumn{4}{c}{$K=2$ (NN2+$\calD_2$)}                                                      \\
\multicolumn{4}{c}{}                                                                      &  &  & \multicolumn{4}{c}{}                                                                           \\
\multicolumn{2}{c}{FC1}                          & \multicolumn{2}{c}{FC2}                &  &  & \multicolumn{2}{c}{FC1}                               & \multicolumn{2}{c}{FC2}                \\
{[}0.005,0.015{]} & \multicolumn{1}{c|}{HT(0,1)} & \multicolumn{2}{c}{\multirow{3}{*}{-}} &  &  & {[}0.08,0.16{]}    & \multicolumn{1}{c|}{HT(0,1)}     & \multicolumn{2}{c}{\multirow{3}{*}{-}} \\
                  & \multicolumn{1}{c|}{}        & \multicolumn{2}{c}{}                   &  &  &                    & \multicolumn{1}{c|}{}            & \multicolumn{2}{c}{}                   \\
                  & \multicolumn{1}{c|}{}        & \multicolumn{2}{c}{}                   &  &  &                    & \multicolumn{1}{c|}{}            & \multicolumn{2}{c}{}                   \\
{[}0.02,0.055{]}  & \multicolumn{1}{c|}{BT(1,1)} & \multicolumn{2}{c}{-}                  &  &  & {[}0.24,1.52{]}    & \multicolumn{1}{c|}{BT{[}1,1{]}} & {[}0.08,0.48{]}  & BT{[}1,1{]}         \\
{[}0.06,0.235{]}  & \multicolumn{1}{c|}{LT(1,1)} & {[}0.005,0.4{]}         & LT(1,1)      &  &  & {[}1.6,1.92{]}     & \multicolumn{1}{c|}{LT(1,1)}     & {[}0.56,1.6{]}   & LT(1,1)             \\
{[}0.24,0.4{]}    & \multicolumn{1}{c|}{LT(0,1)} &                         &              &  &  & {[}2,4.8{]}        & \multicolumn{1}{c|}{LT(0,1)}     & {[}1.68,4.8{]}   & LT(0,1)             \\
\multicolumn{4}{c}{}                                                                      &  &  & \multicolumn{4}{c}{}                                                                           \\
\multicolumn{4}{c}{$K=5$ (NN2+$\calD_1$)}                                                 &  &  & \multicolumn{4}{c}{$K=5$ (NN2+$\calD_2$)}                                                      \\
\multicolumn{4}{c}{}                                                                      &  &  & \multicolumn{4}{c}{}                                                                           \\
\multicolumn{2}{c}{FC1}                          & \multicolumn{2}{c}{FC2}                &  &  & \multicolumn{2}{c}{FC1}                               & \multicolumn{2}{c}{FC2}                \\
{[}0.005,0.015{]} & \multicolumn{1}{c|}{HT(0,1)} & \multicolumn{2}{c}{\multirow{3}{*}{-}} &  &  & {[}0.08,0.48{]}    & \multicolumn{1}{c|}{HT(0,0)}     & \multicolumn{2}{c}{\multirow{3}{*}{-}} \\
{[}0.02,0.035{]}  & \multicolumn{1}{c|}{HT(4,1)} & \multicolumn{2}{c}{}                   &  &  & {[}0.56,0.56{]}{]} & \multicolumn{1}{c|}{HT(0,1)}     & \multicolumn{2}{c}{}                   \\
{[}0.04,0.045{]}  & \multicolumn{1}{c|}{HT(0,5)} & \multicolumn{2}{c}{}                   &  &  & {[}0.64,0.88{]}    & \multicolumn{1}{c|}{HT(0,5)}     & \multicolumn{2}{c}{}                   \\
{[}0.05,0.12{]}   & \multicolumn{1}{c|}{BT(1,4)} & {[}0.005,0.125{]}       & BT(1,4)      &  &  & {[}0.96,2.24{]}    & \multicolumn{1}{c|}{BT(1,4)}     & {[}0.08,2.16{]}  & BT(0,4)$\to$BT(1,4) \\
{[}0.125,0.4{]}   & \multicolumn{1}{c|}{LT(1,4)} & {[}0.03,0.38{]}         & LT(1,4)      &  &  & {[}2.32,2.48{]}    & \multicolumn{1}{c|}{LT(1,4)}     & {[}2.24,4.8{]}   & LT(1,4)             \\
                  & \multicolumn{1}{c|}{}        & {[}0.385,0.4{]}         & LT(0,5)      &  &  & {[}2.56,4.8{]}     & \multicolumn{1}{c|}{LT(0,4)}     &                  &                     \\
\multicolumn{4}{c}{}                                                                      &  &  & \multicolumn{4}{c}{}                                                                           \\
\multicolumn{4}{c}{$K=8$ (NN2+$\calD_1$)}                                                 &  &  & \multicolumn{4}{c}{$K=8$ (NN2+$\calD_2$)}                                                      \\
\multicolumn{4}{c}{}                                                                      &  &  & \multicolumn{4}{c}{}                                                                           \\
\multicolumn{2}{c}{FC1}                          & \multicolumn{2}{c}{FC2}                &  &  & \multicolumn{2}{c}{FC1}                               & \multicolumn{2}{c}{FC2}                \\
{[}0.005,0.025{]} & \multicolumn{1}{c|}{HT(0,1)} & \multicolumn{2}{c}{\multirow{3}{*}{-}} &  &  & {[}0.08,0.8{]}     & \multicolumn{1}{c|}{HT(0,0)}     & \multicolumn{2}{c}{\multirow{3}{*}{-}} \\
{[}0.03,0.06{]}   & \multicolumn{1}{c|}{HT(7,1)} & \multicolumn{2}{c}{}                   &  &  & {[}0.88,0.96{]}    & \multicolumn{1}{c|}{HT(0,1)}     & \multicolumn{2}{c}{}                   \\
{[}0.065,0.065{]} & \multicolumn{1}{c|}{HT(0,8)} & \multicolumn{2}{c}{}                   &  &  & {[}1.04,1.12{]}    & \multicolumn{1}{c|}{HT(0,8)}     & \multicolumn{2}{c}{}                   \\
{[}0.07,0.135{]}  & \multicolumn{1}{c|}{BT(1,7)} & {[}0.005,0.095{]}       & BT(1,7)      &  &  & {[}1.2,2.72{]}     & \multicolumn{1}{c|}{BT(1,7)}     & {[}0.08,2.4{]}   & BT(1,7)             \\
{[}0.14,0.4{]}    & \multicolumn{1}{c|}{LT(1,7)} & {[}0.1,0.24{]}          & LT(1,7)      &  &  & {[}2.8,3.04{]}     & \multicolumn{1}{c|}{LT(1,7)}     & {[}2.48,4.64{]}  & LT(1,7)             \\
                  & \multicolumn{1}{c|}{}        & {[}0.245,0.4{]}         & LT(0,8)      &  &  & {[}3.12,4.8{]}     & \multicolumn{1}{c|}{LT(0,7)}     & {[}4.72,4.8{]}   & LT(0,8)            
\\\bottomrule\\
\end{tabular}

\footnotesize{* The interval is the range of tuning parameters which tune data SNR.}
\end{sidewaystable}

\begin{sidewaystable}[htbp]
\caption{~~The results of ESD types in different layers with different SNR in different batchsizes. }
\label{tb:batch}

\begin{tabular}{cccccccccccc}\toprule
\multicolumn{6}{c}{Batchsize$=32$ (NN1+$\calD_1$)}                                                                                                                                                                          &  &  & \multicolumn{4}{c}{Batchsize$=32$ (NN2+$\calD_2$)}                                                                       \\
\multicolumn{6}{c}{}                                                                                                                                                                                                        &  &  & \multicolumn{4}{c}{}                                                                                                     \\
\multicolumn{2}{c}{FC2}                                                       & \multicolumn{2}{c}{FC3}                                                          & \multicolumn{2}{c}{FC4}                                  &  &  & \multicolumn{2}{c}{FC1}                                                         & \multicolumn{2}{c}{FC2}                \\
{[}0.01,0.17{]}               & \multicolumn{1}{c|}{HT(0,1)}                  & {[}0.01,0.44{]}                  & \multicolumn{1}{c|}{RC}                       & \multicolumn{2}{c}{\multirow{3}{*}{-}}                   &  &  & \multirow{3}{*}{{[}0.08,0.8{]}} & \multicolumn{1}{c|}{\multirow{3}{*}{HT(0,0)}} & \multicolumn{2}{c}{\multirow{3}{*}{-}} \\
{[}0.18,0.46{]}               & \multicolumn{1}{c|}{HT(7,1)}                  & {[}0.45,0.52{]}                  & \multicolumn{1}{c|}{HT(7,1)}                  & \multicolumn{2}{c}{}                                     &  &  &                                 & \multicolumn{1}{c|}{}                         & \multicolumn{2}{c}{}                   \\
{[}0.47,0.55{]}               & \multicolumn{1}{c|}{HT(0,8)}                  & {[}0.52,0.59{]}                  & \multicolumn{1}{c|}{HT(0,8)}                  & \multicolumn{2}{c}{}                                     &  &  &                                 & \multicolumn{1}{c|}{}                         & \multicolumn{2}{c}{}                   \\
{[}0.56,0.84{]}               & \multicolumn{1}{c|}{BT{(}1,7{)}}              & {[}0.6,0.78{]}                   & \multicolumn{1}{c|}{BT(1,7)}                  & {[}0.15,0.45{]}               & HT(0,7)                  &  &  & {[}0.88,3.6{]}                  & \multicolumn{1}{c|}{BT{(}0,7{)}}              & {[}0.08,3.28{]}  & RC$\to$BT{(}1,7{)}  \\
{[}0.85,1.08{]}               & \multicolumn{1}{c|}{LT(1,7)}                  & {[}0.79,1.12{]}                  & \multicolumn{1}{c|}{LT(1,7)}                  & {[}0.46,2{]}                  & LT(0,7)                  &  &  & {[}3.68,4.8{]}                  & \multicolumn{1}{c|}{LT(0,7)}                  & {[}3.36,4.8{]}   & LT(1,7)             \\
{[}1.09,2{]}                  & \multicolumn{1}{c|}{LT(0,8)}                  & {[}1.13,2{]}                     & \multicolumn{1}{c|}{LT(0,8)}                  &                               &                          &  &  &                                 & \multicolumn{1}{c|}{}                         &                  &                     \\
\multicolumn{6}{c}{}                                                                                                                                                                                                        &  &  & \multicolumn{4}{c}{}                                                                                                     \\
\multicolumn{6}{c}{Batchsize$=64$ (NN1+$\calD_1$)}                                                                                                                                                                          &  &  & \multicolumn{4}{c}{Batchsize$=64$ (NN2+$\calD_2$)}                                                                       \\
\multicolumn{6}{c}{}                                                                                                                                                                                                        &  &  & \multicolumn{4}{c}{}                                                                                                     \\
\multicolumn{2}{c}{FC2}                                                       & \multicolumn{2}{c}{FC3}                                                          & \multicolumn{2}{c}{FC4}                                  &  &  & \multicolumn{2}{c}{FC1}                                                         & \multicolumn{2}{c}{FC2}                \\
{[}0.01,0.19{]}               & \multicolumn{1}{c|}{HT(0,1)}                  & \multirow{3}{*}{{[}0.01,0.32{]}} & \multicolumn{1}{c|}{\multirow{3}{*}{HT(0,8)}} & \multicolumn{2}{c}{\multirow{3}{*}{-}}                   &  &  & {[}0.08,0.8{]}                  & \multicolumn{1}{c|}{HT(0,0)}                  & \multicolumn{2}{c}{\multirow{3}{*}{-}} \\
{[}0.2,0.25{]}                & \multicolumn{1}{c|}{HT(7,1)}                  &                                  & \multicolumn{1}{c|}{}                         & \multicolumn{2}{c}{}                                     &  &  & {[}0.88,0.96{]}                 & \multicolumn{1}{c|}{HT(0,1)}                  & \multicolumn{2}{c}{}                   \\
{[}0.26,0.36{]}               & \multicolumn{1}{c|}{HT(0,8)}                  &                                  & \multicolumn{1}{c|}{}                         & \multicolumn{2}{c}{}                                     &  &  & {[}1.04,1.12{]}                 & \multicolumn{1}{c|}{HT(0,8)}                  & \multicolumn{2}{c}{}                   \\
{[}0.37,0.72{]}               & \multicolumn{1}{c|}{BT(1,7)}                  & {[}0.33,0.63{]}                  & \multicolumn{1}{c|}{BT(1,7)}                  & \multicolumn{2}{c}{-}                                    &  &  & {[}1.2,2.72{]}                  & \multicolumn{1}{c|}{BT(1,7)}                  & {[}0.08,2.4{]}   & BT(1,7)             \\
{[}0.73,1.01{]}               & \multicolumn{1}{c|}{LT(1,7)}                  & {[}0.64,1.15{]}                  & \multicolumn{1}{c|}{LT(1,7)}                  & {[}0.01,2{]}                  & LT(0,7)                  &  &  & {[}2.8,3.04{]}                  & \multicolumn{1}{c|}{LT(1,7)}                  & {[}2.48,4.64{]}  & LT(1,7)             \\
{[}1.02,1.9{]}                & \multicolumn{1}{c|}{LT(0,8)}                  & {[}1.16,1.95{]}                  & \multicolumn{1}{c|}{LT(0,8)}                  &                               &                          &  &  & {[}3.12,4.8{]}                  & \multicolumn{1}{c|}{LT(0,7)}                  & {[}4.72,4.8{]}   & LT(0,8)             \\
{[}1.95,2{]}                  & \multicolumn{1}{c|}{LT(0,7)}                  & {[}2,2{]}                        & \multicolumn{1}{c|}{LT(0,7)}                  &                               &                          &  &  &                                 & \multicolumn{1}{c|}{}                         &                  &                     \\
\multicolumn{6}{c}{}                                                                                                                                                                                                        &  &  & \multicolumn{4}{c}{}                                                                                                     \\
\multicolumn{6}{c}{Batchsize$=256$ (NN1+$\calD_1$)}                                                                                                                                                                         &  &  & \multicolumn{4}{c}{Batchsize$=256$ (NN2+$\calD_2$)}                                                                      \\
\multicolumn{6}{c}{}                                                                                                                                                                                                        &  &  & \multicolumn{4}{c}{}                                                                                                     \\
\multicolumn{2}{c}{FC2}                                                       & \multicolumn{2}{c}{FC3}                                                          & \multicolumn{2}{c}{FC4}                                  &  &  & \multicolumn{2}{c}{FC1}                                                         & \multicolumn{2}{c}{FC2}                \\
{[}0.01,0.16{]}               & \multicolumn{1}{c|}{HT(0,0)}                  & \multicolumn{2}{c|}{\multirow{3}{*}{-}}                                          & \multicolumn{2}{c}{\multirow{3}{*}{-}}                   &  &  & \multicolumn{2}{c|}{\multirow{3}{*}{-}}                                         & \multicolumn{2}{c}{\multirow{3}{*}{-}} \\
{[}0.17,0.22{]}               & \multicolumn{1}{c|}{HT(0,8)}                  & \multicolumn{2}{c|}{}                                                            & \multicolumn{2}{c}{}                                     &  &  & \multicolumn{2}{c|}{}                                                           & \multicolumn{2}{c}{}                   \\
                              & \multicolumn{1}{c|}{}                         & \multicolumn{2}{c|}{}                                                            & \multicolumn{2}{c}{}                                     &  &  & \multicolumn{2}{c|}{}                                                           & \multicolumn{2}{c}{}                   \\
{[}0.23,0.55{]}               & \multicolumn{1}{c|}{BT(0,7)$\to$BT(1,7)}      & {[}0.01,0.4{]}                   & \multicolumn{1}{c|}{BT(1,7)}                  & \multicolumn{2}{c}{-}                                    &  &  & \multicolumn{2}{c|}{-}                                                          & \multicolumn{2}{c}{-}                  \\
\multirow{3}{*}{{[}0.56,2{]}} & \multicolumn{1}{c|}{\multirow{3}{*}{LT(1,7)}} & {[}0.41,1.01{]}                  & \multicolumn{1}{c|}{LT(1,7)}                  & \multirow{3}{*}{{[}0.01,2{]}} & \multirow{3}{*}{LT(0,7)} &  &  & {[}0.08,1.2{]}                  & \multicolumn{1}{c|}{LT(0,0)}                  & {[}0.08,4.64{]}  & LT(1,7)             \\
                              & \multicolumn{1}{c|}{}                         & {[}1.02,1.95{]}                  & \multicolumn{1}{c|}{LT(0,8)}                  &                               &                          &  &  & {[}1.28,4.8{]}                  & \multicolumn{1}{c|}{LT(0,7)}                  & {[}4.72,4.8{]}   & LT(0,8)             \\
                              & \multicolumn{1}{c|}{}                         & {[}2,2{]}                        & \multicolumn{1}{c|}{LT(0,7)}                  &                               &                          &  &  &                                 & \multicolumn{1}{c|}{}                         &                  &                    
\\\bottomrule\\
\end{tabular}
\footnotesize{* The interval is the range of tuning parameters which tune data SNR.}
\end{sidewaystable}


\section{Experiments with Real Data}
\label{sec:realdata}

The previous results are based on synthetic Gaussian data.
Here we conduct experiments with real data sets
to show the impact of complex features in data on weight matrices spectra. 
The DNNs chosen for these experiments are 
LeNet, MiniAlexNet and VGG11
\citep{lecun1998gradient,krizhevsky2012imagenet}, which are the
most classic and representative  DNNS in pattern recognition.
We consider two  data sets, the MNIST and CIFAR10. 
Note that in  \citeauthor{martin2018implicit}'s work, the data sets such as CIFAR10, CIFAR100 and Image1000  all bring in HT type spectra due to complex features unlike the MNIST data set. In our  experiments, we select  MiniAlexNet instead of the more extensive AlexNet to reduce computing complexity. Table~\ref{Summaryrealdata} gives the type of spectra obtained in the trained NNs with each data set.


\begin{table}[htbp]
\centering
\caption{~~Spectra types obtained in the four combinations (Data Set)$\times$NN with batch size 16. Numbers in parentheses are the corresponding detection rates on test data.}
\label{Summaryrealdata}
\begin{tabular}{cccc}
\hline
\diagbox{Data Set}{NN}                & LeNet      & MiniAlexNet & VGG11      \\ \cline{2-4} 
\multicolumn{1}{c|}{MNIST}   & LT (99\%) & LT  (99\%)  & LT  (99\%) \\
\multicolumn{1}{c|}{CIFAR10} & HT (64\%)  & HT (76\%)   & HT (81\%)  \\ \hline
\end{tabular}
\end{table}
In training with MNIST, the spectra of weight matrices are always LT type, independently of the used NN;  for
CIFAR10, the spectra are always HT type  in all networks. The testing accuracies of detection on MNIST are  99\% in all networks, while  those on  CIFAR10 are   64\% with
LeNet, 76\% with  MiniAlexNet and 81\% with VGG11 respectively.
The differences in testing accuracy give evidence that CIFAR10 has  much more complex features than MNIST, and the complex features indicate more classification difficulty in CIFAR10 than MNIST. In the real data experiments, it shows that training on CIFAR10 is more likely to bring in the HT, which gives new evidence on the impact of classification difficulty on weight matrices spectra. Moreover,  
The complex features will also bring in the complex correlations during training procedure and thus HT emerges. Full details for these experimental results are
reported in Section~\ref{sec:realdataresult}.

\subsection{Experimental Design}

\begin{figure}[htbp]

\begin{minipage}[t]{1.07\linewidth}
\centering

\hspace{-1.3cm}\includegraphics{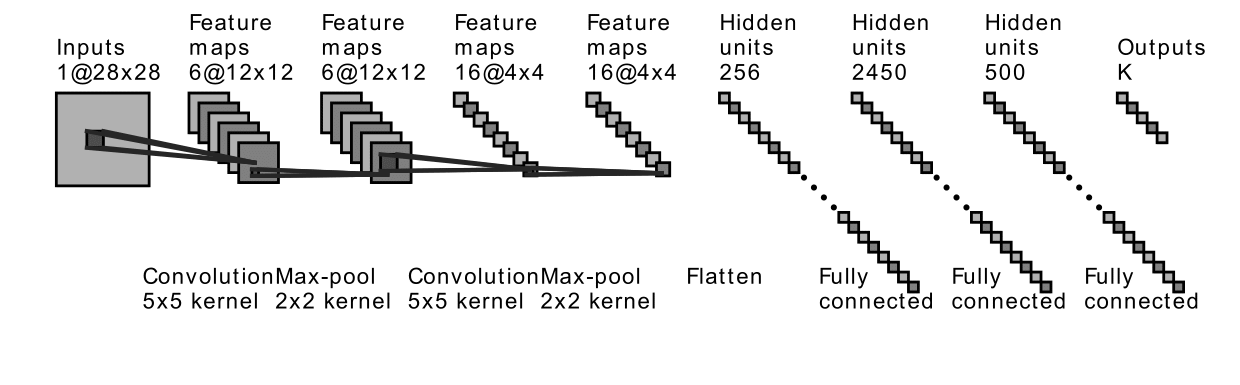}

\end{minipage}

\begin{minipage}[t]{1.07\linewidth}
\centering
\hspace{-1.3cm}\includegraphics{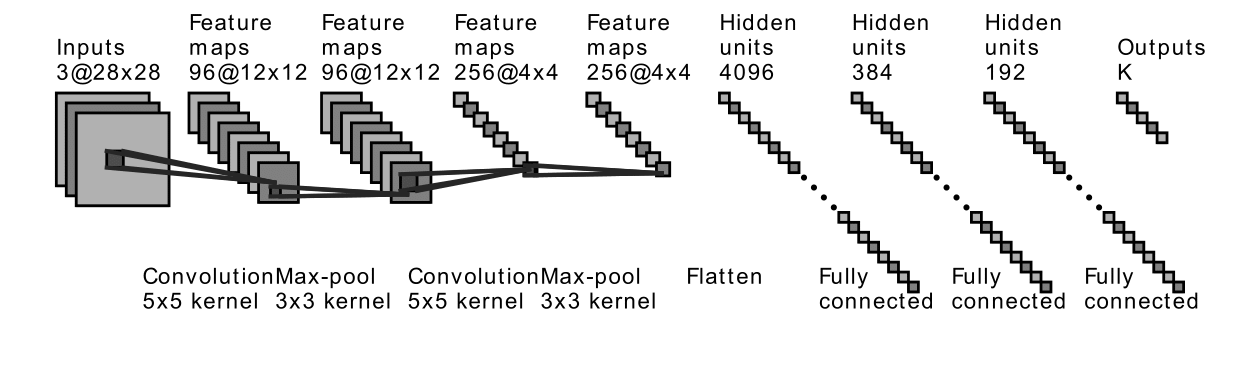}
\end{minipage}
\caption{~~ The structure of LeNet (top) and MiniAlexNet (Bottom). The input data sets are MNIST with size $1\times28\times 28$ or CIFAR10 with resize $3\times28\times 28$, the fully connected layers lay behind convolutional layers.}
\label{fig: structure}
\end{figure}

The structures of LeNet and MiniAlexNet are shown in Figure \ref{fig: structure}. Readers could refer the structure of VGG in \cite{2014Very}. The data sets we use are MNIST and CIFAR10. We tune batch sizes to have different practical architectures, then check spectra type in the different data sets under such different practical architectures. As before, we save trained models for the first 10 epochs and every four epochs afterward. The optimization methodology is the same as previously (See Section \ref{sec:optimization}).

\newgeometry{top=2.5cm}



\subsection{ Results on real data experiments}
\label{sec:realdataresult}
\begin{figure}[htbp]
\centering

\subfigure[16]{
\begin{minipage}[t]{0.19\linewidth}
\centering
\includegraphics[width=1.1in]{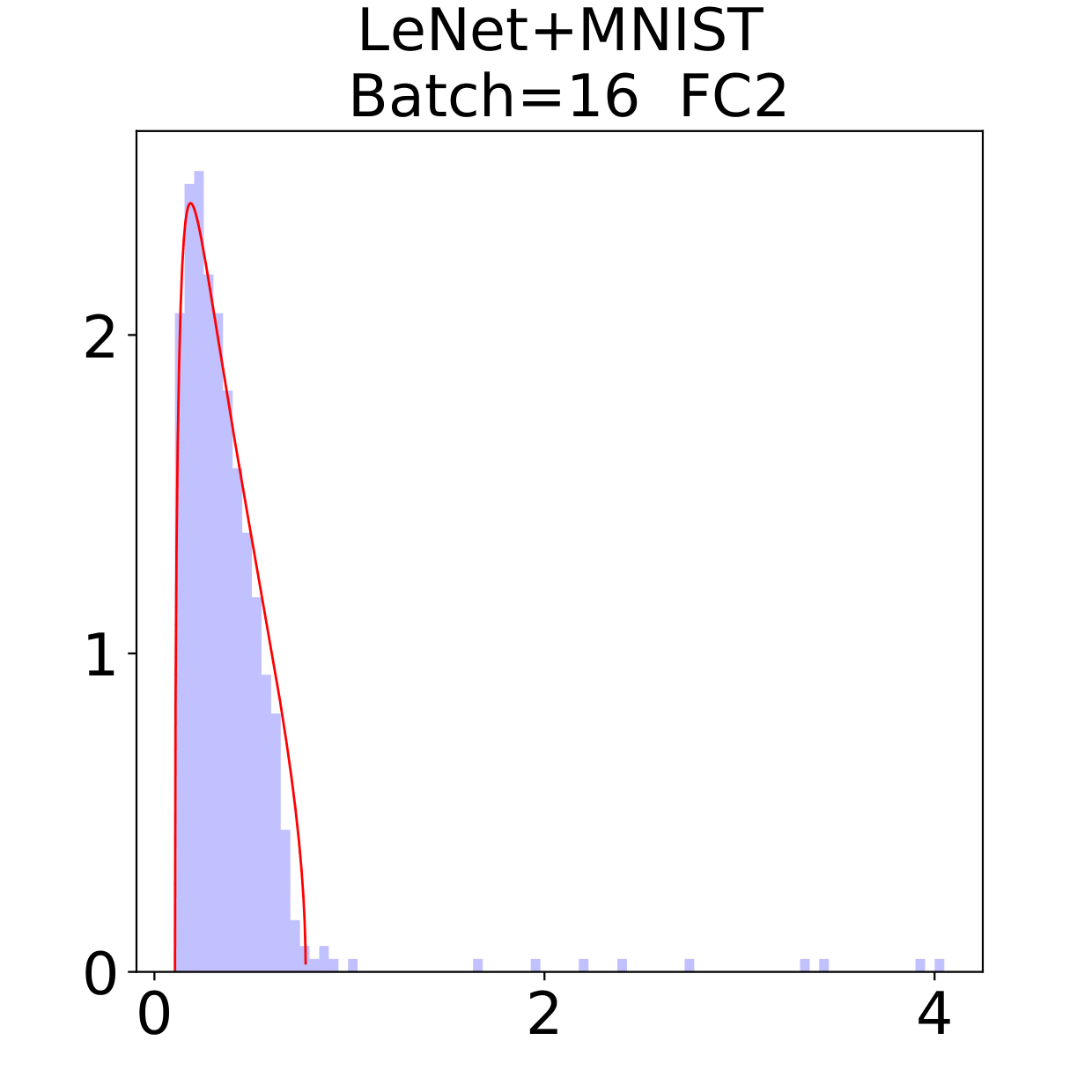}
\end{minipage}%
}%
\subfigure[32]{
\begin{minipage}[t]{0.19\linewidth}
\centering
\includegraphics[width=1.1in]{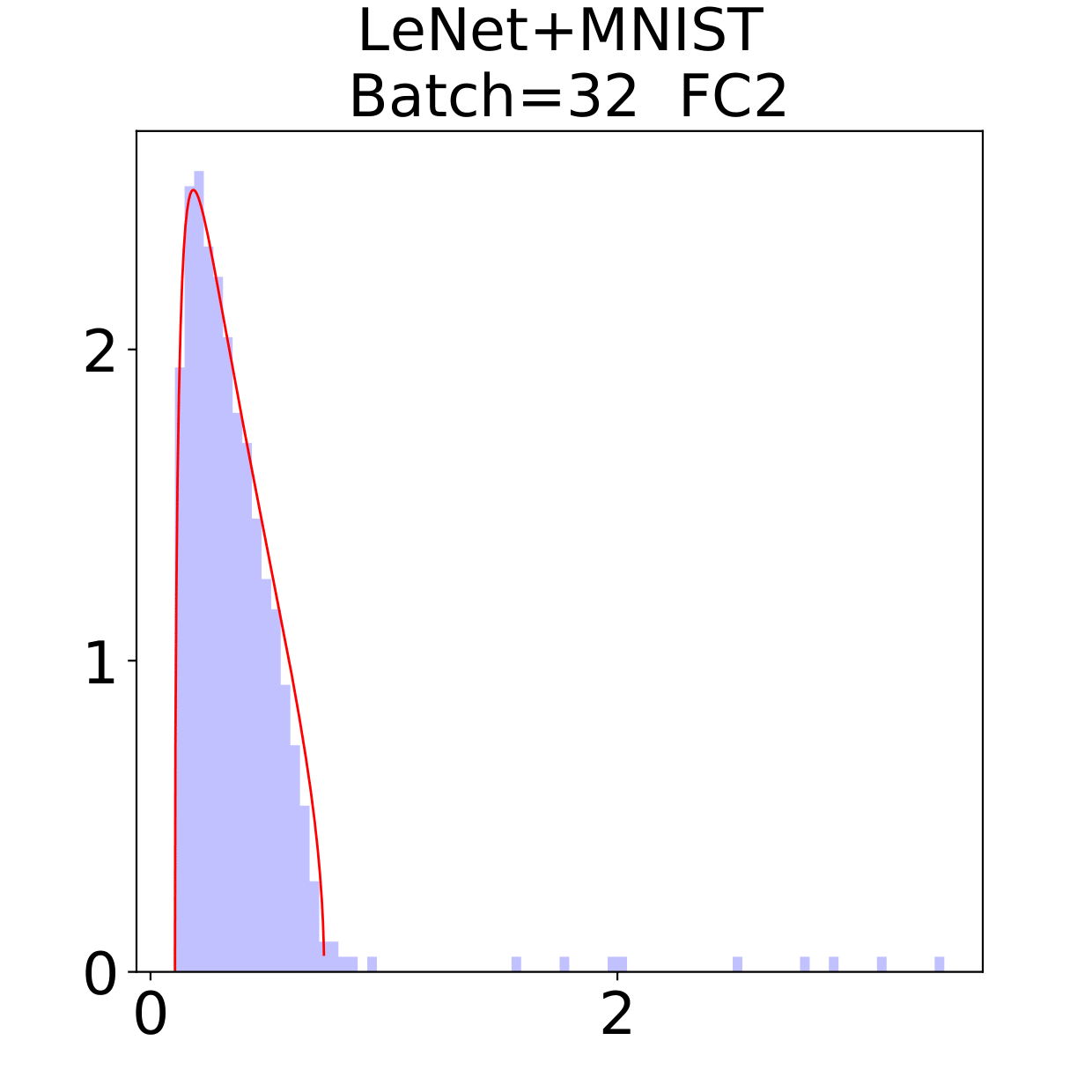}
\end{minipage}%
}%
\subfigure[64]{
\begin{minipage}[t]{0.19\linewidth}
\centering
\includegraphics[width=1.1in]{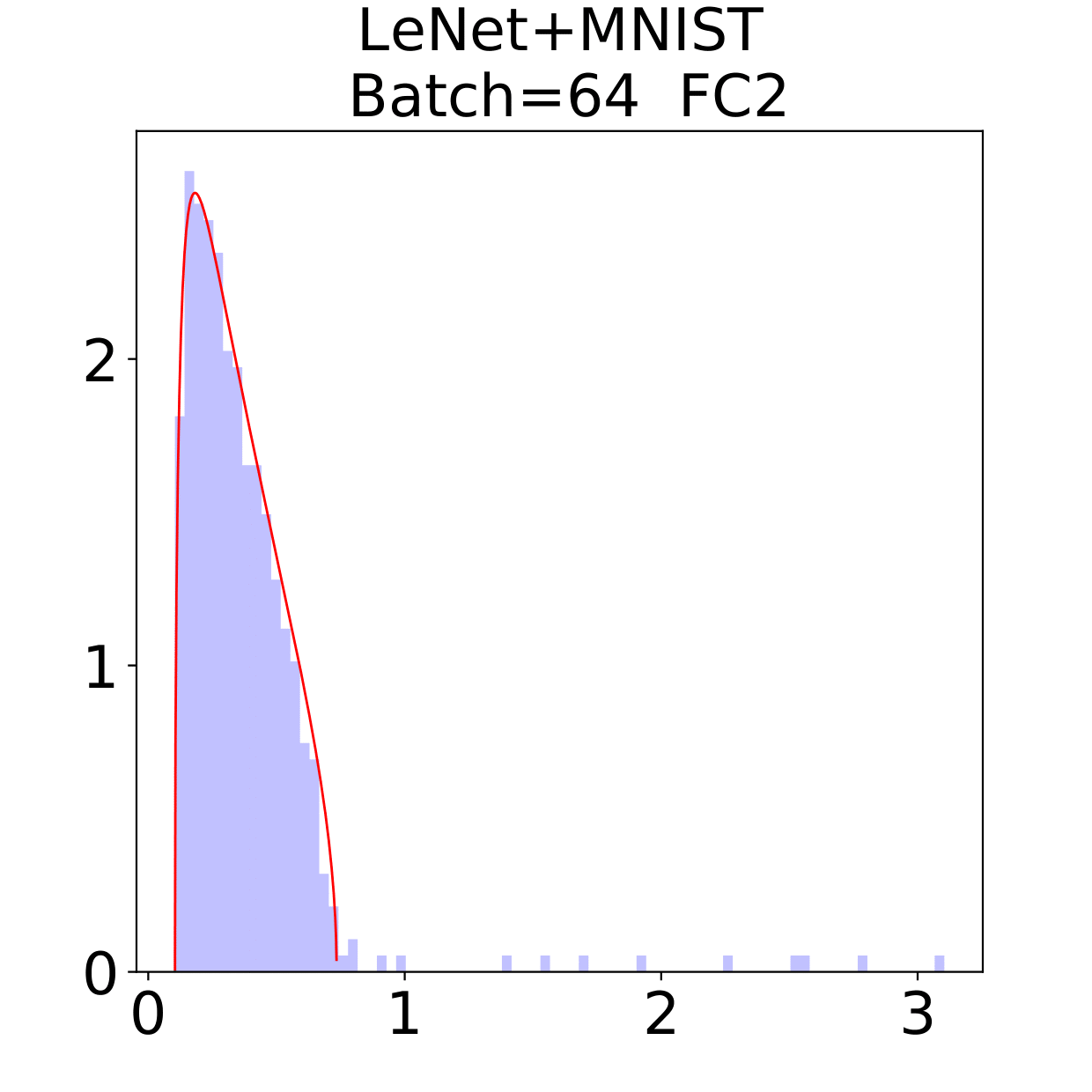}
\end{minipage}
}%
\subfigure[128]{
\begin{minipage}[t]{0.19\linewidth}
\centering
\includegraphics[width=1.1in]{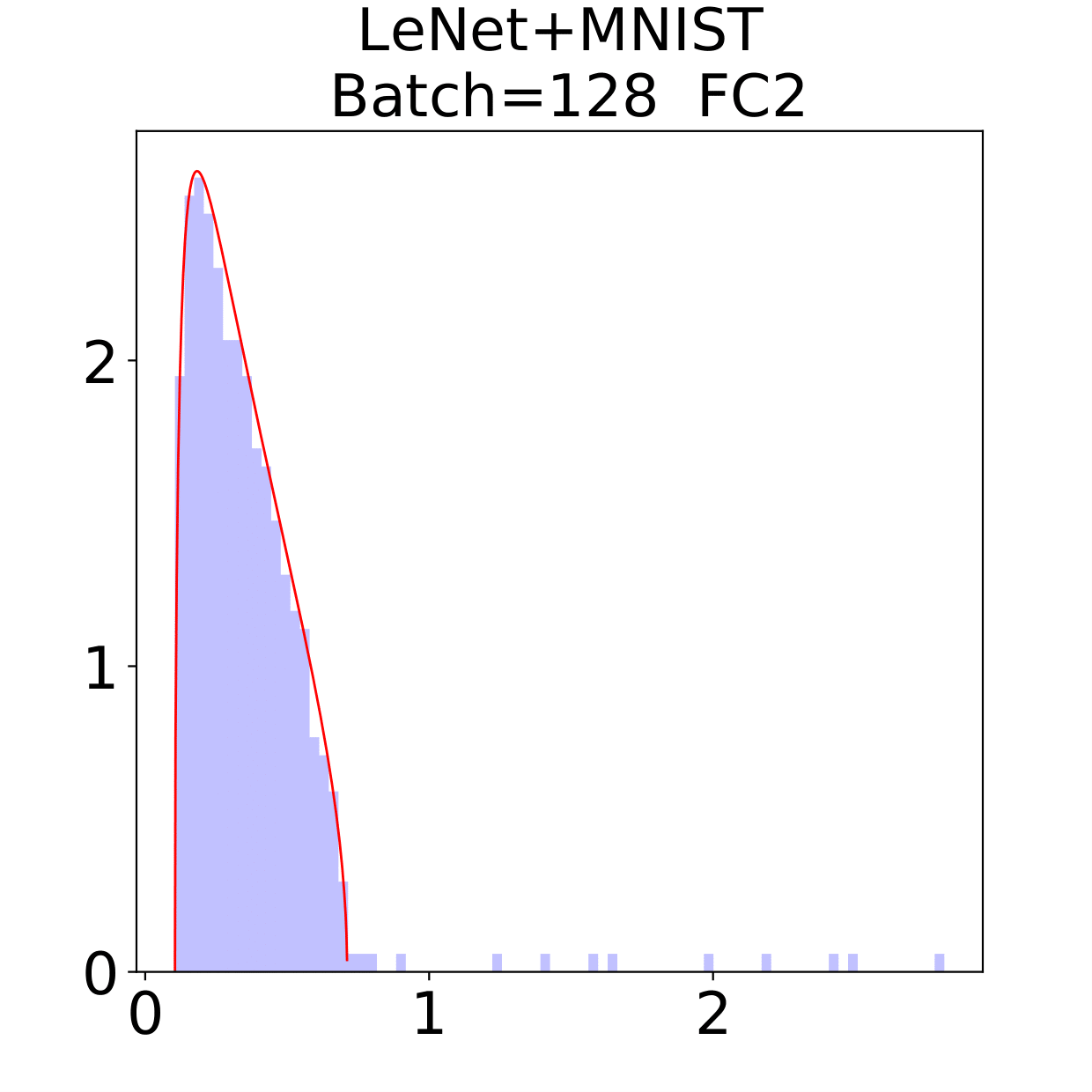}
\end{minipage}
}%
\subfigure[256]{
\begin{minipage}[t]{0.19\linewidth}
\centering
\includegraphics[width=1.1in]{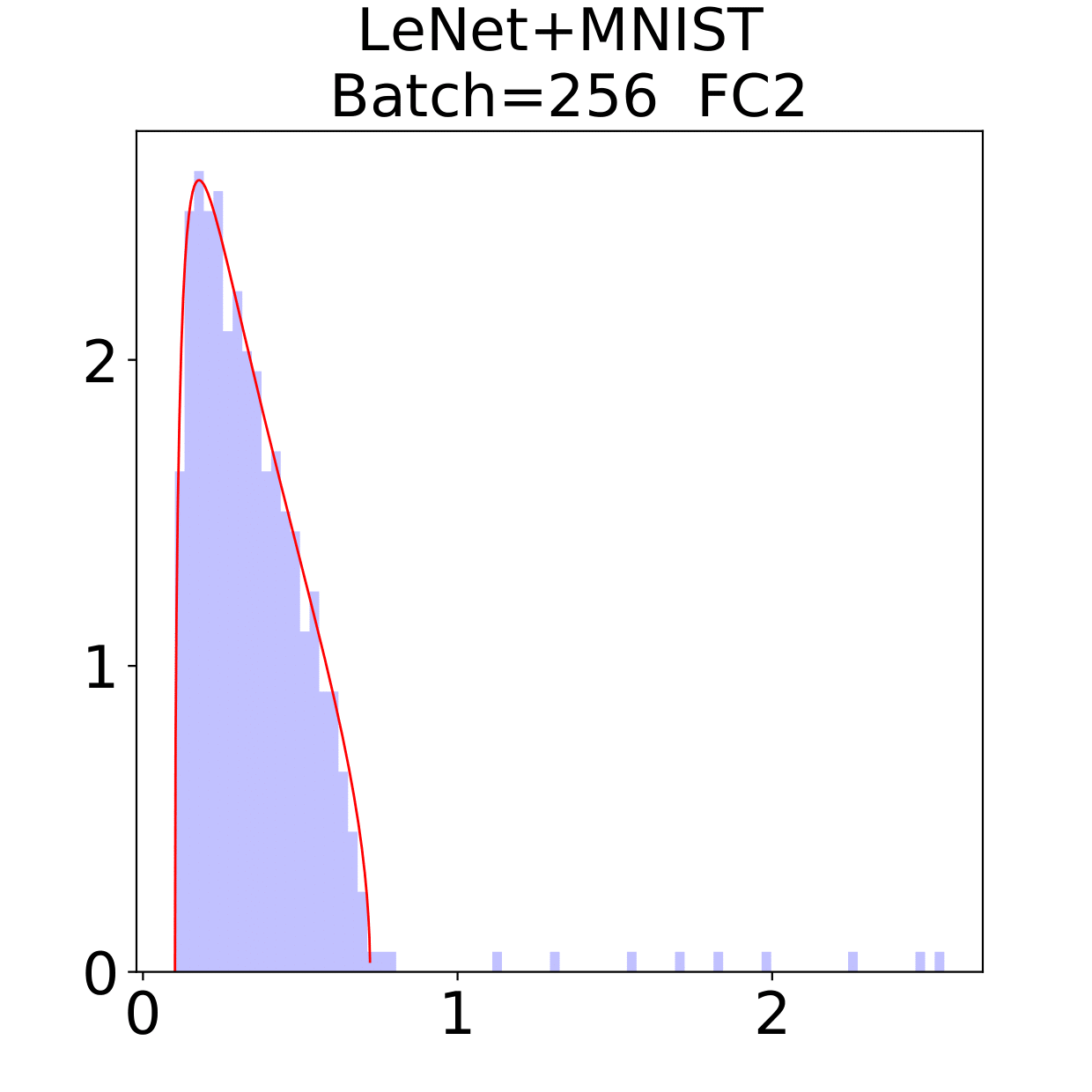}
\end{minipage}
}%

\subfigure[16]{
\begin{minipage}[t]{0.19\linewidth}
\centering
\includegraphics[width=1.1in]{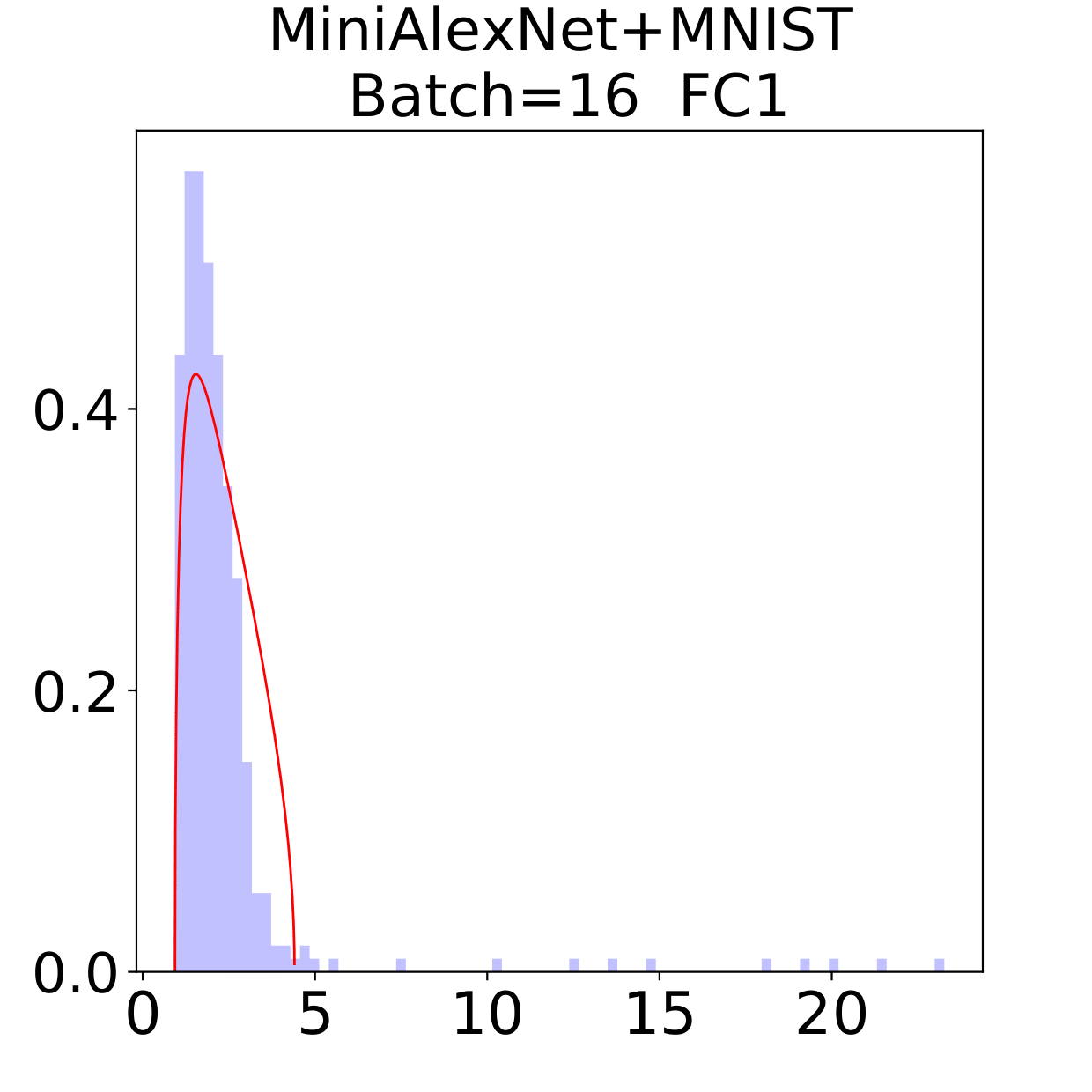}
\end{minipage}%
}%
\subfigure[32]{
\begin{minipage}[t]{0.19\linewidth}
\centering
\includegraphics[width=1.1in]{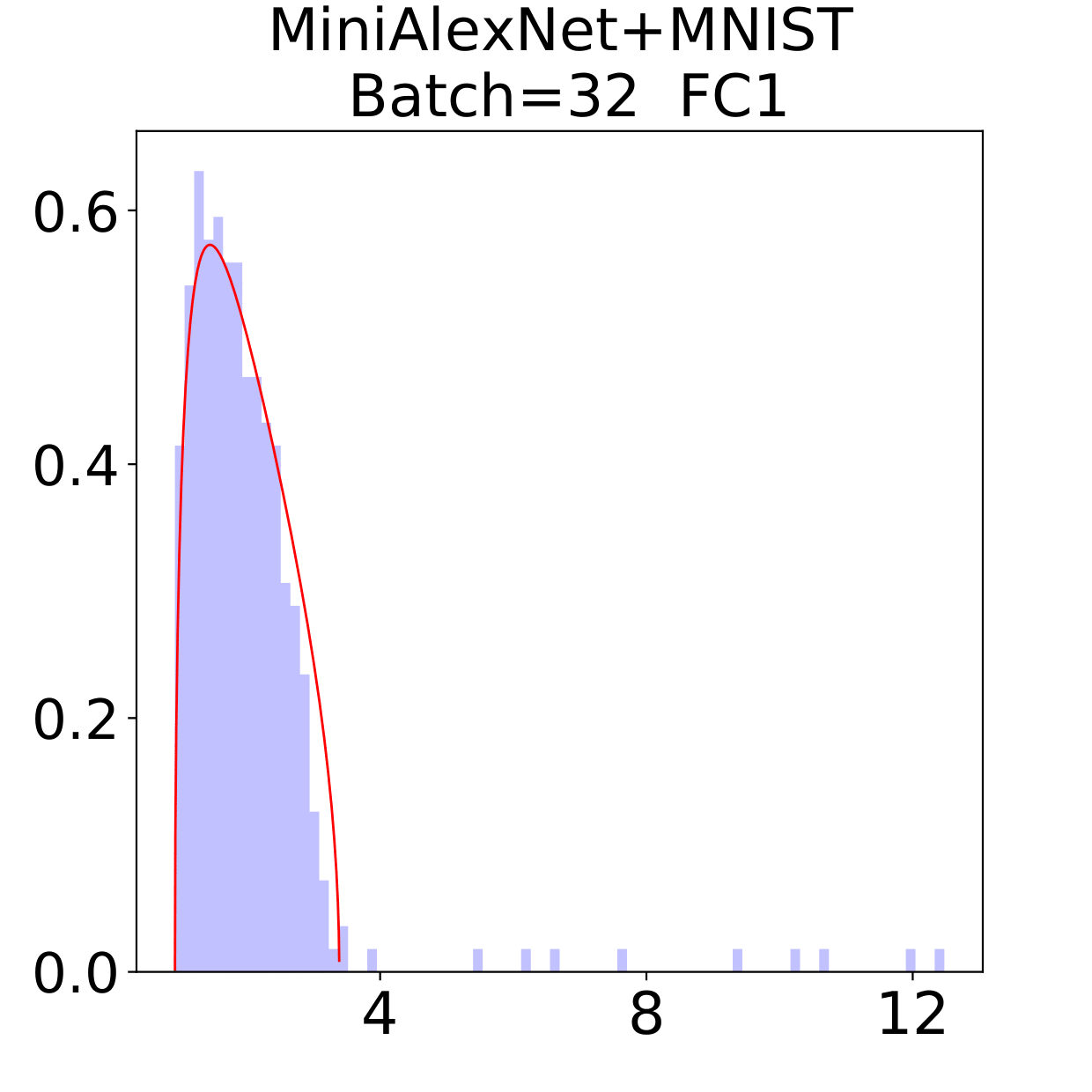}
\end{minipage}%
}%
\subfigure[64]{
\begin{minipage}[t]{0.19\linewidth}
\centering
\includegraphics[width=1.1in]{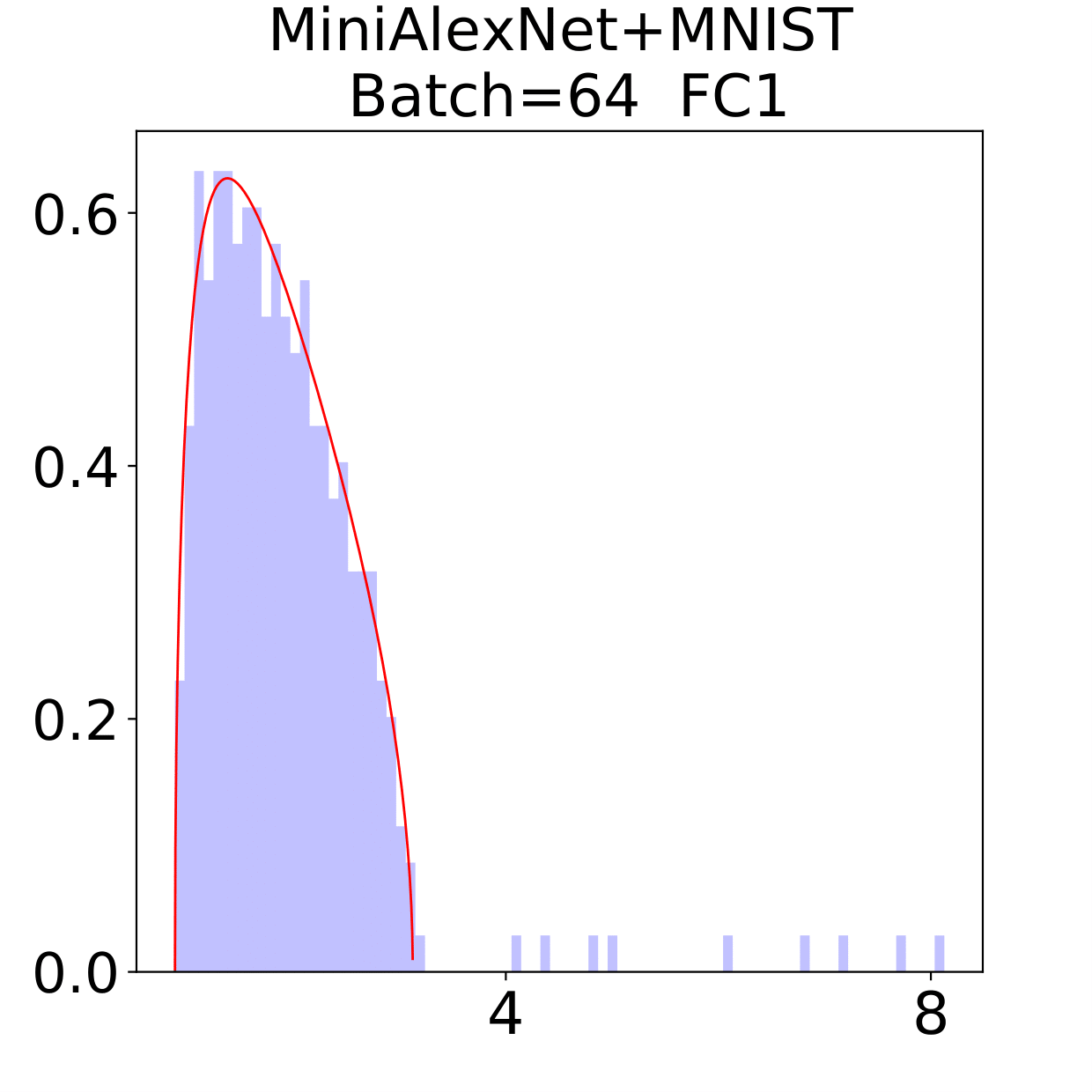}
\end{minipage}
}%
\subfigure[128]{
\begin{minipage}[t]{0.19\linewidth}
\centering
\includegraphics[width=1.1in]{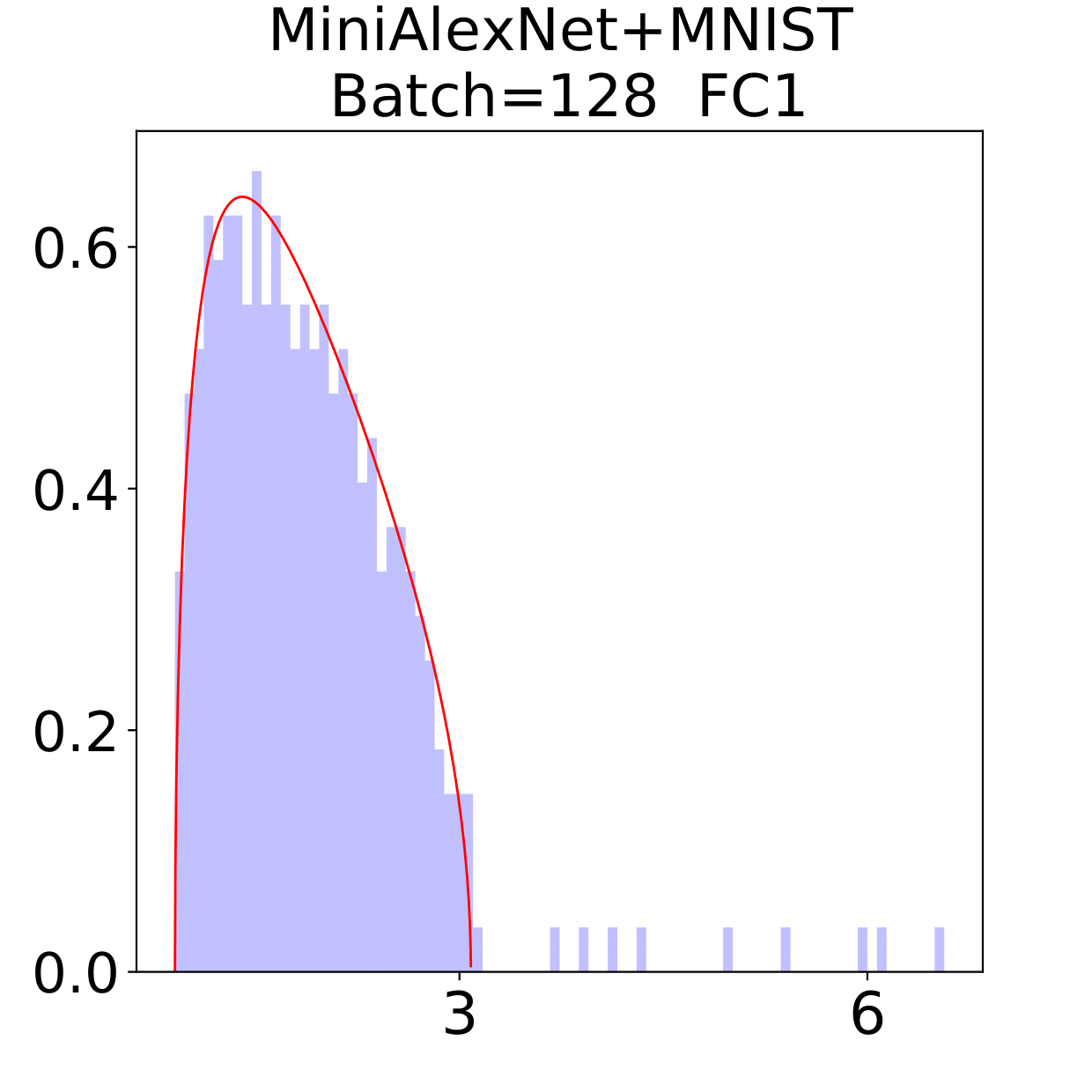}
\end{minipage}
}%
\subfigure[256]{
\begin{minipage}[t]{0.19\linewidth}
\centering
\includegraphics[width=1.1in]{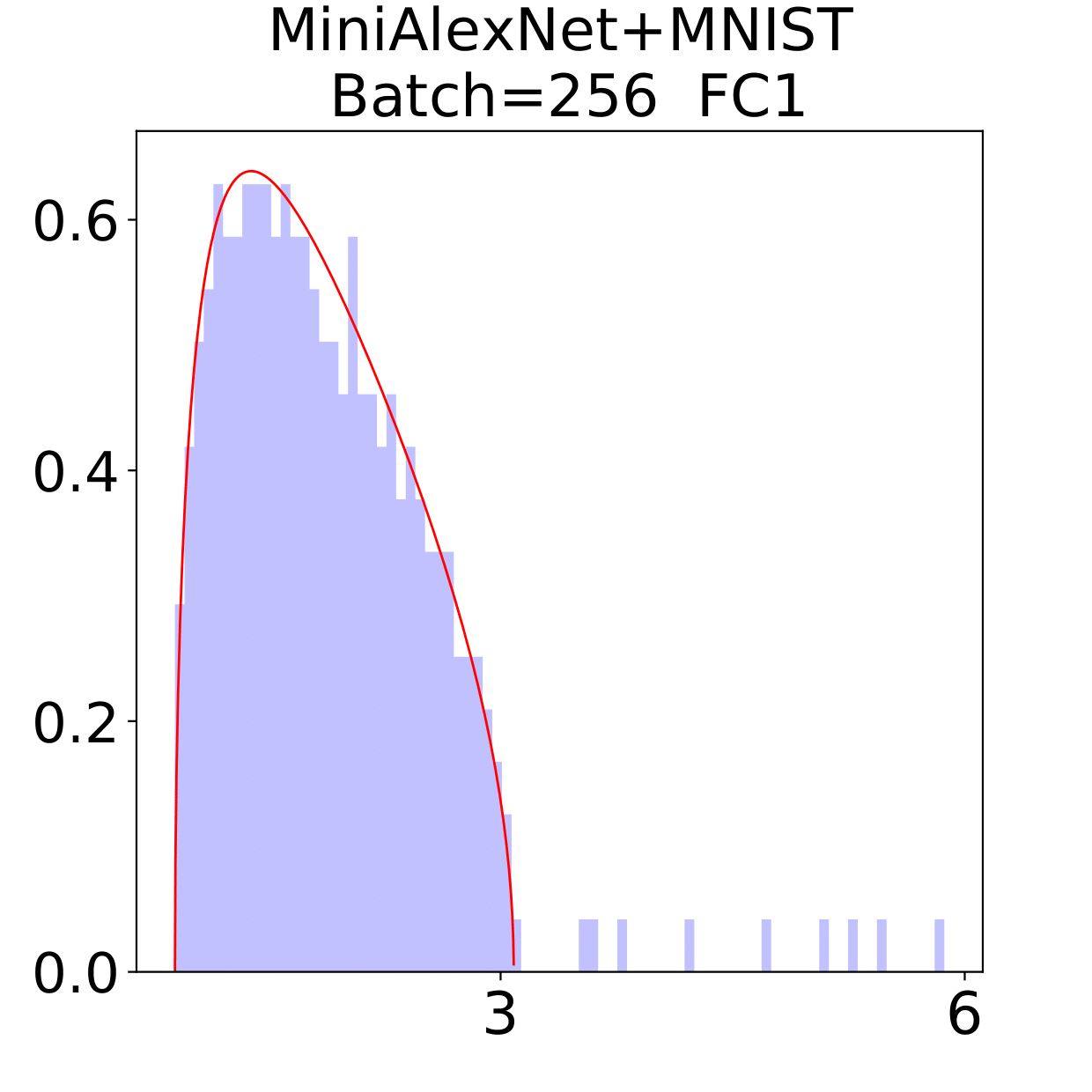}
\end{minipage}
}%

\subfigure[16]{
\begin{minipage}[t]{0.19\linewidth}
\centering
\includegraphics[width=1.1in]{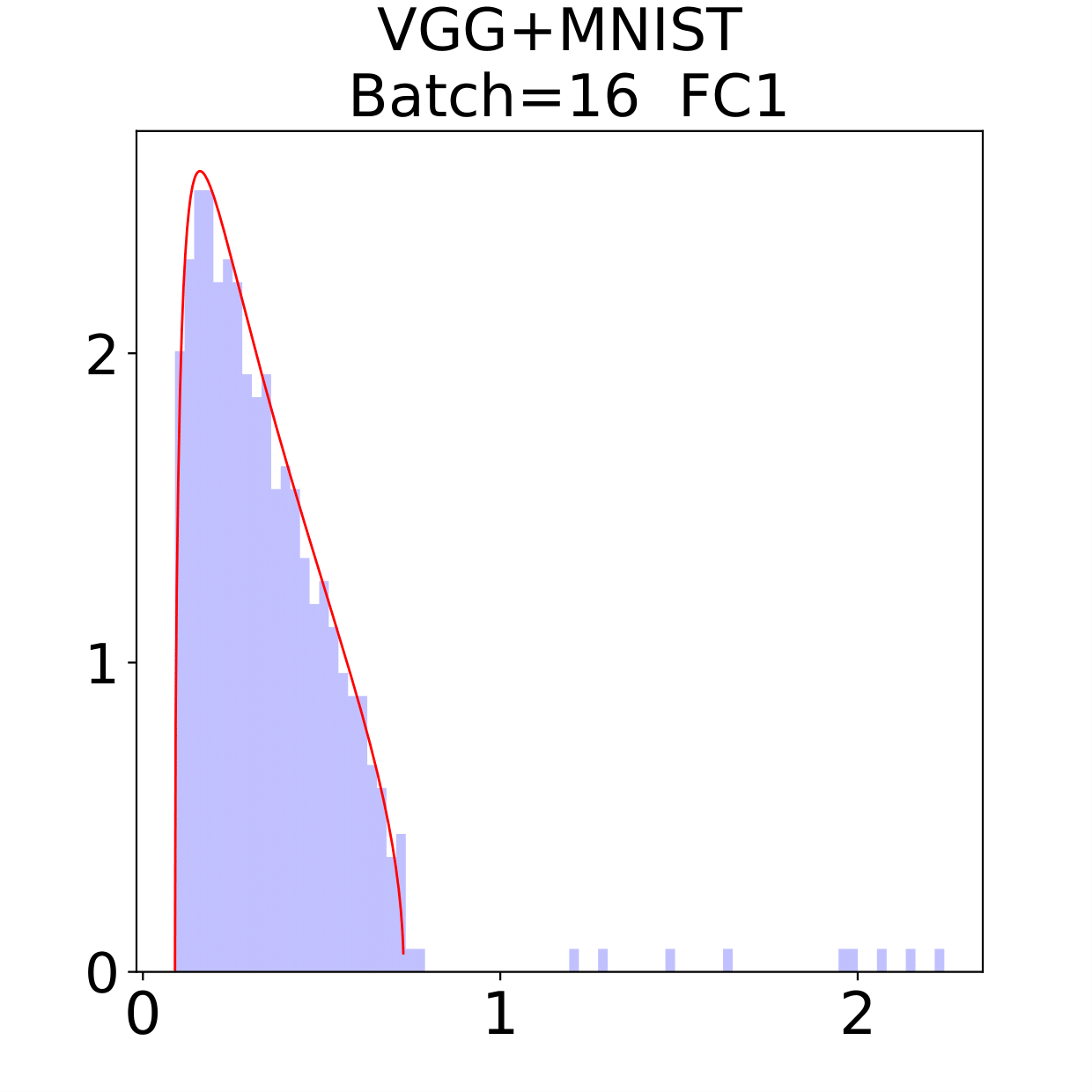}
\end{minipage}%
}%
\subfigure[32]{
\begin{minipage}[t]{0.19\linewidth}
\centering
\includegraphics[width=1.1in]{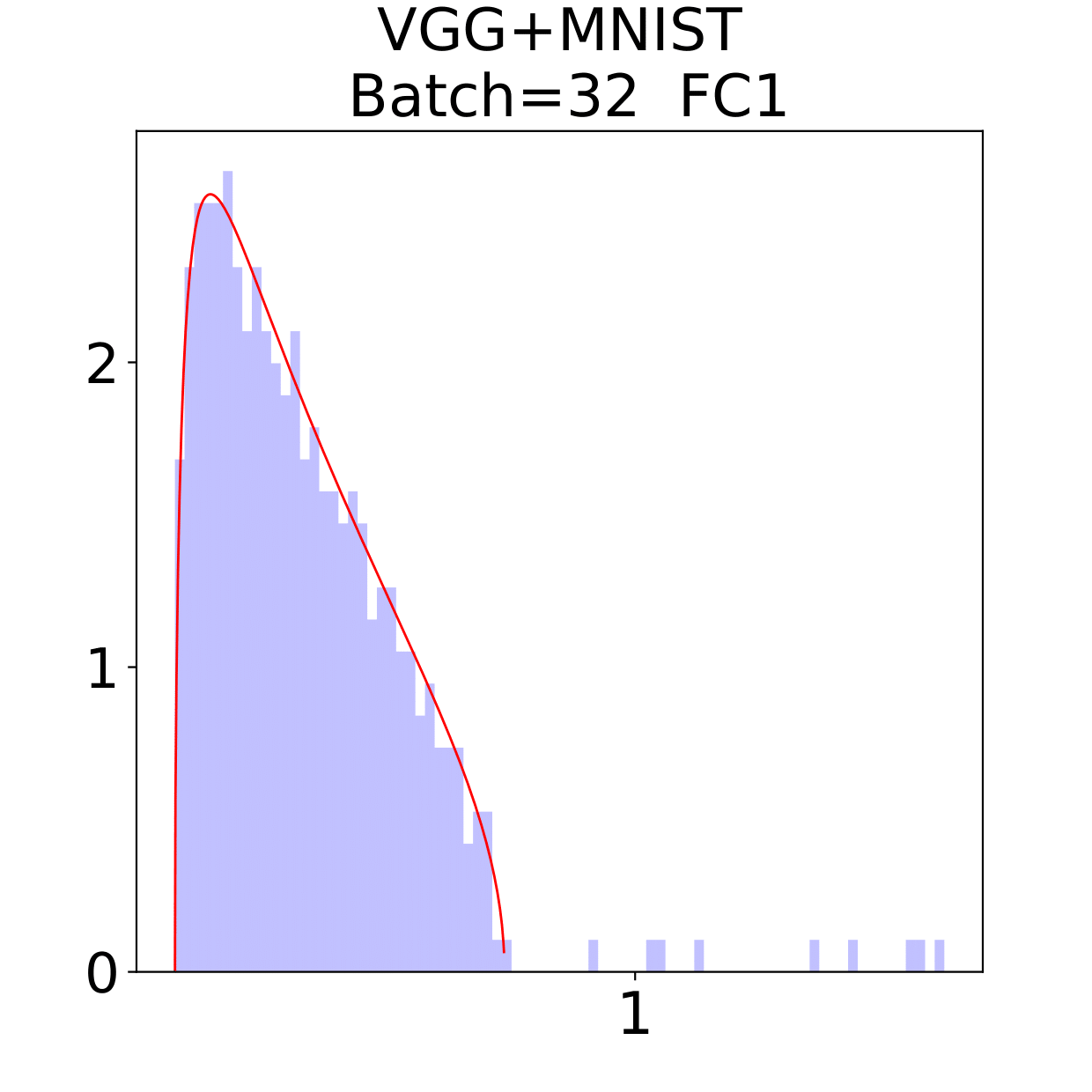}
\end{minipage}%
}%
\subfigure[64]{
\begin{minipage}[t]{0.19\linewidth}
\centering
\includegraphics[width=1.1in]{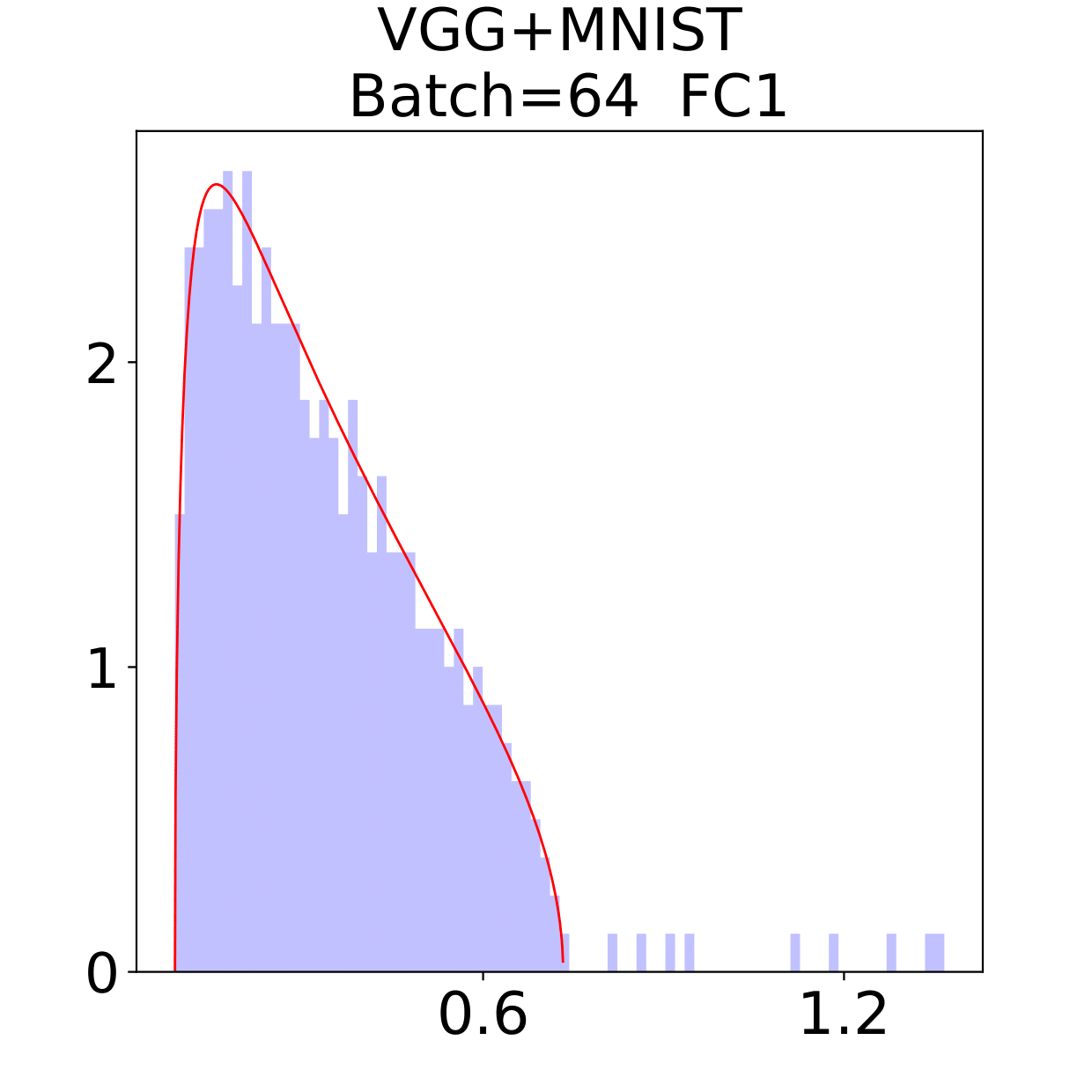}
\end{minipage}
}%
\subfigure[128]{
\begin{minipage}[t]{0.19\linewidth}
\centering
\includegraphics[width=1.1in]{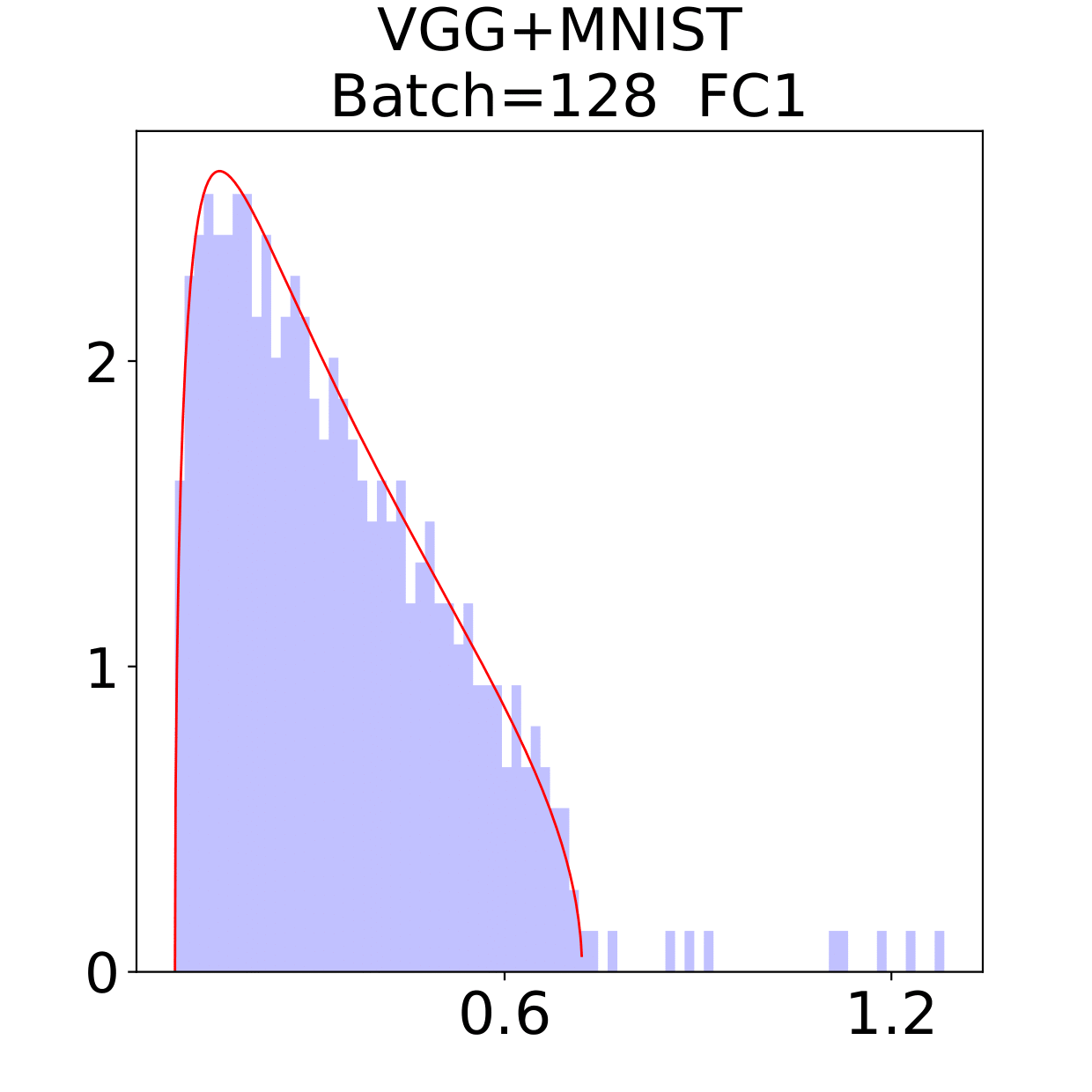}
\end{minipage}
}%
\subfigure[256]{
\begin{minipage}[t]{0.19\linewidth}
\centering
\includegraphics[width=1.1in]{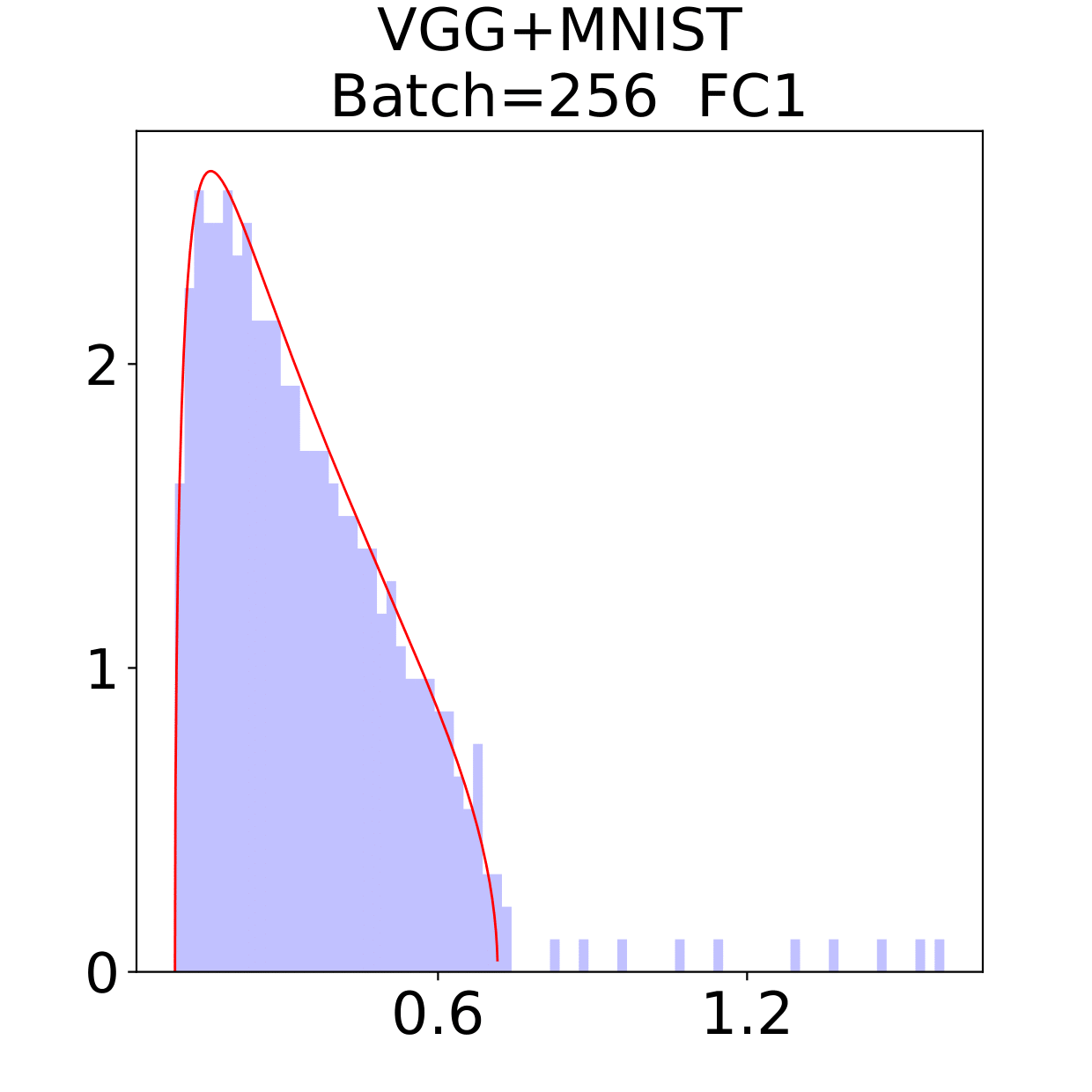}
\end{minipage}
}%

\centering
\caption{~~ Training on MNIST: Weight matrix spectra at final epoch 248. LeNet: (a)-(e); MiniAlexNet: (f)-(j); VGG : (k)-(o). Columns show experiments with different batch sizes.}
\label{RealDataMNIST}
\end{figure}

\begin{figure}[htbp]
\centering

\subfigure[16]{
\begin{minipage}[t]{0.19\linewidth}
\centering
\includegraphics[width=1.1in]{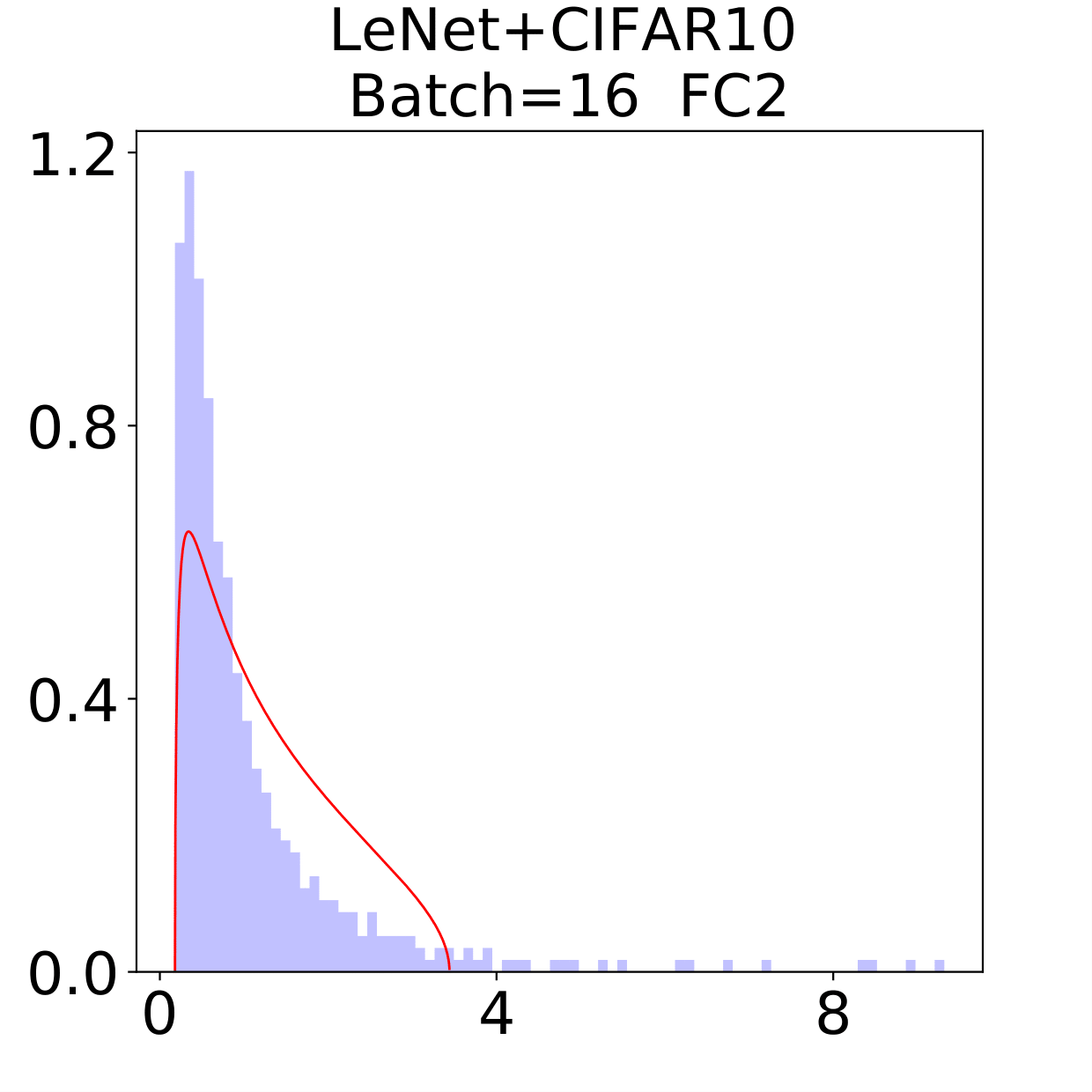}
\end{minipage}%
}%
\subfigure[32]{
\begin{minipage}[t]{0.19\linewidth}
\centering
\includegraphics[width=1.1in]{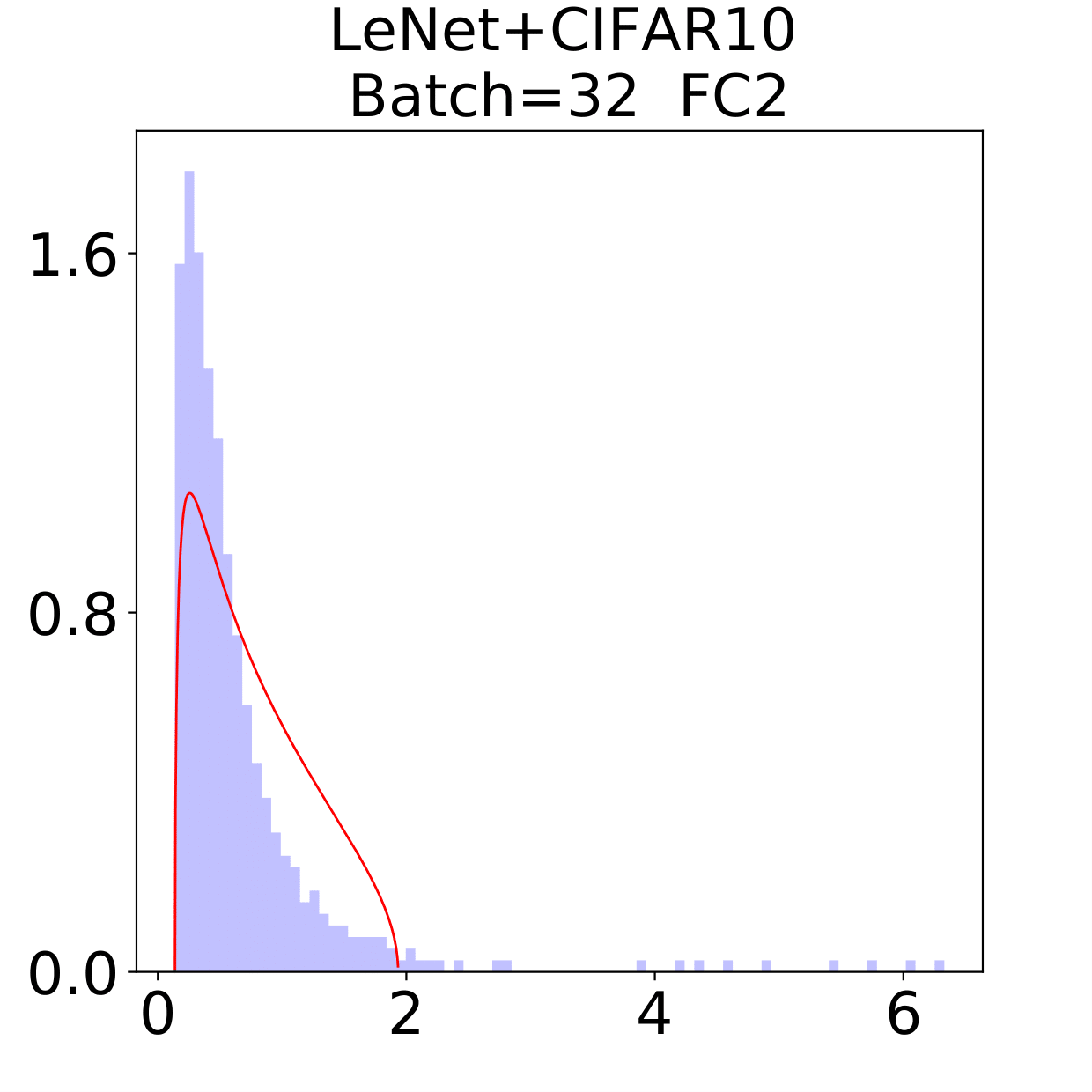}
\end{minipage}%
}%
\subfigure[64]{
\begin{minipage}[t]{0.19\linewidth}
\centering
\includegraphics[width=1.1in]{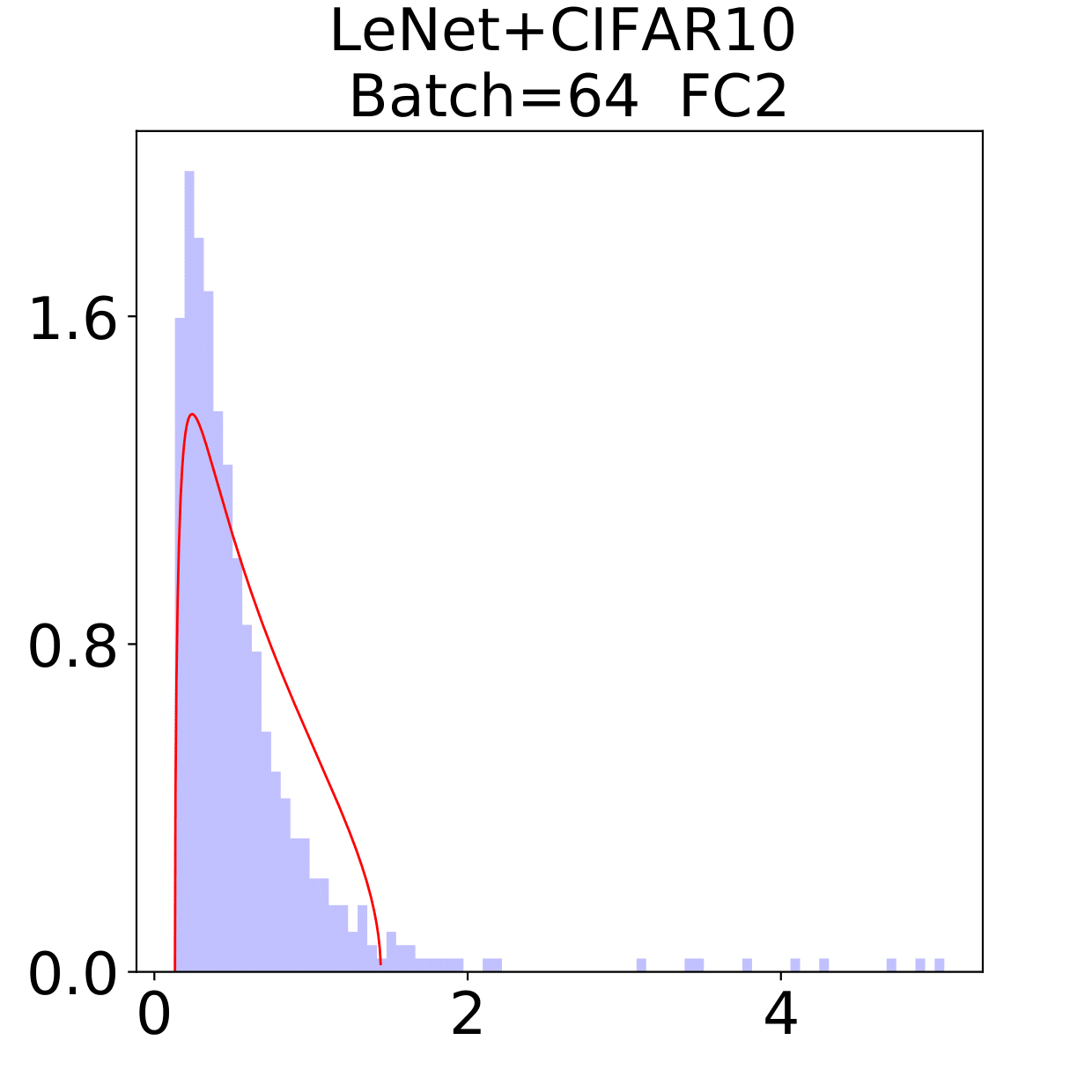}
\end{minipage}
}%
\subfigure[128]{
\begin{minipage}[t]{0.19\linewidth}
\centering
\includegraphics[width=1.1in]{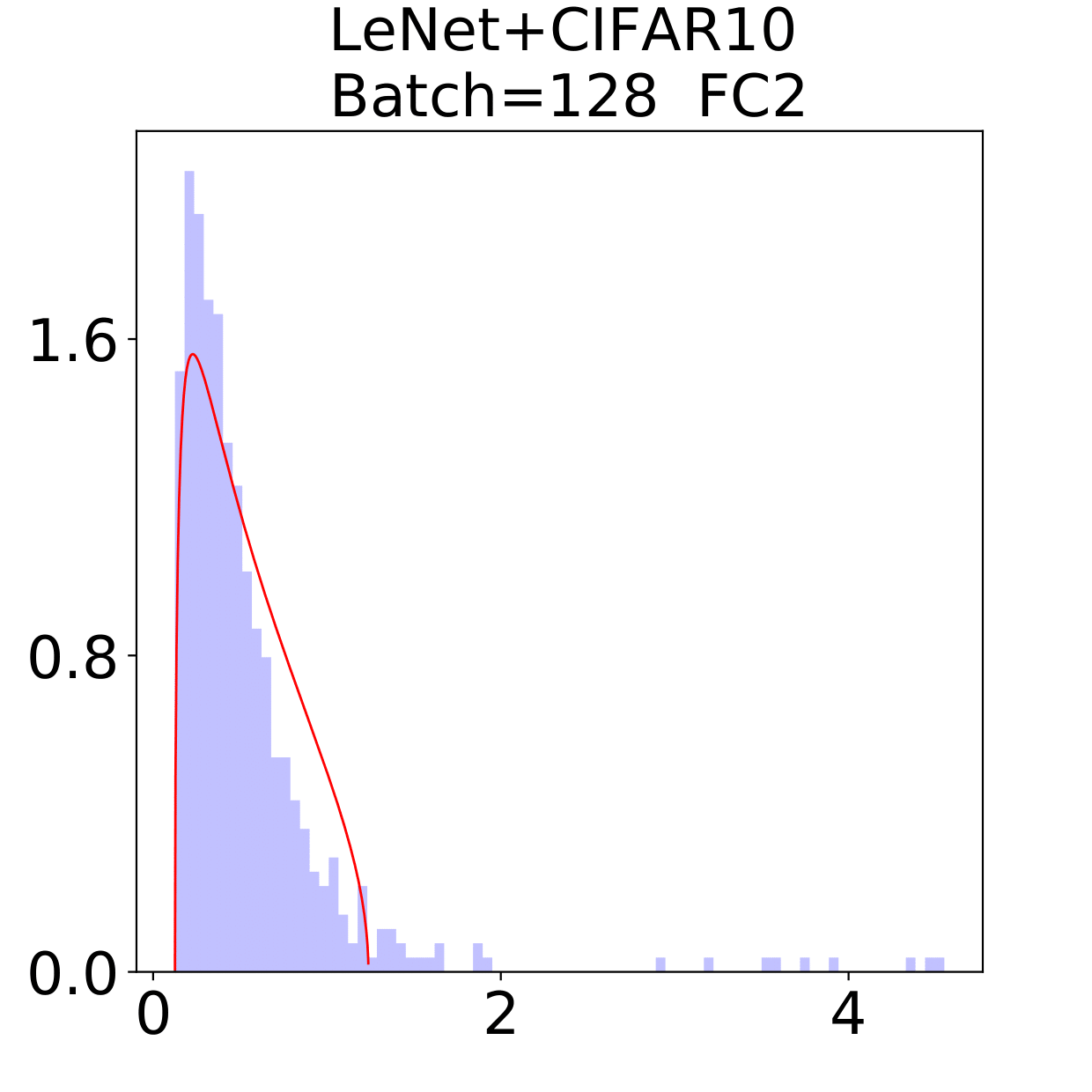}
\end{minipage}
}%
\subfigure[256]{
\begin{minipage}[t]{0.19\linewidth}
\centering
\includegraphics[width=1.1in]{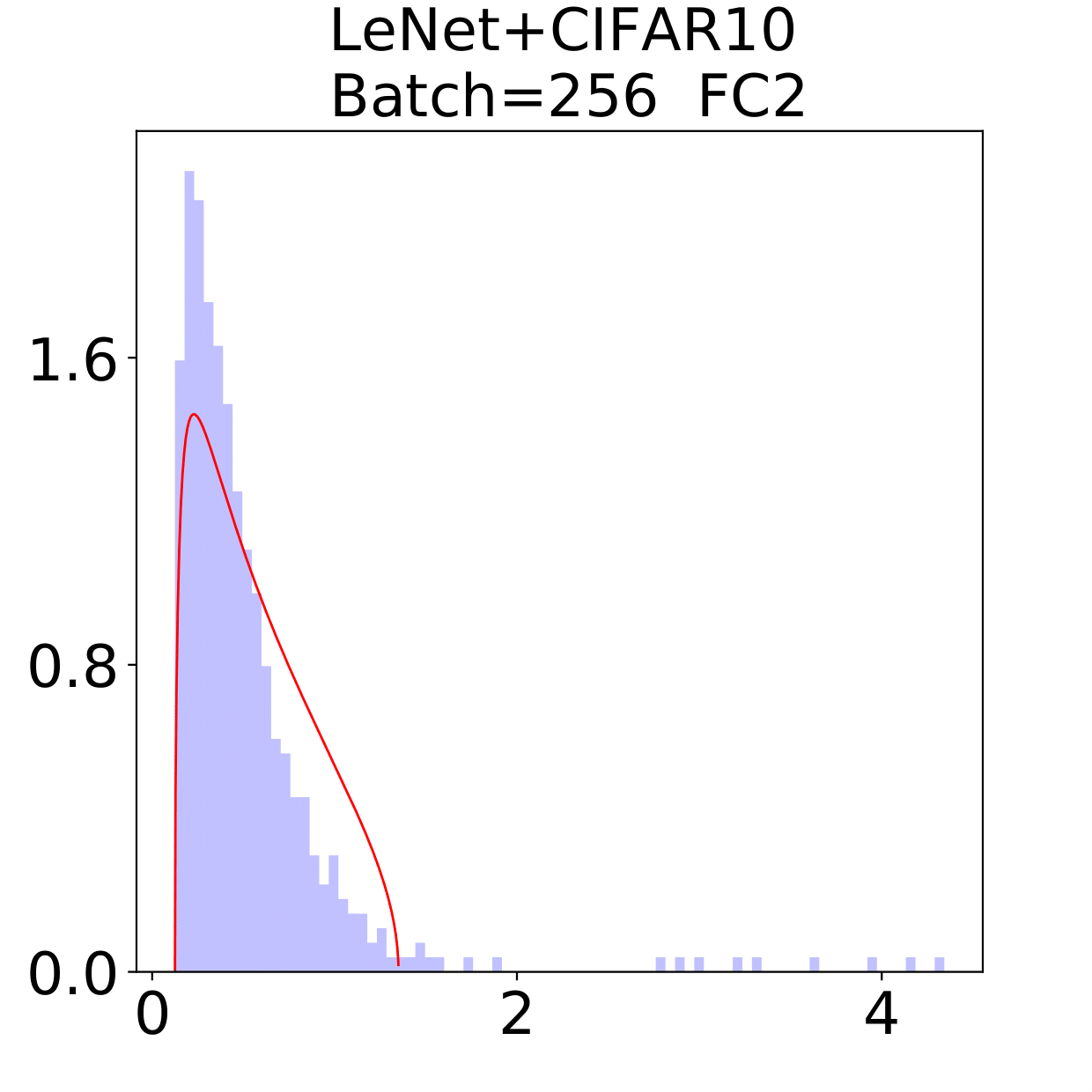}
\end{minipage}
}%

\subfigure[16]{
\begin{minipage}[t]{0.19\linewidth}
\centering
\includegraphics[width=1.1in]{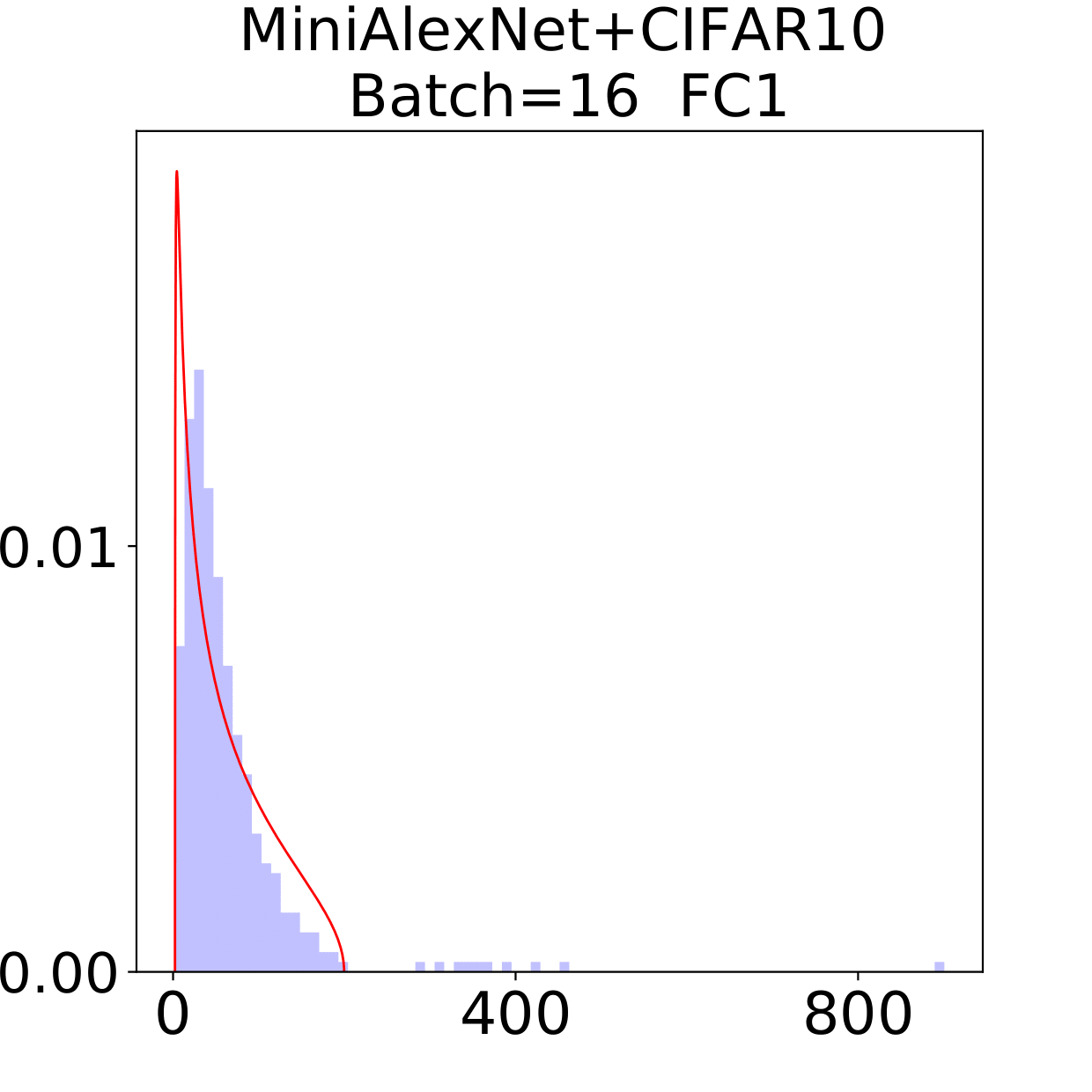}
\end{minipage}%
}%
\subfigure[32]{
\begin{minipage}[t]{0.19\linewidth}
\centering
\includegraphics[width=1.1in]{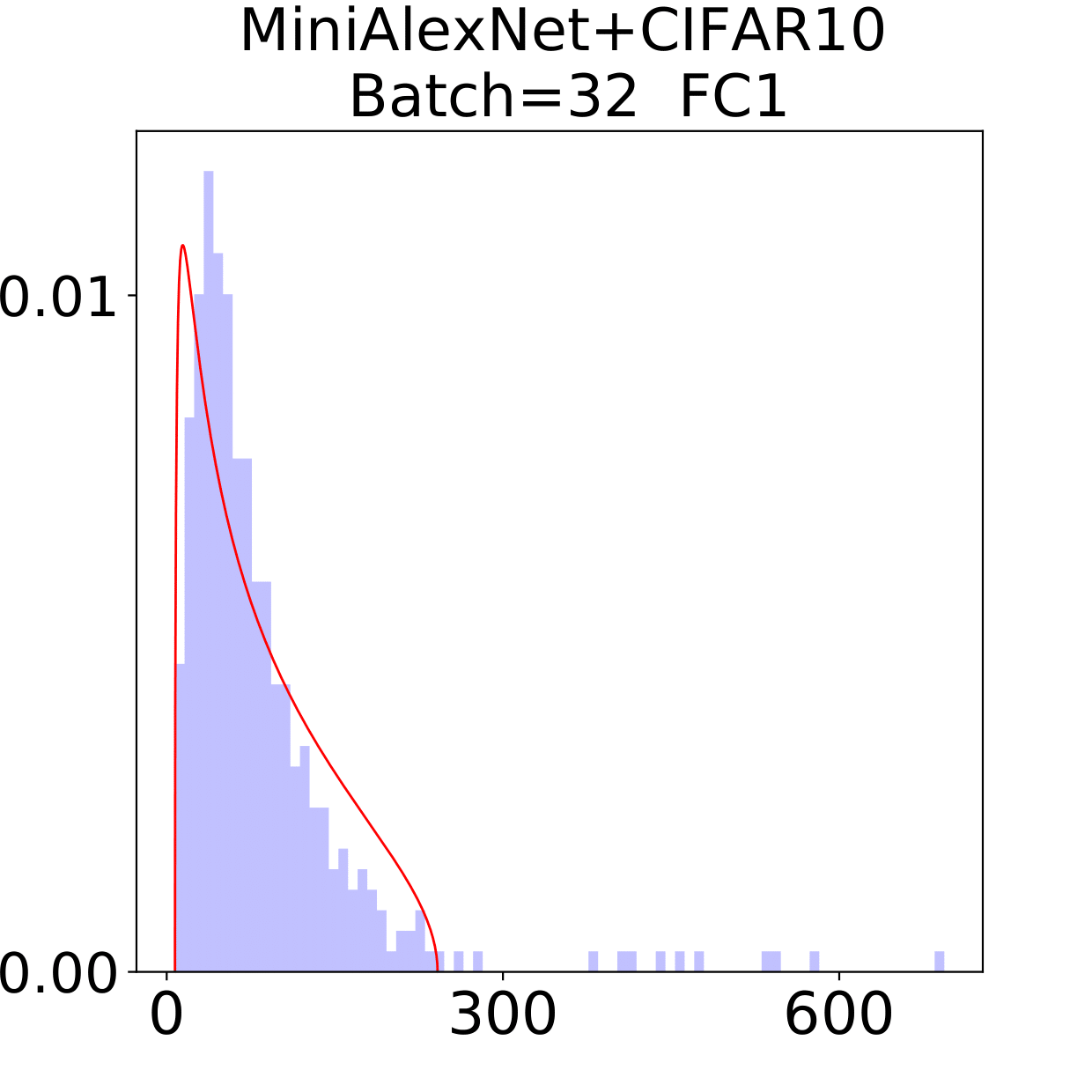}
\end{minipage}%
}%
\subfigure[64]{
\begin{minipage}[t]{0.19\linewidth}
\centering
\includegraphics[width=1.1in]{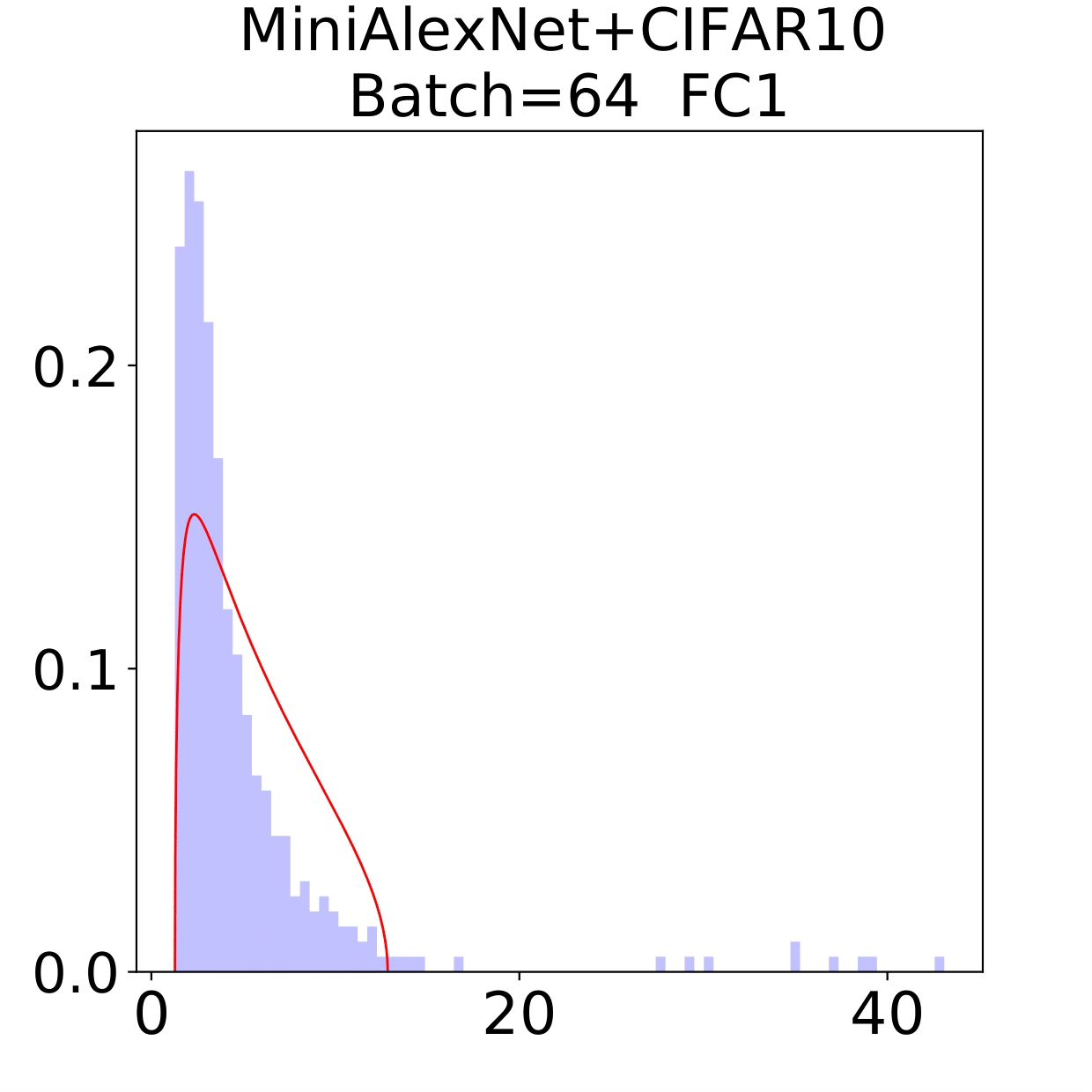}
\end{minipage}
}%
\subfigure[128]{
\begin{minipage}[t]{0.19\linewidth}
\centering
\includegraphics[width=1.1in]{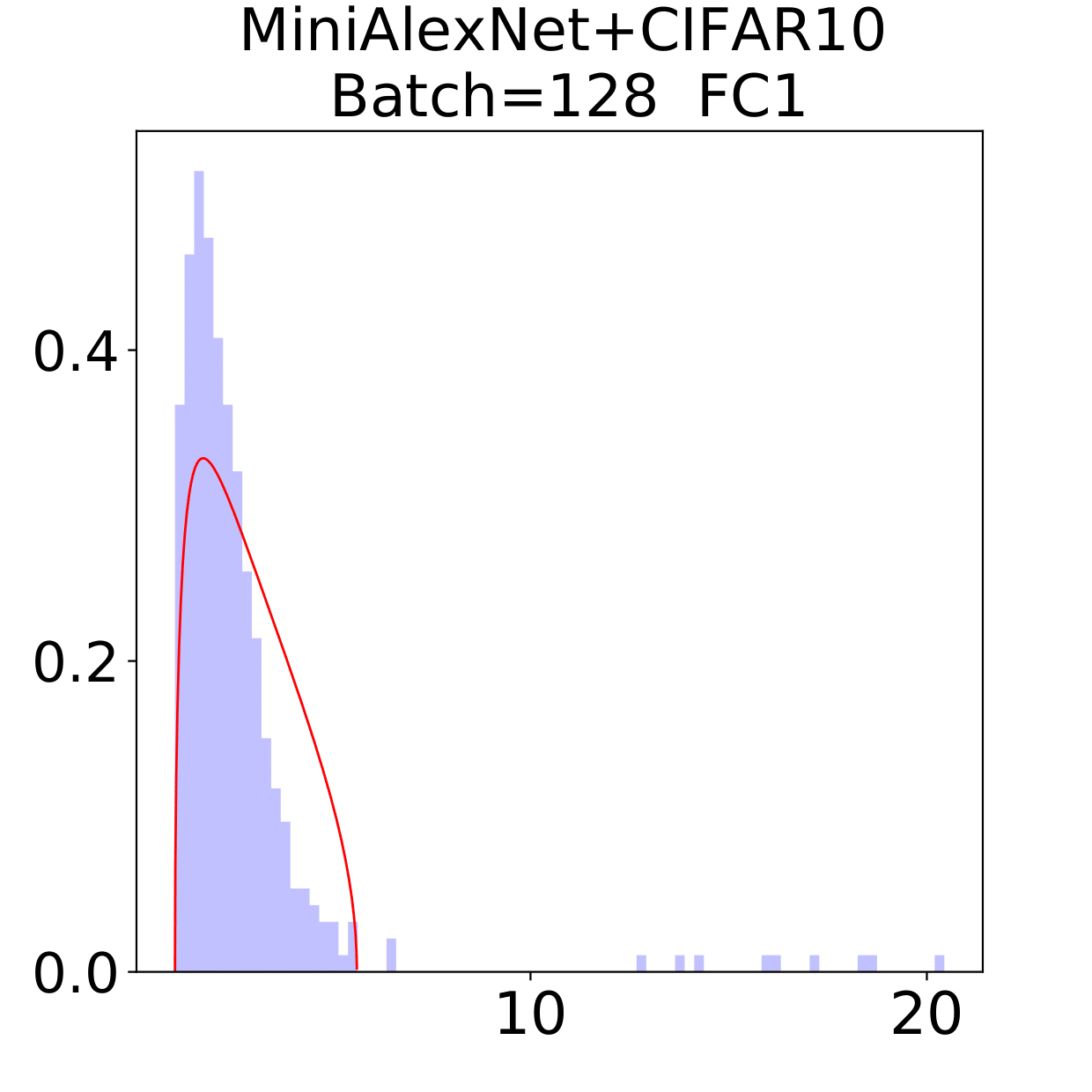}
\end{minipage}
}%
\subfigure[256]{
\begin{minipage}[t]{0.19\linewidth}
\centering
\includegraphics[width=1.1in]{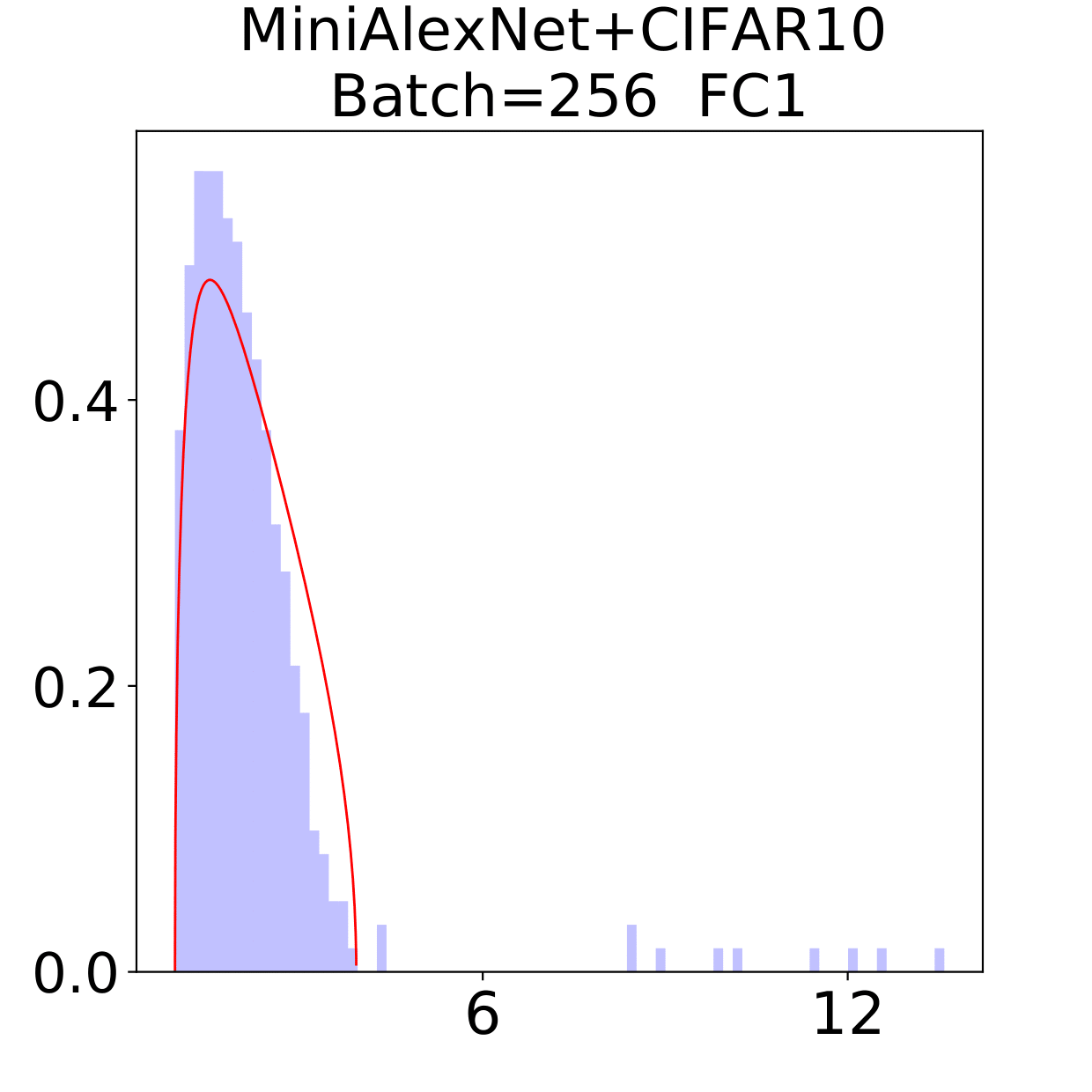}
\end{minipage}
}%

\subfigure[16]{
\begin{minipage}[t]{0.19\linewidth}
\centering
\includegraphics[width=1.1in]{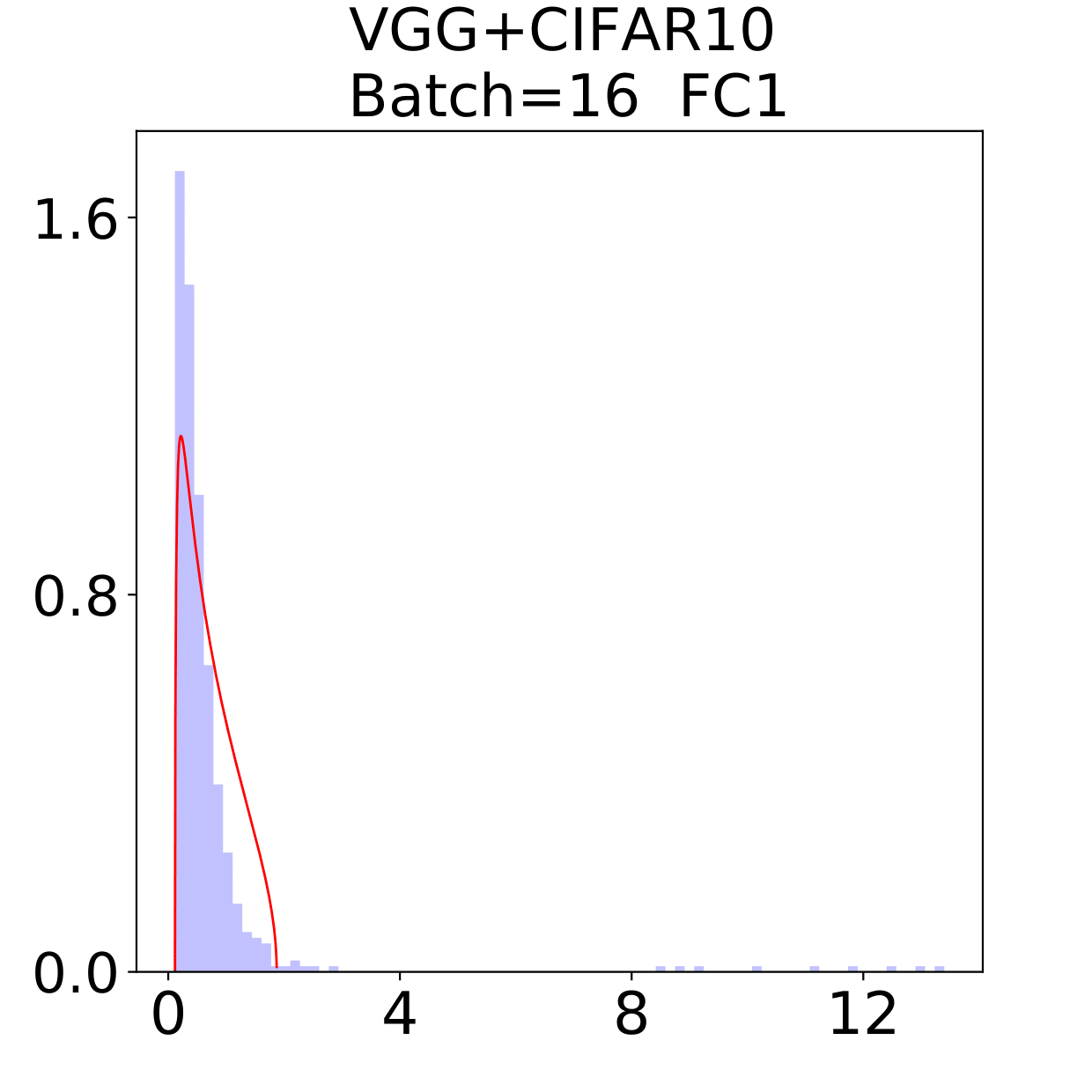}
\end{minipage}%
}%
\subfigure[32]{
\begin{minipage}[t]{0.19\linewidth}
\centering
\includegraphics[width=1.1in]{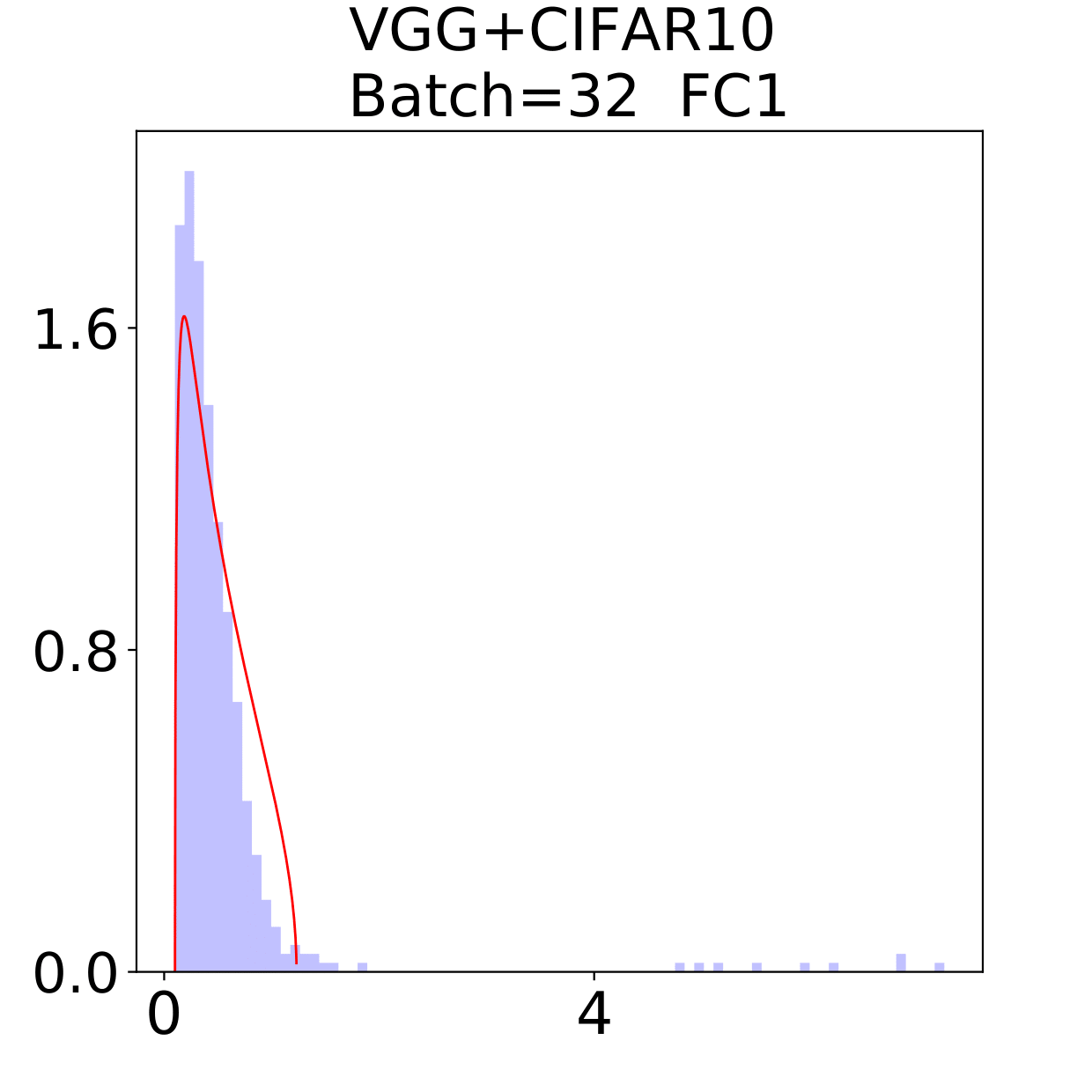}
\end{minipage}%
}%
\subfigure[64]{
\begin{minipage}[t]{0.19\linewidth}
\centering
\includegraphics[width=1.1in]{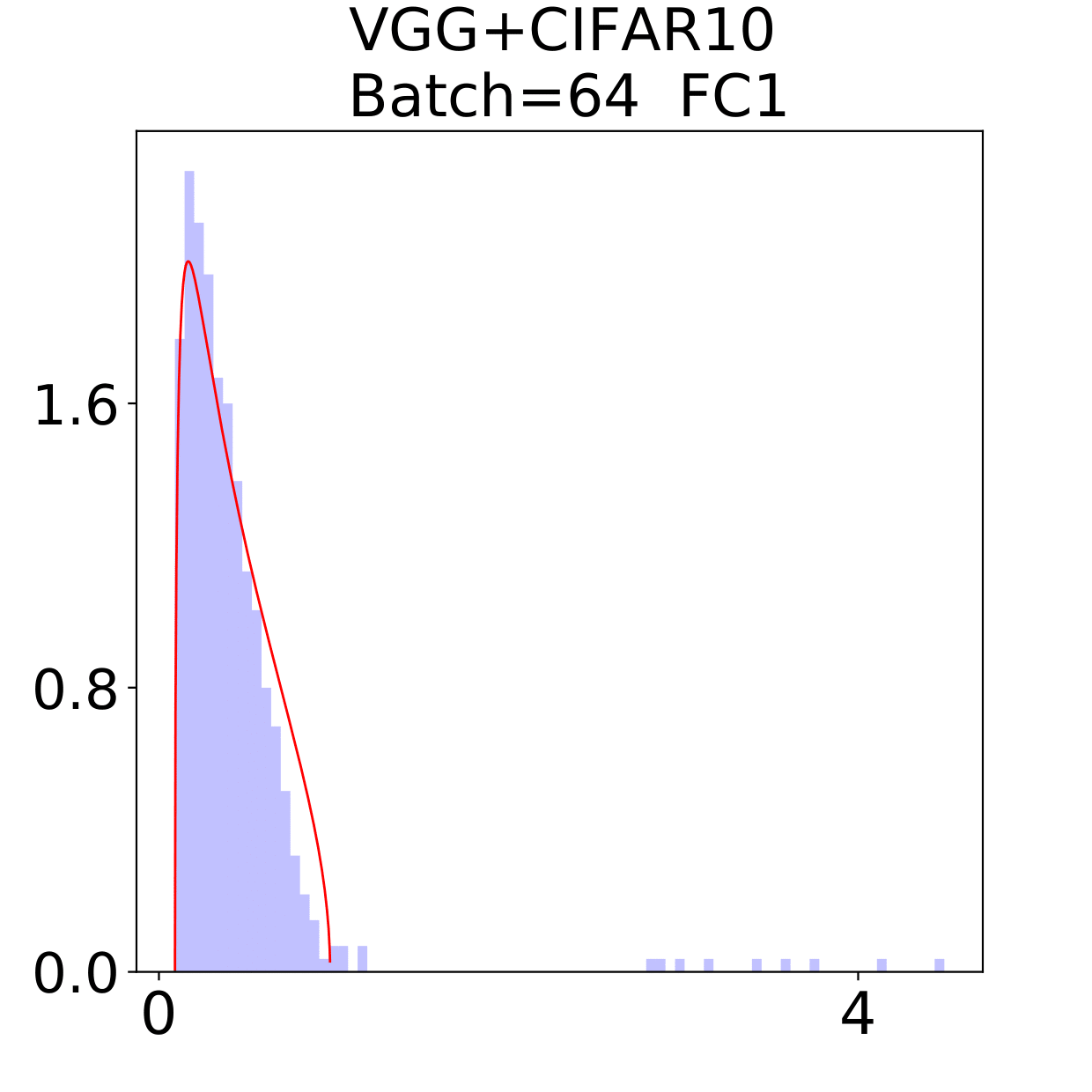}
\end{minipage}
}%
\subfigure[128]{
\begin{minipage}[t]{0.19\linewidth}
\centering
\includegraphics[width=1.1in]{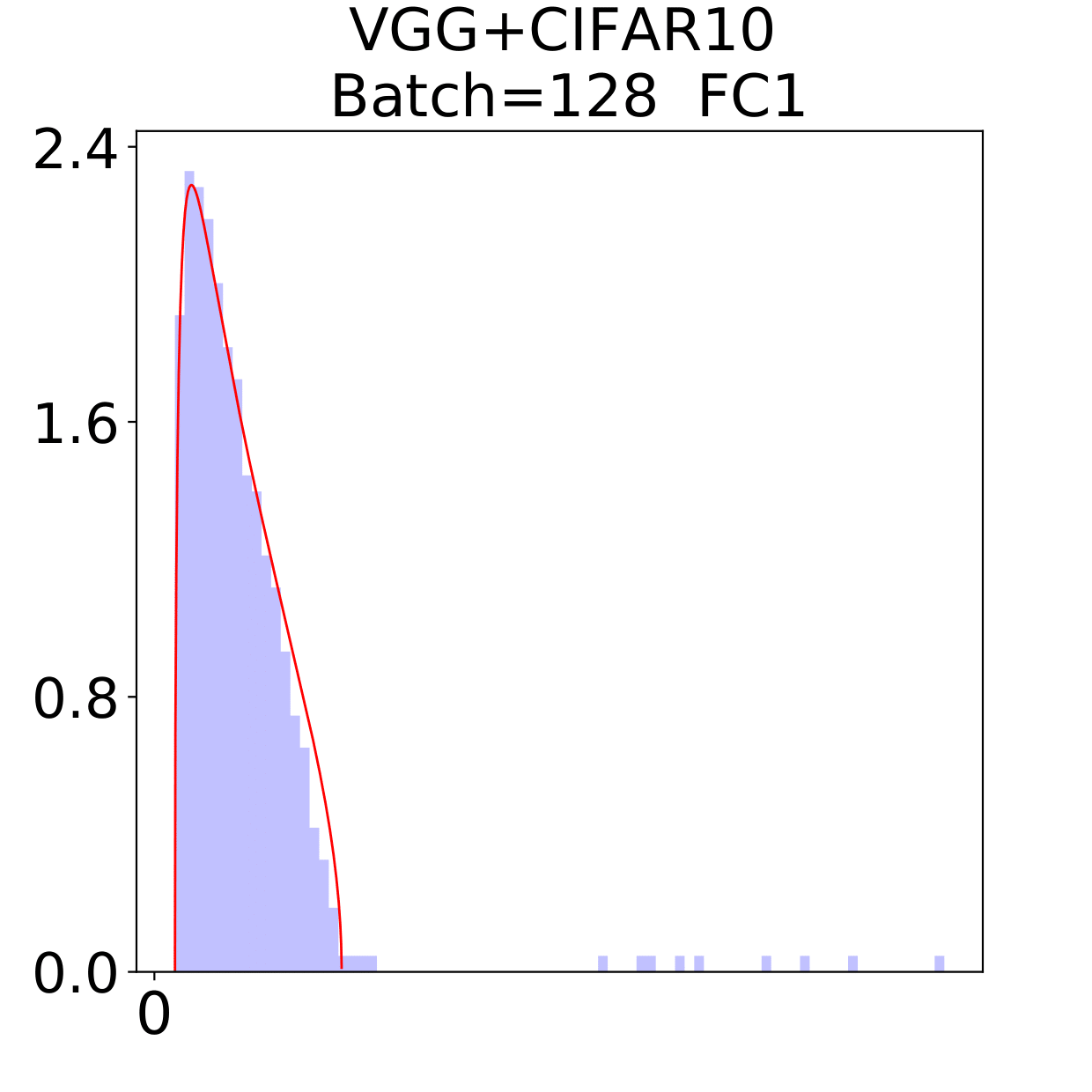}
\end{minipage}
}%
\subfigure[256]{
\begin{minipage}[t]{0.19\linewidth}
\centering
\includegraphics[width=1.1in]{picture/RealData/VGGMNIST/FC1_256.png}
\end{minipage}
}%

\centering
\caption{~~ Training on CIFAR10: Weight matrix spectra at final epoch 248. LeNet: (a)-(e); MiniAlexNet: (f)-(j); VGG : (k)-(o). Columns show experiments with different batch sizes.}
\label{RealDataCIFAR}
\end{figure}

In this section, we check the weight matrices spectra on various NNs with two different data sets, see whether the complex features in data sets have great impact on the formation of HT. At the final epoch, the ESDs in LeNet, MiniAlexNet and VGG essentially reflect the  characteristic features of the data in MNIST and CIFAR10.  We concentrate on weight matrices in Fully connected layers and display the results in Figure \ref{RealDataMNIST}-\ref{RealDataCIFAR}.

The weight matrix has the structure $2450\times 500$ in LeNet, $4096\times 384$ in MiniAlexNet and $2048\times 500$ in VGG. As figures \ref{RealDataMNIST} and \ref{RealDataCIFAR} show, the spectra are very different under the two training data sets on all neural networks. The type detection method is illustrated in section \ref{sec:threetypes}.  In LeNet+MNIST, the spectra are BT type except batch size 256 with LT type, these BT types are slightly different from MP Law; In LeNet+CIFAR10, the spectra are also BT type but similar to HT except batch size 16. In MiniAlexNet+MNIST, the spectra are LT type except batch size 16 and 32 with BT type; In MiniAlexNet+CIFAR10, the spectra are HT and BT which visibly different from MP Law. In VGG+MNIST, the spectra are all LT type; In VGG+CIFAR10, the spectra transit from BT (visibly similar to HT) to LT with increment of batch size.

\newgeometry{top=2.5cm}

Although different hyper parameters and model architectures may have great impact on weight matrices spectra type, we could not ignore the impact from data itself. Different data classification difficulty also has great impact on the weight matrices spectra type. Compared with Figure \ref{RealDataMNIST} and \ref{RealDataCIFAR}, the NNs trained on CIFAR10 are more likely to have HT, nevertheless the increment of batch size makes the spectra gradually approximate to the LT type. The complex features in CIFAR10 could increase the data classification difficulty and bring in HT, and the weight matrices spectra display HT shapes, which implicitly reflects the status of whole training procedure. Especially when HT emerges, the indication of some implicit regularization happened gives us new insights that spectra could be regarded as training information encoder. According to this, we propose spectral criterion for early stopping in section \ref{sec:specCriterion}.

\newgeometry{top=2.5cm}

\section{Spectral Criterion}
\label{sec:specCriterion}

As a regularization technology in Deep Learning, early stopping is
adopted to improve generalization accuracy of a DNN.
People may use
testing data set to obtain convenient stopping time in practice, but when we
model the data set, it is a trade off to separate data set into
training and testing. Sometimes as \cite{martin2021predicting} pointed out, it is more expensive to acquire the testing
data set.
There are also situations where practioners of Deep Learning are laid to use
pre-trained and existing DNNs without access to test data.

\vspace{2pt}

So an important question we address here is:
Without any testing data set, shall we early stop or not? And  how to define an  early stopping time? 

\vspace{2pt}

The spectra of weight matrices encode information during the training time, according to this we construct an application to guide the early stopping through a {\bf Spectral Criterion}. Precisely, we introduce a distance between an observed weight matrix spectrum and the reference MP Law. When this distance is judged large enough, we obtain evidence for the formation of a HT or BT type spectrum, thus the implicit regularization in the DNN leads to the decision of stopping the training process. Note that this spectral criterion for early stopping does not need any test data.

We now describe this spectral criterion in more detail.  Consider a
$n\times N$ ($n\le N$) weight matrix $W$ and let
$X_1,X_2,...,X_n$ be the $n$ non-zero eigenvalues of the square matrix
$WW^T$. (These are also, by definition, the squares of the singular values of the matrix $W$. The initialization of $W$ has been rescaled with $1/\sqrt{N}$.) We then construct a histogram estimator $\hp_M(x)$ for the joint
density of the eigenvalues using $M$ bins.   
Next, let $p_{c,\sigma^2}(x)$ be the reference Mar\v{c}enko-Pastur density (Appendix \ref{sec: RMT}) 
depending on a scale parameter $\sigma^2$ and a shape parameter
$0< c<1$, with a compact support $[a,b]$  ($0<a<b$). In practice, the parameters $c$ and $\sigma^2$ in the reference MP density $p_{c,\sigma^2}(x)$ are also estimated by using $X_1,...,X_n$. This leads to an estimated MP density function $p_{\hc,\hsigma^2}(x)$.
The estimation of distance between the distribution of the $n$ eigenvalues and the
MP density is defined as $L_1$ distance
\begin{align}
\label{eq: estimateDistance}
    \hs_n  =\int_a^b |\hp_M(x)-p_{\hc,\hsigma^2}(x)|dx.
\end{align}  
Under the null hypothesis that the eigenvalues $\{X_i\}$ follow the MP
law, we have a precise rate for $\hs_n\to0$, which leads to our spectral criterion.

\vskip 3mm

\noindent {\bf Spectral  criterion.} \quad  Set $M=2\lfloor n^{\frac13}\rfloor$ and consider a threshold value $s_*=C*\sqrt{\log  n}/n^{\frac13}$  with $C=0.4$, 
For each training epoch,  calculate
$\hs_n$ in equation (\ref{eq: estimateDistance}) by the {\bf Algorithm} in
Appendix~\ref{sec:algo}. The training is stopped if $\hs_n > s_*$.
\\[1mm]
(To gain more robustness in this stopping procedure, in all the
experiments, we will stop the training at three consecutive epochs
where $\hs_n > s_*$ happen (instead of at the first such epoch).)

\vskip 3mm

The whole details on the determination of the distance value $\hs_n$ and
the threshold value $s_*$ are given later in Section~\ref{sec:specCriterion-details}. One may ask {\bf why the appearance of a HT spectrum is a good early stopping time?} Let us elaborate more on the important reasons for the early stopping
rule above based on the appearance of a HT spectrum.   
\cite{martin2021predicting} mentioned that MP Law spectra could not
evaluate the performance of the trained model, but Heavy Tail spectra
could. The Heavy Tail spectra may correspond to better or worse test performance. From information encoder perspective, we argue that the emergence of Heavy Tail or Rank Collapse in weight matrices could be viewed in two ways:
\begin{itemize}
  \item  Indication of the poor quality in the training data or the poor ability in the whole system: in synthetic data experiments, the poor training data or system quality will lead to instability or overfitting during the whole training dynamics. So the emergence of Heavy Tail can be treated as an \textbf{alarm} for these hidden and problematic issues in the network.
  \item  Indication of a regularized structure that has acquired considerable information from the training data: from another point of view, the HT phenomenon is far from the initial MP Laws introduced with random weight initialization. Its emergence can be viewed as an indication of a well-trained structure that has already captured sufficient information from the input data. Such structure will somehow ensure the testing accuracy of the whole system, and additional training will not bring much improvement.
\end{itemize}

The spectra criterion is validated in both synthetic and real data experiments. Evidence for this {\bf spectral criterion} is developed in details with extensive experimental results in Sections~\ref{sec: earlystopsimu}
and \ref{sec: earlystopreal}. In synthetic data experiments, the spectral criterion provides an early stopping epoch where the testing accuracy is much higher than
the final testing accuracy, even when the training accuracy is still increasing. In real data experiments, the spectral criterion could also offer high-quality stopping time, ensuring testing accuracy and cutting off a large unnecessary training time.


\subsection{Technical details of the spectral criterion}
\label{sec:specCriterion-details}

Consider  $n$ data points 
$X_1,X_2,...,X_n$, supported on an interval $[a,b]$, with
$0<a<b$. Consider a mesh net on the interval on $M$ bins of binsize
$(b-a)/M$, 
$$
B_j=\left(  a+ (j-1)\frac{b-a}{M}, a+ j\frac{b-a}{M}  \right],\quad  1\le j\le M.
$$
The histogram  estimator for the density function of the data is 
$$
\hp_M(x)=\frac{M}{n(b-a)}\sum\limits_{i=1}^nI(X_i\in B(x)),
$$
$B(x)$ is the bin $x$ belongs to.

With respect to Random Matrix Theory Results given in Appendix \ref{sec: RMT}, the density function of the 
standard MP Law $MP_{c,\sigma^2}$ is
$$
p_{c,\sigma^2}(x)=MP_{c,\sigma^2}(x)=\frac{1}{2\pi c\sigma^2x}\sqrt{(b-x)(x-a)},
$$
with $a=\sigma^2(1-\sqrt{c})^2$ and $b=\sigma^2(1+\sqrt{c})^2$. 
We thus use the following $L_1$ distance between the two density
functions to measure the departure of the data points $\{X_i\}$ from
the MP law:
\begin{align}
\label{eq: Distance}
s_n=\int_{a}^{b}|\hp_M(x)-p_{c,\sigma^2}(x)|dx.
\end{align}

We have the following estimated rate for $s_n$ under the null hypothesis
that the data points follow the MP-law. 
\begin{proposition}
\label{prop1}
Suppose $\{X_i\}_{i=1}^n$ are generated independently from $p_{c,\sigma^2}(x)=MP_{c,\sigma^2}$, then the distance in eq\ref{eq: Distance} satisfies
$$
s_n=O_p(\frac{1}{M}+\sqrt{\frac{M\log n}{n}}).
$$
\end{proposition}

$O_p$ is the standard convergence notation in probability. The proof is given in Appendix~\ref{sec:proof-prop1}.
Due to the fact that MP density $p_{c,\sigma^2}(x)$ has unbounded derivatives at its
edge points $\{a, b\}$, the above estimated rate is obtained via a special adaptation of the existing rate from the literature.

In practice, we do not know the parameters $c$ and $\sigma^2$ of the
reference MP density $p_{c,\sigma^2}(x)$.  Then we use the observed extreme  statistics
$\hat a=X_{(1)}$,  and $\hat b=X_{(n)}$ to estimate $a$ and $b$,
respectively. These lead to corresponding estimates $\hat c$ and
$\hat\sigma^2$  for the parameters
$c$ and $\sigma^2$, respectively.  
The MP density function with estimated parameters is then 
$$
p_{\hc,\hsigma^2}(x)=\frac{1}{2\pi \hat c\hat\sigma^2x}\sqrt{(\hat b-x)(x-\hat a)}
I(\hat a\leq x\leq \hat b).
$$
Finally, the  $L_1$ distance between the data set $\{X_1,\ldots,X_n\}$
and the MP law is  estimated
by 
\begin{equation}
  \hs_n=\int_{a}^{b}|\hp_M(x)-p_{\hc,\hsigma^2}(x)|dx.\nonumber
\end{equation}
The following proposition guarantees a  convergence rate for the
estimator $\hs_n$,

\begin{proposition}
\label{prop2}
As $\hs_n$ defined above, we have
\begin{equation}\label{eq:rate-snhat}
  \hs_n=O_p(\frac{1}{n^{1/3}}+\frac{1}{M}+\sqrt{\frac{M\log n}{n}}).
\end{equation}
\end{proposition}

The proof of the proposition is given in~Appendix~\ref{sec:proof-prop2}.

The proposition is next used to define a rejection region for the null
hypothesis.
Consider  $M=O(n^{\frac{1}{3}})$.   From \eqref{eq:rate-snhat}, under
the null hypothesis, $\hs_n $ will converge to zero at the 
optimal rate of   $O_P(\sqrt{\log n}/n^{\frac13})$.   In contrast,
under a  deviation of ESDs in weight matrices such as emergence of
Heavy Tails,  $\hs_n$ will no longer tend to $0$. This result, combined with the previous finding of  data-effective
regularization in weight matrices  spectra, permits the definition of
the spectral criterion for early stopping of the training
process in a DNN introduced in section \ref{sec:specCriterion}.

\begin{figure}[htbp]
\centering
\subfigure{
\begin{minipage}[t]{0.2\linewidth}
\centering
\includegraphics[width=1.0in]{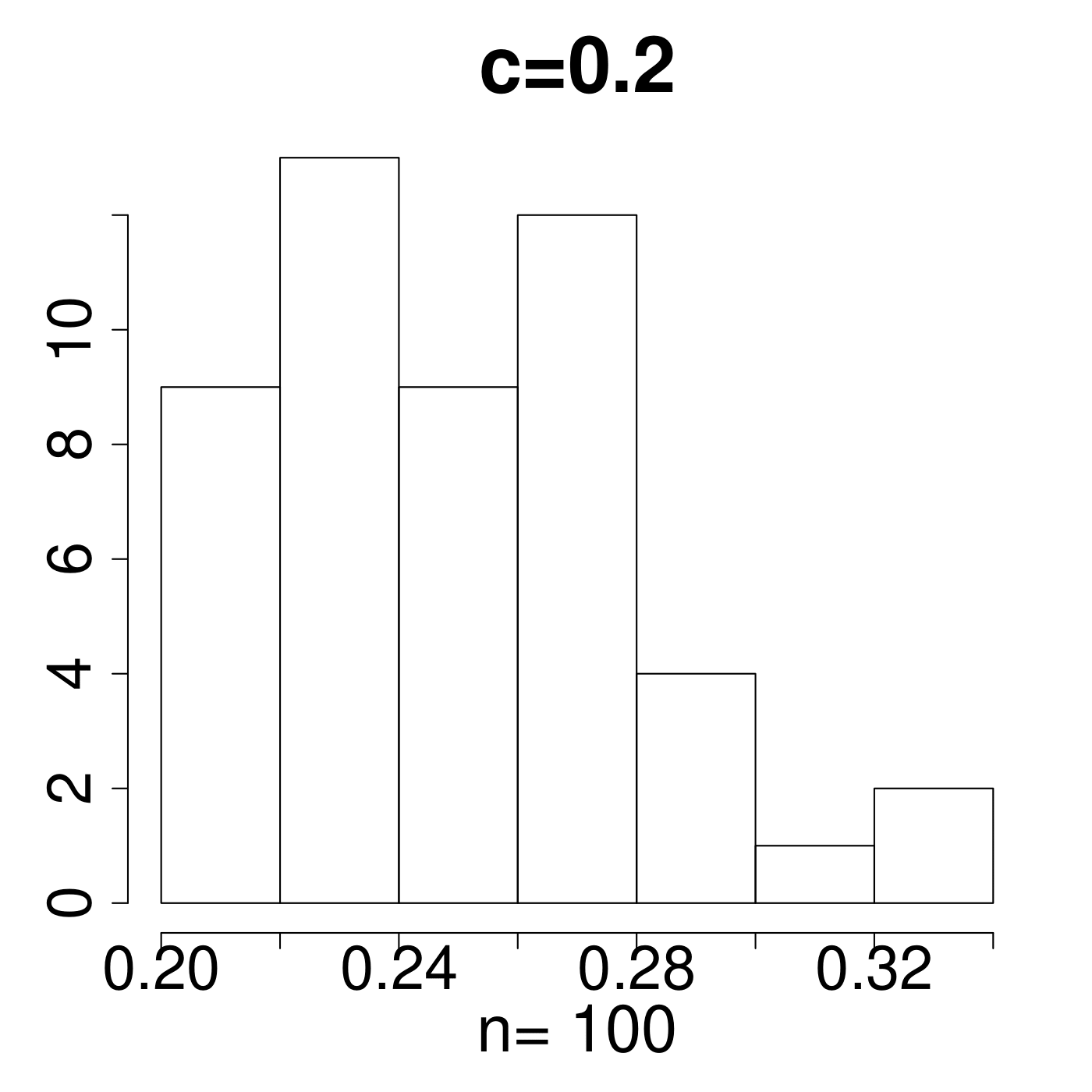}
\end{minipage}%
}%
\subfigure{
\begin{minipage}[t]{0.2\linewidth}
\centering
\includegraphics[width=1.0in]{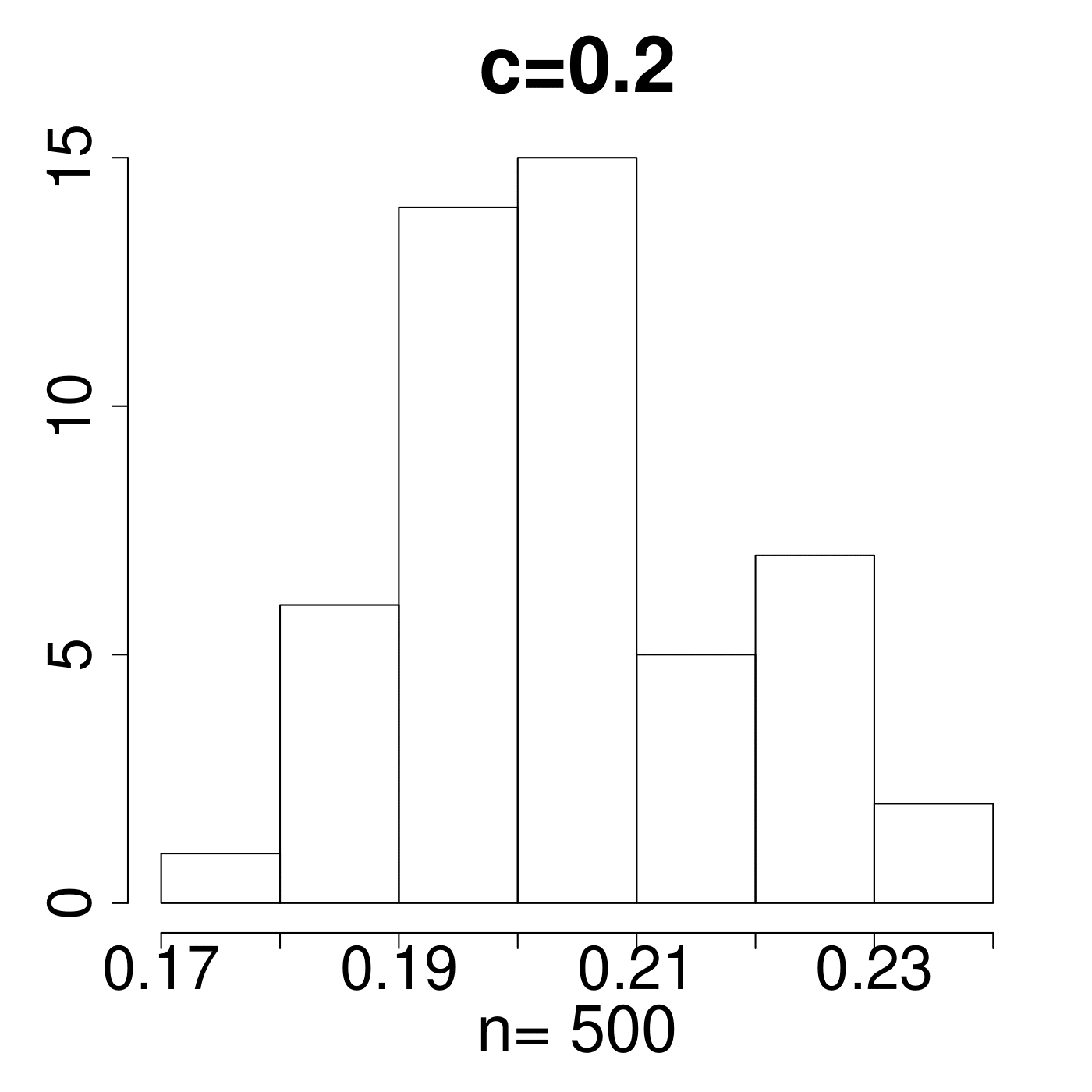}
\end{minipage}%
}%
\subfigure{
\begin{minipage}[t]{0.2\linewidth}
\centering
\includegraphics[width=1.0in]{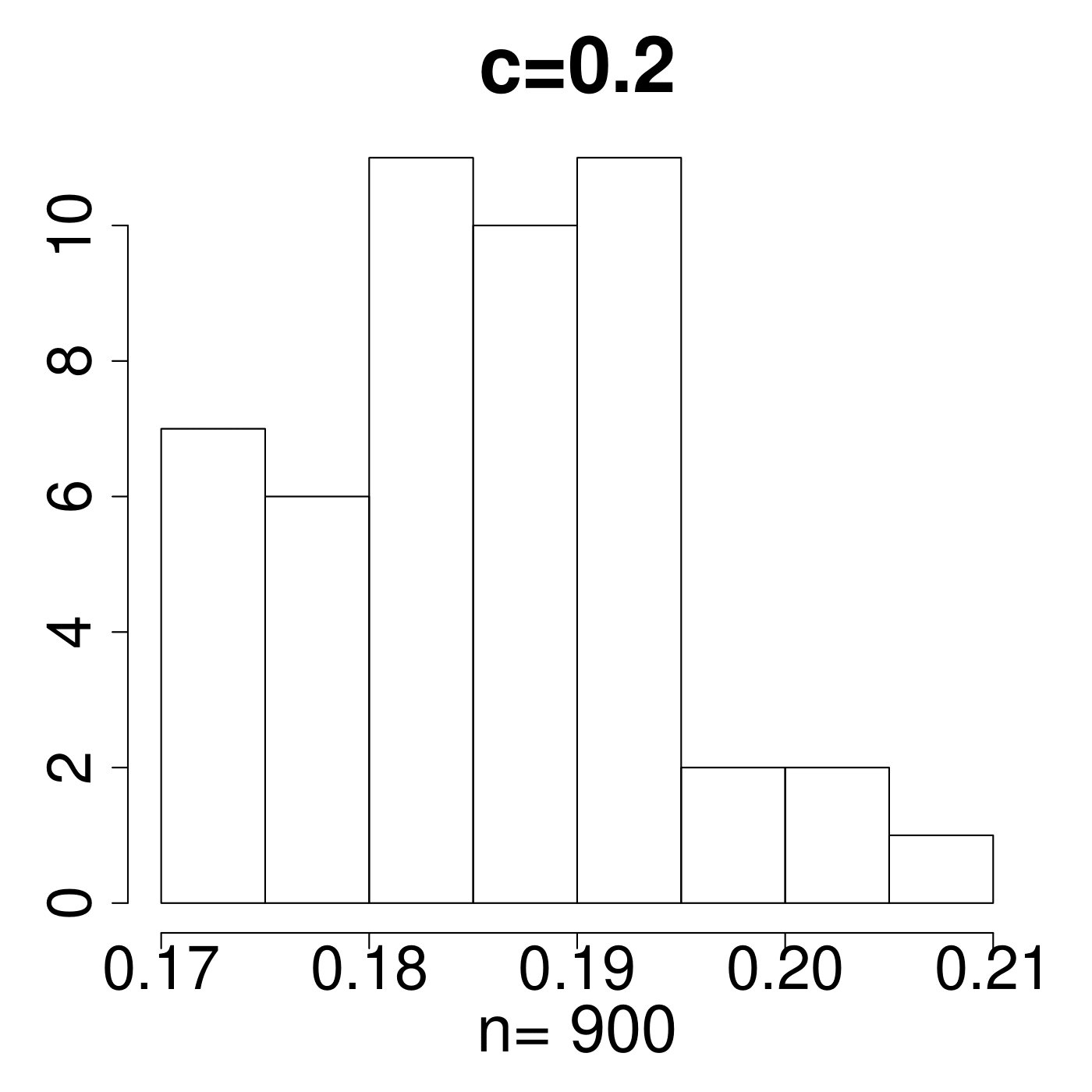}
\end{minipage}
}%
\subfigure{
\begin{minipage}[t]{0.2\linewidth}
\centering
\includegraphics[width=1.0in]{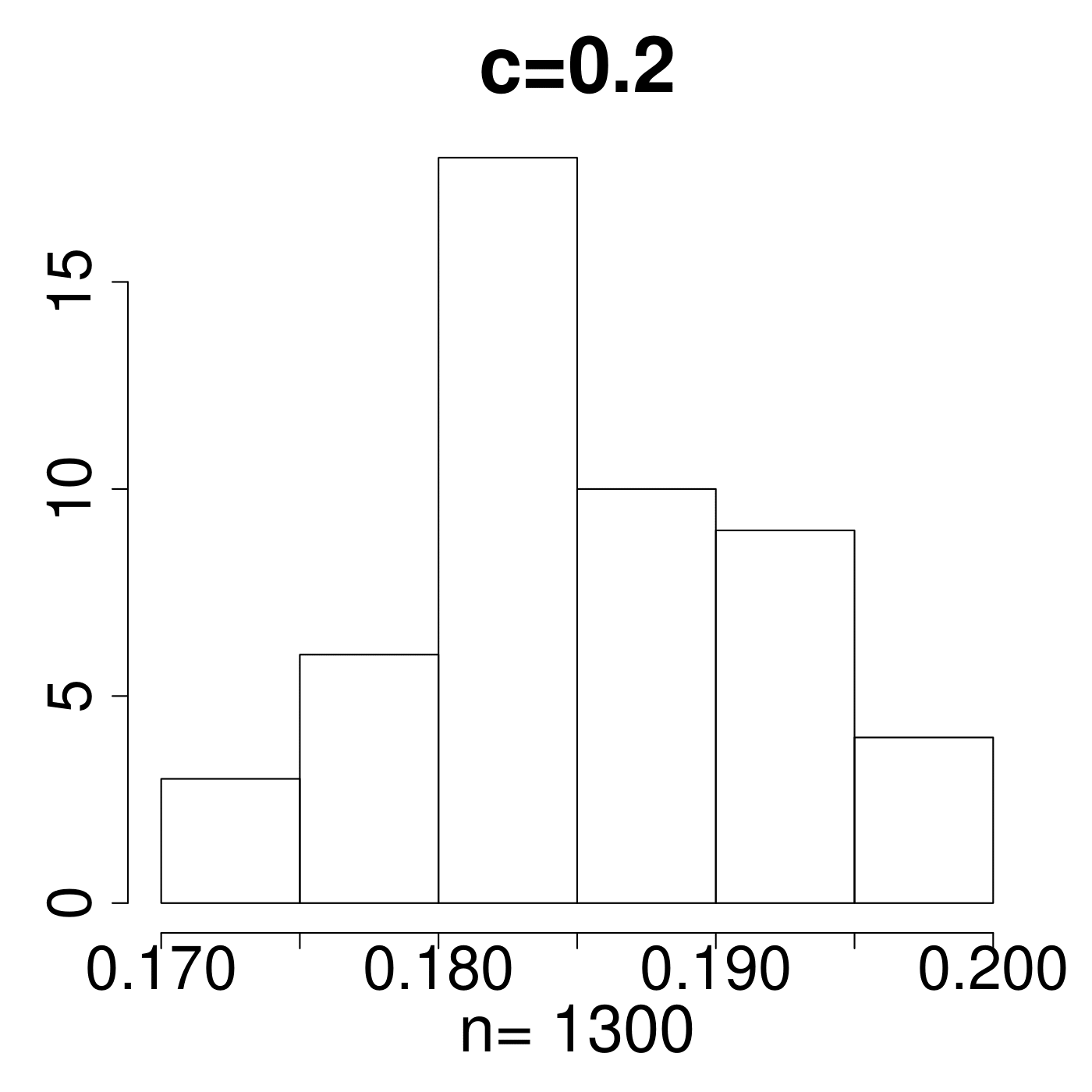}
\end{minipage}
}%

\subfigure{
\begin{minipage}[t]{0.2\linewidth}
\centering
\includegraphics[width=1.0in]{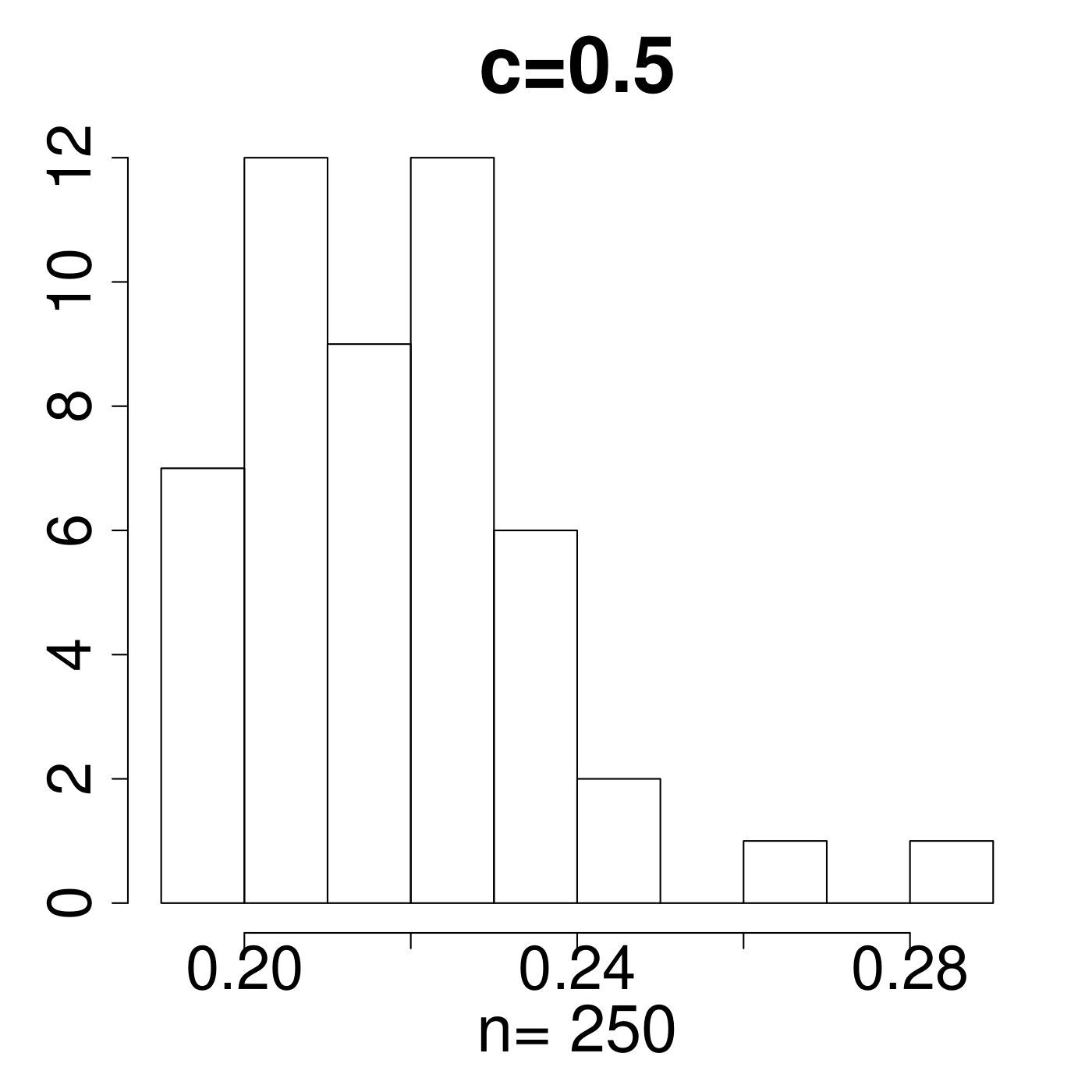}
\end{minipage}%
}%
\subfigure{
\begin{minipage}[t]{0.2\linewidth}
\centering
\includegraphics[width=1.0in]{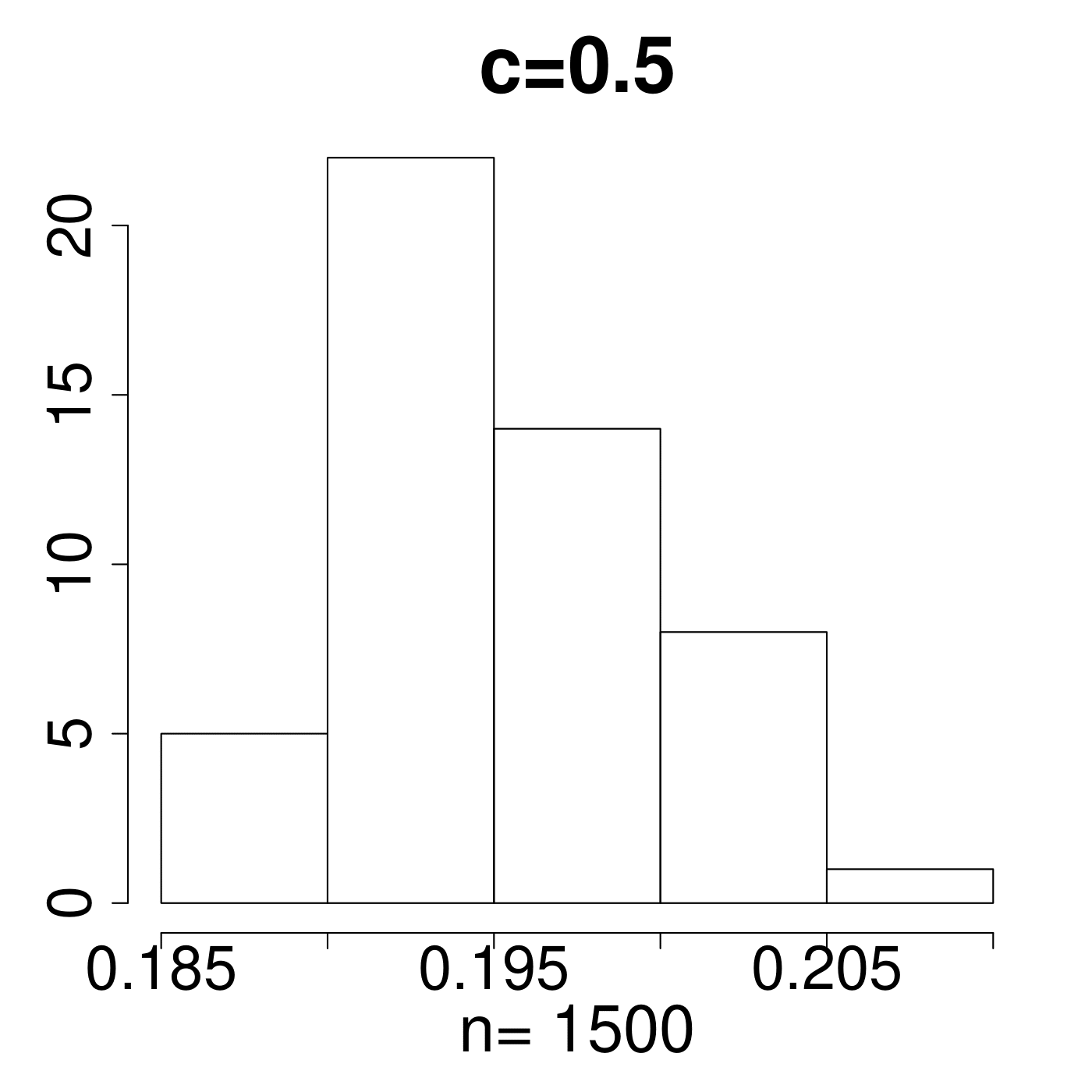}
\end{minipage}%
}%
\subfigure{
\begin{minipage}[t]{0.2\linewidth}
\centering
\includegraphics[width=1.0in]{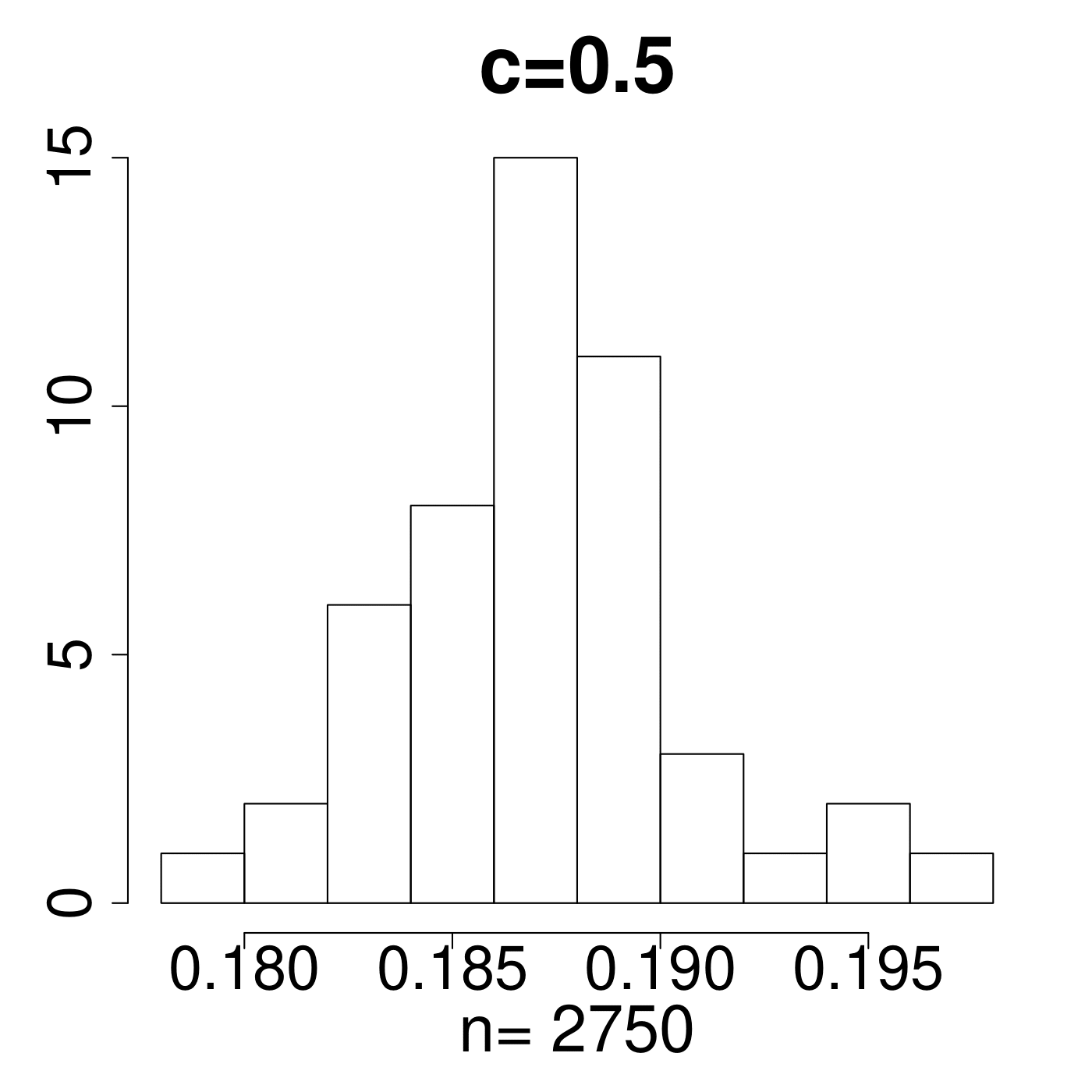}
\end{minipage}
}%
\subfigure{
\begin{minipage}[t]{0.2\linewidth}
\centering
\includegraphics[width=1.0in]{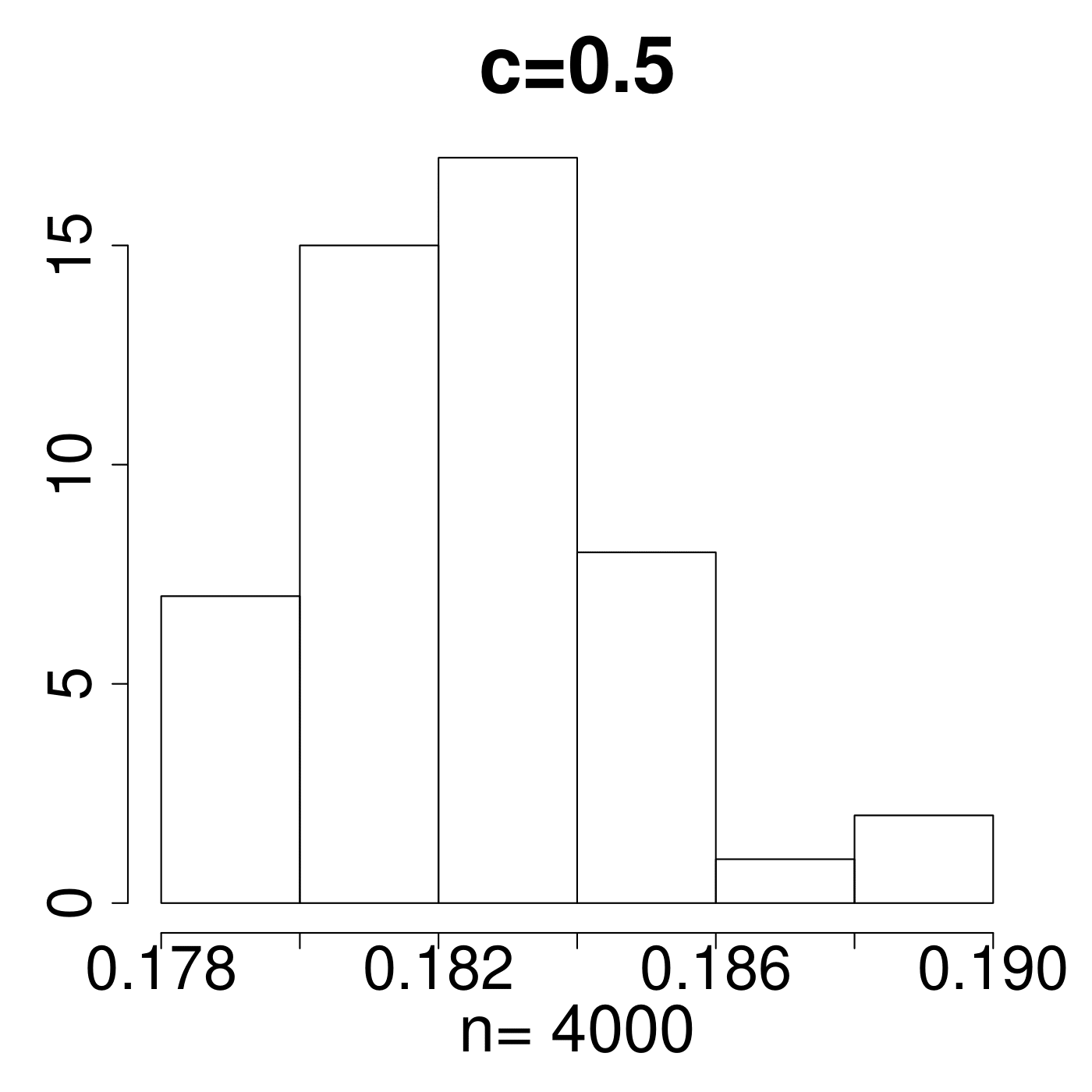}
\end{minipage}
}%

\subfigure{
\begin{minipage}[t]{0.2\linewidth}
\centering
\includegraphics[width=1.0in]{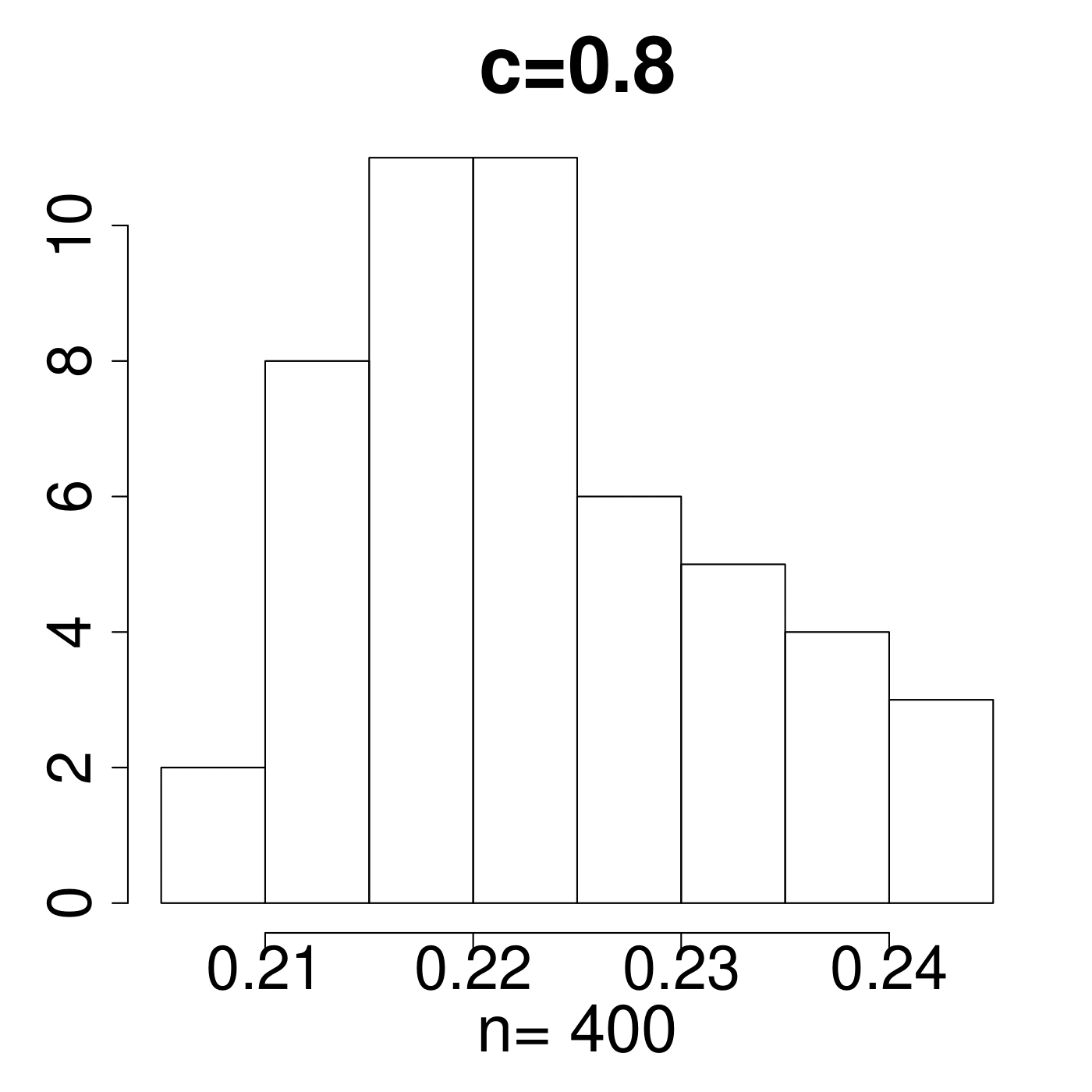}
\end{minipage}%
}%
\subfigure{
\begin{minipage}[t]{0.2\linewidth}
\centering
\includegraphics[width=1.0in]{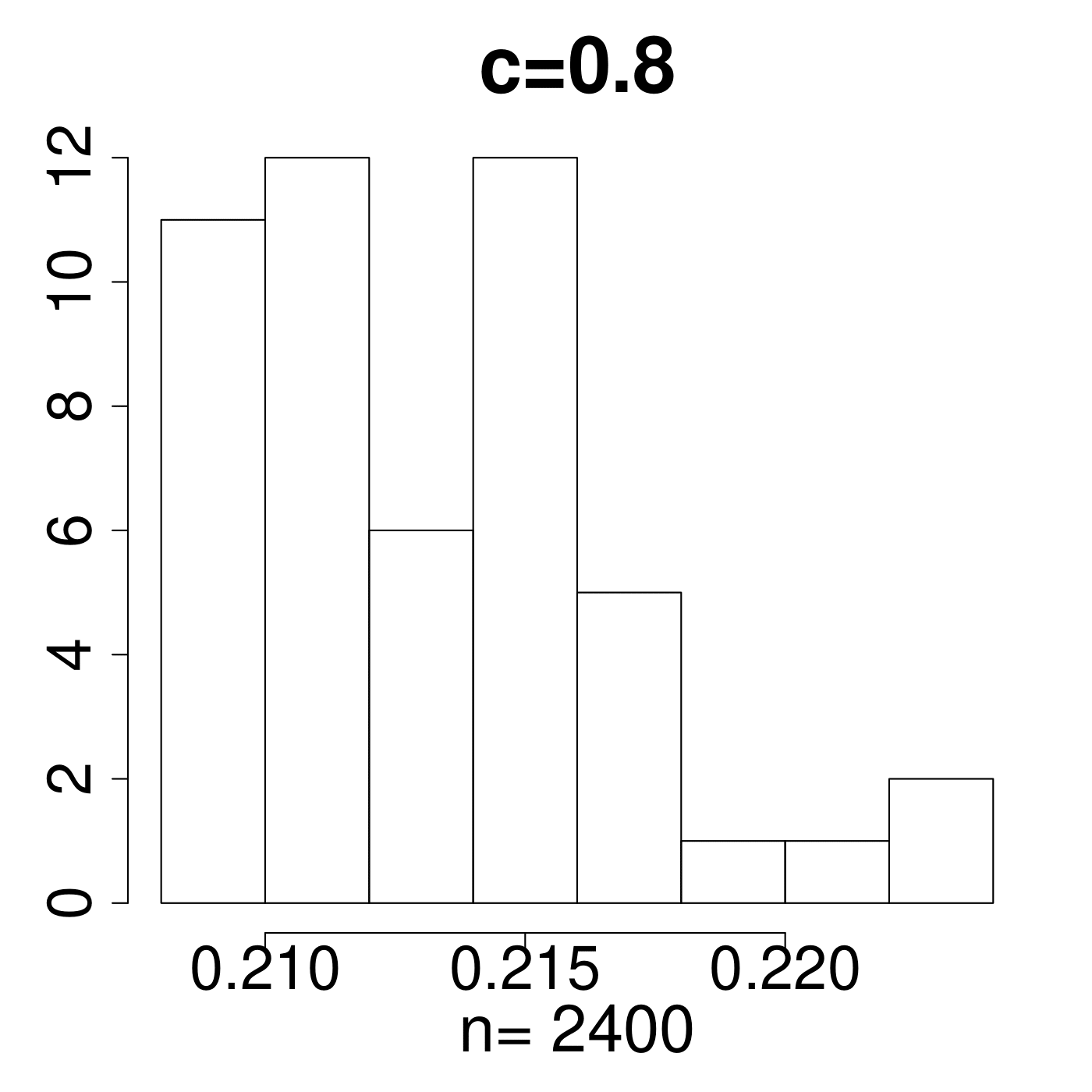}
\end{minipage}%
}%
\subfigure{
\begin{minipage}[t]{0.2\linewidth}
\centering
\includegraphics[width=1.0in]{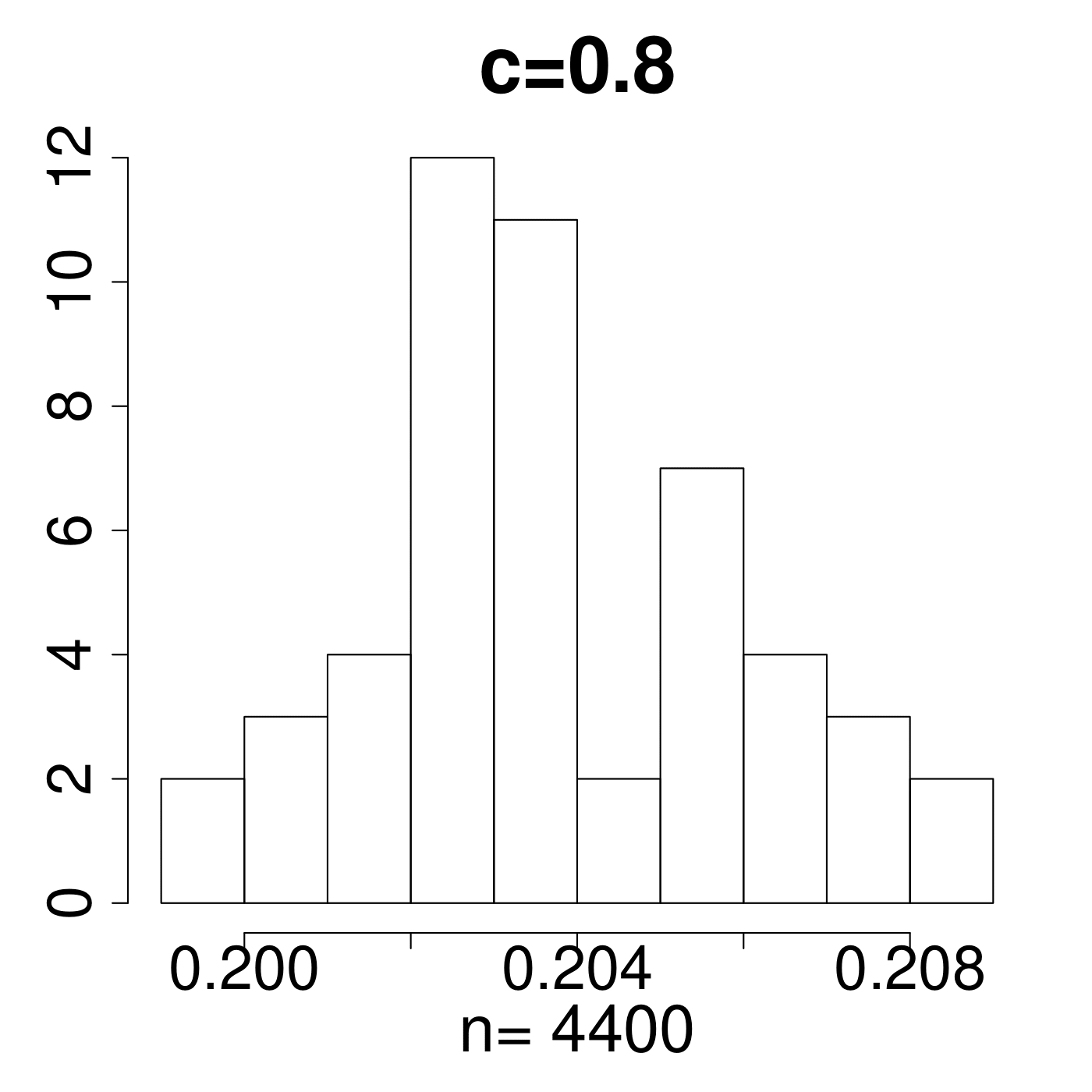}
\end{minipage}
}%
\subfigure{
\begin{minipage}[t]{0.2\linewidth}
\centering
\includegraphics[width=1.0in]{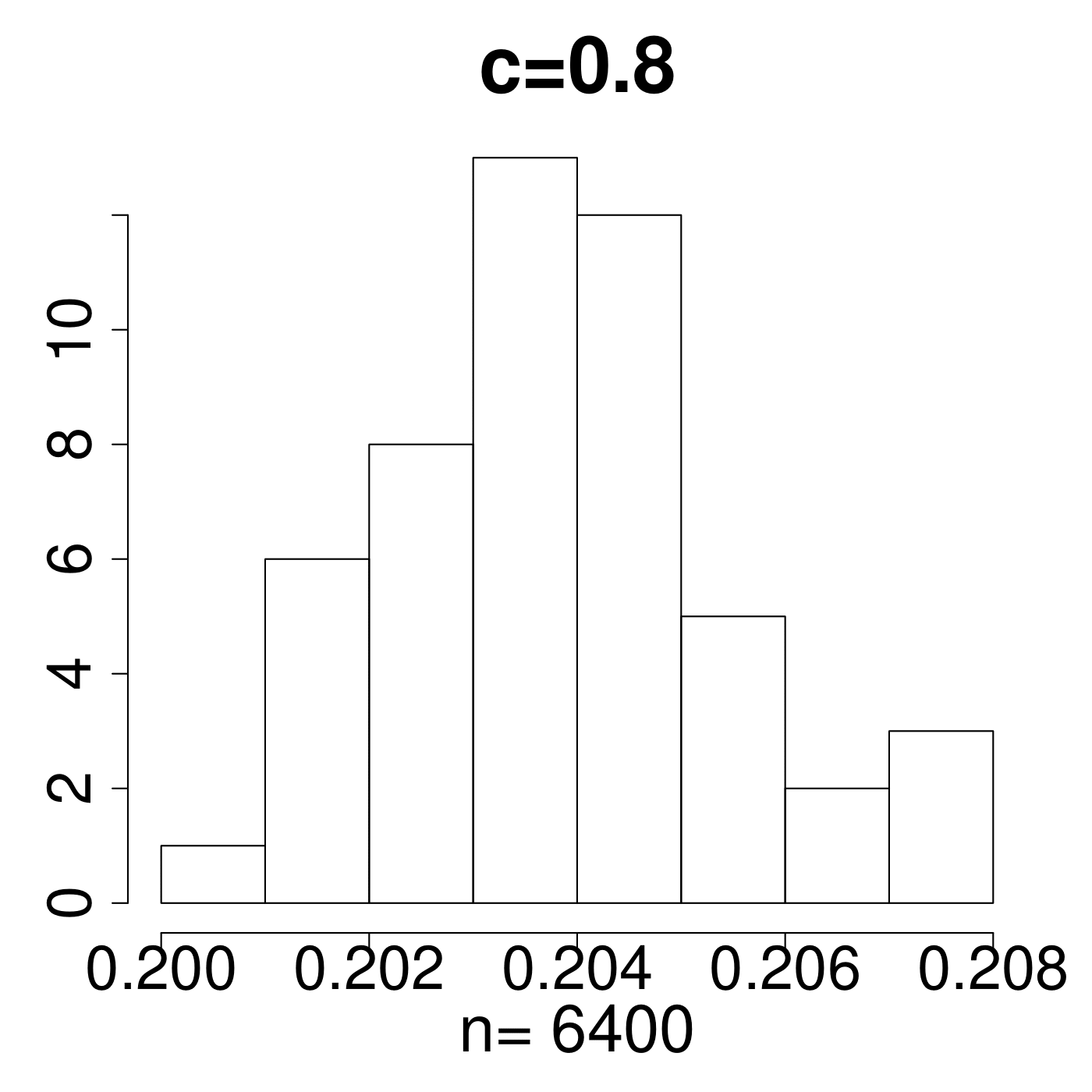}
\end{minipage}
}%

\centering
\caption{~~ Histograms of $\hs_nn^{\frac13}/\sqrt{\log n}$ from different $c$ and $n$: For each pair of $(c,n)$, the eigenvalues are generated from standard MP Law with 50 repetitions that lead to 50 values of the statistic.}
\label{valueofC}
\end{figure}

\begin{remark}
  In the simulation for different MP Laws, we generate eigenvalues and get histograms of $\hs_nn^{\frac13}/\sqrt{\log n}$. As shown in Figure \ref{valueofC}, the value $\hs_nn^{\frac13}/\sqrt{\log n}$ always lies in the interval [0.15,0.25]. 
We empirically suggest the critical line with $C=0.4$ from theoretical simulations with comparison to the extreme value 0.35 displayed in Figure \ref{valueofC}. The value of C is set a little higher due to the fact that any slight deviation from MP law will make the value $\hs_nn^{\frac13}/\sqrt{\log n}$ in a high level. The satisfactory results selected by the {spectral criterion} from $C=0.4$ to $C=0.6$ are all acceptable in our experiments. Basically, any value of $C$ in the range of $[0.4,0.6]$ can be recommended for the spectral criterion from our experimental results.

\end{remark}

\subsection{Early stopping in synthetic data experiments}\label{sec: earlystopsimu}

Because of huge amount of data under analysis, we conduct experiments, stock the relevant  data, and then check the results offline.
The epochs where we save trained NNs for different architectures are
all fixed at
0, 1, 2,..., 9, 10, 12, 16, 20,..., 248, the latter epochs having an
increment of four.
Even in such sparse data reservation, the total data we obtained is
larger than 1TB.

\begin{figure}[htbp]
\centering
\subfigure[NN1$\calD_1$]{
\begin{minipage}[t]{0.2\linewidth}
\centering
\includegraphics[width=1.2in]{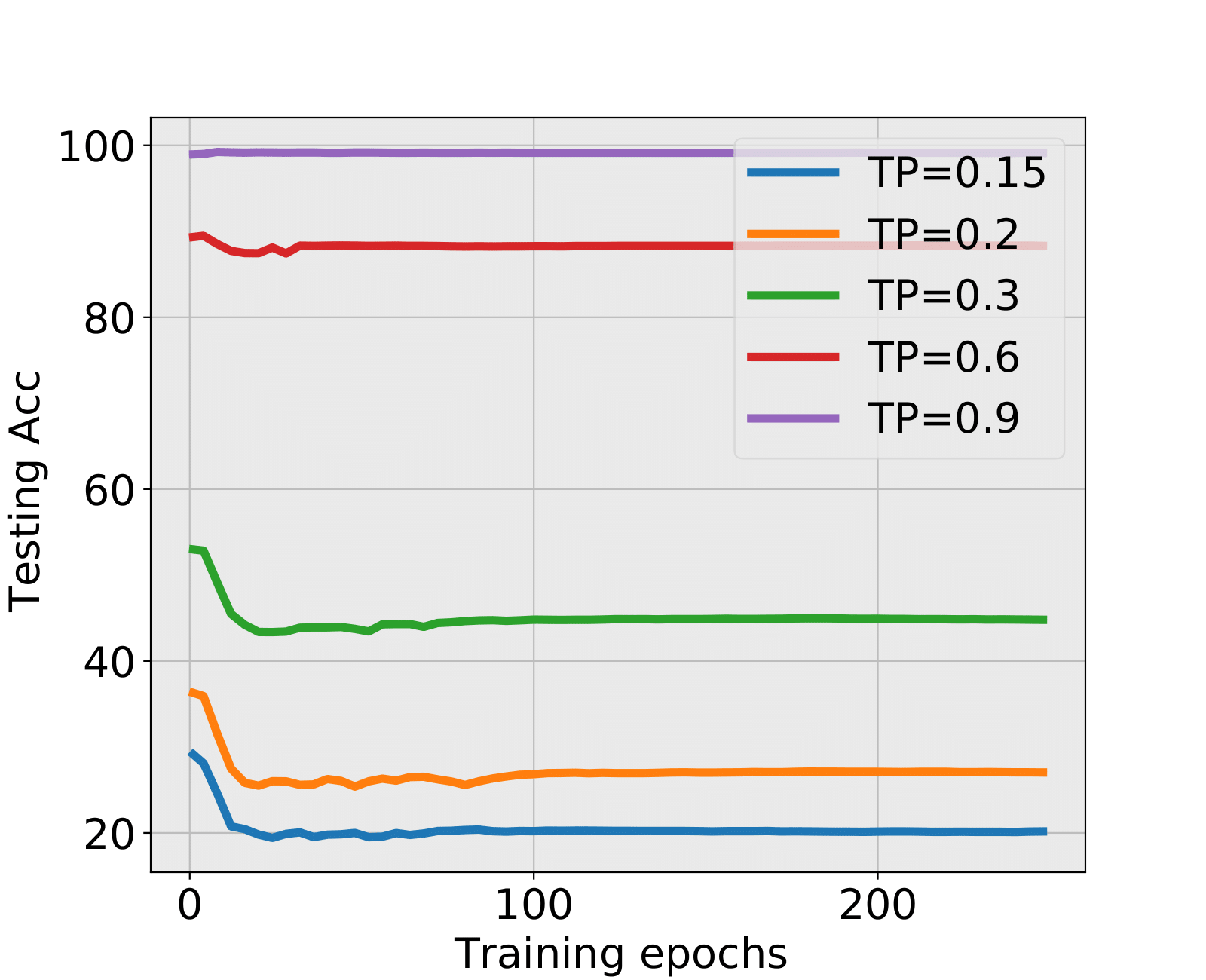}
\end{minipage}%
}%
\subfigure[NN1$\calD_2$]{
\begin{minipage}[t]{0.2\linewidth}
\centering
\includegraphics[width=1.2in]{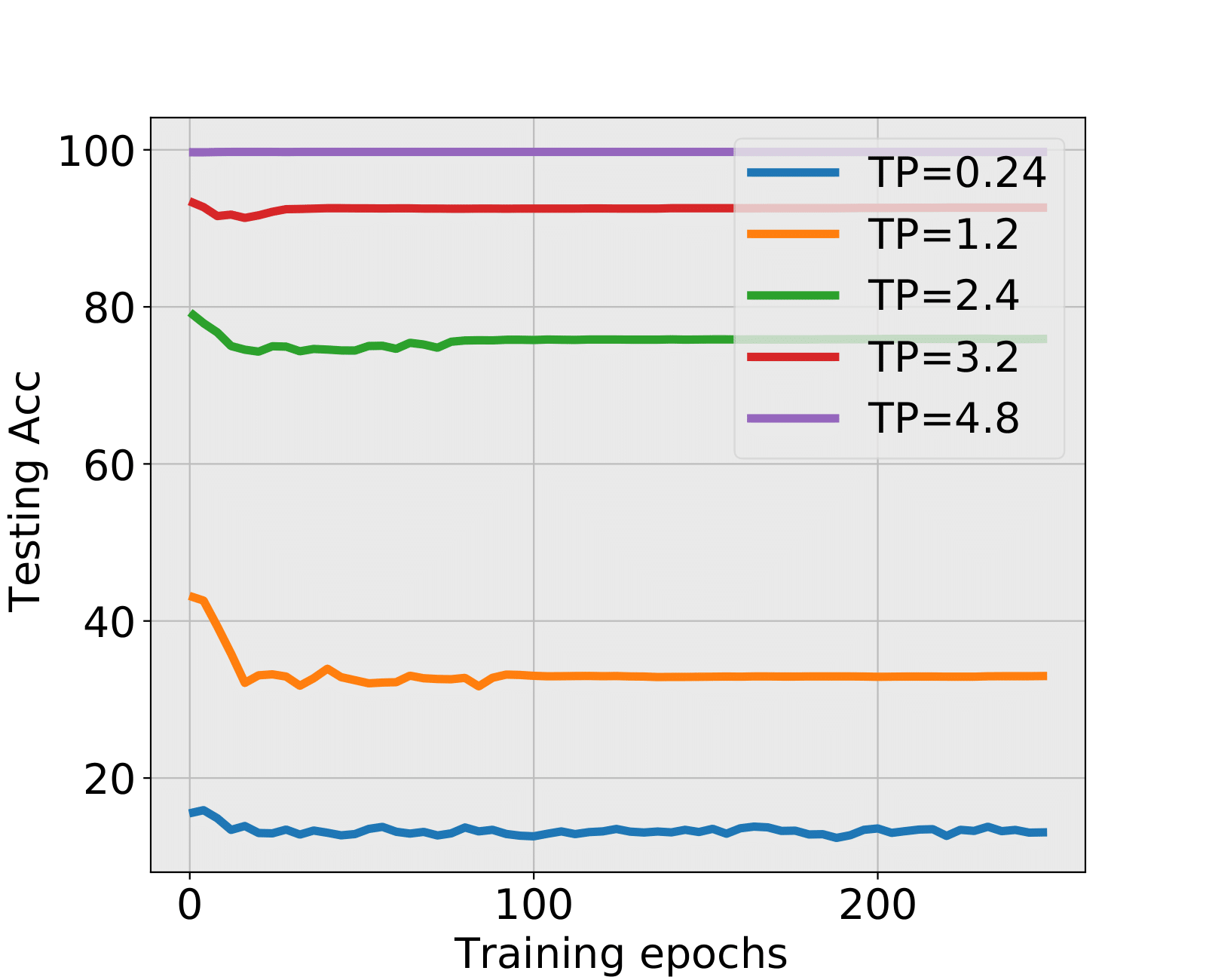}
\end{minipage}%
}%
\subfigure[NN2$\calD_1$]{
\begin{minipage}[t]{0.2\linewidth}
\centering
\includegraphics[width=1.2in]{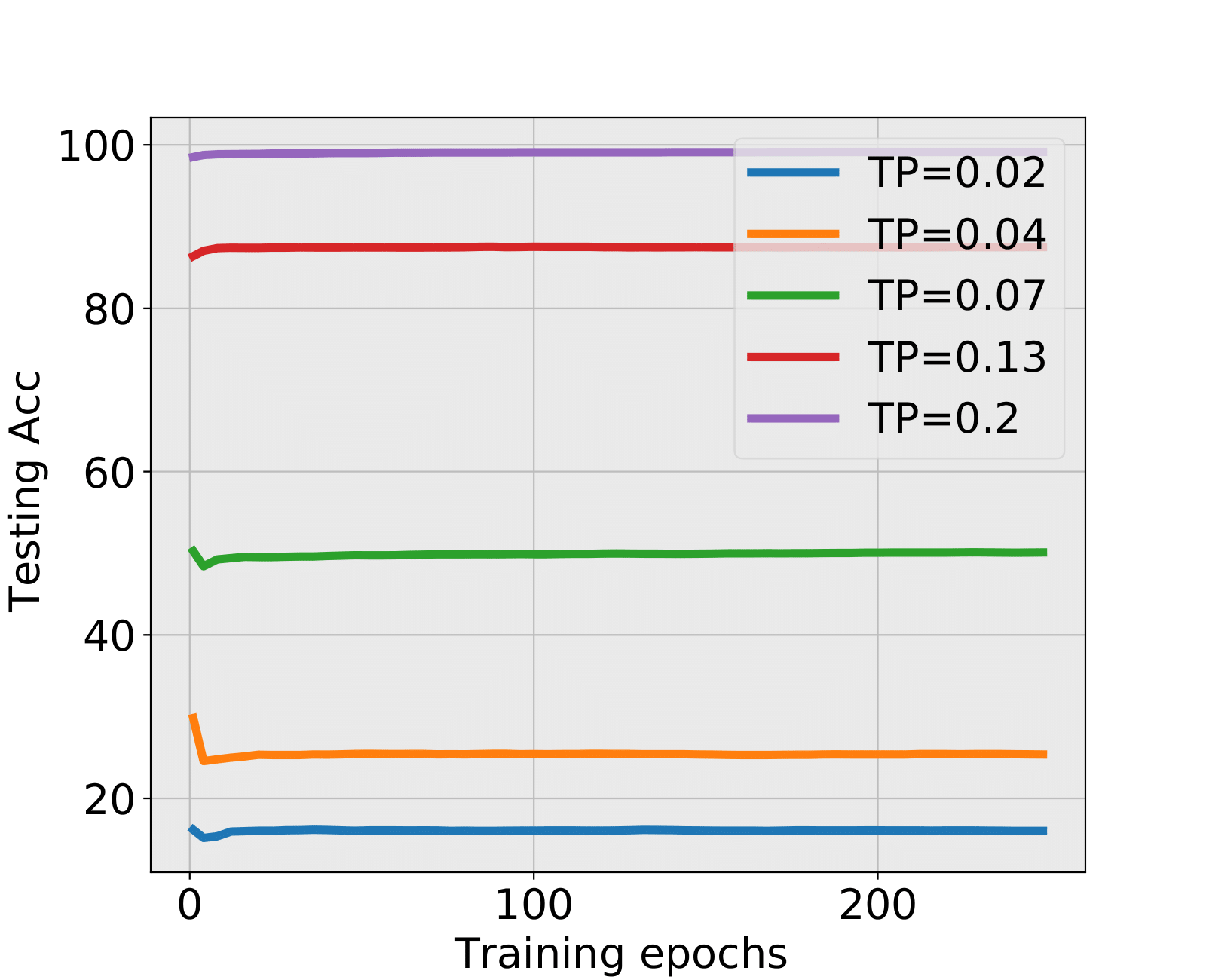}
\end{minipage}
}%
\subfigure[NN2$\calD_2$]{
\begin{minipage}[t]{0.2\linewidth}
\centering
\includegraphics[width=1.2in]{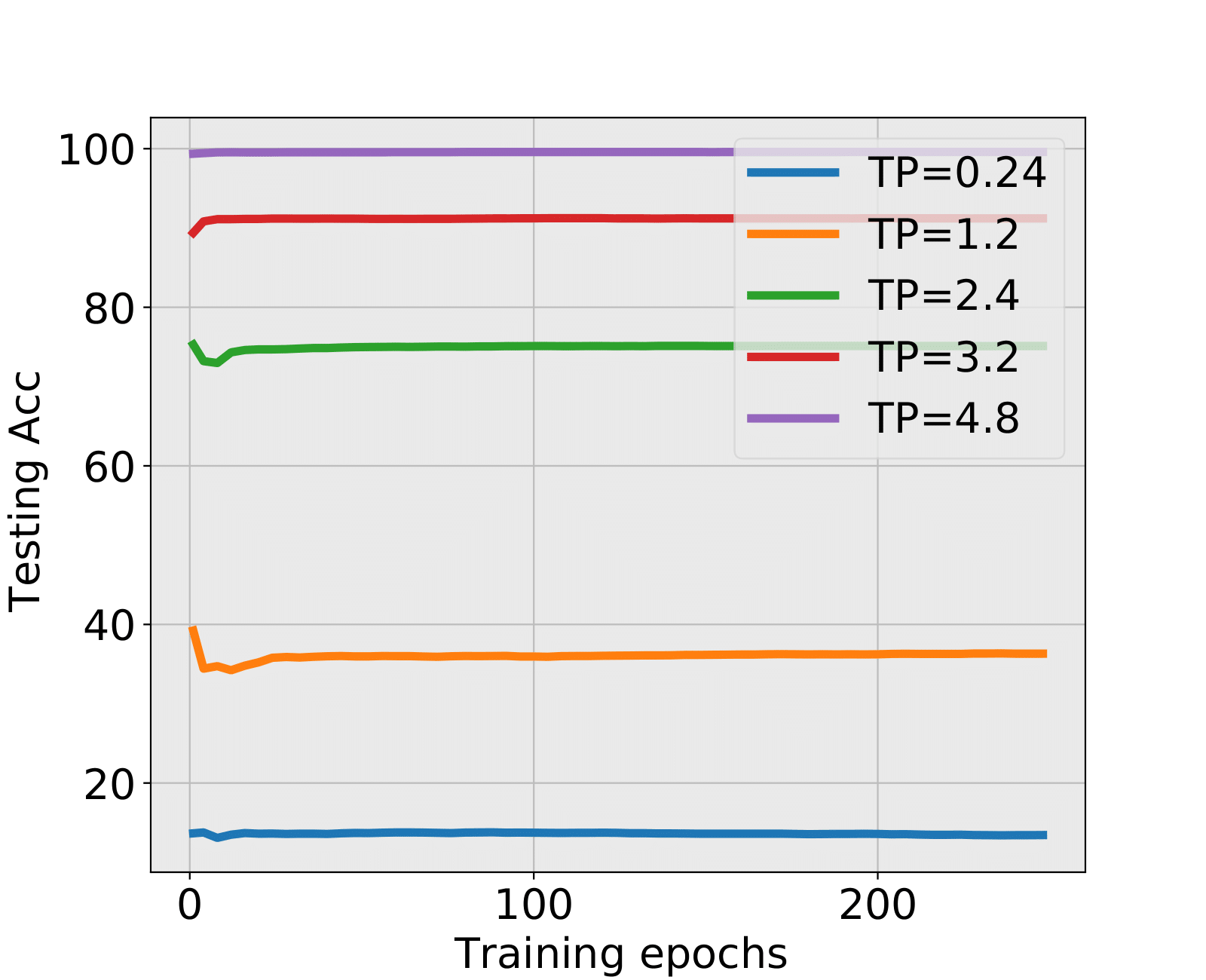}
\end{minipage}
}%

\subfigure[NN1$\calD_1$]{
\begin{minipage}[t]{0.2\linewidth}
\centering
\includegraphics[width=1.2in]{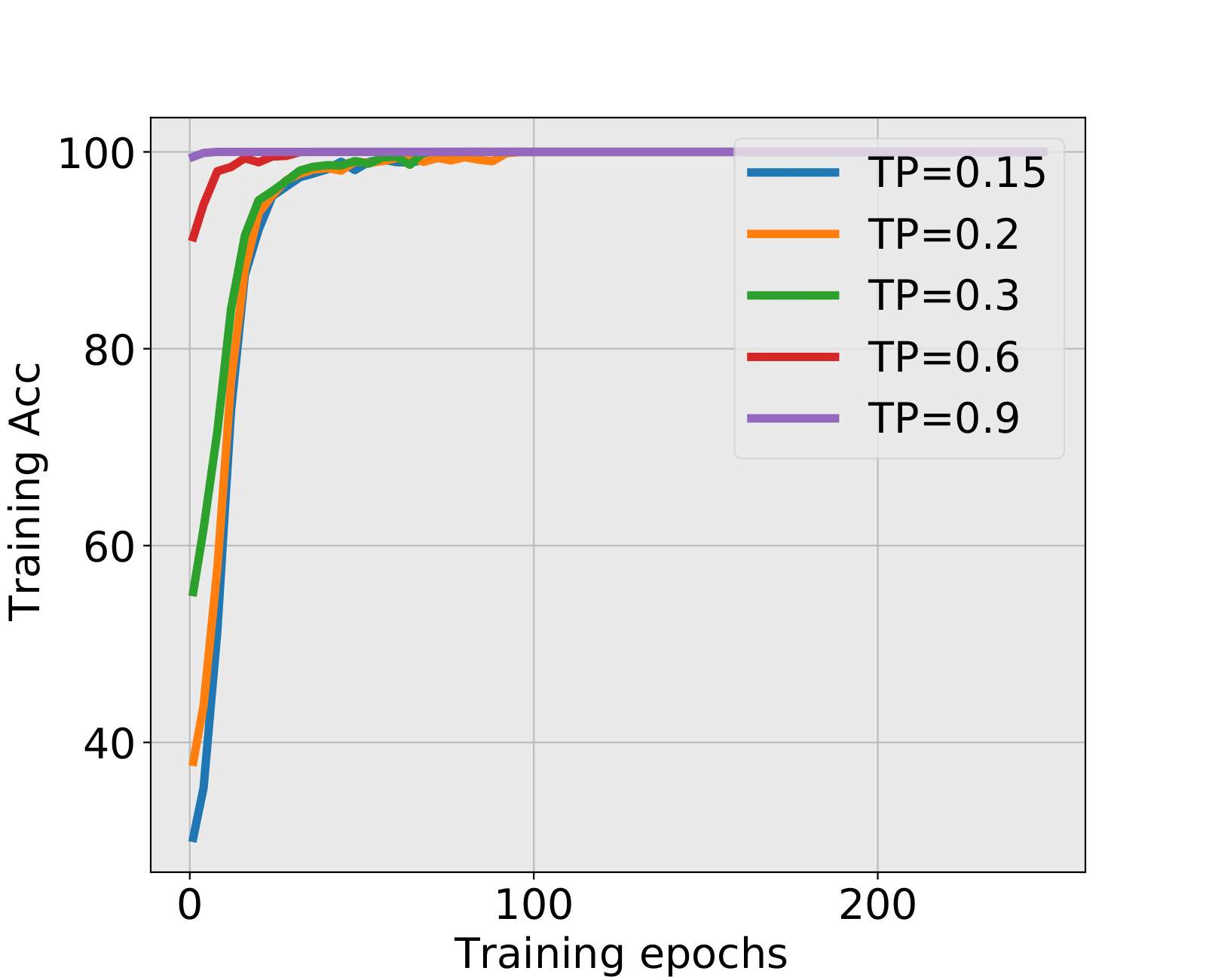}
\end{minipage}%
}%
\subfigure[NN1$\calD_2$]{
\begin{minipage}[t]{0.2\linewidth}
\centering
\includegraphics[width=1.2in]{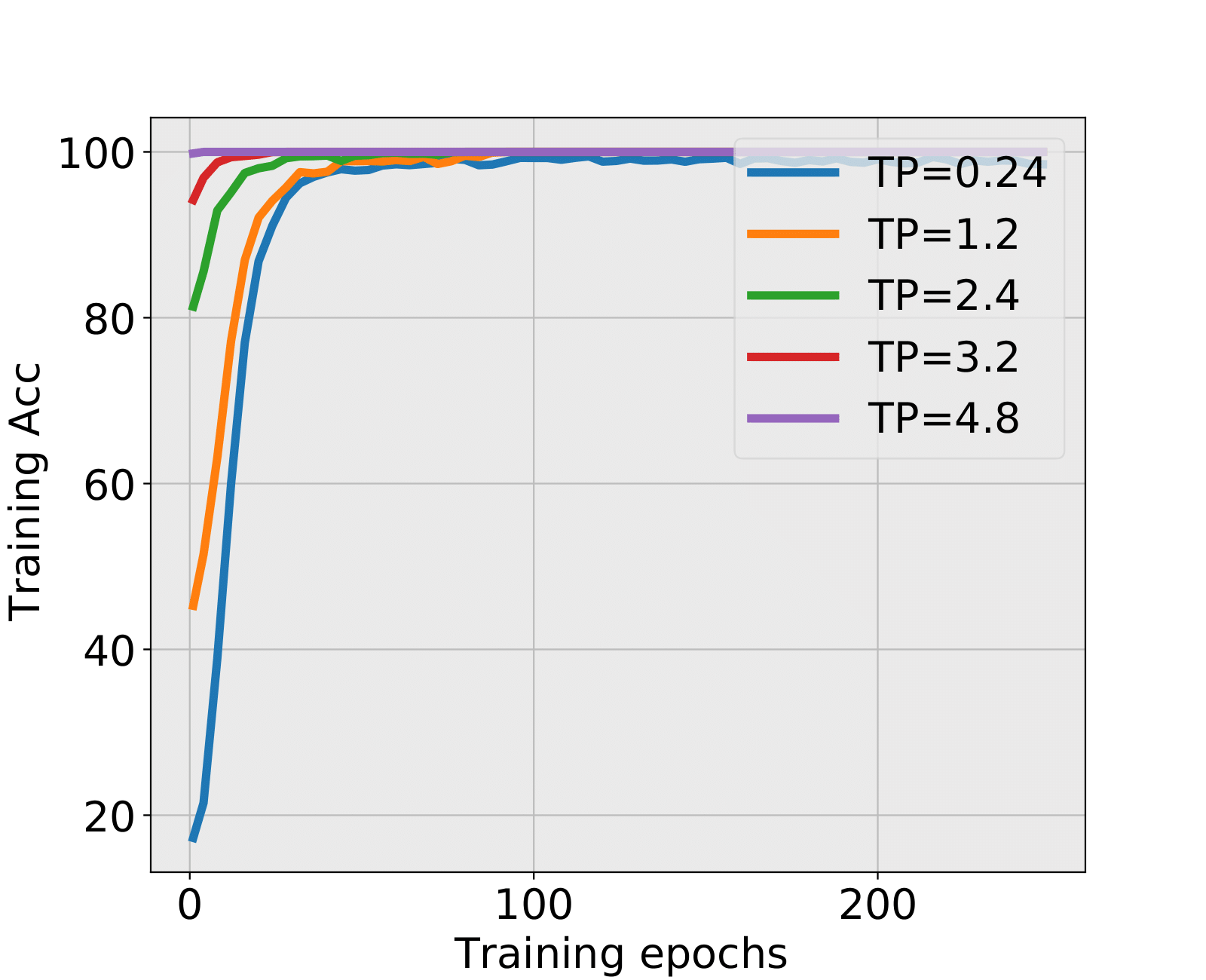}
\end{minipage}%
}%
\subfigure[NN2$\calD_1$]{
\begin{minipage}[t]{0.2\linewidth}
\centering
\includegraphics[width=1.2in]{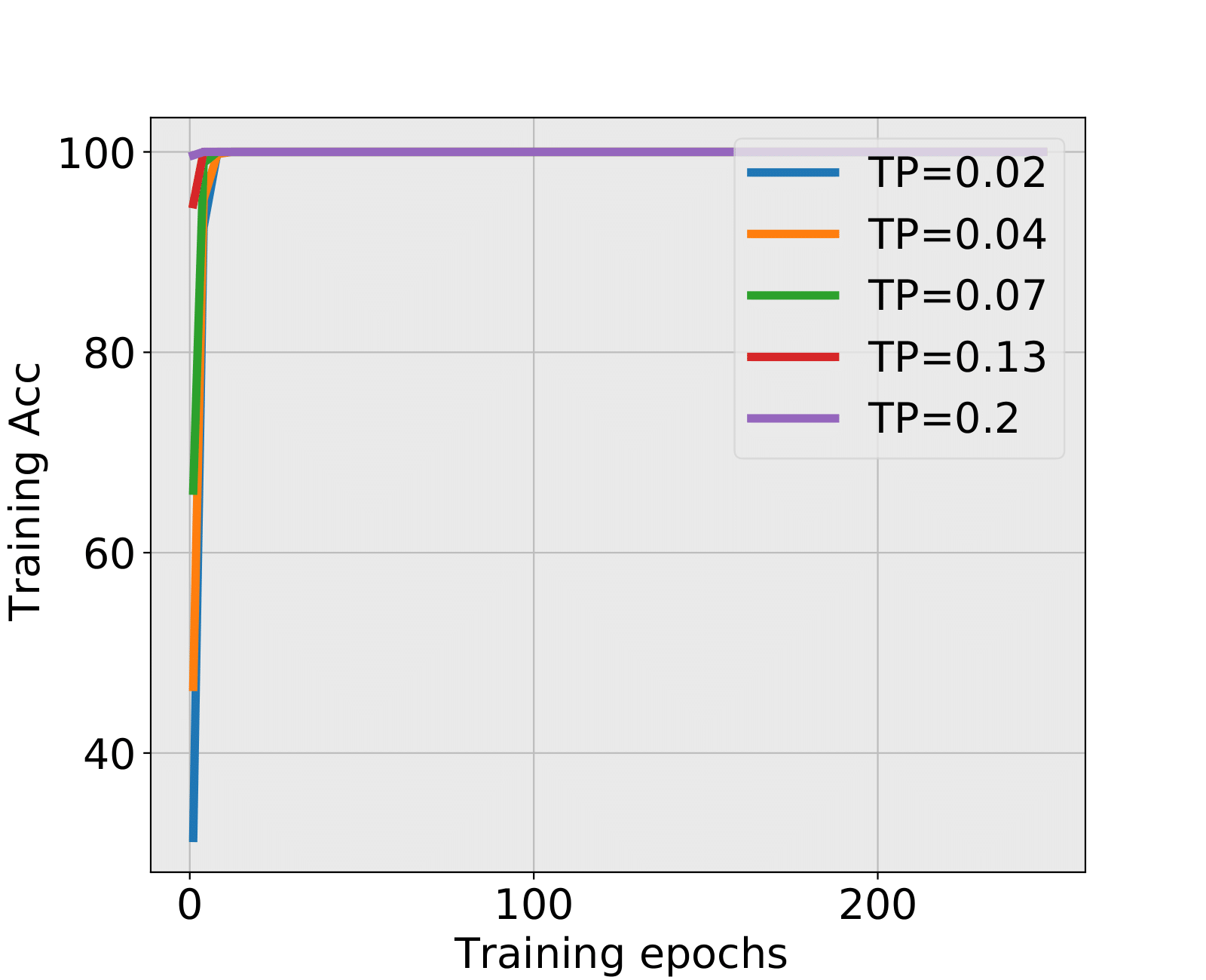}
\end{minipage}
}%
\subfigure[NN2$\calD_2$]{
\begin{minipage}[t]{0.2\linewidth}
\centering
\includegraphics[width=1.2in]{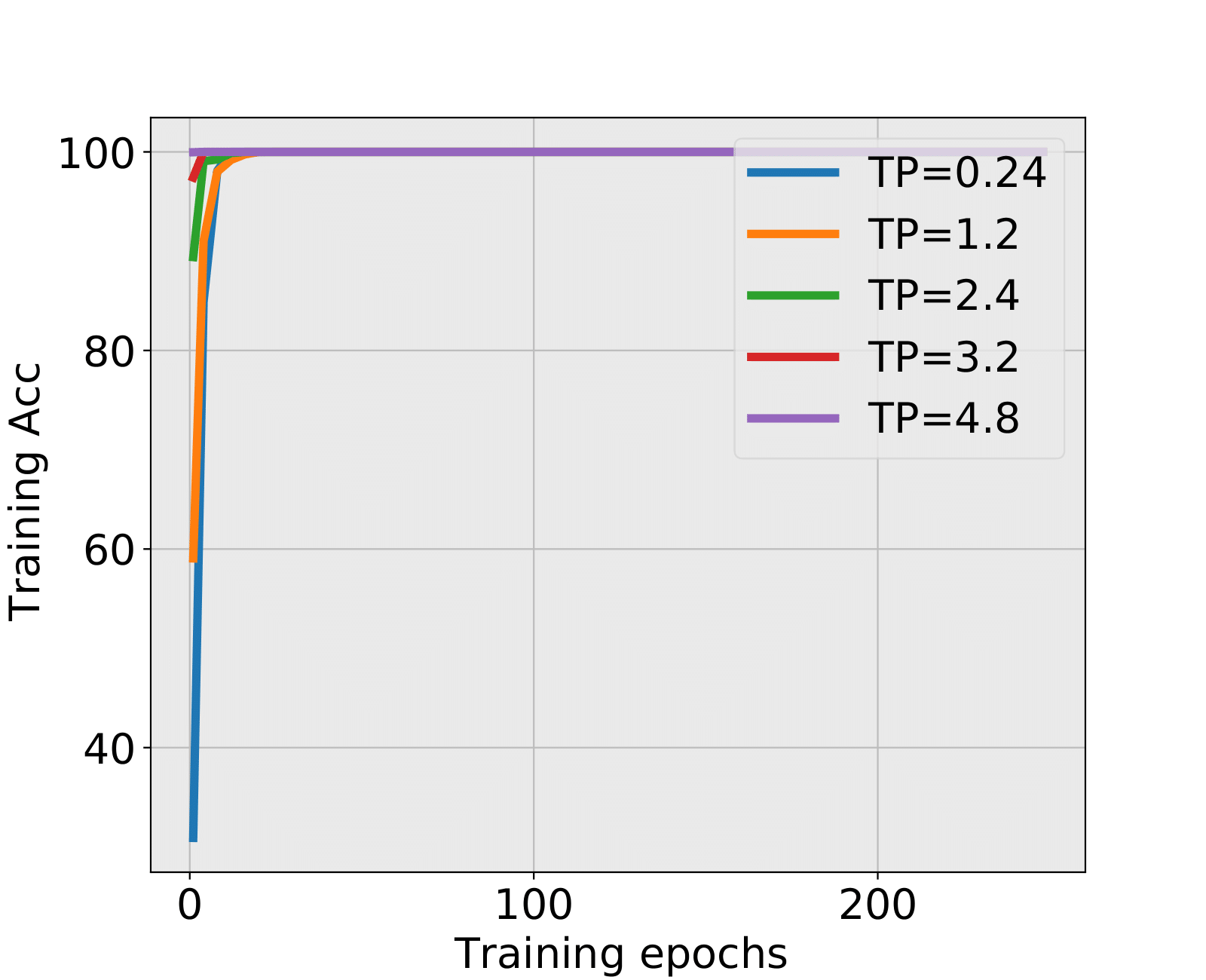}
\end{minipage}
}%
\centering
\caption{~~ Testing and Training Accuracy: We begin the line at epoch=1. Testing accuracy: (a)-(d); Training accuracy: (e)-(h). y-axis is the accuracy value, x-axis is the training epochs. Different line represents different SNRs in data sets.}
\label{spectrasimu}
\end{figure}

We apply the {spectra criterion} developed in
Section~\ref{sec:specCriterion} to the stocked training epochs, and
decide the early stopping time if the criterion is met.
When this happens, we compare  the  test accuracy  of the
corresponding NN with that of the NN trained till the final epoch
(248th). This comparison serves to measure the quality of the early
stopping using the spectral criterion. Experiment results with $K=8$ are shown in Table~\ref{tbl:earlystoppingsimu} and Figure \ref{spectrasimu}.

\bigskip 

\noindent{\bf Comments}:
\begin{itemize}
\item NN1+$\calD_1$ and NN1+$\calD_2$: During the first 20 epochs when SNR is low, the testing accuracy is decreasing while the training accuracy is increasing. Spectral criterion detects such hidden and problematic issues and recommends to early stop. It is truly remarkable that almost all early stopped NNs have higher test accuracies than the corresponding NNs trained till the end. The advantage is particularly important when the SNR is low. When SNR is large, there might be no alarm by the spectral criterion, see the situation of TP=0.9 in NN1+$\calD_1$ and TP=4.8 in NN1+$\calD_2$. This is in fact a consistency of the spectral criterion, no early stopping is needed, and the fully trained NNs have indeed higher test accuracies.

\item NN2+$\calD_1$ and NN2+$\calD_2$: Spectral Criterion detects stopping time under low SNR, nonetheless the testing accuracy is a little lower than the final testing accuracy. As the differences are very small,  huge training time is cut off, and testing accuracy is ensured due to the emergence of well-trained structure that already seized sufficient information. 


\end{itemize}

\begin{table}[htbp!]
\caption{~~ Early stopping results in synthetic data experiments with $C=0.4$: stopping epochs selected by spectral criterion in different layers' weight matrices and their testing accuracy (Test Acc). The symbol ”-” means no early stopping epoch is found by the spectral criterion.\label{tbl:earlystoppingsimu}}
\centering

\begin{tabular}{|c|c|c|c|c|c|c|c|}
\multicolumn{8}{c}{The combination NN1+$\calD_1$}\\
  \hline
\multirow{2}{*}{\begin{tabular}[c]{@{}c@{}}Typical\\ TP\end{tabular}} & \multicolumn{4}{c|}{spectral criterion $C=0.4$}  & \multicolumn{3}{c|}{Final Epoch 248} \\ \cline{2-8} 
                                                                      & epoch(FC2)  & Test Acc & epoch(FC3)  & Test Acc & FC1       & FC2      & Test Acc      \\ \hline
0.15                                                                  & 7           & 25.84\%  & 10          & 23.23\%  & HT        & HT       & 20.17\%       \\ \hline
0.2                                                                   & 7           & 32.70\%  & 12          & 27.48\%  & HT        & HT       & 27.03\%       \\ \hline
0.3                                                                   & 7           & 49.36\%  & 12          & 45.48\%  & HT        & HT       & 44.80\%       \\ \hline
0.6                                                                   & 8           & 88.52\%  & 32          & 88.32\%  & BT       & BT      & 88.30\%       \\ \hline
0.9                                                                   & \multicolumn{2}{c|}{-} & \multicolumn{2}{c|}{-} & LT        & LT       & 99.13\%       \\ \hline
\end{tabular}

\vskip2mm
\begin{tabular}{|c|c|c|c|c|c|c|c|}
\multicolumn{8}{c}{The combination NN1+$\calD_2$}\\
  \hline
\multirow{2}{*}{\begin{tabular}[c]{@{}c@{}}Typical\\ TP\end{tabular}} & \multicolumn{4}{c|}{spectral criterion $C=0.4$}  & \multicolumn{3}{c|}{Final Epoch 248} \\ \cline{2-8} 
                                                                      & epoch(FC2)  & Test Acc & epoch(FC3)  & Test Acc & FC1       & FC2      & Test Acc      \\ \hline
0.24                                                                  & 9           & 14.69\%  & 16          & 13.89\%  & HT        & HT       & 13.08\%       \\ \hline
1.2                                                                   & 7           & 38.61\%  & 12          & 35.84\%  & HT        & HT       & 32.98\%       \\ \hline
2.4                                                                   & 7           & 77.19\%  & 16          & 74.55\%  & BT        & BT       & 75.92\%       \\ \hline
3.2                                                                   & 9           & 92.11\%  & \multicolumn{2}{c|}{-} & BT        & LT      & 92.64\%       \\ \hline
4.8                                                                   & \multicolumn{2}{c|}{-} & \multicolumn{2}{c|}{-} & LT        & LT       & 99.73\%       \\ \hline
\end{tabular}

\vskip2mm
\begin{tabular}{|c|c|c|c|c|c|c|c|}
\multicolumn{8}{c}{The combination NN2+$\calD_1$}\\
\hline
\multirow{2}{*}{\begin{tabular}[c]{@{}c@{}}Typical\\ TP\end{tabular}} & \multicolumn{4}{c|}{spectral criterion $C=0.4$}  & \multicolumn{3}{c|}{Final Epoch 248} \\ \cline{2-8} 
                                                                      & epoch(FC1)  & Test Acc & epoch(FC2)  & Test Acc & FC1       & FC2      & Test Acc      \\ \hline
0.02                                                                  & 6           & 14.89\%  & 7           & 15.84\%  & HT        & BT      & 16.02\%       \\ \hline
0.04                                                                  & 8           & 24.78\%  & 7           & 23.34\%  & HT        & BT      & 25.38\%       \\ \hline
0.07                                                                  & 5           & 48.31\%  & 6           & 48.63\%  & BT        & BT      & 50.12\%       \\ \hline
0.13                                                                  & 6           & 87.03\%  & \multicolumn{2}{c|}{-} & BT       & LT       & 87.50\%       \\ \hline
0.2                                                                   & \multicolumn{2}{c|}{-} & \multicolumn{2}{c|}{-} & LT        & LT       & 99.14\%       \\ \hline
\end{tabular}

\vskip2mm
\begin{tabular}{|c|c|c|c|c|c|c|c|}
\multicolumn{8}{c}{The combination NN2+$\calD_2$}\\
  \hline
\multirow{2}{*}{\begin{tabular}[c]{@{}c@{}}Typical\\ TP\end{tabular}} & \multicolumn{4}{c|}{spectral criterion $C=0.4$}  & \multicolumn{3}{c|}{Final Epoch 248} \\ \cline{2-8} 
                                                                      & epoch(FC1)  & Test Acc & epoch(FC2)  & Test Acc & FC1       & FC2      & Test Acc      \\ \hline
0.24                                                                  & 10          & 13.08\%  & 6           & 12.89\%  & HT        & BT       & 13.44\%       \\ \hline
1.2                                                                   & 12          & 34.22\%  & 5           & 34.63\%  & BT        & BT       & 36.31\%       \\ \hline
2.4                                                                   & 5           & 72.59\%  & 16          & 74.61\%  & BT       & BT      & 75.12\%       \\ \hline
3.2                                                                   & \multicolumn{2}{c|}{-} & \multicolumn{2}{c|}{-} & LT        & LT       & 91.20\%       \\ \hline
4.8                                                                   & \multicolumn{2}{c|}{-} & \multicolumn{2}{c|}{-} & LT        & LT       & 99.59\%         \\ \hline
\end{tabular}
\end{table}

The spectral criterion is valid when overfitting appears in training. In such situation, the training and testing accuracies do not have the same tendency. As training epochs increase, Figure \ref{spectrasimu} shows that training accuracy tends to 100\% while testing accuracy is highly related with the tuning parameter $\delta$ or $t$. Without testing data, the  spectral criterion could propose an early stopping time even when the  training accuracy is increasing. 

When the spectral criterion gives different epochs in different layers, there is a question that \textbf{how to decide a stopping time?} From our experimental results, we empirically suggest that any epoch after the time some layer hits the critical value $s_*$ is suitable to stop, and it is strongly recommended to stop training if there is more than one layer hitting the critical value. For example, TP=0.15 in NN1+$\calD_1$, epochs in 7-10 are all suitable early stopping time with a guaranteed test accuracy.

\newpage
\subsection{Early stopping in real data experiments}\label{sec: earlystopreal}

In real data experiments, we still follow the settings in section \ref{sec: earlystopsimu} to evaluate the quality of early stopping time using the spectral criterion by checking LeNet/MiniAlexNet+MNIST/CIFAR10. Experiment results are shown in Table~\ref{tbl:earlystopingreal}.

\bigskip 

\noindent{\bf Comments}:
\begin{itemize}
\item LeNet+MNIST and LeNet+CIFAR10: Testing accuracy and training accuracy are both increasing during the training process. The FC2 layer in LeNet hits the critical value $s_*$ first, and provides a time which could stop. For MNIST, the test accuracy in the early stopped epoch only has the negligible difference $0.1\%$ with the final test accuracy; For CIFAR10, the test accuracy is lower but still guaranteed compared with the final test accuracy. We check the FC1 layer for CIFAR10, find that the ``strongly suggested'' stopping epochs in batch sizes 16 and 32 have much higher test accuracies, and in larger batch sizes, we could stop for saving time or keep training for a higher test accuracy. 

\item  MiniAlexNet+MNIST and MiniAlexNet+CIFAR10: The FC1 layer always hits the critical value first. For MNIST, the test accuracy still has negligible difference with the final test accuracy in small batch sizes 16 and 32, and no early stopping epoch is found by the spectral criterion in large batch sizes. For CIFAR10, it is the most representative experiment because the training explosion happens in batch sizes 16 and 32. We figure out whether spectral criterion gives alarm and performs well. The answer is yes. The spectral criterion strongly suggested to stop before training explosion and has a quite high test accuracy. In large batch sizes, the test accuracies are also ensured with a large amount of cutting training time.

\end{itemize}

Table \ref{tbl:earlystopingreal} shows the robustness and well performance of the spectral criterion. Actually, $C=0.4$ makes the critical value quite strict, even generated from MP Law, $\hs_nn^{\frac13}/\sqrt{\log n}$ could achieve the extreme value 0.35, any slight deviation from MP Law could hit the critical value, which we still consider it as MP Law or MP Law + Bleeding out. 

\begin{figure}[htbp]
\centering
\subfigure[LeMNIST]{
\begin{minipage}[t]{0.2\linewidth}
\centering
\includegraphics[width=1.2in]{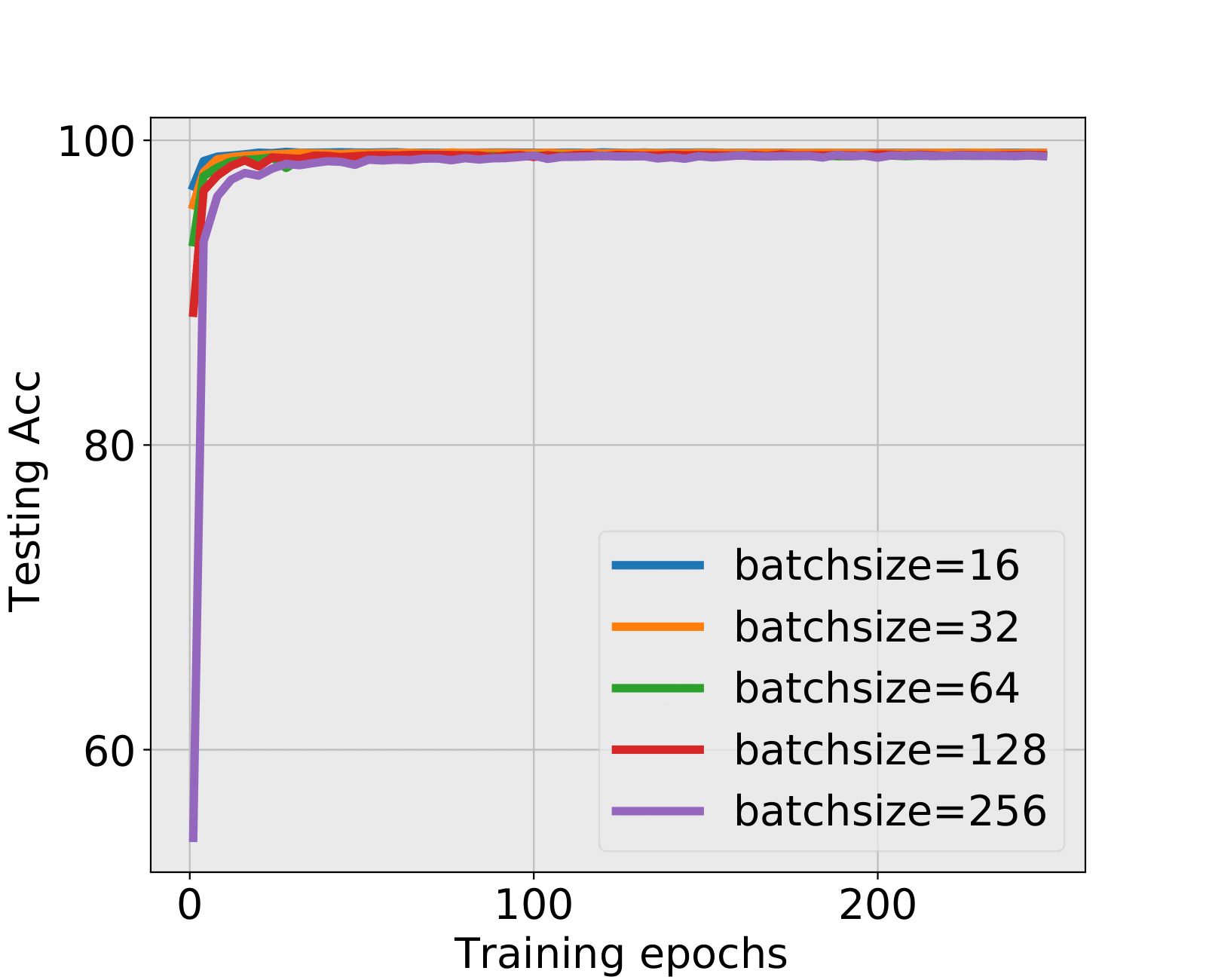}
\end{minipage}%
}%
\subfigure[LeCIFAR]{
\begin{minipage}[t]{0.2\linewidth}
\centering
\includegraphics[width=1.2in]{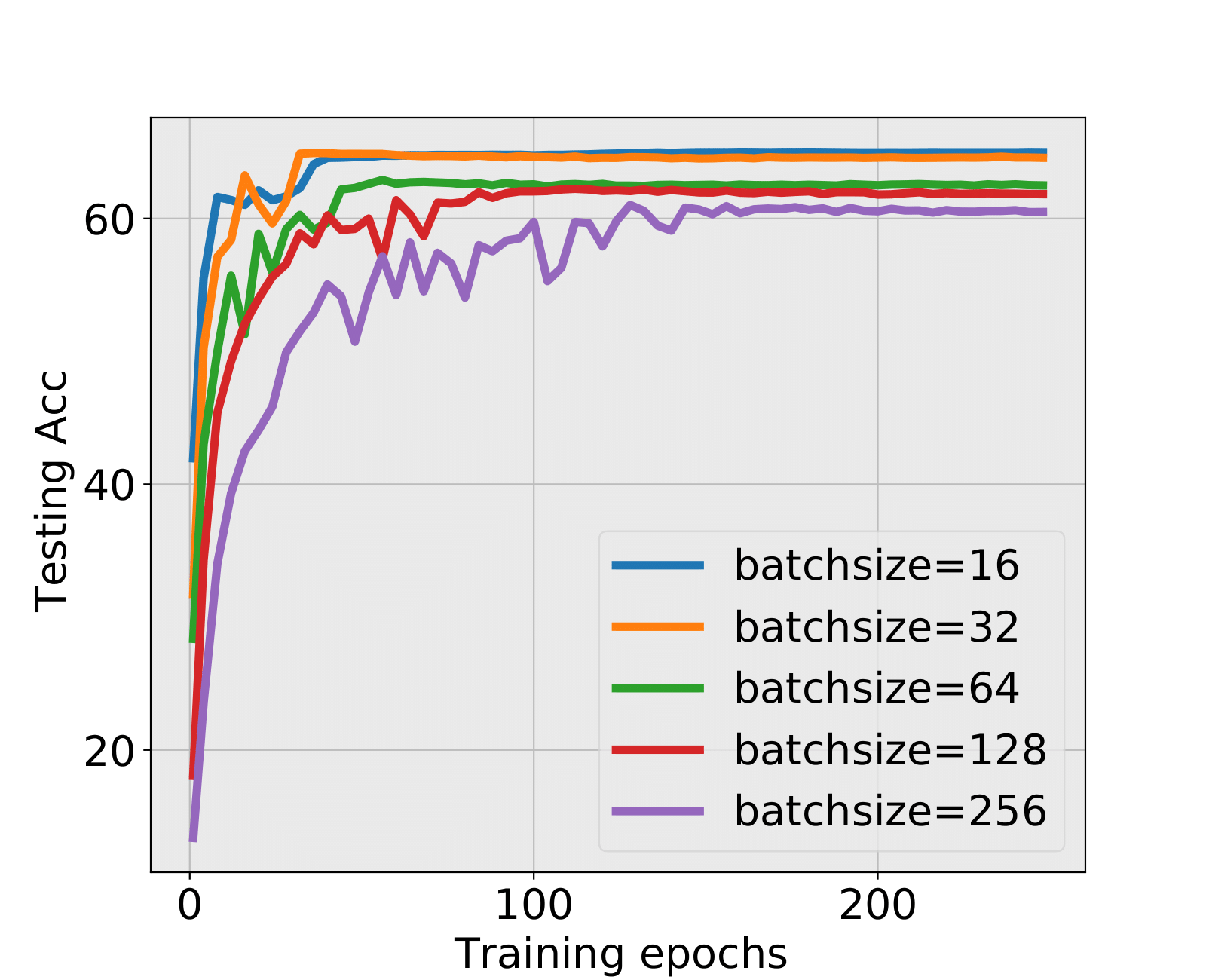}
\end{minipage}
}%
\subfigure[AlexMNIST]{
\begin{minipage}[t]{0.2\linewidth}
\centering
\includegraphics[width=1.2in]{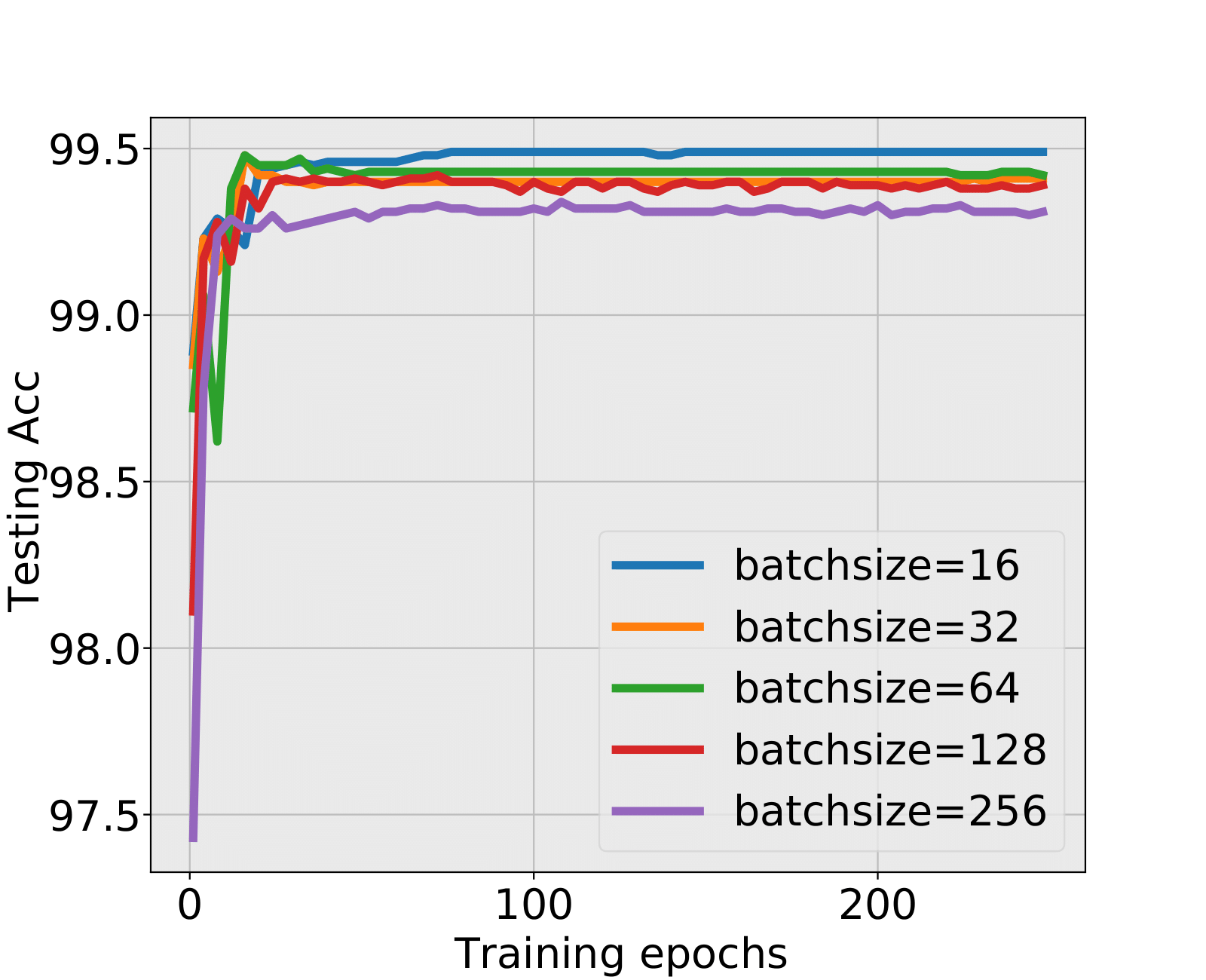}
\end{minipage}%
}%
\subfigure[AlexCIFAR]{
\begin{minipage}[t]{0.2\linewidth}
\centering
\includegraphics[width=1.2in]{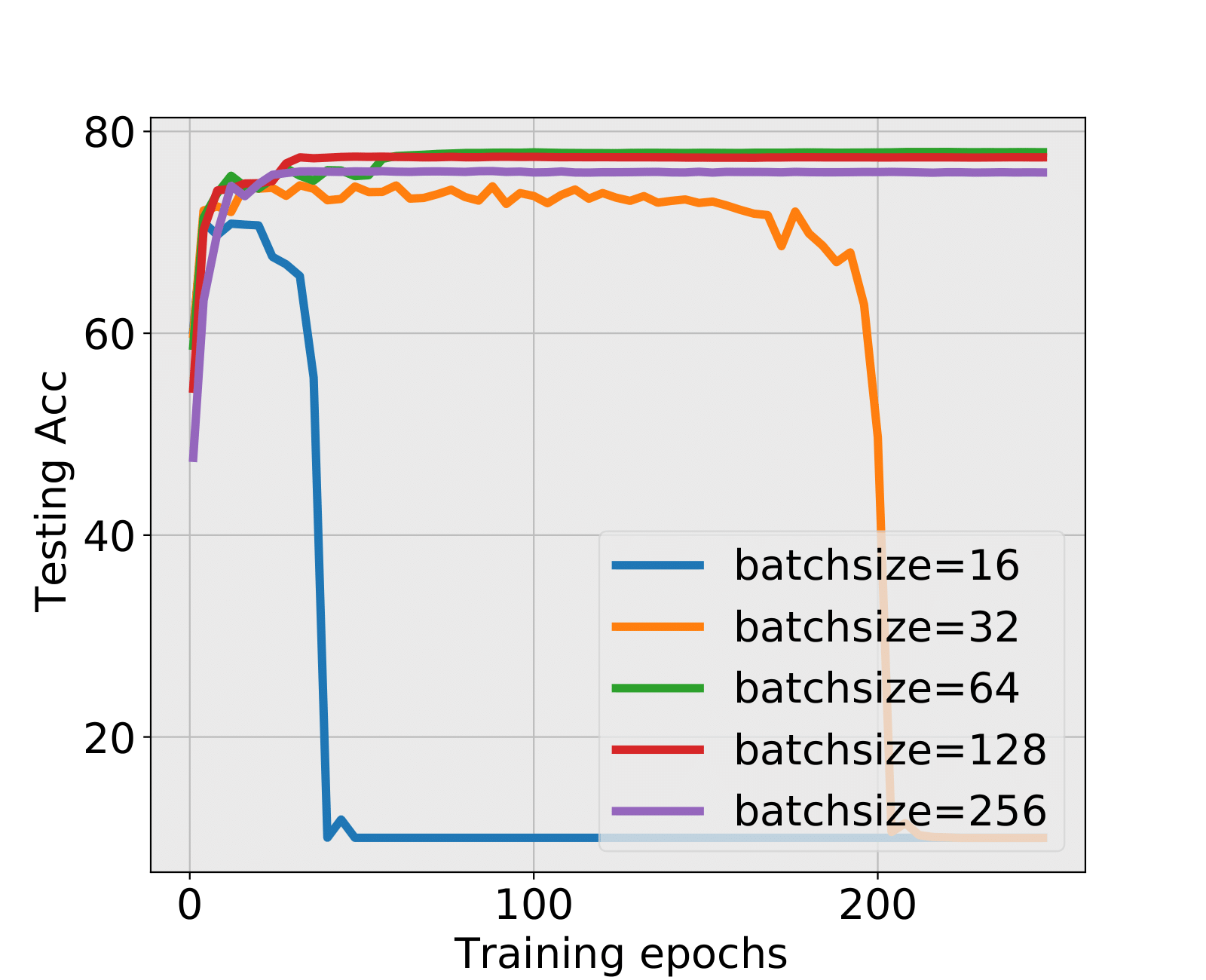}
\end{minipage}
}%

\subfigure[LeMNIST]{
\begin{minipage}[t]{0.2\linewidth}
\centering
\includegraphics[width=1.2in]{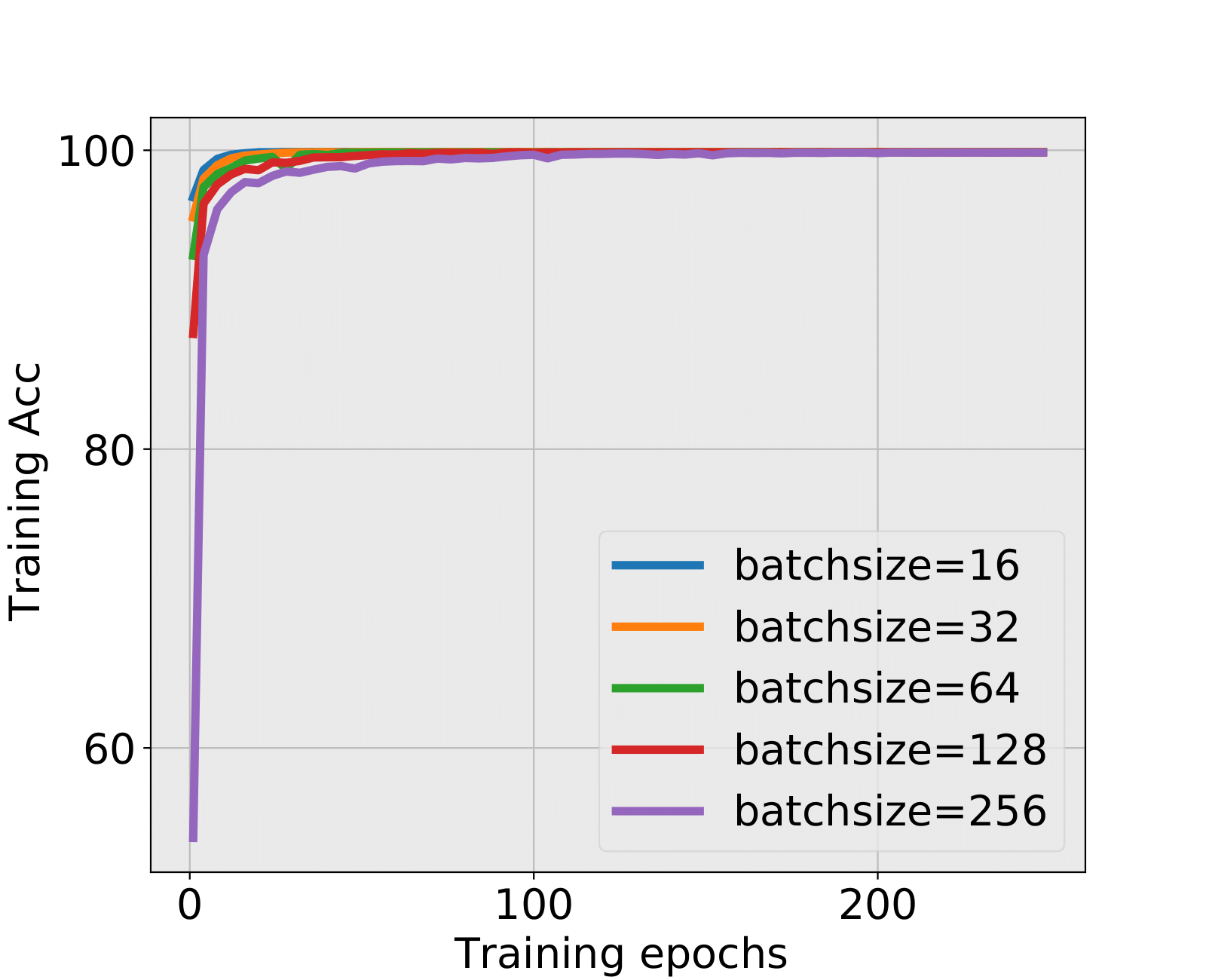}
\end{minipage}%
}%
\subfigure[LeCIFAR]{
\begin{minipage}[t]{0.2\linewidth}
\centering
\includegraphics[width=1.2in]{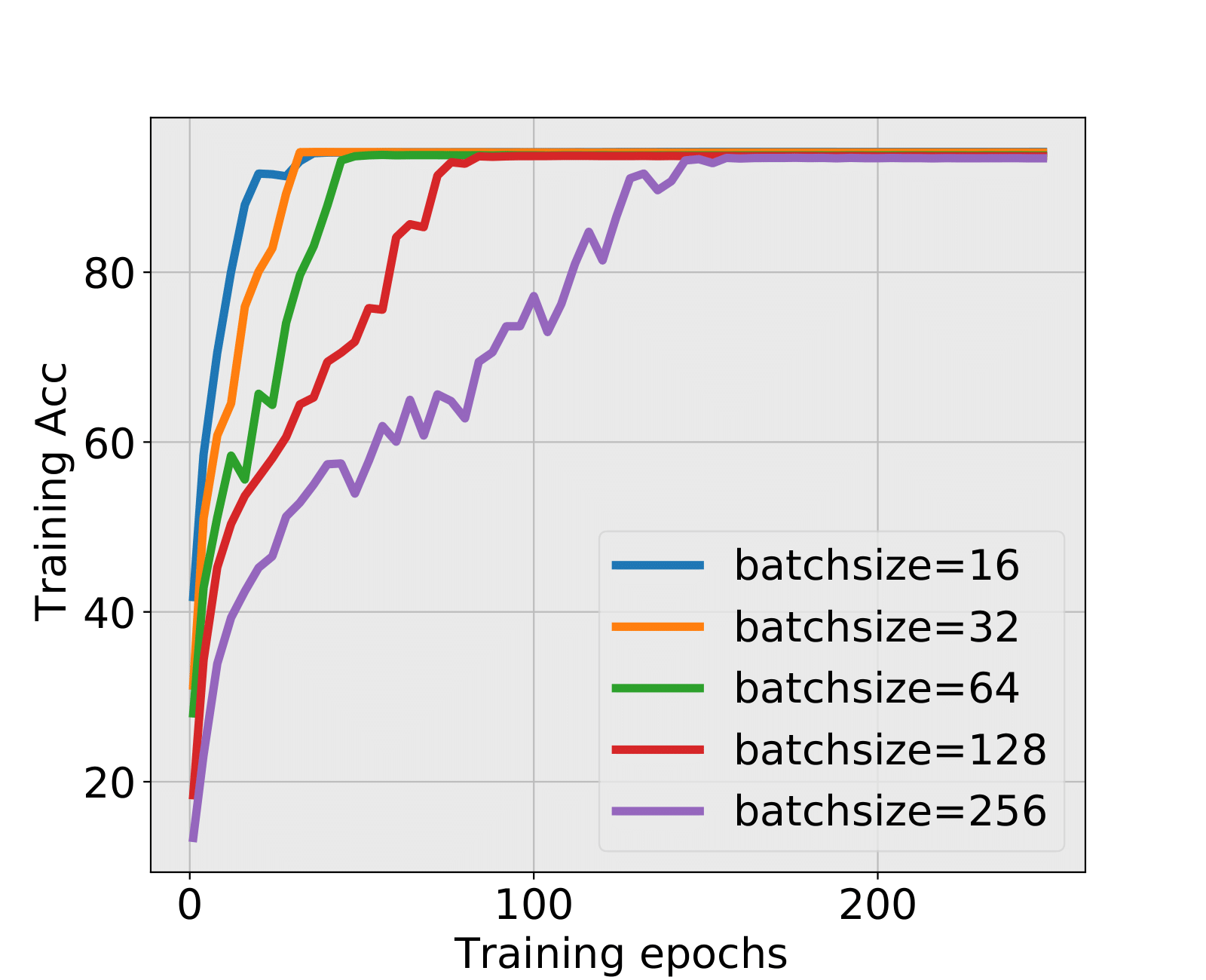}
\end{minipage}
}%
\subfigure[AlexMNIST]{
\begin{minipage}[t]{0.2\linewidth}
\centering
\includegraphics[width=1.2in]{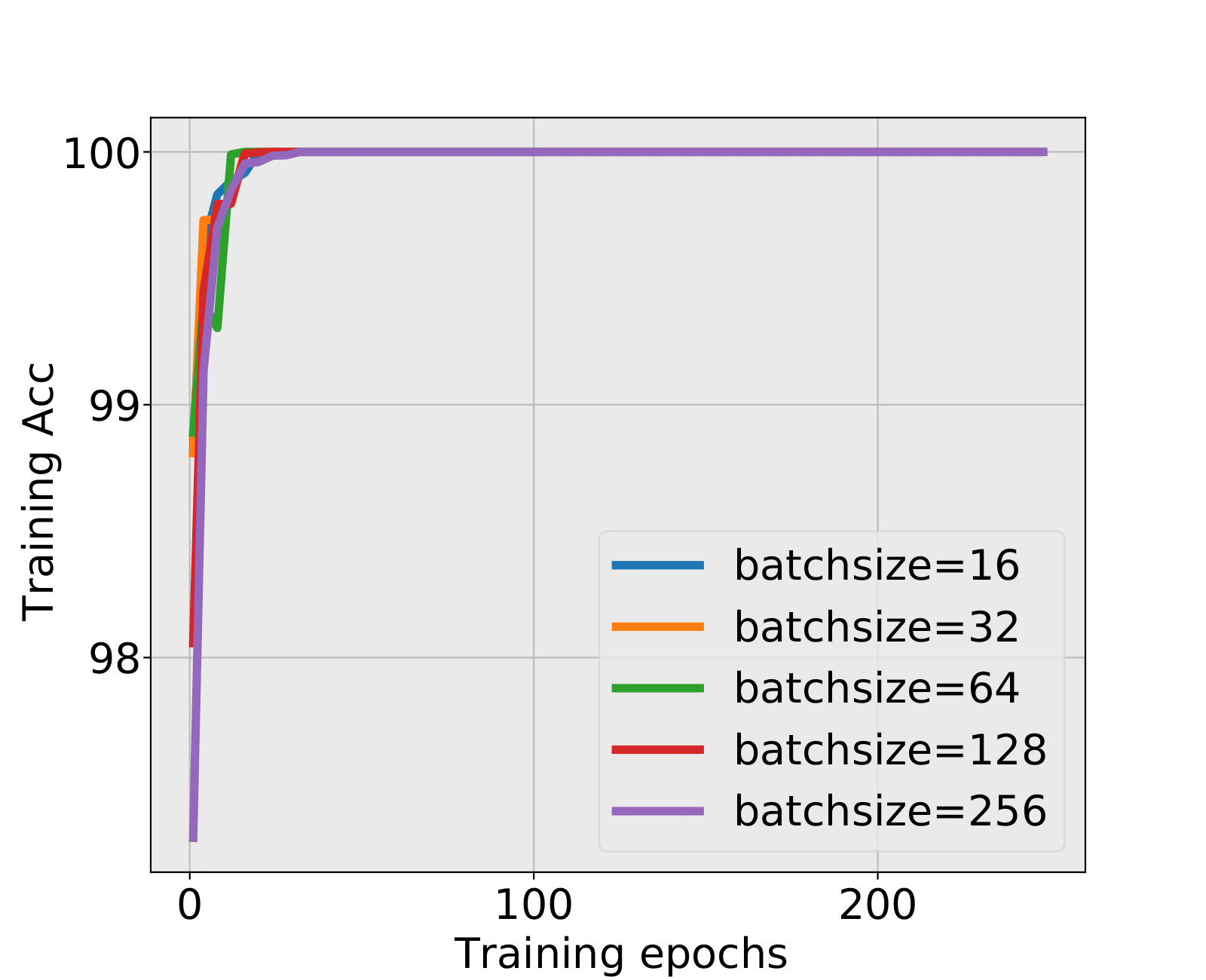}
\end{minipage}%
}%
\subfigure[AlexCIFAR]{
\begin{minipage}[t]{0.2\linewidth}
\centering
\includegraphics[width=1.2in]{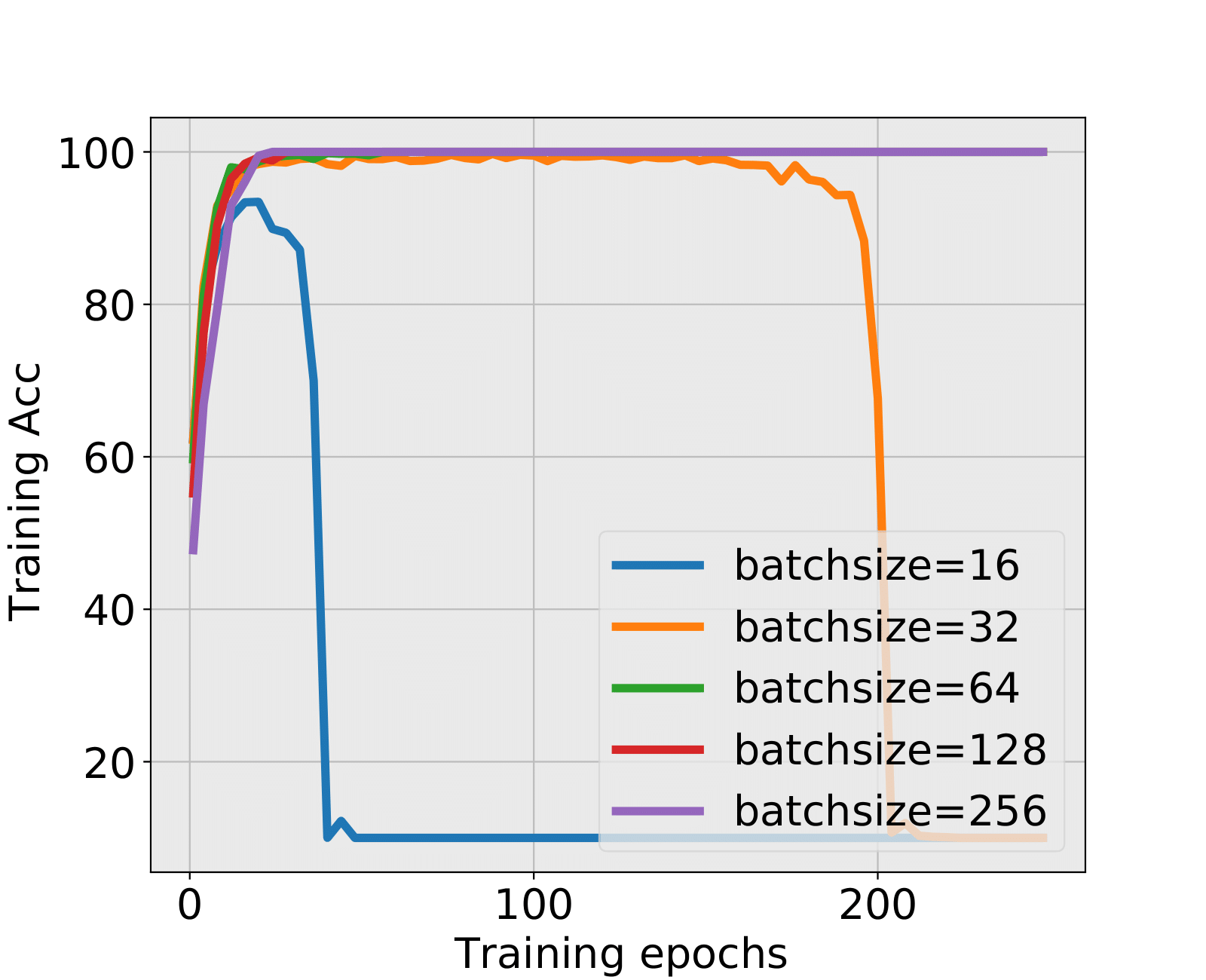}
\end{minipage}
}%

\centering
\caption{~~ Testing and Training Accuracy: We begin the line at epoch=1. Testing accuracy: (a)-(d); Training accuracy: (e)-(h). y-axis is the accuracy value, x-axis is the training epochs. Different line represents different batch sizes. (The notation ``LeMNIST'' means ``LeNet+MNIST'', same to ``LeCIFAR'', ``AlexMNIST'' and ``AlexCIFAR'').}
\label{spectrareal}
\end{figure}

We tune the constant $C=0.6$ to give further check. Details are shown in Appendix \ref{sec: specrit}.
An intersting thing is that when $C=0.6$, there is nearly no early stopping or training alarm on MNIST, and high quality epochs are still offered in the training of CIFAR10. We also note that with $C=0.6$, the \textbf{spectral criterion} predicts the explosion quite accurately in MiniAlexNet+CIFAR10, as the first hit on the critical line is epochs 28 and 188, the explosion begins. (We select epoch 36,196 for smoothness consideration, seen in Figure \ref{spectrareal}(d) and Table \ref{tbl:earlystopingreal1}.)

\begin{table}[htbp]
\centering
\caption{~~ Early stopping results in real data experiments with $C=0.4$: stopping epochs selected by spectral criterion in different layers' weight matrices and their testing accuracy (Test Acc). The symbol ”-” means no early stopping epoch is found by the spectral criterion.\label{tbl:earlystopingreal}}

\begin{tabular}{|c|c|c|c|c|c|c|c|}
\multicolumn{8}{c}{The combination LeNet+MNIST}\\
\hline
\multirow{2}{*}{batchsize} & \multicolumn{4}{c|}{spectral criterion $C=0.4$}   & \multicolumn{3}{c|}{Final Epoch 248} \\ \cline{2-8} 
                           & epoch(FC1)  & Test Acc & epoch(FC2)  & Test Acc & FC1       & FC2      & Test Acc      \\ \hline
16                         & \multicolumn{2}{c|}{-} & 16          & 99.08\%  & LT        & BT      & 99.17\%       \\ \hline
32                         & \multicolumn{2}{c|}{-} & 40          & 99.13\%  & LT        & BT      & 99.17\%       \\ \hline
64                         & \multicolumn{2}{c|}{-} & 68          & 98.98\%  & LT        & BT      & 98.98\%       \\ \hline
128                        & \multicolumn{2}{c|}{-} & 124         & 98.91\%  & LT        & BT       & 99.03\%       \\ \hline
256                        & \multicolumn{2}{c|}{-} & \multicolumn{2}{c|}{-} & LT        & LT       & 98.96\%       \\ \hline
\end{tabular}
\vskip 2mm
\begin{tabular}{|c|c|c|c|c|c|c|c|}
\multicolumn{8}{c}{The combination LeNet+CIFAR10}\\
\hline
\multirow{2}{*}{batchsize} & \multicolumn{4}{c|}{spectral criterion $C=0.4$}  & \multicolumn{3}{c|}{Final Epoch 248} \\ \cline{2-8} 
                           & epoch(FC1)  & Test Acc & epoch(FC2) & Test Acc & FC1       & FC2      & Test Acc      \\ \hline
16                         & 24          & 61.37\%  & 8          & 61.62\%  & BT       & HT       & 64.99\%       \\ \hline
32                         & 60          & 64.78\%  & 10         & 57.94\%  & BT        & HT       & 64.57\%       \\ \hline
64                         & \multicolumn{2}{c|}{-} & 28         & 59.19\%  & LT        & BT       & 62.49\%       \\ \hline
128                        & \multicolumn{2}{c|}{-} & 60         & 61.38\%  & LT        & BT       & 61.83\%       \\ \hline
256                        & \multicolumn{2}{c|}{-} & 84         & 54.23\%  & LT        & BT       & 60.49\%       \\ \hline
\end{tabular}

\vskip 2mm
\begin{tabular}{|c|c|c|c|c|c|c|c|}
\multicolumn{8}{c}{The combination MiniAlexNet+MNIST}\\
\hline
\multirow{2}{*}{batchsize} & \multicolumn{4}{c|}{spectral criterion $C=0.4$}   & \multicolumn{3}{c|}{Final Epoch 248} \\ \cline{2-8} 
                           & epoch(FC1)  & Test Acc & epoch(FC2)  & Test Acc & FC1       & FC2      & Test Acc      \\ \hline
16                         & 4           & 99.23\%  & \multicolumn{2}{c|}{-} & BT       & LT       & 99.49\%       \\ \hline
32                         & 20          & 99.42\%  & \multicolumn{2}{c|}{-} & BT        & LT       & 99.41\%       \\ \hline
64                         & \multicolumn{2}{c|}{-} & \multicolumn{2}{c|}{-} & LT        & LT       & 99.42\%       \\ \hline
128                        & \multicolumn{2}{c|}{-} & \multicolumn{2}{c|}{-} & LT        & LT       & 99.39\%       \\ \hline
256                        & \multicolumn{2}{c|}{-} & \multicolumn{2}{c|}{-} & LT        & LT       & 99.31\%       \\ \hline
\end{tabular}

\vskip 2mm
\begin{tabular}{|c|c|c|c|c|c|c|c|}
\multicolumn{8}{c}{The combination MiniAlexNet+CIFAR10}\\
\hline
\multirow{2}{*}{batchsize} & \multicolumn{4}{c|}{spectral criterion $C=0.4$}              & \multicolumn{3}{c|}{Final Epoch 248} \\ \cline{2-8} 
                           & epoch(FC1) & Test Acc & epoch(FC2)  & Test Acc & FC1     & FC2     & Test Acc         \\ \hline
16                         & 3          & 69.05\%  & 9           & 72.02\%  & HT      & RC      & 10\%(explode)    \\ \hline
32                         & 4          & 72.17\%  & 16          & 74.64\%  & HT      & RC      & 10\%(explode)    \\ \hline
64                         & 5          & 71.61\%  & 28          & 76.35\%  & BT      & BT      & 77.94\%          \\ \hline
128                        & 10         & 74.14\%  & \multicolumn{2}{c|}{-} & BT      & LT      & 77.43\%          \\ \hline
256                        & 24         & 75.70\%  & \multicolumn{2}{c|}{-} & BT     & LT      & 75.93\%          \\ \hline
\end{tabular}
\end{table}

\section{Conclusion}
\pagestyle{plain}

Data classification difficulty has a great impact on the spectra of weight matrices. We study it from three aspects: SNR, class numbers and complex features, and find that more difficult to classify, higher probability HT emerges. Further, in line with \cite{martin2018implicit}, HT could be regarded as a training information encoder and indicates some implicit regularization. Such implicit regularization in the weight
matrices provides  a new way of  understanding  the whole training procedure.  We apply this phenomenon to derive an early stopping
procedure in order to avoid over-training (in case of poor data
quality) and cut off large training time. From the encoded information, spectral criterion could even provide an early stopped time when the training accuracy is increasing. 

Spectral Analysis provides a new way for the understanding of Deep Learning. Once understood spectra in DNNs better, more practical guidance for the whole learning procedures may be provided. Our study in weight matrices spectra also leads to several questions to explore in the future, such as how SGD generates Heavy Tail in the poor data quality but Light Tail in the high data quality in DNNs.





\newpage
\bibliography{ref}
\newpage


\appendix
\begin{center}
\setlength{\baselineskip}{25pt}
\vspace{0.2in}
\noindent{\fontsize{18pt}{1em}\selectfont \textbf{Supplementary materials to `Impact of classification difficulty on the weight matrices spectra in Deep Learning'  }}\\[12pt]

{\fontsize{14pt}{1.2em}\selectfont
Xuran Meng, Jianfeng Yao
\\[10pt]
\textit{\small{Department of Statistics and Actuarial Science, University of Hong Kong,\\
Hong Kong SAR, China}}
\\[10pt]
\textsuperscript{*} \textit{\small{To whom correspondence should be addressed: jeffyao@hku.hk}} \\[10pt]
\date{}
}
\end{center}

\section{Preliminaries on RMT}\label{sec: RMT}
\pagestyle{plain}
In this section, we review the results we use from RMT. This section is aim to give readers some references to know about RMT. The reference is \cite{yao2015sample}. RMT provides us lots of analytic results for both square or rectangular large matrices.  The well-known results in RMT: Marchenko-Pauster (MP) Law which shows eigenvalues' distribution of rectangular matrices, and Tracy-Widom Law which describes how the max eigenvalue distributed. When analyze DNNs weight matrices, MP Law, which is applicable to rectangular matrices, will give guidance on the analysis in spectrum of DNNs weight matrices.

\vspace{4pt}

\subsection{MP Law and stieltjes Transform}
MP Law is given as follows, 
\begin{theorem}
\label{thm: mpLaw}
(Marchenko Pauster Law) Suppose that the entries $\{w_{ij}\}$ of the matrix $W\in \bbR^{n\times p}$ are $i.i.d.$ complex random variables with mean zero and variance $\sigma^2$, and $n/p\to c\in (0,\infty)$.  Define $S_p=\frac{1}{p}WW^*$, $\lambda_1\geq\lambda_2\geq...\geq\lambda_n$ is the eigenvalue of $S_p$, then the ESD of $S_p$
$$
F^{S_p}\Def\frac{1}{n}\sum_jI(x<\lambda_j)\cvas F_{c,\sigma^2}(x) 
$$
as $p\to\infty$, here $I(\cdot)$ is the indicator function, $F_{c,\sigma^2}$ is the MP Law has the form
$$
F_{c,\sigma^2}(x)=\left\{\begin{aligned}
    &\frac{1}{2\pi xc\sigma^2}\sqrt{(b-x)(x-a)},\ \ \ \ &\text{if}\ a\leq x\leq b,\\
    &0,&\text{otherwise},
\end{aligned}\right.
$$
with an additional point mass of value $1-1/c$ at the origin if $c>1$, where $a=\sigma^2(1-\sqrt{c})^2$ and $b=\sigma^2(1+\sqrt{c})^2$.
\end{theorem}

The powerful mathematical tool to achieve the MP Law is stieltjes Transform, let $\nu$ be a finite measure on the real line with support $\Gamma_\nu$, its stieltjes transform is
$$
s_\nu(z)=\int \frac{1}{x-z}\nu(dx),\ \ z\in\bbC\backslash \Gamma_\nu.
$$
To reduct $\nu$ from Stieltjes transform, we have the equation
$$
\nu(x)=\lim\limits_{v\to0^+}\frac{1}{\pi}\Im s_\nu(x+iv).
$$
Details can be refered in \citep{yao2015sample}.

The Stieltjes transform of standard MP Law satisfies equation
$$
s(z)=\frac{1}{1-z-\{c+czs(z)\}},
$$
we can get results in a more generalized settings:
\begin{theorem}
\label{thm: GmpLaw}
(Generalized MP Law) Let $S_p$ be the matrix defined in Theorem \ref{thm: mpLaw}, $(T_p)$ is a sequence of nonnegative Hermitian matrices of size $n\times n$ which is deterministic or independent with $S_p$, and the sequence $(H_p=F^{T_p})$ of the ESD of $(T_p)$ almost surely weakly converges to a nonrandom probability measure $H(t)$, then the ESD $(F^{S_pT_p})$ of $S_pT_p$ weakly converges to a nonrandom probability measure $F_{c,H}$, that its Stieltjes transform is implicitly defined by the equation
$$
s(z)=\int\frac{1}{t(1-c-czs(z))-z}dH(t),\ \ z\in\bbC^+
$$
\end{theorem}

\vspace{4pt}

\subsection{Spike Model}

Suppose that the matrix $T_p$ has the limit ESD $H(t)$, the element $w_{ij}$ in $W$ satisfies
$$
Ew_{11}=0,Ew_{11}^2=1,Ew_{11}^4<+\infty
$$
define 
$$
\psi(\alpha)=\psi_{c,H}(\alpha)=\alpha+c\int\frac{t\alpha}{\alpha-t}dH(t),
$$
we say the eigenvalue $\alpha_j$ of $T_p$ with fixed index $j$ is a distant spike eigenvalue if
$$
\psi'(\alpha_j)>0.
$$
Then we have the following theorem,
\begin{theorem}
 The entries $\{w_{ij}\}$ of the matrix $W\in \bbR^{n\times p}$ are $i.i.d.$ complex random variables with mean zero, variance 1 and finite fourth moment, $n/p\to c\in (0,\infty)$.  $S_p=\frac{1}{p}WW^*$, $(T_p)$ is a sequence of nonnegative Hermitian matrices of size $n\times n$ which is deterministic or independent with $S_p$, and the sequence $(H_p=F^{T_p})$ of the ESD of $(T_p)$ almost surely weakly converges to a nonrandom probability measure $H(t)$,  $\lambda_1\geq\lambda_2\geq...\geq\lambda_n$ is the eigenvalue of $S_pT_p$, $\alpha_1\geq\alpha_2\geq...\geq\alpha_n$ is the eigenvalue of $T_p$, suppose that $\alpha_j$ with fixed index $j$ is a distant spike eigenvalue, then as $p\to+\infty$,\newline
 \text{(1)} if $\alpha_1<+\infty$, then $\lambda_j\cvas\psi(\alpha_j)$,\newline
 
\noindent\text{(2)} if $\alpha_j\to+\infty$, then $\lambda_j/\alpha_j\cvas1$.
\end{theorem}

\vspace{4pt}

\subsection{Tracy-Widom Law}

\begin{theorem}
(Tracy-Widom Law) Let $\{w_{i,j}\}$, $i,j=1,2,...,$ be a double array of i.i.d. complex-valued random variables with mean $0$, variance $1$ and finite fourth-order moment. Consider the sample covariance matrix $S_p$ defined same as Theorem \ref{thm: mpLaw}, the eigenvalues of $S_p$: $\lambda_1\geq\lambda_2\geq...\geq\lambda_n$ in a decreasing order. When $n/p\to c$, we have
$$
\lambda_1\cvas (1+\sqrt{c})^2,\quad (order\ 0)
$$
further, define $\mu_{pn}=\frac{1}{p}\{(p-1)^{\frac12}+n^{\frac12}\}^2$ and $\sigma_{pn}=\frac{1}{p}[(p-1)^{\frac12}+n^{\frac12}]\{(p-1)^{-\frac12}+n^{-\frac12}\}^{\frac13}$,
$$
\frac{\lambda_1-\mu_{pn}}{\sigma_{pn}}\cvd F_1,\quad (order\ 1)
$$
where $F_1$ is the Tracy-Widom Law of order 1 whose distribution function is given by 
$$
F_1(s)=\exp{\left\{\int_s^{+\infty}q(x)+(x-s)^2q^2(x)dx\right\}},\quad s\in\bbR,
$$
where $q$ solves the Painleve \uppercase\expandafter{\romannumeral2} differential equation
$$
q''(x)=xq(x)+2q^3(x).
$$
with boundary condition
$$
q(s)\sim Ai(s),~s\to+\infty,
$$
here $Ai(s)$ is the airy function.
\end{theorem}

\section{Technique Proof}
In this section, we prove the proposition in section \ref{sec:specCriterion}. 
\subsection{\textbf{Proof of proposition \ref{prop1}}} \label{sec:proof-prop1}

From definition, $\hp_M\left(x\right)=\frac{M}{n\left(b-a\right)}\sum\limits_{i=1}^nI\left(X_i\in B\left(x\right)\right)$, 
$$\begin{aligned}
s_n&=\int_{a}^{b}\left|\hp_M\left(x\right)-p\left(x\right)\right|dx\\
&\leq \int_{a}^{b}\left|E\hp_M\left(x\right)-p\left(x\right)\right|dx+\int_{a}^{b}\left|\hp_M\left(x\right)-E\hp_M\left(x\right)\right|dx.
\end{aligned}$$

\vspace{2pt}

We first prove $\int_{a}^{b}\left|E\hp_M\left(x\right)-p\left(x\right)\right|dx=O\left(\frac{1}{M}\right)$.

\vspace{2pt}

There exists $x^*\in B_1$ such that $P\left(X_i\in B_1\right)=C\int_{a}^{a+\frac{b-a}{M}}\frac{\sqrt{\left(b-x\right)\left(x-a\right)}}{x}dx=C*\frac{b-a}{M}*\frac{\sqrt{\left(b-x^*\right)\left(x^*-a\right)}}{x^*}\leq C*\frac{1}{M}*\sqrt{\frac{1}{M}}$, thus
$$\begin{aligned}
P\left(X_i\in B_1\right)&=O\left(\frac{1}{M^{\frac{3}{2}}}\right),
\end{aligned}$$
similar to get $P\left(X_i\in B_M\right)=O\left(\frac{1}{M^{\frac{3}{2}}}\right)$, then
$$\begin{aligned}
\int_{B_1\cup B_M}\left|E\hp_M\left(x\right)-p\left(x\right)\right|dx&\leq \int_{B_1\cup B_M}E\hp_M\left(x\right)+p\left(x\right)dx\\
&=\int_{B_1\cup B_M}\frac{M}{b-a}P\left(X_i\in B_1\cup B_M\right)dx+P\left(X_i\in B_1\cup B_M\right)\\
&=O\left(\frac{1}{M^{\frac{3}{2}}}\right).
\end{aligned}$$

For $\forall x\in B_l$, $l\in\{2,3,...,M-1\}$, there exists $x^*_l\in B_l$ satisfies
$$\begin{aligned}
E\hp_M\left(x\right)&=\frac{M}{b-a}P\left(X_i\in\left(B\left(x\right)\right)\right)\\
&=\frac{\int_{a+\frac{l}{M}\left(b-a\right)}^{a+\frac{l+1}{M}\left(b-a\right)}p\left(u\right)du}{\left(b-a\right)/M}=p\left(x^*_l\right),
\end{aligned}$$
then for $\forall x\in B_l$, there exists $x^{**}_l$ between $x$ and $x^*_l$, such that 
$$\begin{aligned}
\left|E\hp_M\left(x\right)-p\left(x\right)\right|&=\left|p\left(x^*_l\right)-p\left(x\right)\right|= \left|p'\left(x^{**}_l\right)\left(x^{*}_l-x\right)\right|\\
&\leq C*\frac{\left|x^{**}_l-\left(x^{**}_l-a\right)\left(b-x^{**}_l\right)\right|}{\sqrt{\left(x^{**}_l-a\right)\left(b-x^{**}_l\right)}\left(x^{**}_l\right)^2}*\left|x^*_l-x\right|\\
&\leq\frac{C}{\sqrt{lM}},
\end{aligned}$$
$C$ is an absolute value. Then 
$$\begin{aligned}
\int_{B_1^C\cap B_M^C}\left|E\hp_M\left(x\right)-p\left(x\right)\right|dx&=\sum\limits_{l=2}^{M-1}\int_{B_l}\left|E\hp_M\left(x\right)-p\left(x\right)\right|dx\\
&\leq \sum\limits_{l=2}^{M-1}\int_{B_l}\frac{1}{\sqrt{l}}\cdot \frac{C}{\sqrt M}dx\\
&=\sum\limits_{l=2}^{M-1}\frac{1}{\sqrt{l}}\cdot \frac{C}{\sqrt M}\cdot\frac{1}{M}\\
&=O\left(\frac{1}{M}\right).
\end{aligned}$$
Thus
$$\begin{aligned}
\int_{a}^{b}\left|E\hp_M\left(x\right)-p\left(x\right)\right|dx&=\int_{B_1\cup B_M}\left|E\hp_M\left(x\right)-p\left(x\right)\right|dx\\
&+\int_{B_1^C\cap B_M^C}\left|E\hp_M\left(x\right)-p\left(x\right)\right|dx\\
&=O\left(\frac1M\right).
\end{aligned}$$

We next prove that $\int_{a}^{b}\left|\hp_M\left(x\right)-E\hp_M\left(x\right)\right|dx=O_p\left(\sqrt{\frac{M\log n}{n}}\right)$,

$$\begin{aligned}
P\left(\sup\limits_x\left|\hp_M\left(x\right)-E\hp_M\left(x\right)\right|>\varepsilon\right)&=P\left(M\cdot\max\limits_{l=1,...,M}\frac{1}{n}\left|\sum\limits_{i=1}^nI\left(X_i\in B_l\right)-nP\left(X_i\in B_l\right)\right|>\varepsilon\right)\\
&=P\left(\max\limits_{l=1,...,M}\frac{1}{n}\left|\sum\limits_{i=1}^nI\left(X_i\in B_l\right)-nP\left(X_i\in B_l\right)\right|>\frac{\varepsilon}{M}\right)\\
&\leq \sum\limits_{l=1}^{M}P\left(\frac{1}{n}\left|\sum\limits_{i=1}^nI\left(X_i\in B_l\right)-nP\left(X_i\in B_l\right)\right|>\frac{\varepsilon}{M}\right).
\end{aligned}$$
Note that $P\left(X_i\in B_l\right)=O\left(1/M\right)$, by Berstein inequality, 
$$\begin{aligned}
\sum\limits_{l=1}^{M}P\left(\frac{1}{n}\left|\sum\limits_{i=1}^nI\left(X_i\in B_l\right)-nP\left(X_i\in B_l\right)\right|>\frac{\varepsilon}{M}\right)
&\leq \sum\limits_{l=1}^{M}e^{-C\frac{\frac{n^2\varepsilon^2}{M^2}}{\frac{n}{M}+\frac{n\varepsilon}{3M}}}\\
&\leq Me^{-C\frac{n\varepsilon}{M}},
\end{aligned}$$
for some constant $C>0$, set $\varepsilon=\sqrt{\frac{M\log n}{n}}$, we achieve 
$$
\sup\limits_{x\in\left[a,b\right]}\left|\hp_M\left(x\right)-E\hp_M\left(x\right)\right|=O_p\left(\sqrt{\frac{M\log n}{n}}\right),
$$
which implies given $\forall$ $\varepsilon_0>0$, there $\exists R>0$, such that
$$
P\left(\sqrt{\frac{n}{M\log n}}\sup\limits_{x\in\left[a,b\right]}|\hp_M\left(x\right)-E\hp_M\left(x\right)|>R\right)<\varepsilon_0,
$$
for $\int_{a}^{b}\left|\hp_M\left(x\right)-E\hp_M\left(x\right)\right|dx$, there exists $x_n$ satisfies
$$
\left|\hp_M\left(x_n\right)-E\hp_M\left(x_n\right)\right|\left(b-a\right)\geq \int_{a}^{b}\left|\hp_M\left(x\right)-E\hp_M\left(x\right)\right|dx,
$$
then
$$\begin{aligned}
\varepsilon_0&>P\left(\sqrt{\frac{n}{M\log n}}\sup\limits_{x\in\left[a,b\right]}|\hp_M\left(x\right)-E\hp_M\left(x\right)|>R\right)\\
&\geq P\left(\sqrt{\frac{n}{M\log n}}|\hp_M\left(x_n\right)-E\hp_M\left(x_n\right)|\left(b-a\right)>R\left(b-a\right)\right)\\
&\geq P\left(\sqrt{\frac{n}{M\log n}}\int_{a}^{b}\left|\hp_M\left(x\right)-E\hp_M\left(x\right)\right|dx>R\left(b-a\right)\right).\\
\end{aligned}$$
Thus we achieve
$$
\int_{a}^{b}\left|\hp_M\left(x\right)-E\hp_M\left(x\right)\right|dx=O_p\left(\sqrt{\frac{M\log n}{n}}\right),
$$
which proves the proposition.
\newpage

\subsection{\textbf{Proof of proposition \ref{prop2}}} \label{sec:proof-prop2}

Recall $X_{\left(i\right)}$ is the order statistics. For $\forall \varepsilon>0$, $\exists R>0$ such that 
$$\begin{aligned}
P\left(n^{\frac{2}{3}}\left(X_{\left(1\right)}-a\right)>R\right)&=\prod\limits_{i=1}^nP\left(X_{i}-a>R/n^{\frac23}\right)\\
&=\left(1-\int_a^{a+\frac{R}{n^{\frac23}}}p\left(x\right)dx\right)^n\\
&\sim\left(1-C\int_a^{a+\frac{R}{n^{\frac23}}}\sqrt{x-a}dx\right)^n\\
&=\left(1-CR^{\frac32}/n\right)^n\\
&<\varepsilon,
\end{aligned}$$
as $n\to\infty$, then $X_{\left(1\right)}-a=O_p\left(n^{-\frac{2}{3}}\right)$, similar to get $b-X_{\left(n\right)}=O_p\left(n^{-\frac23}\right)$, which implies same rate as Tracy Widom Law.

$$
\hs_n\leq s_n+\int_a^b|\hp\left(x\right)-p\left(x\right)|dx,
$$
The only thing is to prove
$$
\int_a^b|\hp\left(x\right)-p\left(x\right)|dx=O_p\left(n^{-\frac13}\right).
$$

Indeed, set $C_0=2\pi\sigma^2c$, $\hC_0=\frac{\pi\left(\sqrt{X_{\left(n\right)}}-\sqrt{X_{\left(1\right)}}\right)^2}{2}$, we have
$$\begin{aligned}
\int_a^b\left|\hp\left(x\right)-p\left(x\right)\right|dx&=\int_a^b\left|\frac{\sqrt{\left(b-x\right)\left(x-a\right)}}{C_0x}-\frac{\sqrt{\left(X_{\left(n\right)}-x\right)\left(x-X_{\left(1\right)}\right)}}{\hC_0x}I\left(\left[X_{\left(1\right)},X_{\left(n\right)}\right]\right)\right|dx\\
&\leq \int_a^b\left|\frac{1}{\hC_0}\right|\cdot\left|\frac{\sqrt{\left(b-x\right)\left(x-a\right)}}{x}-\frac{\sqrt{\left(X_{\left(n\right)}-x\right)\left(x-X_{\left(1\right)}\right)}}{x}I\left(\left[X_{\left(1\right)},X_{\left(n\right)}\right]\right)\right|dx\\
&+\left|\frac{1}{C_0}-\frac{1}{\hC_0}\right|\Def P_1+P_2,
\end{aligned}$$
where $P_1=\int_a^b\left|\frac{\sqrt{\left(b-x\right)\left(x-a\right)}}{C_0x}-\frac{\sqrt{\left(X_{\left(n\right)}-x\right)\left(x-X_{\left(1\right)}\right)}}{\hC_0x}I\left(\left[X_{\left(1\right)},X_{\left(n\right)}\right]\right)\right|dx$, $P_2=\left|\frac{1}{C_0}-\frac{1}{\hC_0}\right|$. Note that $\hC_0\cvp C_0>0$, by continuous mapping, $\frac{1}{\hC_0}\cvp \frac{1}{C_0}>0$, thus $\frac{1}{\hC_0}=O_p\left(1\right)$. Then for $P_2=\left|\frac{1}{C_0}-\frac{1}{\hC_0}\right|$, we have 
$$\begin{aligned}
\left|\frac{1}{C_0}-\frac{1}{\hC_0}\right|&=\frac{\pi\left(\left(\sqrt{b}-\sqrt{a}~\right)^2-\left(\sqrt{X_{\left(n\right)}}-\sqrt{X_{\left(1\right)}}\right)^2\right)}{2C_0\hC_0}\\
&\leq O_p\left(1\right)\cdot O_p\left(\sqrt{b}-\sqrt{X_{\left(n\right)}}+\sqrt{X_{\left(1\right)}}-\sqrt{a}\right)\\
&=O_p\left(n^{-\frac{1}{3}}\right).
\end{aligned}$$

For the first part, the integration can be separated as 
$$\begin{aligned}
P_1&=\int_a^{X_{\left(1\right)}}\left|\frac{1}{\hC_0}\right|\cdot\left|\frac{\sqrt{\left(b-x\right)\left(x-a\right)}}{x}-\frac{\sqrt{\left(X_{\left(n\right)}-x\right)\left(x-X_{\left(1\right)}\right)}}{x}I\left(\left[X_{\left(1\right)},X_{\left(n\right)}\right]\right)\right|dx\\
&+\int_{X_{\left(1\right)}}^{X_{\left(n\right)}}\left|\frac{1}{\hC_0}\right|\cdot\left|\frac{\sqrt{\left(b-x\right)\left(x-a\right)}}{x}-\frac{\sqrt{\left(X_{\left(n\right)}-x\right)\left(x-X_{\left(1\right)}\right)}}{x}I\left(\left[X_{\left(1\right)},X_{\left(n\right)}\right]\right)\right|dx\\
&+\int_{X_{\left(n\right)}}^b\left|\frac{1}{\hC_0}\right|\cdot\left|\frac{\sqrt{\left(b-x\right)\left(x-a\right)}}{x}-\frac{\sqrt{\left(X_{\left(n\right)}-x\right)\left(x-X_{\left(1\right)}\right)}}{x}I\left(\left[X_{\left(1\right)},X_{\left(n\right)}\right]\right)\right|dx\\
&\Def P_{11}+P_{12}+P_{13},
\end{aligned}$$
there exists $x^*\in \left(a,X\left(1\right)\right)$ such that 
$$\begin{aligned}
P_{11}=\left|\frac{1}{\hC_0}\right|\cdot\left|\frac{\sqrt{\left(b-x^*\right)\left(x^*-a\right)}}{x^*}\right|\left(X_{\left(1\right)}-a\right)=O_p\left(n^{-\frac23}\right),
\end{aligned}$$
similar to get 

$$\begin{aligned}
P_{13}=O_p\left(n^{-\frac23}\right),
\end{aligned}$$

$$\begin{aligned}
P_{12}&=O_p\left(\int_{X_{\left(1\right)}}^{X_{\left(n\right)}}\left|\frac{\left(b-a-X_{\left(n\right)}+X_{\left(1\right)}\right)x-ab+X_{\left(n\right)}X_{\left(1\right)}}{x\left(\sqrt{\left(b-x\right)\left(x-a\right)}+\sqrt{\left(X_{\left(n\right)}-x\right)\left(x-X_{\left(1\right)}\right)}\right)}\right|dx\right)\\
&=O_p\left(\int_{X_{\left(1\right)}}^{X_{\left(n\right)}}\left|\frac{b-a-X_{\left(n\right)}+X_{\left(1\right)}}{\sqrt{\left(b-x\right)\left(x-a\right)}}\right|dx\right)+O_p\left(\int_{X_{\left(1\right)}}^{X_{\left(n\right)}}\left|\frac{ab-X_{\left(n\right)}X_{\left(1\right)}}{x\sqrt{\left(b-x\right)\left(x-a\right)}}\right|dx\right),
\end{aligned}$$
by simple integration calculation, we have 
$$
P_{12}=O_p\left(n^{-\frac13}\right).
$$

Thus, 
$$
\int_a^b|\hp\left(x\right)-p\left(x\right)|dx=O_p\left(n^{-\frac13}\right),
$$
which proves the proposition.

\section{Algorithm} \label{sec:algo}
In this  section, we add the algorithm to detect the spikes and the deviation from MP Law. Algorithm \ref{Detectspikes} gives an automatic method of detecting spikes. The key point in the algorithm is to detect the large gap between the bulk and the spike. The method is comparing difference between each two ordered eigenvalues. The gap between the spike and bulk is much larger than the average level. 
\begin{algorithm}
\caption{Auto Detection of spikes}\label{Detectspikes}
\begin{algorithmic}
\Require Eigenvalues $\{\lambda_i\}_{i=1}^N$, $\alpha\gets7$.
\State Sort $\lambda_i$ with descending order, $\lambda_i>\lambda_{i+1}$
\For{$i=1$; $i<N$; $i++$}
\State $\beta_i\gets \lambda_i-\lambda_{i+1}$
\EndFor

\State $\beta=\sum_{i=1}^{N-1}\beta_i/(N-1)$
\State $r\gets$ $\alpha\cdot\beta$

\For{$i=1$; $i<N/2$; $i++$}
\If{$\beta_i>r$}
\State HS $\gets i$
\EndIf
\EndFor

\For{$i=N-1$; $i>N/2$; $i--$}
\If{$\beta_i>r$}
\State TS $\gets N-i$
\EndIf
\EndFor
\State \Return HS,TS
\end{algorithmic}
\end{algorithm}

Algorithm \ref{Detectspikes} gives an automatic method of detecting spikes. $\alpha$ is the tuning parameter. When the gap between spikes and bulk is larger than $\alpha$ multiply the average difference level, the gap will be detected by Algorithm \ref{Detectspikes}. HS and TS represents the number of spikes larger or smaller than the value of bulk respectively.

After detecting the spikes in Algorithm \ref{Detectspikes}, the deviation measurement between ESDs in weight matrices and standard MP Law is given by Algorithm \ref{DetectSn}. As described in \ref{sec:threetypes}, this algorithm is also used to detect the difference of BT and LT. Algorithm \ref{DetectSn} gives the value of $\hs_n$, discussed in section \ref{sec:specCriterion}, a standard metric for us to measure the deviation from standard MP Law. We also apply this algorithm into detecting the indicated early stopping time.

\newgeometry{top=3cm}
\begin{algorithm}
\caption{Get deviation measurement $\hs_n$}\label{DetectSn}
\begin{algorithmic}
\Require Eigenvalues $\{\lambda_i\}_{i=1}^N$, $\alpha\gets7$.
\State Sort $\lambda_i$ with descending order, $\lambda_i>\lambda_{i+1}$
\State HS,TS $\gets$ Algorithm \ref{Detectspikes}($\{\lambda_i\}_{i=1}^N$, $\alpha$)
\State $n\gets N-$HS$-$TS

\For{$i=1$; $i\leq n$; $i++$}
\State $\gamma_i\gets \lambda_{i+\text{HS}}$            \Comment Get eigenvalues lying in the bulk
\EndFor

\State $M\gets2\lfloor n^{\frac13}\rfloor$   \Comment The number of Bins
\State $H\gets \lfloor n/M\rfloor$
\State $f(x)\gets \frac{2\sqrt{(\gamma_1-x)(x-\gamma_n)}}{\pi(\sqrt{\gamma_1}-\sqrt{\gamma_n})^2x}$
\State $\hs_n=0$
\For{$i=1$; $i<M$; $i++$}
\State $a,b\gets (i-1)H+1,iH+1$
\State $L\gets(b-a)/n/(\gamma_a-\gamma_b)$
\State $s\gets\int_{\gamma_b}^{\gamma_a}|f(x)-L|dx$
\State $\hs_n\gets\hs_n+s$
\EndFor
\State $a,b\gets (M-1)H+1,n$
\State $L\gets(b-a)/n/(\gamma_a-\gamma_b)$
\State $s\gets\int_{\gamma_b}^{\gamma_a}|f(x)-L|dx$
\State $\hs_n\gets\hs_n+s$

\State \Return $\hs_n$
\end{algorithmic}
\end{algorithm}

\section{spectral criterion with $C=0.6$}\label{sec: specrit}
\vspace{-3pt}
In this section, we add extra experimental results of spectral criterion with $C=0.6$. The constant $C$ in the spectral criterion is set a little higher due to the fact that any slight deviation from standard MP Law will suggest to stop if we set $C$ low. Thus the value of $C$ should be determined in some reasonable ways. Combined with simulation results, we add the experimental results of $C=0.6$ to check the spectral criterion.

A higher value of $C$ means that the training time suggested by the criterion will be longer. However, we could not give the value of $C$ too large in an unreasonable way. $C=0.6$ is added below to give us a reasonable interval of $C$. Summarized from the experiments below, $C=0.6$ also gives us good properties of spectral criterion. The spectral criterion detects the problematic issues in the numeric experiments and suggests us to stop even when the training accuracy is increasing. In the real data experiments, the spectral criterion predicts the training explosion quite accurately. Combined with the results of $C=0.4$ and $C=0.6$, the reasonable interval value of $C$ we suggest is [0.4,0.6]. 

\begin{table}[htbp]
\centering
\caption{~~ Early stopping results in numeric experiments with $C=0.6$: stopping epochs selected by spectral criterion in different layers' weight matrices and their testing accuracy (Test Acc). The symbol ”-” means no early stopping epoch is found by the spectral criterion.\label{tbl:earlystoppingsimu1}}
\begin{tabular}{|c|c|c|c|c|c|c|c|}
\multicolumn{8}{c}{The combination NN1+$\calD_1$}\\
\hline
\multirow{2}{*}{\begin{tabular}[c]{@{}c@{}}Typical\\ TP\end{tabular}} & \multicolumn{4}{c|}{spectral criterion $C=0.6$}          & \multicolumn{3}{c|}{Final Epoch 248} \\ \cline{2-8} 
                                                                      & epoch(FC2)  & Test Acc & epoch(FC3)  & Test Acc & FC1       & FC2      & Test Acc      \\ \hline
0.15                                                                  & 8           & 24.58\%  & 16          & 20.44\%  & HT        & HT       & 20.17\%       \\ \hline
0.2                                                                   & 8           & 31.50\%  & 16          & 25.83\%  & HT        & HT       & 27.03\%       \\ \hline
0.3                                                                   & 8           & 49.09\%  & 12          & 45.48\%  & HT        & HT       & 44.80\%       \\ \hline
0.6                                                                   & 9           & 87.96\%  & \multicolumn{2}{c|}{-} & BT       & BT      & 88.30\%       \\ \hline
0.9                                                                   & \multicolumn{2}{c|}{-} & \multicolumn{2}{c|}{-} & LT        & LT       & 99.13\%       \\ \hline
\end{tabular}
\vskip 2mm

\begin{tabular}{|c|c|c|c|c|c|c|c|}
\multicolumn{8}{c}{The combination NN1+$\calD_2$}\\
\hline
\multirow{2}{*}{\begin{tabular}[c]{@{}c@{}}Typical\\ TP\end{tabular}} & \multicolumn{4}{c|}{spectral criterion $C=0.6$}  & \multicolumn{3}{c|}{Final Epoch 248} \\ \cline{2-8} 
                                                                      & epoch(FC2)  & Test Acc  & epoch(FC3)  & Test Acc & FC1       & FC2      & Test Acc      \\ \hline
0.24                                                                  & 9           & 14.69\%   & \multicolumn{2}{c|}{-} & HT        & HT       & 13.08\%       \\ \hline
1.2                                                                   & 8           & 39.31\%   & 16          & 32.11\%  & HT        & HT       & 32.98\%       \\ \hline
2.4                                                                   & 8           & 76.75\%   & 20          & 74.29\%  & HT        & HT       & 75.92\%       \\ \hline
3.2                                                                   & 10          & 91.94\%   & \multicolumn{2}{c|}{-} & HT        & LT      & 92.64\%       \\ \hline
4.8                                                                   & \multicolumn{2}{c|}{-}  & \multicolumn{2}{c|}{-} & LT        & LT       & 99.73\%       \\ \hline
\end{tabular}

\vskip 2mm

\begin{tabular}{|c|c|c|c|c|c|c|c|}
\multicolumn{8}{c}{The combination NN2+$\calD_1$}\\
\hline
\multirow{2}{*}{\begin{tabular}[c]{@{}c@{}}Typical\\ TP\end{tabular}} & \multicolumn{4}{c|}{spectral criterion $C=0.6$}  & \multicolumn{3}{c|}{Final Epoch 248} \\ \cline{2-8} 
                                                                      & epoch(FC1)  & Test Acc  & epoch(FC2)  & Test Acc & FC1       & FC2      & Test Acc      \\ \hline
0.02                                                                  & \multicolumn{2}{c|}{-}  & \multicolumn{2}{c|}{-} & HT        & BT      & 16.02\%       \\ \hline
0.04                                                                  & \multicolumn{2}{c|}{-}  & \multicolumn{2}{c|}{-} & HT        & BT      & 25.38\%       \\ \hline
0.07                                                                  & \multicolumn{2}{c|}{-}  & \multicolumn{2}{c|}{-} & HT        & BT      & 50.12\%       \\ \hline
0.13                                                                  & \multicolumn{2}{c|}{-}  & \multicolumn{2}{c|}{-} & BT       & LT       & 87.50\%       \\ \hline
0.2                                                                   & \multicolumn{2}{c|}{-}  & \multicolumn{2}{c|}{-} & LT        & LT       & 99.14\%       \\ \hline
\end{tabular}
\vskip 2mm

\begin{tabular}{|c|c|c|c|c|c|c|c|}
\multicolumn{8}{c}{The combination NN2+$\calD_2$}\\
\hline
\multirow{2}{*}{\begin{tabular}[c]{@{}c@{}}Typical\\ TP\end{tabular}} & \multicolumn{4}{c|}{spectral criterion $C=0.6$}  & \multicolumn{3}{c|}{Final Epoch 248} \\ \cline{2-8} 
                                                                      & epoch(FC1)  & Test Acc  & epoch(FC2)  & Test Acc & FC1       & FC2      & Test Acc      \\ \hline
0.24                                                                  & 12          & 13.48\%   & 7           & 13.19\%  & HT        & HT       & 13.44\%       \\ \hline
1.2                                                                   & 24          & 35.80\%   & 5           & 34.63\%  & HT        & HT       & 36.31\%       \\ \hline
2.4                                                                   & 6           & 73.80\%   & 36          & 74.86\%  & BT       & BT      & 75.12\%       \\ \hline
3.2                                                                   & \multicolumn{2}{c|}{-}  & \multicolumn{2}{c|}{-} & LT        & LT       & 91.20\%       \\ \hline
4.8                                                                   & \multicolumn{2}{c|}{-}  & \multicolumn{2}{c|}{-} & LT        & LT       & 99.59\%         \\ \hline
\end{tabular}
\end{table}

\vspace{-30pt}
\begin{table}[htbp]
\centering
\caption{~~  Early stopping results in real data experiments with $C=0.6$.\label{tbl:earlystopingreal1}}
\setlength{\textwidth}{1.5mm}{
\begin{tabular}{|c|c|c|c|c|c|c|c|}
\multicolumn{8}{c}{The combination LeNet+MNIST}\\
\hline
\multirow{2}{*}{batchsize} & \multicolumn{4}{c|}{spectral criterion $C=0.6$}                 & \multicolumn{3}{c|}{Final Epoch 248} \\ \cline{2-8} 
                           & epoch(FC1)  & Test Acc & epoch(FC2)  & Test Acc & FC1       & FC2      & Test Acc      \\ \hline
16                         & \multicolumn{2}{c|}{-} & 32          & 99.19\%  & LT        & BT      & 99.17\%       \\ \hline
32                         & \multicolumn{2}{c|}{-} & \multicolumn{2}{c|}{-} & LT        & BT      & 99.17\%       \\ \hline
64                         & \multicolumn{2}{c|}{-} & \multicolumn{2}{c|}{-} & LT        & BT      & 98.98\%       \\ \hline
128                        & \multicolumn{2}{c|}{-} & \multicolumn{2}{c|}{-} & LT        & BT       & 99.03\%       \\ \hline
256                        & \multicolumn{2}{c|}{-} & \multicolumn{2}{c|}{-} & LT        & LT       & 98.96\%       \\ \hline
\end{tabular}}
\vskip 2mm
\begin{tabular}{|c|c|c|c|c|c|c|c|}
\multicolumn{8}{c}{The combination LeNet+CIFAR10}\\
\hline
\multirow{2}{*}{batchsize} & \multicolumn{4}{c|}{spectral criterion $C=0.6$}              & \multicolumn{3}{c|}{Final Epoch 248} \\ \cline{2-8} 
                           & epoch(FC1)  & Test Acc & epoch(FC2) & Test Acc & FC1       & FC2      & Test Acc      \\ \hline
16                         & 28          & 61.66\%  & 8          & 61.62\%  & BT       & HT       & 64.99\%       \\ \hline
32                         & \multicolumn{2}{c|}{-} & 20         & 61.06\%  & BT        & HT       & 64.57\%       \\ \hline
64                         & \multicolumn{2}{c|}{-} & 32         & 60.27\%  & LT        & BT       & 62.49\%       \\ \hline
128                        & \multicolumn{2}{c|}{-} & 60         & 61.38\%  & LT        & BT       & 61.83\%       \\ \hline
256                        & \multicolumn{2}{c|}{-} & 92         & 58.33\%  & LT        & BT       & 60.49\%       \\ \hline
\end{tabular}

\vskip 2mm
\begin{tabular}{|c|c|c|c|c|c|c|c|}
\multicolumn{8}{c}{The combination MiniAlexNet+MNIST}\\
\hline
\multirow{2}{*}{batchsize} & \multicolumn{4}{c|}{spectral criterion $C=0.6$}                 & \multicolumn{3}{c|}{Final Epoch 248} \\ \cline{2-8} 
                           & epoch(FC1)  & Test Acc & epoch(FC2)  & Test Acc & FC1       & FC2      & Test Acc      \\ \hline
16                         & 5           & 98.64\%  & \multicolumn{2}{c|}{-} & BT       & LT       & 99.49\%       \\ \hline
32                         & \multicolumn{2}{c|}{-} & \multicolumn{2}{c|}{-} & BT        & LT       & 99.41\%       \\ \hline
64                         & \multicolumn{2}{c|}{-} & \multicolumn{2}{c|}{-} & LT        & LT       & 99.42\%       \\ \hline
128                        & \multicolumn{2}{c|}{-} & \multicolumn{2}{c|}{-} & LT        & LT       & 99.39\%       \\ \hline
256                        & \multicolumn{2}{c|}{-} & \multicolumn{2}{c|}{-} & LT        & LT       & 99.31\%       \\ \hline
\end{tabular}

\vskip 2mm
\setlength{\textwidth}{1.5mm}{
\begin{tabular}{|c|c|c|c|c|c|c|c|}
\multicolumn{8}{c}{The combination MiniAlexNet+CIFAR10}\\
\hline
\multirow{2}{*}{batchsize} & \multicolumn{4}{c|}{spectral criterion $C=0.6$}              & \multicolumn{3}{c|}{Final Epoch 248} \\ \cline{2-8} 
                           & epoch(FC1) & Test Acc & epoch(FC2)  & Test Acc & FC1     & FC2     & Test Acc         \\ \hline
16                         & 4          & 71.01\%  & 36(RC)      & 55.6\%   & HT      & RC      & 10\%(explode)    \\ \hline
32                         & 4          & 72.17\%  & 196(RC)     & 62.84\%  & HT      & RC      & 10\%(explode)    \\ \hline
64                         & 6          & 73.03\%  & \multicolumn{2}{c|}{-} & BT      & BT      & 77.94\%          \\ \hline
128                        & 12         & 74.31\%  & \multicolumn{2}{c|}{-} & BT      & LT      & 77.43\%          \\ \hline
256                        & 28         & 75.87\%  & \multicolumn{2}{c|}{-} & BT     & LT      & 75.93\%          \\ \hline
\end{tabular}}
\end{table}

\end{document}